\documentclass[nohyperref]{article}

\usepackage[hidelinks]{hyperref}
\usepackage{bookmark}
\hypersetup{
    colorlinks=false,
}
\usepackage{amssymb}
\usepackage{mathtools}
\usepackage{amsthm}
\theoremstyle{plain}

\theoremstyle{definition}

\theoremstyle{remark}

\usepackage[textsize=tiny]{todonotes}
\usepackage{microtype}
\usepackage{graphicx}
\usepackage[skip=0pt,belowskip=-2pt]{caption}
\usepackage{subcaption}
\usepackage{booktabs} % for professional tables
\usepackage{xcolor}
\usepackage{multirow}
\usepackage{float}
\usepackage{dblfloatfix}
\usepackage{chngcntr}
\usepackage{wrapfig}

\usepackage[algoruled,lined,linesnumbered]{algorithm2e} 

\usepackage{amsfonts, bm, bbm, mathtools, amsmath}

\newcommand{\bbR}{\mathbb{R}}

\newcommand{\calS}{\mathcal{S}}
\newcommand{\calA}{\mathcal{A}}

\newcommand{\bfa}{\mathbf{a}}

\newcommand{\bfw}{\mathbf{w}}
\newcommand{\bfu}{\mathbf{u}}
\newcommand{\bfz}{\mathbf{z}}
\newcommand{\bfs}{\mathbf{s}}
\newcommand{\bfx}{\mathbf{x}}
\newcommand{\bfv}{\mathbf{v}}

\usepackage[accepted]{icml2022}

\icmltitlerunning{Towards Evaluating Adaptivity of Model-Based Reinforcement Learning Methods}
\begin{document}

\twocolumn[
\icmltitle{Towards Evaluating Adaptivity of\\ Model-Based Reinforcement Learning Methods}
           
\icmlsetsymbol{equal}{*}

\begin{icmlauthorlist}
\icmlauthor{Yi Wan}{equal,UofA,Mila}
\icmlauthor{Ali Rahimi-Kalahroudi}{equal,Mila,Udem}
\icmlauthor{Janarthanan Rajendran}{Mila,Udem}
\icmlauthor{Ida Momennejad}{Microsoft}
\icmlauthor{Sarath Chandar}{Mila,Poly,Cifar}
\icmlauthor{Harm van Seijen}{Microsoft}
\end{icmlauthorlist}

\icmlaffiliation{Mila}{Mila - Quebec AI Institute}
\icmlaffiliation{Poly}{Ecole Polytechnique de Montreal}
\icmlaffiliation{Cifar}{Canada CIFAR AI Chair}
\icmlaffiliation{Udem}{Universite de Montreal}
\icmlaffiliation{UofA}{University of Alberta}
\icmlaffiliation{Microsoft}{Microsoft}

\icmlcorrespondingauthor{Yi Wan}{wan6@ualberta.ca}
\icmlcorrespondingauthor{Ali Rahimi-Kalahroudi}{ali-rahimi.kalahroudi@mila.quebec}

\icmlkeywords{Reinforcement Learning, Planning, Model-based Reinforcement Learning, Continual Learning}

\vskip 0.3in
]

\printAffiliationsAndNotice{\icmlEqualContribution} % otherwise use the standard text.

\begin{abstract}
In recent years, a growing number of deep model-based reinforcement learning (RL) methods have been introduced. The interest in deep model-based RL is not surprising, given its many potential benefits, such as higher sample efficiency and the potential for fast adaption to changes in the environment. However, we demonstrate, using an improved version of the recently introduced Local Change Adaptation (LoCA) setup, that well-known model-based methods such as PlaNet and DreamerV2 perform poorly in their ability to adapt to local environmental changes.  Combined with prior work that made a similar observation about the other popular model-based method, MuZero, a trend appears to emerge, suggesting that current deep model-based methods have serious limitations. We dive deeper into the causes of this poor performance, by identifying elements that hurt adaptive behavior and linking these to underlying techniques frequently used in deep model-based RL. We empirically validate these insights in the case of linear function approximation by demonstrating that a modified version of linear Dyna achieves effective adaptation to local changes. Furthermore, we provide detailed insights into the challenges of building an adaptive nonlinear model-based method, by experimenting with a nonlinear version of Dyna.

\end{abstract}

\section{Introduction}

While model-free reinforcement learning (RL) is still the dominant approach in deep RL, more and more research on deep model-based RL appears \citep{wang2019benchmarking,moerland2020model}. 
This is hardly surprising, as model-based RL (MBRL), which leverages estimates of the reward and transition model, could hold the key to some persistent challenges in deep RL, 
such as sample efficiency and effective adaptation to environment changes. Although deep model-based RL has gained momentum, there are questions to be raised about the proper way to evaluate its progress. A common performance metric is sample-efficiency in a single task \citep{wang2019benchmarking}, which has several disadvantages. First, it conflates progress due to model-based RL with other factors such as representation learning. More importantly, it is unclear whether in an single-task setting, model-based RL is always more sample-efficient than model-free RL because learning a policy directly is not necessarily slower than learning a model and planning with the model. By contrast, when it comes to solving multiple tasks that share (most of) the dynamics, it is arguable that model-based RL has a clear advantage.

Instead of measuring the sample-efficiency of an algorithm in a single task, \citet{vanseijen-LoCA} developed the Local Change Adaptation (LoCA) setup to measure the agent's ability to adapt when the task changes. This approach, inspired by approaches used in neuroscience for measuring model-based behavior in humans and rodents, is designed to measure how quickly an RL algorithm can adapt to a local change in the reward using its learned environment model. They used this to show that the deep model-based method MuZero \cite{schrittwieser2019mastering}, which achieves great sample-efficiency on Atari, was not able to effectively adapt to a local change in the reward, even on simple tasks.

This paper builds out this direction further. First, we improve the original LoCA setup, such that it 
is simpler, less sensitive to its hyperparameters
and can be more easily applied to stochastic environments. Our improved setup is designed to make a binary classification of model-based methods: those that can effectively adapt to local changes in the environment and those that cannot. 

We apply our improved setup to the MuJoCo Reacher domain and use it to evaluate two continuous-control model-based methods, PlaNet and DreamerV2 \cite{hafner2019learning, hafner2019dream, hafner2020mastering}. Both methods turn out to adapt poorly to local changes in the environment. Combining these results with the results from \citet{vanseijen-LoCA}, which showed a similar shortcoming of MuZero, a trend appears to emerge, suggesting that modern deep model-based methods are unable to adapt effectively to local changes in the environment.

We take a closer look at what separates model-based methods that adapt poorly from model-based methods that adapt effectively, by evaluating various tabular model-based methods. This leads us to define two failure modes that prohibit adaptivity. The first failure mode is linked to MuZero, potentially justifying its poor adaptivity to local changes. Further analysis of the Planet and the DreamerV2 methods enables us to identify two more failure modes that are unique to approximate (i.e., non-tabular) model-based methods.

Using the insights about important failure modes, we set off to design adaptive model-based methods that rely on function approximation. First, we demonstrate that by making small modifications to the classical linear Dyna method. The resulting algorithm adapts effectively in a 
% difficult 
challenging setting (sparse reward and stochastic transitions). We then perform experiments with a nonlinear version of our adaptive linear Dyna algorithm. For the nonlinear Dyna algorithm, we are not able to achieve effective adaptation, as our third and forth identified failure modes appear to be difficult to overcome.

\section{The Improved LoCA Setup}\label{sec: Measuring Adaptivity of Model-Based Methods}

In this section, we present the LoCA setup introduced by \citet{vanseijen-LoCA}, as well as our improved version that is simpler and more robust. The LoCA setup consists of a task configuration and an experiment configuration.  The LoCA setup is inspired by how model-based behavior is identified in behavioral neuroscience (e.g., see \citet{daw2011model}).

The task configuration is the same for the original and our improved version of the LoCA setup and is discussed next. The original experiment configuration is discussed in Section \ref{sec:Loca experiment - original}; Section \ref{sec:Loca experiment - improved} discusses our improved version.

\subsection{Task Configuration}\label{sec: Simplified LoCA Setup}

\begin{figure}[th]
\centering
\includegraphics[width=\columnwidth]{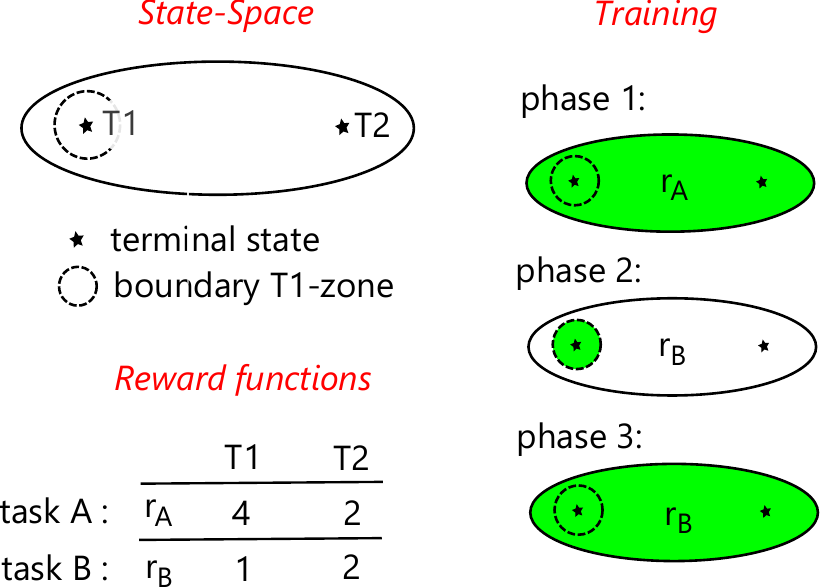}
\caption{\emph{Left:} LoCA task configuration. \emph{Right:} The initial state distribution during training across the three phases of a LoCA experiment.}

\label{fig:simplified LoCA setup}
\end{figure}

The LoCA task configuration considers two different tasks (i.e., reward functions) in the same environment. A method's adaptivity is determined by measuring how effective it can adapt  from the first to the second task. The task configuration only specifies some specific features that the environment should have. In practise, many different domains can be used as the basis for a LoCA environment, ranging from tabular environments with discrete actions to environments with high-dimensional, continuous state and action spaces.

A LoCA environment contains two terminal states, T1 and T2 (see left of Figure \ref{fig:simplified LoCA setup}). Around T1 is a local area that, once entered, the agent is unable to move out of without terminating the episode, regardless of its policy. We refer to this local area as the \emph{T1-zone}. The boundary of the T1-zone can be viewed as a one-way passage. The reward function for task A, $r_A$, and task B, $r_B$, are 0 everywhere, except upon transitions to a terminal state. A transition to T1 results in a reward of 4 under $r_A$ and 1 under $r_B$; transitioning to T2 results in a reward of 2 under both $r_A$ and $r_B$. 
The discount factor $0 < \gamma < 1$ is the same for task A and task B. Note that, while $r_A$ and $r_B$ only differ locally, the optimal policy changes for almost all states: for task A, the optimal policy points towards T1 for the majority of the state-space, while for task B it points towards T2 (except for states within the T1-zone).  

\subsection{Experiment Configuration - Original Version}
\label{sec:Loca experiment - original}

An experiment consists of three different training phases. During Phase 1, the reward function is $r_A$; upon transitioning to Phase 2, the reward function changes to $r_B$ and remains unchanged upon transitioning to Phase 3. 

Crucially, the initial state distribution during training is different for the different phases  (see right of Figure \ref{fig:simplified LoCA setup}). In Phases 1 and 3, the initial state is drawn uniformly at random from the full state space\footnote{Other initial-state distributions are possible too, as long as the distribution enables experiences from across the full state-space.}; in Phase 2, it is drawn from the T1-zone. As the agent cannot escape the T1-zone without terminating the episode, during Phase 2 only samples from the T1-zone are observed. 

The key question that determines adaptivity is whether or not a method can adapt effectively to the new reward function during phase 2. That is, can it change its policy from pointing towards T1 to pointing towards T2 across the state-space, while only observing samples from the local area around T1. Effective adaptation in Phase 2 implies the performance is optimal out of the gate in Phase 3. In the original LoCA setup, evaluation only occurs in Phase 3. If a method does not perform optimal right out of the gate (i.e., does not adapt effectively), the amount of additional training needed in Phase 3 to get optimal performance is used as a measure for how far off the behavior of a method is from ideal adaptive behavior. 

Evaluation occurs by freezing learning periodically during training (i.e., not updating its weights, internal models or replay buffer), and executing its policy for a number of evaluation episodes. During evaluation, an initial-state distribution is used that covers a small area of the state-space roughly in the middle of T1 and T2. The fraction of episodes the agent ends up in the high-reward terminal (T2 for Phase 3) within a certain cut-off time is used as measure for how good its policy is. This measure is called the \emph{top-terminal fraction} and its values are always between 0 (poor performance) and 1 (good performance).

The regret of this metric during Phase 3 with regards to optimal performance is called the \emph{LoCA regret}. A LoCA regret of 0 implies the agent has a top-terminal fraction of 1 out of the gate in Phase 3 and means the agent adapts effectively to local changes.

The original experiment configuration has a number of disadvantages. It involves various hyperparameters, such as the exact placement of the initial-state distribution for evaluation and the cut-off time for reaching T2 that can affect the value of the LoCA metric considerably. 
Furthermore, in stochastic environments, even an optimal policy could end up at the wrong terminal by chance. Hence, in stochastic environments a LoCA regret of 0 is not guaranteed even for adaptive methods. 

Finally, measuring the amount of training needed in Phase 3 to determine how far off a method is from `ideal' behavior is questionable. Adaptivity in Phase 2 is fundamentally different from adaptivity in Phase 3. Adaptivity in Phase 2 requires a method to propagate newly observed reward information to parts of the state-space not recently visited such that the global policy changes to the optimal one under the new reward function, an ability classically associated with model-based learning. By contrast, adaptivity in Phase 3, where the agent observes samples throughout the full state-space, is a standard feature of most RL algorithms, regardless whether they are model-based or not. And while leveraging a learned environment model can reduce the amount of (re)training required, so do many other techniques not unique to model-based learning, such as representation learning or techniques to improve exploration. So beyond determining whether a method is adaptive (LoCA regret of 0) or not (LoCA regret larger than 0), the LoCA regret does not tell much about model-based learning.

\subsection{Experiment Configuration- Improved Version}
\label{sec:Loca experiment - improved}

To address the shortcomings of the original LoCA experiment configuration, we introduce an improved version. In this improved version we evaluate performance by simply measuring the average return over the evaluation episodes and comparing it with the average return of the corresponding optimal policy. Furthermore, as initial-state distribution, the full state-space is used instead of some area in between T1 and T2. Finally, we evaluate the performance during all phases, instead of only the third phase. 

Under our new experiment configuration, we call a method adaptive if it is able to reach (near) optimal expected return in Phase 2 (after sufficiently long training), while also reaching (near) optimal expected return in Phase 1. If a method is able to reach (near) optimal expected return in Phase 1, but not in Phase 2, we call the method non-adaptive. Finally, if a method is unable to reach near-optimal expected return in Phase 1, even after training for a long time, we do not make an assessment of its adaptivity.

Using the expected return to evaluate the quality of the policy instead of a top-terminal makes evaluation a lot more flexible and robust. Not only does it remove the cut-off time hyperparameter, it enables the use of the full state-space as initial-state distribution and can be applied without modification to stochastic environments. These scenarios were tricky for the top-terminal fraction, as it had as requirement that adaptive methods should be able to get a top-terminal fraction of 1.

Finally, our improved LoCA setup no longer tries to assess how far off from ideal adaptive behavior a method's behavior is---a measure conflated by various confounders, as discussed above. Instead, only a binary assessment of adaptivity is made, simplifying evaluation.

Note that with our improved evaluation methodology, Phase 3 is no longer required to evaluate the adaptivity of a method. However, it can be useful to rule out two---potentially easily fixable---causes for poor adaptivity in Phase 2. In particular, if  after training for a long time in Phase 3 the performance plateaus at some suboptimal level, two things can be the case.
First, a method might simply not be able to get close-to-optimal performance in task B regardless of the samples it observes. Second, some methods are designed with the assumption of a stationary environment and cannot adapt to \emph{any} changes in the reward function. This could happen, for example, if a method decays exploration and/or learning rates such that learning becomes less effective over time. In both cases, tuning learning hyperparameters or making minor modifications to the method might help. 

\section{Adaptive Versus Non-Adaptive MBRL}\label{sec: tabular}

The single-task sample efficiency of a method says little about its ability to adapt effectively to local changes in the environment. Moreover, seemingly small discrepancies among different MBRL methods can greatly affect their adaptivity.
In Section \ref{sec: three MBRL methods}, we will illustrate both these points by evaluating three different tabular MBRL methods in the GridWorldLoCA domain (Figure \ref{fig:gridworld}), introduced by \citet{vanseijen-LoCA}, using the LoCA setup \footnote{For simplicity, from now on, we use `the LoCA setup' to denote the \emph{improved} LoCA setup unless specified otherwise.}. In Section \ref{sec: failure modes} we discuss in more detail why some of the tabular MBRL methods fail to adapt effectively. The code of all of the experiments presented in this paper is available at \url{https://github.com/chandar-lab/LoCA2}.

\subsection{Tabular MBRL Experiment}
\label{sec: three MBRL methods}

Each of the three methods we introduce learns an estimate of the environment model (i.e., the transition and reward function). We consider two 1-step models and one 2-step model. The 1-step model consists of $\hat p(s'|s,a)$ and $\hat r(s,a)$ that estimate the 1-step transition and expected reward function, respectively.
Upon observing sample $(S_t, A_t, R_t, S_{t+1})$, this model is updated according to:
\begin{eqnarray*}
\hat r(S_t,A_t) &\leftarrow& \hat r(S_t,A_t) + \alpha \big( R_t  - \hat r(S_t,A_t)\big)\,, \\
\hat p(\cdot|S_t,A_t) &\leftarrow& \hat p(\cdot|S_t,A_t) + \alpha \big( \langle  S_{t+1} \rangle  - \hat p(\cdot|S_t,A_t)\big)\,,
\end{eqnarray*}
with $\alpha$ the (fixed) learning rate and $\langle  S_{t+1} \rangle$ a one-hot encoding of state $S_{t+1}$. Both $\langle  S_{t+1} \rangle$ and $p(\cdot|S_t,A_t)$ are vectors of length $N$, the total number of states.
Planning consists of performing a single state-value update at each time step based on the model estimates.\footnote{This is a special case of asynchronous value iteration, as discussed for example in Section 4.5 of \cite{sutton2018reinforcement}.} For a 1-step model:
$V(s) \leftarrow \max_{a} \Big(\hat r(s,a)  + \gamma \sum_{s'} \hat p(s'|s,a) V(s') \Big)\,.$

We evaluate two variations of this planning routine: \mbox{\emph{mb-1-r}}, where the state that receives the update is selected at random from all possible states; and \mbox{\emph{mb-1-c}}, where the state that receives the update is the current state. Action selection occurs in an $\epsilon$-greedy way, where the action-values of the current state are computed by doing a lookahead step using the learned model and bootstrapping from the state-values. 

We also evaluate \mbox{\emph{mb-2-r}}, which is similar to \mbox{\emph{mb-1-r}} except that it uses a 2-step model, which estimates---under the agent's behavior policy---a distribution of the state 2 time steps in the future and the expected discounted sum of rewards over the next 2 time steps. The update equations for this method, as well as further experiment details, can be found in Section \ref{app: Experiment Procedure of the tabular LoCA setup}. 

A summary of the methods we evaluate is shown in Table \ref{tab:various model-based methods}. Besides these methods, we also test the performance of the model-free method Sarsa($\lambda$) with $\lambda=0.95$. 
\begin{table}[hbt]
\centering
    \caption{Tabular model-based methods being evaluated.}
    \label{tab:various model-based methods}
    \vspace{5pt}
    \begin{tabular}{l|c|c}
      \toprule % <-- Toprule here
      Method & Model & State receiving value update \\
      \midrule % <-- Midrule here
      \emph{mb-1-r} & 1-step & Randomly selected \\
      \emph{mb-1-c} & 1-step & Current state \\
      \emph{mb-2-r} & 2-step & Randomly selected\\
      \bottomrule % <-- Bottomrule here
    \end{tabular}
\end{table}

\begin{figure*}[t]
\centering
% \begin{subfigure}{.25\textwidth}
%     \centering
%     % \hspace{10mm}
%     \includegraphics[width=\textwidth]{ICML2022_figures/gridworld_ext.pdf}
%     \caption{}
%     \label{fig:gridworld}
% \end{subfigure}%
% \begin{subfigure}{.25\textwidth}
%     \includegraphics[width=1.5\textwidth]{ICML2022_figures/tabular_results.pdf}
%     \caption{}
% \end{subfigure}
\begin{subfigure}{.35\textwidth}
    \centering
    \includegraphics[height=3.0cm]{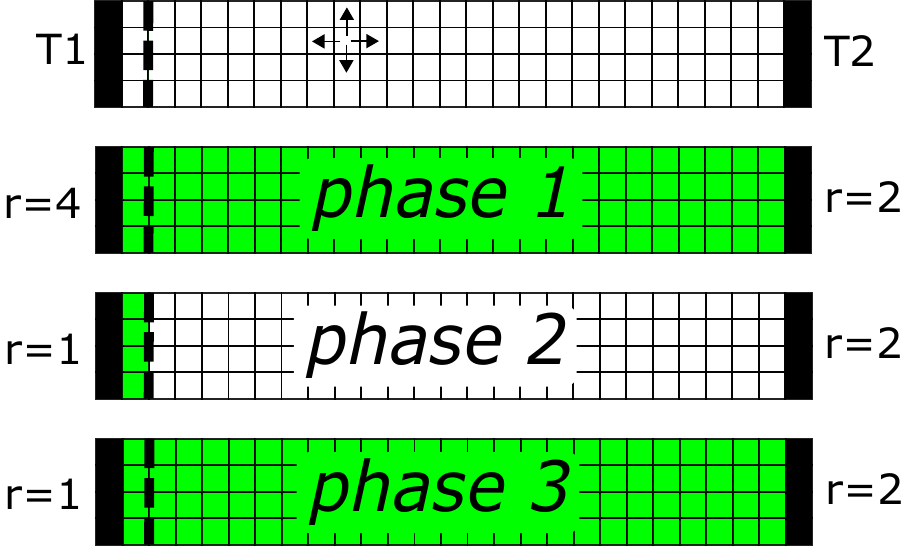}
    \hspace{10mm}
    \caption{}
    \label{fig:gridworld}
\end{subfigure}%
\begin{subfigure}{.55\textwidth}
    \includegraphics[height=4.0cm]{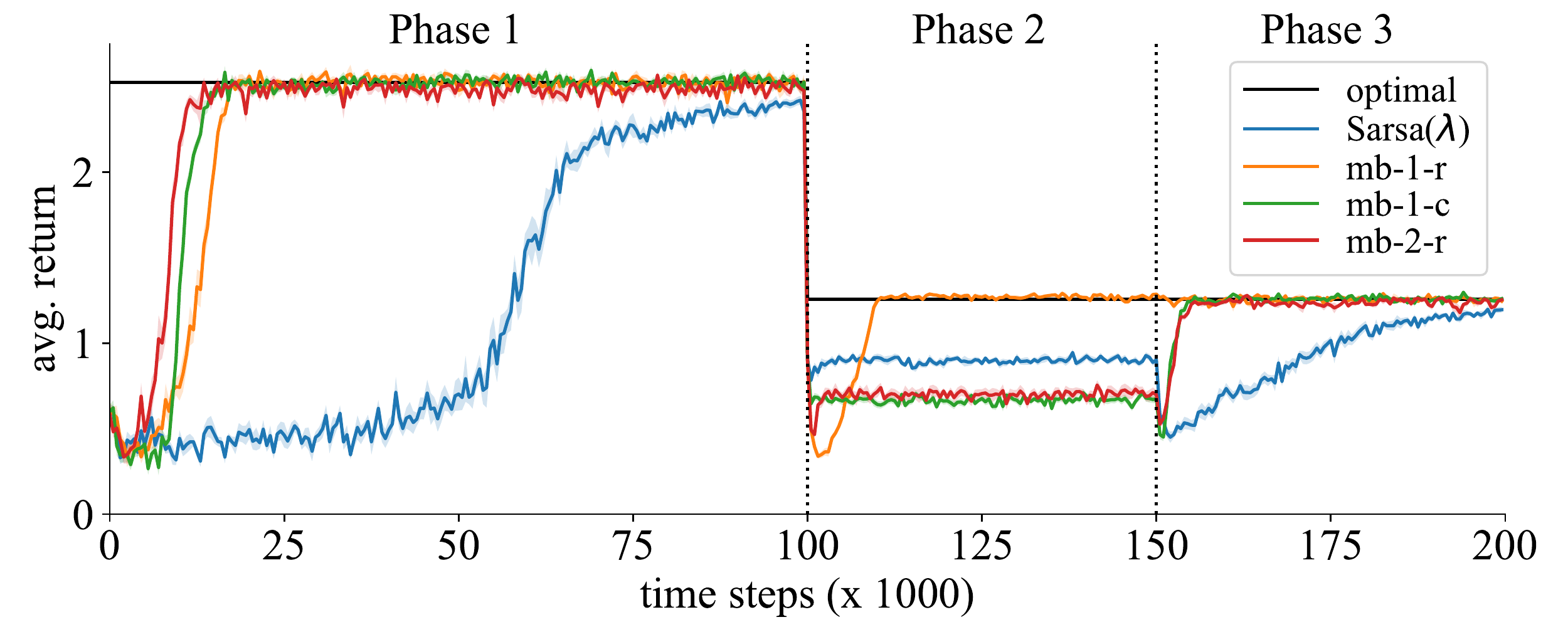}
    \caption{}
\end{subfigure}
\caption{a) GridWorldLoCA domain (top) and the rewards and initial-state training distributions (in green) for the different phases. b) Comparison of the 3 model-based methods from Table \ref{tab:various model-based methods} on the GridworldLoCA domain. While all methods converge to the optimal policy in Phases 1 and 3, only \emph{mb-1-r} converges in Phase 2. We call model-based methods that are able to converge to optimal performance in Phase 2 (locally) adaptive model-based methods.}
\label{fig: tabular results}
\end{figure*} 

We use a stochastic version of the GridWorldLoCA domain, where the action taken results
with 25\% probability in a move in a random direction instead of the preferred direction. The initial-state distribution for training in Phases 1 and 3, as well as the initial-state distribution for evaluation in all three phases is equal to the uniform random distribution across the full state-space; the initial-state distribution for training in phase 2 is equal to the uniform random distribution across the T1-zone.

Figure \ref{fig: tabular results} shows the performance of the various methods, averaged over 10 independent runs. While all three MBRL methods have similar performance in Phase 1 (i.e., similar single-task sample-efficiency), their performance in Phase 2 (i.e., their adaptivity to local changes in the environment) is very different. Specifically, even though the methods are very similar, only \emph{mb-1-r} is able to change its policy to the optimal policy of task B during Phase 2. By contrast, \emph{mb-1-c} and \emph{mb-2-r} lack the flexibility to adapt effectively to the new task. Note that Sarsa($\lambda$) achieves a higher average return in phase 2 than \emph{mb-1-c} and \emph{mb-2-r}. This may seem to be odd, given that it is a model-free method. There is however a simple explanation: the policy of Sarsa($\lambda$) in phase 2 still points to T1, which now results in a reward of 1 instead of 4. By contrast, the policy changes for \emph{mb-1-c} and \emph{mb-2-r} are such that the agent neither moves to T1 nor T2 directly, instead, it mostly moves back and forth during the evaluation period, resulting in a reward of 0 most of time.

\subsection{Discussion}
\label{sec: failure modes}

The tabular MBRL experiment illustrates two different reasons why a model-based method may fail to adapt. 

\emph{\textbf{Failure Mode \#1:} Planning relies on a value function, which only gets updated for the current state.} 

This applies to \emph{mb-1-c}. A local change in the environment or reward function can results in a different value function across the entire state-space. Since \emph{mb-1-c} only updates the value of the current state, during phase 2 only states within the T1-zone are updated. And because evaluation uses an initial-state distribution that uses the full state-space, the average return will be low. 

\emph{\textbf{Failure Mode \#2:} The prediction targets of the learned environment model are implicitly conditioned on the policy used during training.}

This is the case for \emph{mb-2-r}, as the model predicts the state and reward 2 time steps in the future, using only the current state and action as inputs. Hence, there is an implicit dependency on the behavior policy. For \emph{mb-2-r}, the behavior policy is the $\epsilon$-greedy policy, and during the course of training in Phase 1, this policy converges to the (epsilon-)optimal policy under reward function $r_A$. As a consequence, the environment model being learned will converge to a version that is implicitly conditioned on an optimal policy under $r_A$ as well. During training in Phase 2, this dependency remains for states outside of the T1-zone (which are not visited during Phase 2), resulting in poor performance.

These two failure modes, while illustrated using tabular representations, apply to linear and deep representations as well, as the underlying causes are independent of the representation used. In fact, we believe 
that Failure Mode \#1 is in part responsible for the poor adaptivity of MuZero, as shown in \citet{vanseijen-LoCA}. MuZero relies on a value function that gets updated using update targets based on values that are computed only for the visited states from an episode-trajectory, which is similar to Failure Mode \#1, as we explain in more detail in Appendix \ref{app: MuZero}. Furthermore, an example of a linear multi-step MBRL method that Failure Mode \#2 applies to is LS-Sarsa($\lambda$) introduced by \citet{vanseijen2015deeper}.

\section{Evaluating PlaNet and DreamerV2}\label{sec: planet and dreamer}

In this section, we evaluate the adaptivity of two deep model-based methods, PlaNet \cite{hafner2019learning} and the latest version of Dreamer, called DreamerV2 \cite{hafner2020mastering}, using a modified version of the LoCA setup in a variant of the Reacher domain, which involves a continuous-action domain with $64 \times 64 \times 3$ dimensional (pixel-level) states.

\subsection{The ReacherLoCA Domain}
We introduce a variation on the Reacher environment (the easy version) available from the DeepMind Control Suite \cite{tassa2018deepmind}. The Reacher environment involves controlling the angular velocity of two connected bars in a way such that the tip of the second bar is moved on top of a circular target. The reward is 1 at every time step that the tip is on top of the target, and 0 otherwise. An episode terminates after exactly 1000 time steps. The target location and the orientation of the bars are randomly initialized at the start of each episode. 

\begin{figure*}[h!]\label{fig: reacherloca env}
\centering
    \begin{subfigure}[h]{.3\textwidth}
        \centering
        \includegraphics[width=0.75\textwidth]{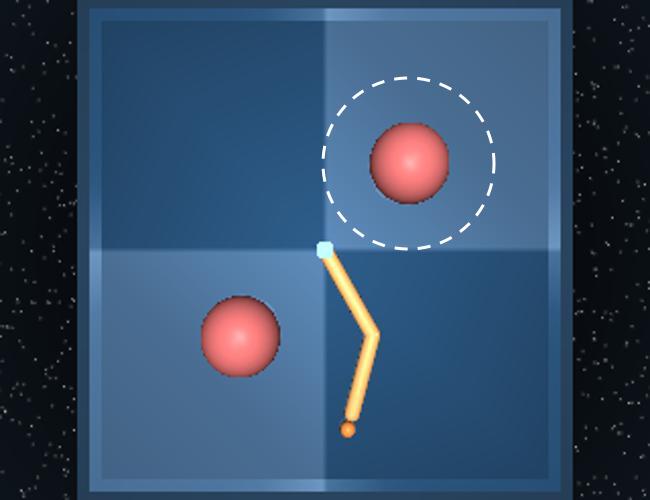}
        \hspace{20mm}
        \caption{The ReacherLoCA Domain}
    \end{subfigure}%
    \begin{subfigure}[h]{.33\textwidth}
        \centering
        \includegraphics[width=0.95\textwidth]{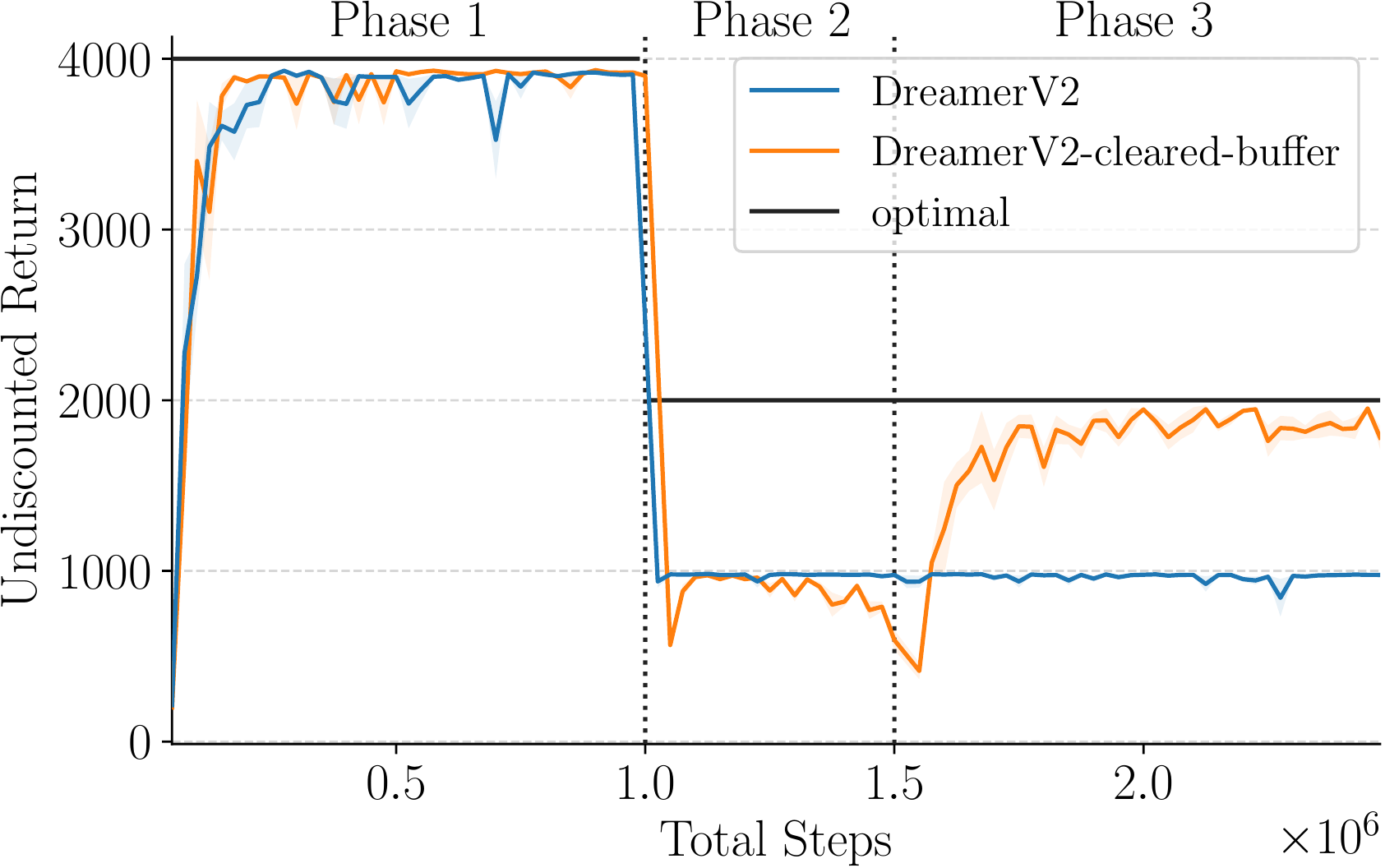}
    \caption{DreamerV2}
    \end{subfigure}%
    \begin{subfigure}[h]{0.33\textwidth}
        \centering
        \includegraphics[width=0.95\textwidth]{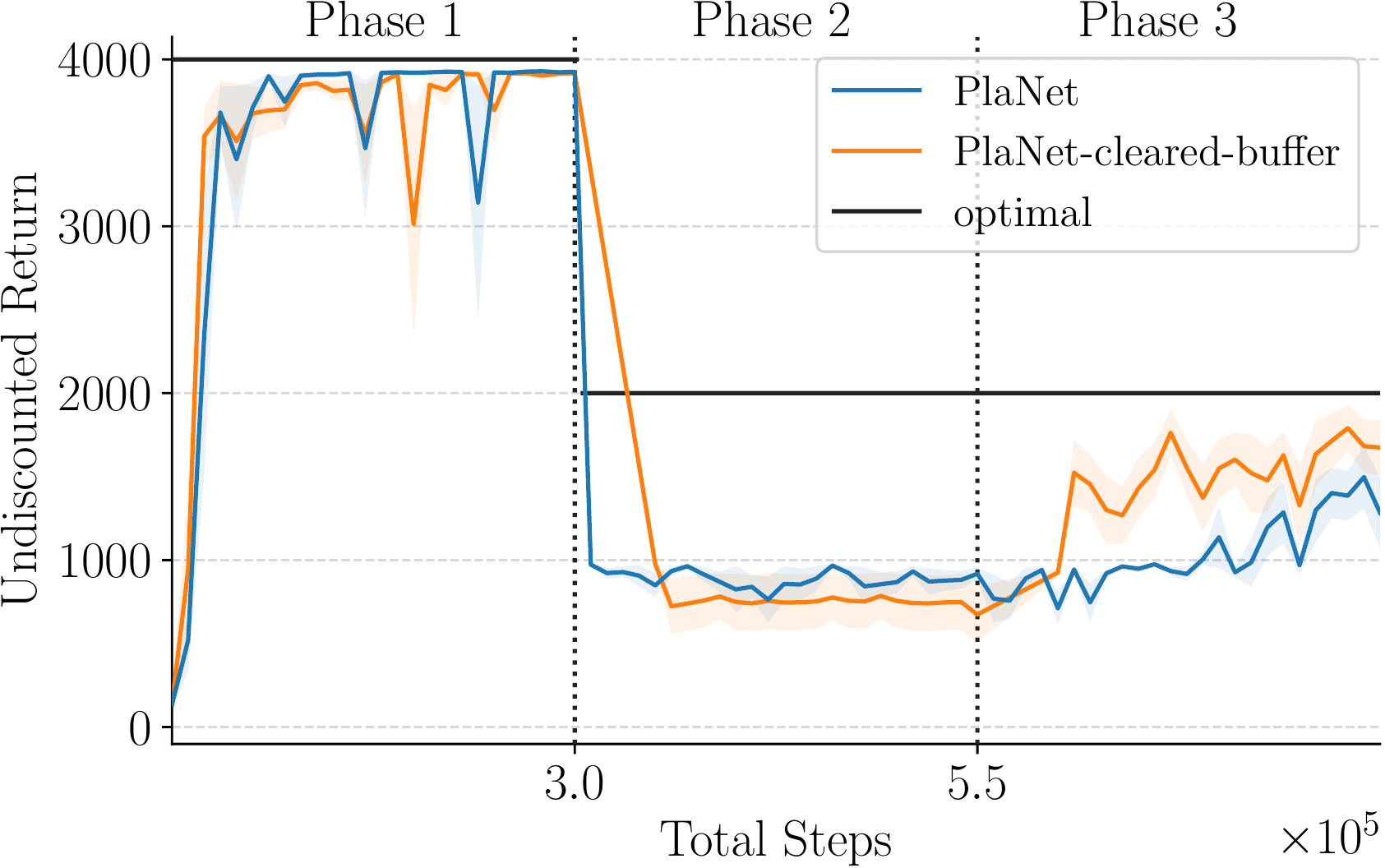}
        \caption{PlaNet}
    \end{subfigure}%
    \caption{a) Illustration of the ReacherLoCA domain. The dashed white circle shows the T1-zone. b) and c) Plots showing the learning curves of DreamerV2 and PlaNet. We show the maximum achievable return at each phase as a baseline. For the setting in which the replay buffer is cleared at the start of each phase, we reinitialized it with 50 random episodes (50000 steps). 
    }
    \label{fig: deep rl results}
\end{figure*}

In our modified domain, ReacherLoCA, we added a second target and fixed both target locations in opposite corners of the domain (Figure \ref{fig: deep rl results}). Furthermore, we created a one-way passage around one of the targets. Staying true to the original Reacher domain, episodes are terminated after exactly 1000 time steps. While this means the target locations are strictly speaking not terminal states, we can apply LoCA to this domain just the same, interpreting the target location with the one-way passage surrounding it as T1 and the other one as T2. Rewards $r_A$ and $r_B$ are applied accordingly. The advantage of staying as close as possible to the original Reacher environment (including episode termination after 1000 time steps) is that we can copy many of the hyperparameters used for PlaNet and DreamerV2 applied to the original Reacher environment and only minor fine-tuning is required.

\begin{figure*}[t]
\centering
\begin{subfigure}[b]{.2\textwidth}
    \centering
    \includegraphics[width=0.99\textwidth]{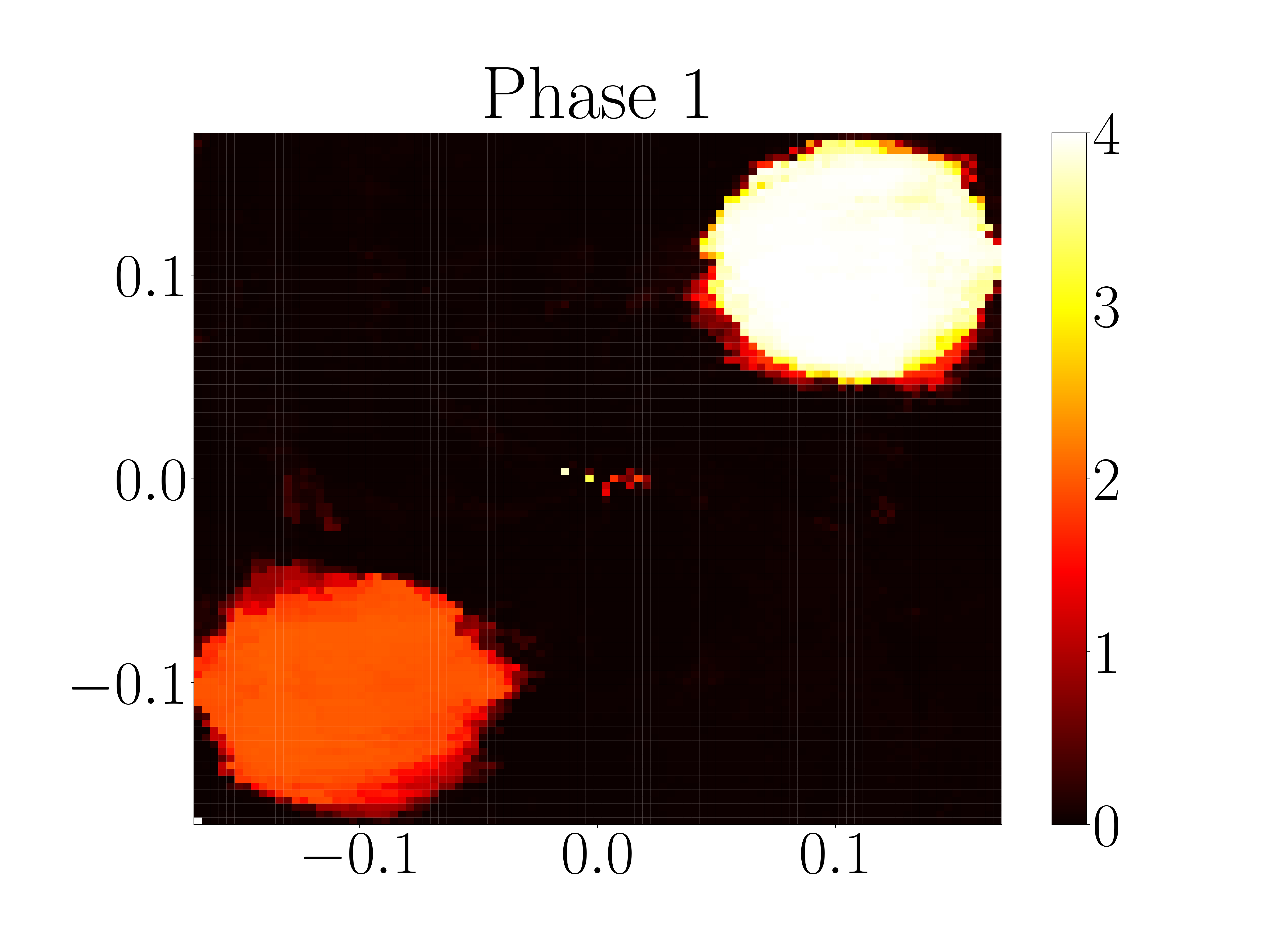}
\end{subfigure}%
\begin{subfigure}[b]{0.2\textwidth}
        \centering
        \includegraphics[width=0.99\textwidth]{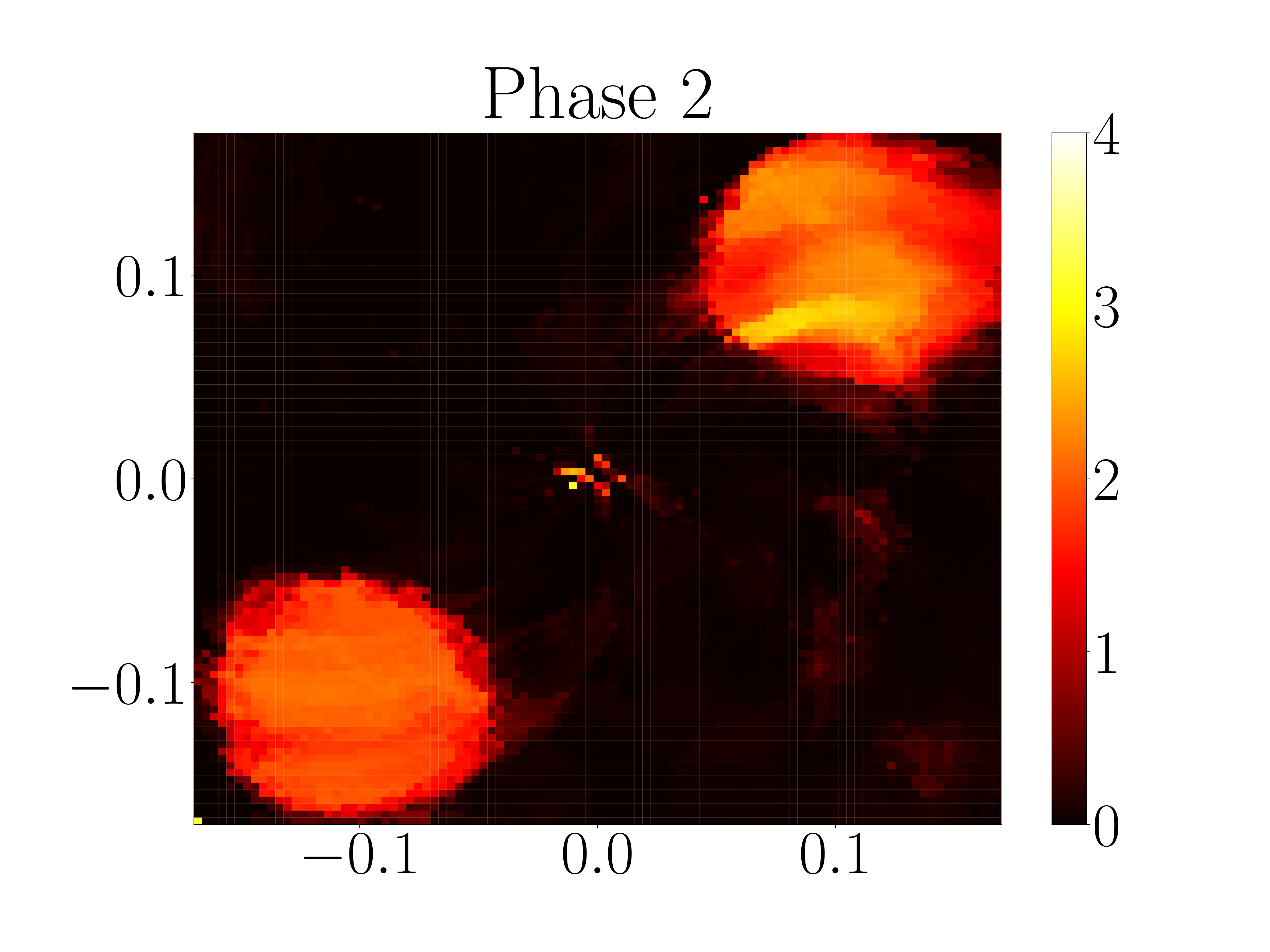}
    \end{subfigure}%
    \begin{subfigure}[b]{0.2\textwidth}
        \centering
        \includegraphics[width=0.99\textwidth]{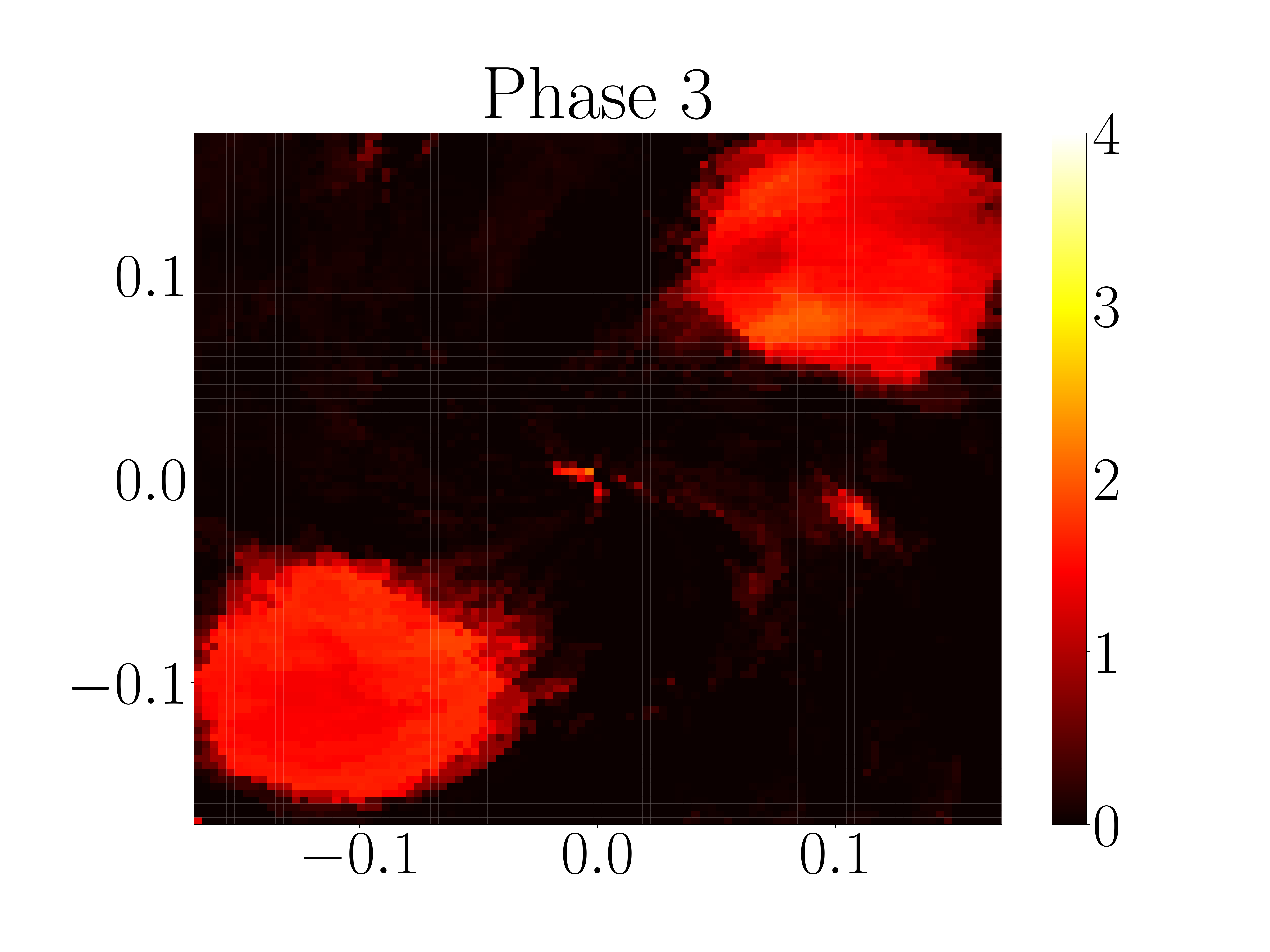}
    \end{subfigure}%
    \begin{subfigure}[b]{.2\textwidth}
        \centering
        \includegraphics[width=0.99\textwidth]{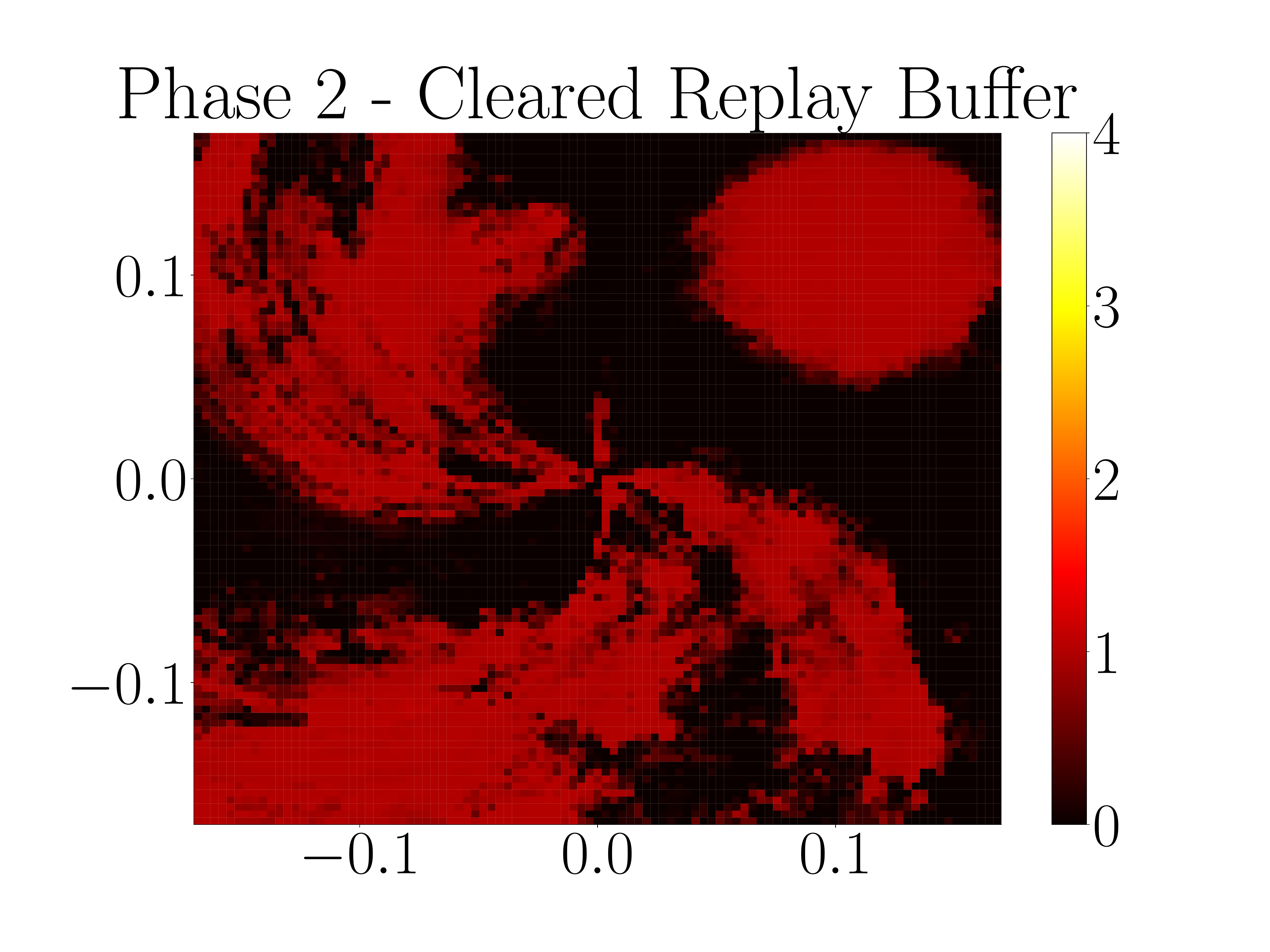}
    \end{subfigure}%
    \begin{subfigure}[b]{0.2\textwidth}
        \centering
        \includegraphics[width=0.99\textwidth]{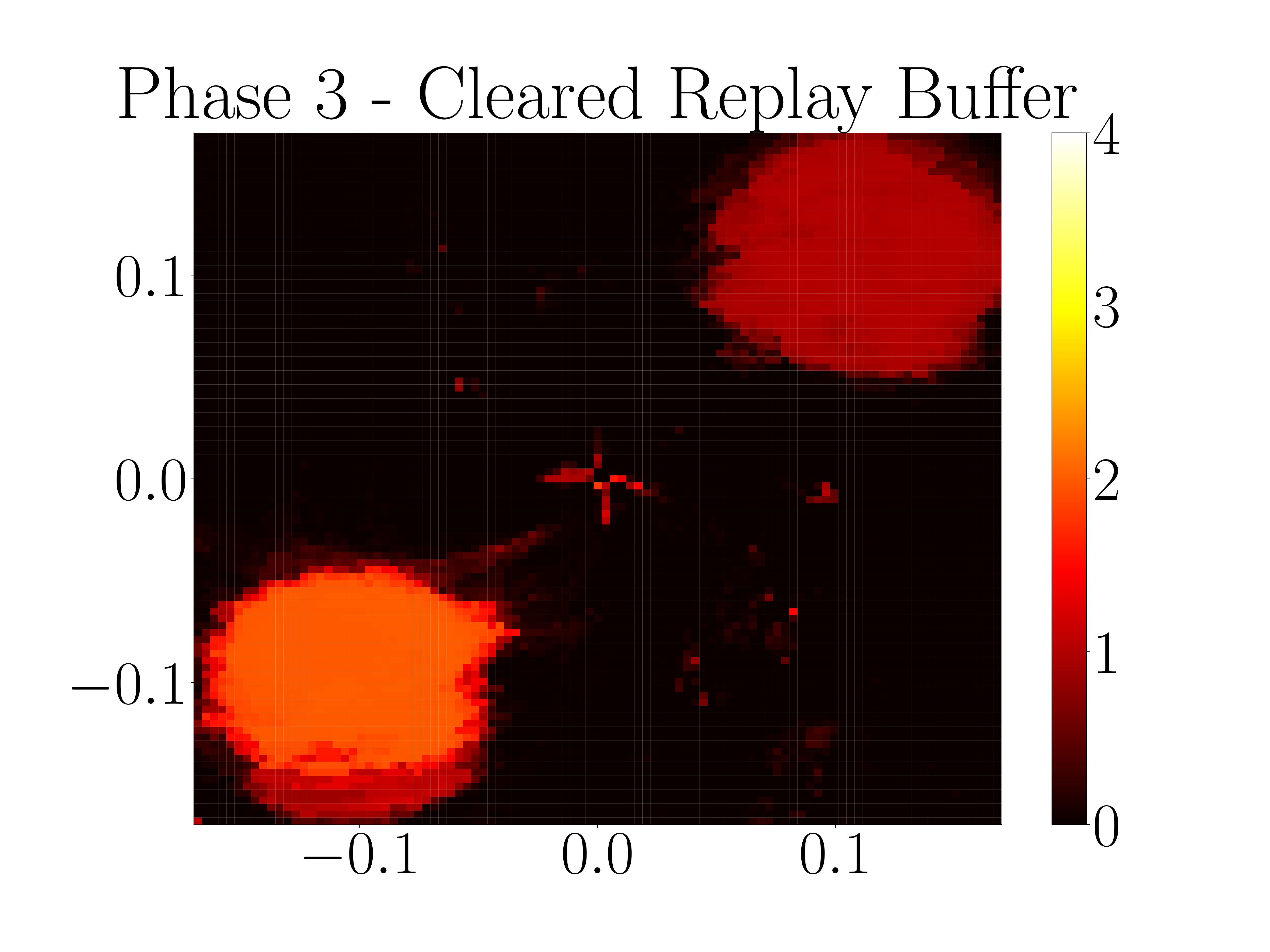}
    \end{subfigure}%
\caption{Visualization of the DreamerV2 agent's estimated reward model generated at the end of each phase. The $x$ and $y$ axes of each heatmap represent the agent's position in the ReacherLoCA domain. 
}
\label{fig: dreamer reward estimates}
\end{figure*}

\subsection{PlaNet and DreamerV2 Experiments}

Note that both PlaNet and DreamerV2 learn from transitions sampled from a large experience replay buffer (they call it experience dataset) that contains all recently visited transitions. When evaluating these algorithms in the ReacherLoCA domain, such a strategy could hurt the agent's ability of adaptation, because in early stage of Phase 2, most of the data that the agent learns from are still from Phase 1 and it takes a long time before all of the task A data in the buffer get removed, if the replay buffer is large.
Therefore a natural question to ask is, in practice, whether the agent can adapt well, with some stale data in the replay buffer. The other interesting question to ask would be, if the agent somehow knows the change of the task and reinitializes its replay buffer when it observes such a change, does the agent perform well in Phase 2?

To answer the above two questions, we tested two variants of each of these two algorithms. For both variants, we used a sufficiently large replay buffer that could contain all the transitions produced during training. For one variant, we reinitialized the replay buffer between two phases by first clearing the replay buffer and then filling replay buffer with certain number of episodes generated by following a random policy. This variant of algorithm requires to know when the environment changes and could therefore take advantage of this prior knowledge to remove 
outdated data. For the other variant, no modification to the replay buffer was applied between two phases. For both algorithms, we did a grid search only over the critical hyperparameters suggested by the corresponding papers. Unless specified otherwise, other hyperparameters were chosen to be the same as those found to be the best by the corresponding papers in the original Reacher environment. Details of the experiment setup are summarized in Sections\ \ref{app: Experiment Setup of the DreamerV2 Experiment} and \ref{app: Experiment Setup of the PlaNet Experiment}. Complete empirical results are presented in Sections \ref{app: Additional DreamerV2 Results} and \ref{app: Additional PlaNet Results}.

For each variant of each of the two algorithms, we drew a learning curve (Figure\ \ref{fig: deep rl results}) corresponding to the best hyperparameter setting found by a grid search (see Sections \ref{app: Additional Information about the DreamerV2 Result Shown in the Middle Panel of Figure DeepRL Results} and \ref{app: Additional Information about the PlaNet Result Shown in the Right Panel of Figure DeepRL Results} for details).
The reported learning curves suggest that neither DreamerV2 nor PlaNet effectively adapted their policy in Phase 2, regardless of the replay buffer being reinitialized or not. Further, when the replay buffer was not reinitialized, both DreamerV2 and PlaNet performed poorly in Phase 3. To obtain a better understanding of the failure of adaptation, we plotted in Figure \ref{fig: dreamer reward estimates} the reward predictions of DreamerV2's world model at the end of each phase. The corresponding results for PlaNet is similar and are presented in Section \ref{app: Additional PlaNet Results}. 
Our empirical results provide affirmative answers to the two questions raised at the beginning of this subsection. 

The first question is addressed by analyzing the predicted rewards without reinitializing the replay buffer. In this case, the predicted reward of T1 overestimates the actual reward (1) in both Phases 2 and 3.
Such an overestimation shows that learning from stale data in the replay buffer apparently hurts model learning. In addition, the corresponding learning curves in Phases 2 and 3 show that the resulting inferior model could be detrimental for planning. The second question is addressed by analyzing the predicted rewards with reinitializing the replay buffer. In this case, the values for T1 at the end of Phase 2 are correct now, but the estimates for the rest of the state space are completely incorrect. This result answers our second question -- with replay buffer reinitialization, the agent had no data outside of the T1-zone to rehearse in Phase 2 and forgot the learned reward model for states outside the T1-zone. Such an issue is called \emph{catastrophic forgetting}, which was originally coined by \citet{mccloskey1989catastrophic}. 
Overall, we conclude that DreamerV2 achieved poor adaptivity to local changes due to two additional failure modes that are different from those outlined in Section \ref{sec: tabular}.

\emph{\textbf{Failure Mode \#3}: Learning from large replay buffers cause interference from the old task.}

\emph{\textbf{Failure Mode \#4}: If the environment model is represented by neural networks, learning from small replay buffers cause model predictions for areas of the state space not recently visited to be off due to catastrophic forgetting.}

\textbf{Remark:} Note that there is a dilemma between the above two failure modes. Also, note that as long as one uses a large replay buffer in a non-stationary environment, Failure Mode \#3 is inevitable. This suggests that, when using neural networks, solving the catastrophic forgetting problem (Failure Mode \#4) is an indispensable need for the LoCA setup and the more ambitious continual learning problem.
Over the past 30 years, significant progress was made towards understanding and solving the catastrophic forgetting problem \citep{french1991using,robins1995catastrophic,french1999catastrophic,goodfellow2013empirical,kirkpatrick2017overcoming,kemker2018measuring}. 
However, a satisfying solution to the problem is still not found and the problem itself is still actively studied currently.

\section{Adaptive MBRL Algorithm with Function Approximation}\label{sec: Dyna}

\begin{figure*}[h]
\centering
\begin{subfigure}{0.33\textwidth}
    \centering
    \includegraphics[width=0.95\textwidth]{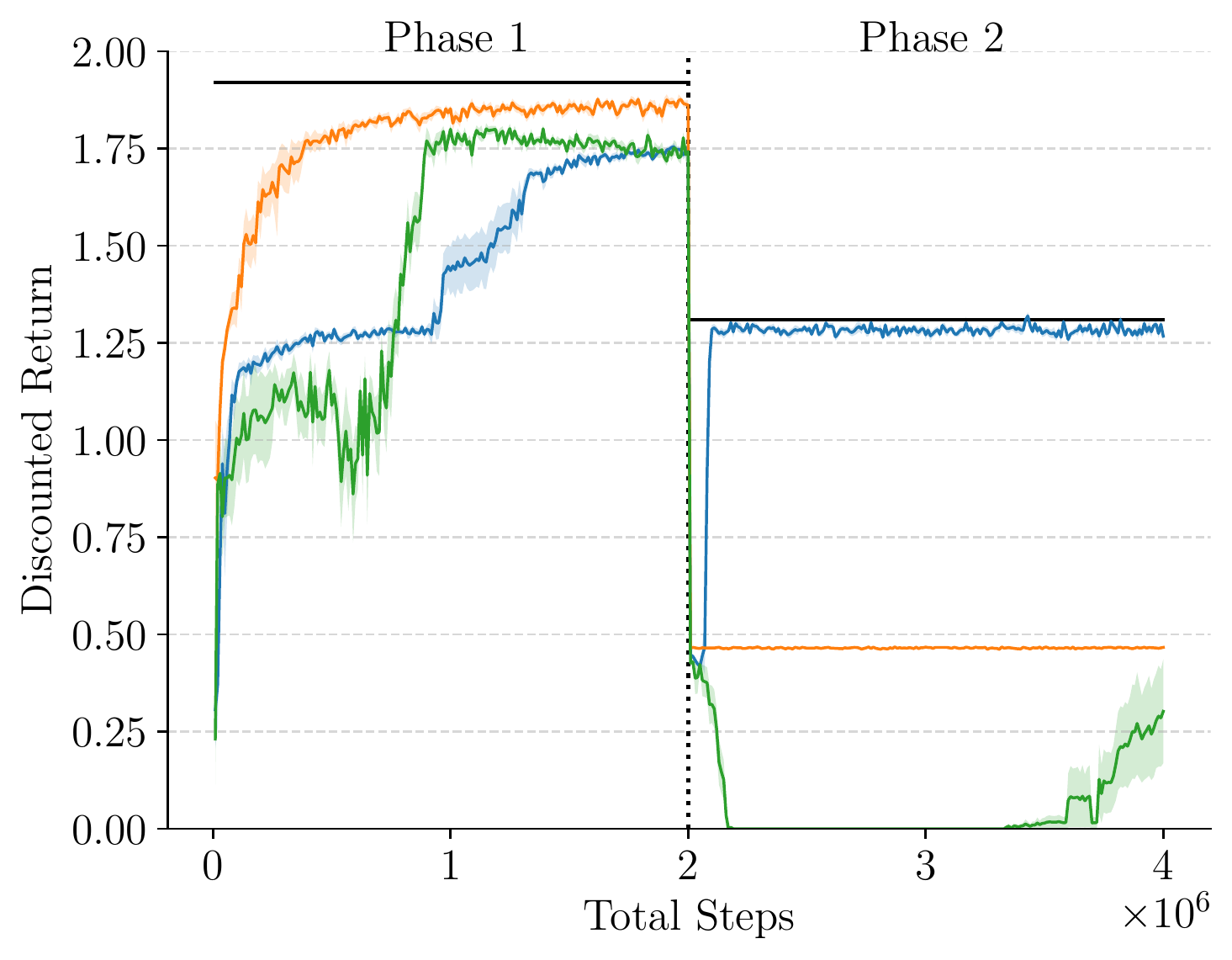}
    \caption{Stochasticity = 0.0}
\end{subfigure}%
\begin{subfigure}{0.33\textwidth}
    \centering
    \includegraphics[width=0.95\textwidth]{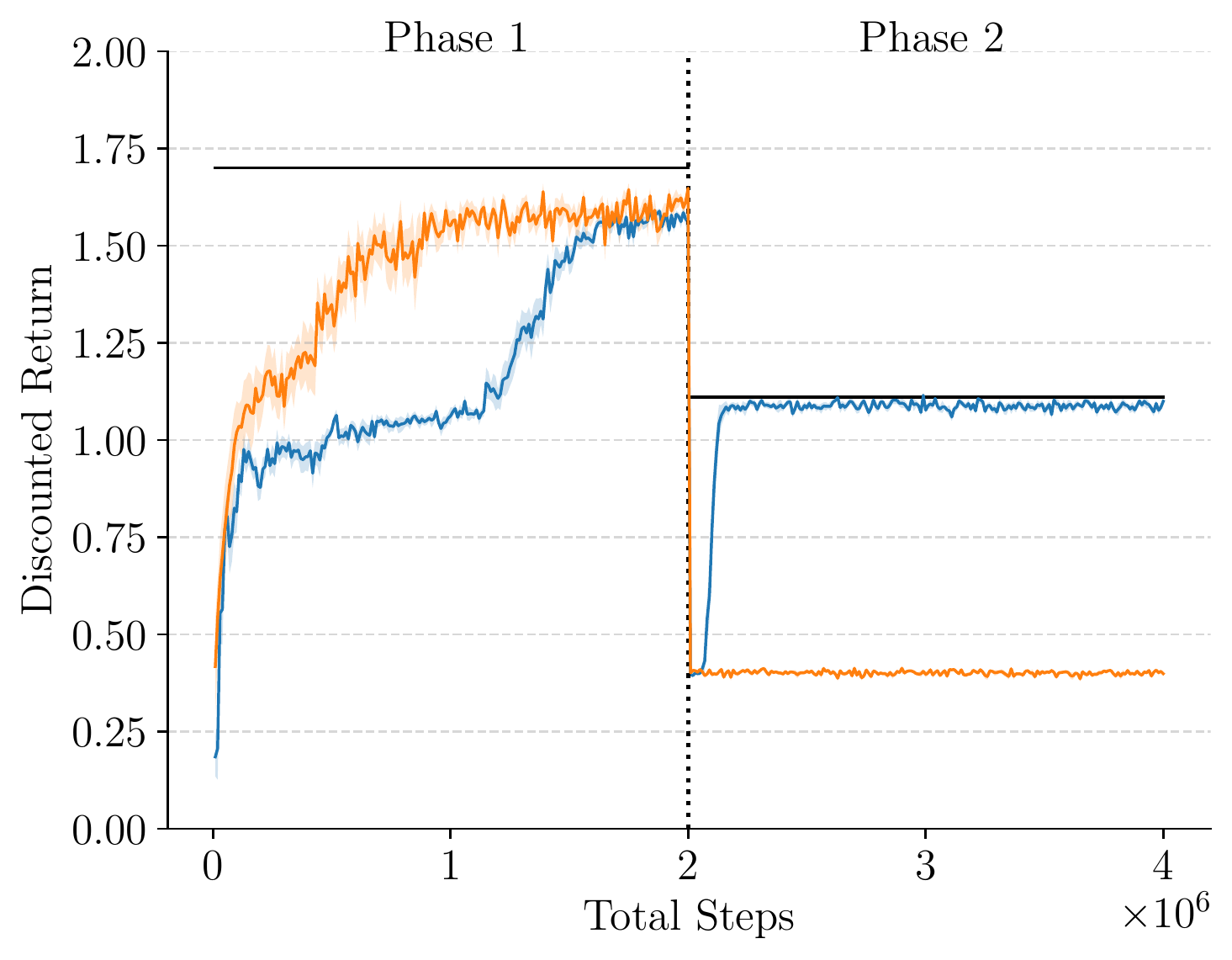}
    \caption{Stochasticity = 0.3}
\end{subfigure}%
\begin{subfigure}{0.33\textwidth}
    \centering
    \includegraphics[width=0.95\textwidth]{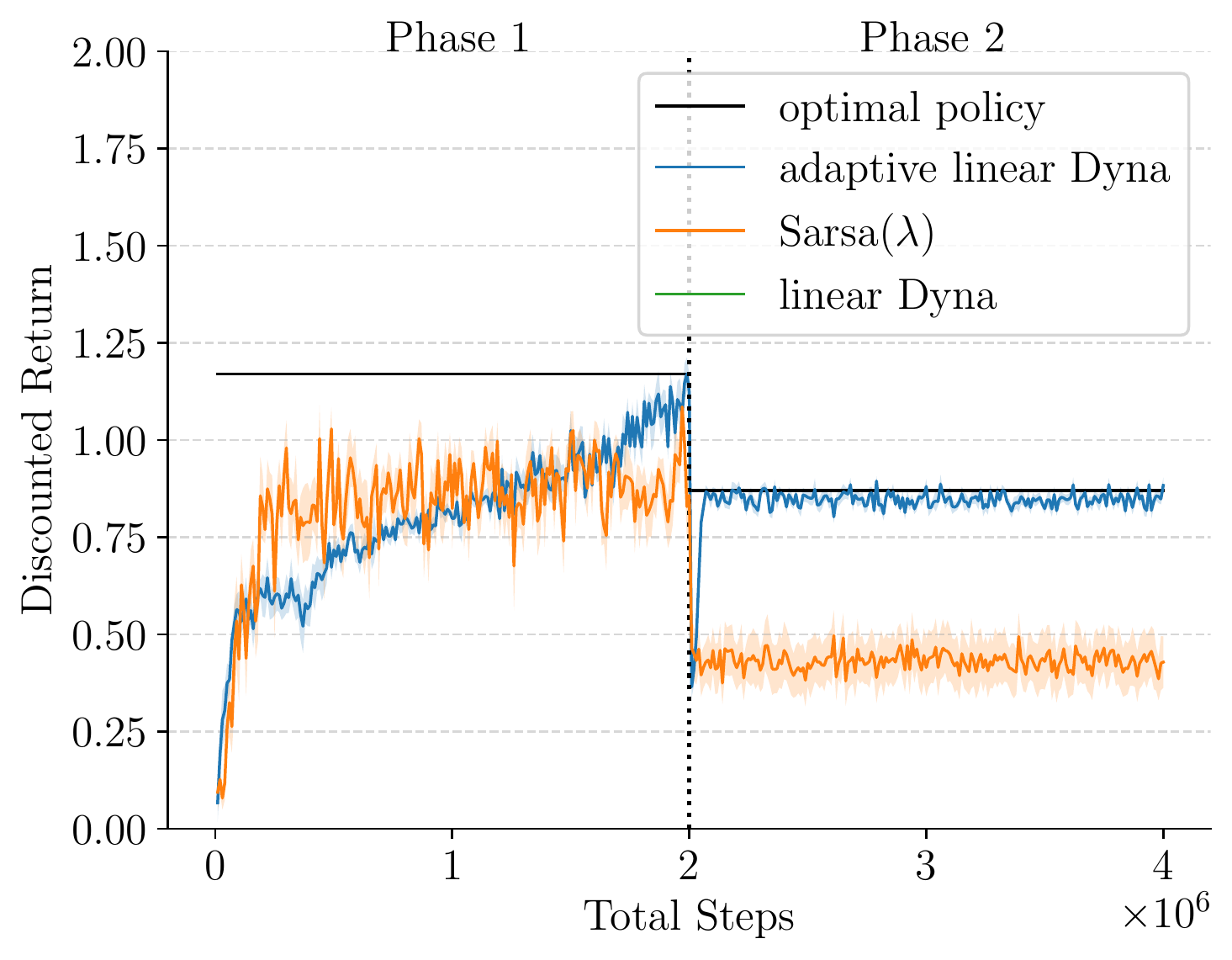}
    \caption{Stochasticity = 0.5}
\end{subfigure}%
\caption{Plots showing that adaptive linear Dyna is adaptive while Sarsa($\lambda$) and linear Dyna are not in the MountaincarLoCA domain. The $x$ axis represents the number of training steps. Each point in a learning curve is the average discounted return obtained by following the agent's greedy policy for 10 runs. The first phase ends and the second phase starts at $2e6$ time step. We tested a broad range of hyperparameters for both algorithms. In each sub-figure, as a baseline, we plotted the best discounted return achieved by Sarsa($\lambda$), after training for sufficiently long in Task A and B, with initial states being sampled from the entire state space (optimal policy).}
\label{fig: linear dyna}
\end{figure*}

In this section, we identify an algorithm that does not fall into the four aforementioned failure modes and understand its behavior using the LoCA setup in a variant of the MountainCar domain \citep{moore1990efficient}. This algorithm, called \emph{adaptive linear Dyna}, is a modified version of the linear Dyna algorithm (Algorithm 4 by \citet{sutton2012dyna}). To overcome the unsolved catastrophic forgetting problem with neural networks, this algorithm takes a step back by using linear function approximation with sparse feature vectors. We provide empirical evidence showing that this algorithm is adaptive in the MountainCarLoCA domain. Additional empirical results show that a nonlinear (neural networks) extension of the linear algorithm did not adapt equally well in the same setup, due to the inferior learned model resulting from catastrophic forgetting or interference from the old task. 

\citet{sutton2012dyna}'s linear Dyna algorithm applies value iteration with a learned linear expectation model, which predicts the expected next state, to update a linear state-value function. 
Using expectation models in this way is sound because value iteration with an expectation model is equivalent to it with an aligned distribution model when using a linear state-value function \citep{wan2019planning}. 
The algorithm does not fall into Failure Mode \#2 because the expectation model is policy-independent.
The algorithm does not fall into the Failure Modes \#3 and \#4 because the model is learned online and the algorithm uses linear function approximation with sparse tile-coded \citep{sutton2018reinforcement} feature vectors. Limiting feature sharing alleviates the catastrophic forgetting problem because learning for one input influences predicting for only few other inputs. 

\citet{sutton2012dyna}'s linear Dyna algorithm does fall into something similar to Failure Mode \#1 because planning is not applied to any of the real feature vectors.
Specifically, planning uses tabular feature vectors (one bit \emph{on} in each vector), while real feature vectors (feature vectors corresponding to real states) are binary vectors with multiple bits on. It is thus unclear if planning with these unreal feature vectors could produce a good policy.
In Figure\ \ref{fig: linear dyna}, we empirically show that original linear Dyna leads to failure of adaptation in Phase 2. 

A natural modification of the algorithm to improve its adaptivity, is to change the way of generating feature vectors for planning. We choose to randomly sample these feature vectors from a buffer containing feature vectors corresponding to recently visited states. We call this buffer the \emph{planning} buffer because feature vectors stored in the buffer are used for planning. While the modification itself is small and simple, the effect of the modification is significant -- the modified algorithm can almost achieve the optimal value in Phase 2. The pseudo code of the algorithm is shown in Algorithm \ref{algo:adaptive linear dyna}. Details of empirical analysis are presented in Section\ \ref{sec: Linear Dyna Experiments}. 

Based on the adaptive linear Dyna algorithm, we propose the nonlinear Dyna Q algorithm (Algorithm \ref{algo:Non-Linear Dyna Q}), in which the value function and the model are both approximated using neural networks. 
Instead of trying to solve the catastrophic interference problem, we adopt the simple replay approach. Specifically, we maintain a \emph{learning} buffer, in addition to the planning buffer, to store recently visited transitions and sample from the learning buffer to generate data for model learning. As discussed previously, this approach falls into a dilemma of learning from stale information (Failure Mode \#3) or forgetting previously learned information (Failure Mode \#4). 
To empirically verify if the dilemma is critical, we varied the size of the learning buffer as well as other hyperparameters to see if there is one parameter setting that supports adaptation. Further, we also tried two strategies of sampling from the learning buffer: 1) randomly sampling from the entire buffer, and 2) sampling half of the data randomly from the entire buffer and the rest randomly only from rewarding data (transitions leading to rewards). Our empirical results in the MountainCarLoCA domain presented in Section\ \ref{sec: Non-linear Dyna Experiments} confirms that this dilemma is critical -- the learned model is inferior, even with the best hyperparameter setting and sampling strategy. 

\subsection{Adaptive Linear Dyna Experiment}\label{sec: Linear Dyna Experiments}

The first experiment was designed to test if the adaptive linear Dyna algorithm can adapt well with proper choices of hyperparameters. We tested our algorithm on a stochastic variant of the MountainCar domain (Section \ref{app: Setups for the Adaptive Linear Dyna Experiment}). To this end, for each level of stochasticity, we did a grid search over the algorithm's hyperparameters and showed the learning curve corresponding to the best hyperparameter setting in Figure~\ref{fig: linear dyna}. The figure also shows, for different levels of stochasticity, learning curves of Sarsa($\lambda$) with the best hyperparameter setting as baselines. The best hyperparameter setting is the one that performs the best in Phase 2, among those that perform well in Phase 1 (See Section \ref{app: Additional Information of Result Shown in Figure Linear Dyna} for details). Further, we show a learning curve of linear Dyna by \citet{sutton2012dyna}. A hyperparameter study of these algorithm is presented in Section \ref{app: Additional Results of the Adaptive Linear Dyna Experiment}.

The learning curves show that for all different levels of stochasticity, adaptive linear Dyna performed well in Phase 1 and adapted quickly and achieved near-optimal return in Phase 2. The learning curve for Sarsa($\lambda$) coincides with our expectation. It performed well in Phase 1, but failed to adapt its policy in Phase 2. Linear Dyna also performed well in Phase 1, but failed to adapt in Phase 2, which is somewhat surprising because the unreal tabular feature vectors were used in both phases. A deeper look at our experiment data (Figure\ \ref{fig: linear dyna plots} and \ref{fig: linear dyna plots tabular}) shows that in Phase 1, although the policy is not apparently inferior, the estimated values are inferior. The estimated values are even worse in Phase 2. We hypothesize that such a discrepancy between Phases 1 and 2 is due to the fact that the model-free learning part of the linear Dyna algorithm helps obtain a relatively accurate value estimation for states over the entire state space in Phase 1, but only for states inside the T1-zone in Phase 2.

The other observation is that when stochasticity = 0.5, Sarsa($\lambda$) is worse than adaptive linear Dyna. Note that there is a very high variance in the learning curve of Sarsa($\lambda$). On the contrary, adaptive linear Dyna achieved a much lower variance, potentially due to its planning with a learned model, which induces some bias and reduces variance. Comparing the two algorithms shows that when the domain is highly stochastic, the variance can be an important factor influencing the performance and planning with a learned model can reduce the variance.

\subsection{Nonlinear Dyna Q Experiment}\label{sec: Non-linear Dyna Experiments}
\begin{figure}[t]
\centering
\begin{subfigure}{.24\textwidth}
    \centering
    \includegraphics[width=0.95\textwidth]{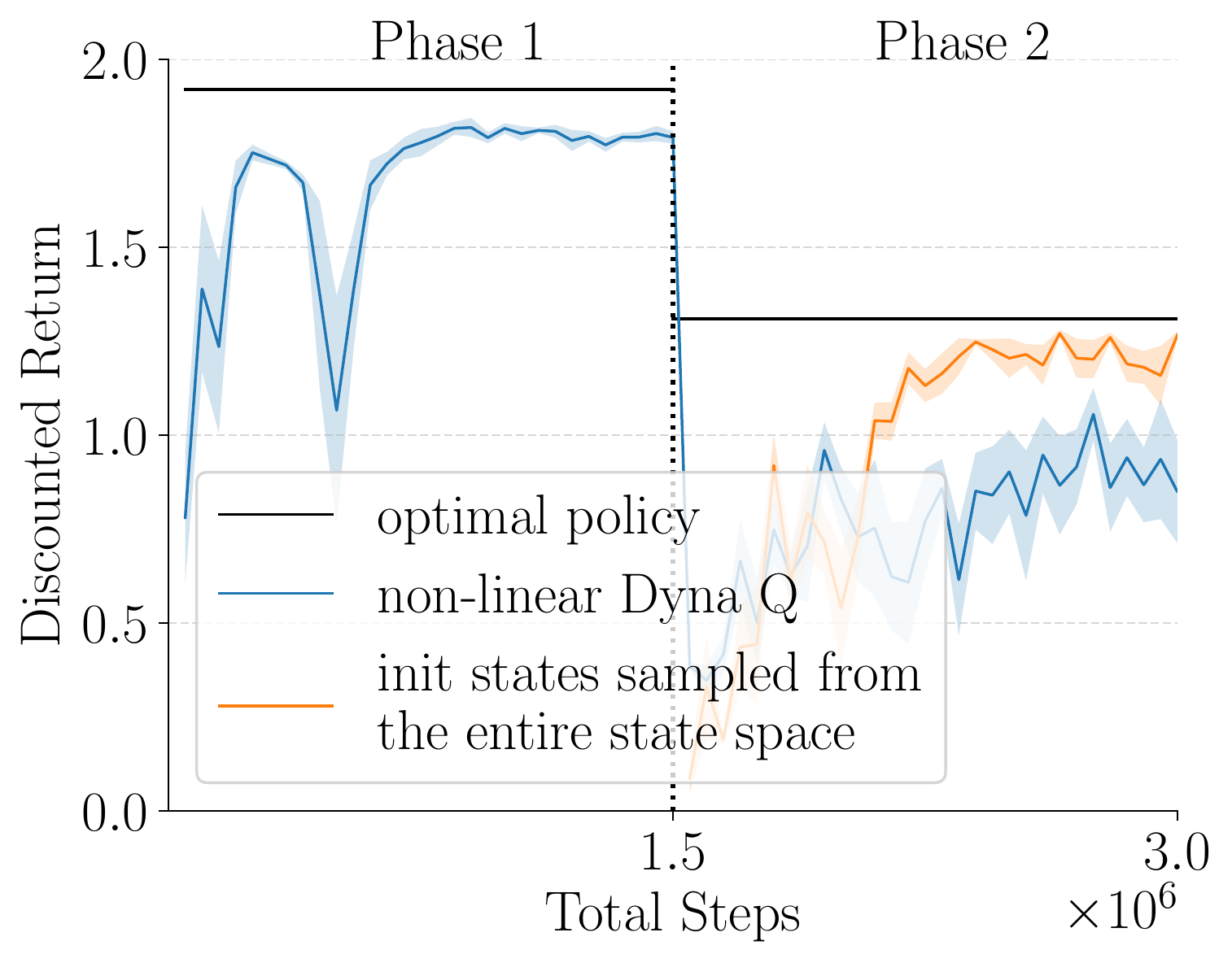}
    \vspace{-0.5mm}
    \caption{Learning Curves}
\end{subfigure}%
\begin{subfigure}{.25\textwidth}
    \centering
    \includegraphics[width=0.95\textwidth]{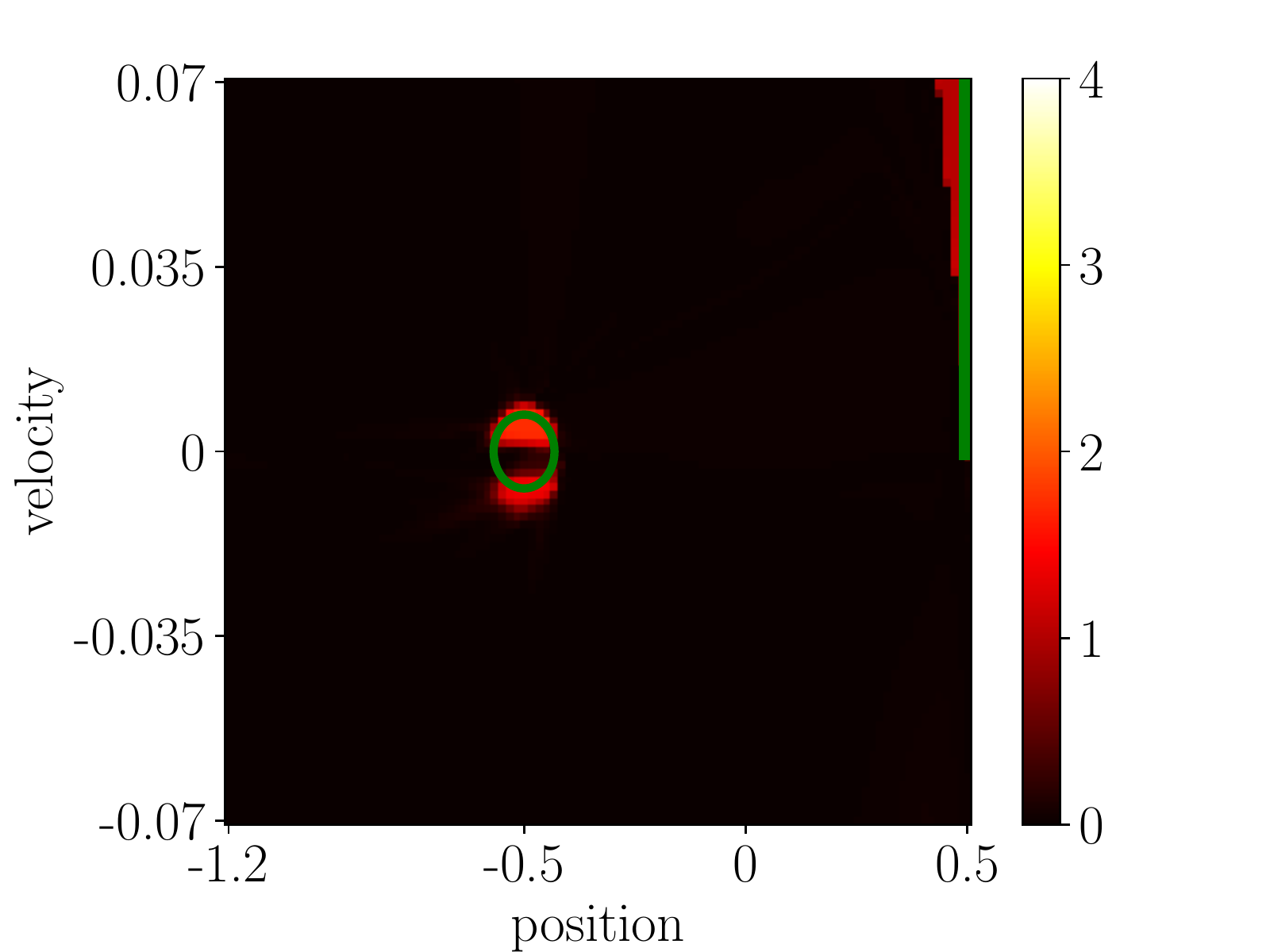}
    \vspace{-0.6mm}
    \vspace{0.15mm}
    \caption{Reward Model}
\end{subfigure}%
\caption{Plots showing learning curves and the estimated reward model of nonlinear Dyna Q at the end of training in the deterministic MountainCarLoCA domain. a) nonlinear Dyna Q struggles to adapt in Phase 2 (blue curve). As a baseline, we show the learning curve produced by applying nonlinear Dyna Q to Task B, with initial states sampled from the entire state space. This shows the best performance this algorithm can achieve for Task B. b) The $x$ and $y$ axes represent the position and the velocity of the car and thus each point represents a state. The green eclipse indicates the T1. The color of each point represents the model's reward prediction of a state (maximized over different actions). 
}
\label{fig: nonlineardyna}
\end{figure}

The second experiment was designed to test if the nonlinear Dyna Q algorithm can successfully adapt in the deterministic MountainCarLoCA domain. 
Tested hyperparameters are specified in Section~\ref{app: Setups for the Nonlinear Dyna Q Experiment}. The hyperparameter setting used to generate the reported learning curve (Figure\ \ref{fig: nonlineardyna}(a)) chosen in a similar way as those chosen for the adaptive linear Dyna experiment.

The learning curve corresponding to the best hyperparameter setting is the one using a large learning buffer, and the sampling strategy that emphasizes rewarding transitions. Nevertheless, even the best hyperparameter setting only produced an inferior policy, as illustrated in the sub-figure (a) of Figure\ \ref{fig: nonlineardyna}. We picked one run with the best hyperparameter setting and plotted the estimated reward model at the end of Phase 2 in Figure\ \ref{fig: nonlineardyna}. Comparing the predicted rewards by the learned model with the true rewarding region marked by the green circle shows catastrophic forgetting severely influences model learning. In fact, we observed that if the model in Phase 2 is trained even longer, eventually T2's reward model will be completely forgotten.

\section{Related Work}
The LoCA setup is designed to measure the adaptivity of an algorithm and serves as a preliminary yet important step towards the ambitious \emph{continual learning} problem \citep{khetarpal2020towards,kessler2021same}. 
In the continual learning problem, the agent interacts with an \emph{non-stationary} environment. The LoCA setup specifies a particular kind of continual learning problem, which involves fully observable environment and a local environment change. 

The LoCA setup is also closely related to the well-known \emph{transfer learning} problem \citep{taylor2009transfer, lazaric2012transfer,zhu2020transfer} because they both need the agent to solve for multiple tasks. Nevertheless, the LoCA setup should not be viewed as a special case of transfer learning. In transfer learning, the agent is informed of which task it is solving, while in the LoCA setup it is not. 

There are several ways in which an algorithm may potentially adapt quickly when the environment changes. For example, the learned feature representation \citep{konidaris2012transfer, barreto2016successor}, the discovered options \citep{sutton1999between,barto2003recent,bacon2018temporal}, or some general meta-knowledge \citep{finn2017model,huisman2021survey} obtained from the old environment may be useful for learning and planning in the new environment. The LoCA setup focuses on evaluating if the agent's fast adaptation relies on its planning with a learned model. 

On the other hand, there are also some existing works developing MBRL algorithms for the continual learning problem or the transfer learning problem \citep{huang2020continual,boloka2021knowledge,zhang2019solar,nguyen2012transferring,lu2020reset}. Although these algorithms have shown great performance in various experiments, none of these algorithms directly resolve the fundamental catastrophic forgetting issue and therefore, generally speaking, they are not likely to demonstrate adaptivity in the LoCA setup.

\section{Discussion and Conclusion}
We introduced an improved version of the LoCA setup, which is simpler, less
sensitive to its hyperparameters and can be more easily applied to stochastic environments. We then studied the adaptivity of two deep MBRL methods using this methodology. Our empirical results, combined with those from  \citet{vanseijen-LoCA}, suggest that several popular modern deep MBRL methods adapt poorly to local changes in the environment.
This is surprising as the adaptivity should be one of the major strengths of MBRL (in behavioral neuroscience, adaptivity to local changes is one of the characteristic features that differentiates model-based from model-free behavior, e.g., see \citet{daw2011model}).

Besides that, we studied the challenges involved with building adaptive model-based methods and identified four important failure modes. These four failure modes were then linked to three modern MBRL algorithms, justifying why they didn't demonstrate adaptivity in experiments using LoCA. The first three of these failure modes can be overcome by using appropriate environment models, planning techniques, and smaller buffer. The fourth failure mode is 
tied to catastrophic forgetting, which is a challenging open problem. 
The most common mitigation technique to the problem, using a large experience replay buffer, is not an option in our case, as it inevitably results in Failure Mode \#3.
Hence, we conclude that the path towards adaptive deep MBRL involves tackling the challenging catastrophic forgetting problem.

\section*{Acknowledgments}

The authors wish to thank Richard Sutton, Arsalan Sharifnassab, and Hadi Nekoei for their valuable feedback during various stages of the work. In addition, we are grateful to Hadi Nekoei and Maryam Hashemzadeh
for taking the time to review our final codebase. We would like to acknowledge Compute Canada and Calcul Quebec for providing computing resources used in this work. YW is supported by a MSR-Mila grant and Amii. SC is supported by a Canada CIFAR AI Chair and an NSERC Discovery Grant.

\newpage
\bibliography{refs}

\begin{thebibliography}{39}
\providecommand{\natexlab}[1]{#1}
\providecommand{\url}[1]{\texttt{#1}}
\expandafter\ifx\csname urlstyle\endcsname\relax
  \providecommand{\doi}[1]{doi: #1}\else
  \providecommand{\doi}{doi: \begingroup \urlstyle{rm}\Url}\fi

\bibitem[Bacon(2018)]{bacon2018temporal}
Bacon, P.-L.
\newblock \emph{Temporal Representation Learning}.
\newblock McGill University (Canada), 2018.

\bibitem[Barreto et~al.(2016)Barreto, Dabney, Munos, Hunt, Schaul, Van~Hasselt,
  and Silver]{barreto2016successor}
Barreto, A., Dabney, W., Munos, R., Hunt, J.~J., Schaul, T., Van~Hasselt, H.,
  and Silver, D.
\newblock Successor features for transfer in reinforcement learning.
\newblock \emph{arXiv preprint arXiv:1606.05312}, 2016.

\bibitem[Barto \& Mahadevan(2003)Barto and Mahadevan]{barto2003recent}
Barto, A.~G. and Mahadevan, S.
\newblock Recent advances in hierarchical reinforcement learning.
\newblock \emph{Discrete event dynamic systems}, 13\penalty0 (1):\penalty0
  41--77, 2003.

\bibitem[Boloka et~al.(2021)Boloka, Makondo, and Rosman]{boloka2021knowledge}
Boloka, T., Makondo, N., and Rosman, B.
\newblock Knowledge transfer using model-based deep reinforcement learning.
\newblock In \emph{2021 Southern African Universities Power Engineering
  Conference/Robotics and Mechatronics/Pattern Recognition Association of South
  Africa (SAUPEC/RobMech/PRASA)}, pp.\  1--6. IEEE, 2021.

\bibitem[Daw et~al.(2011)Daw, Gershman, Seymour, Dayan, and
  Dolan]{daw2011model}
Daw, N.~D., Gershman, S.~J., Seymour, B., Dayan, P., and Dolan, R.~J.
\newblock Model-based influences on humans' choices and striatal prediction
  errors.
\newblock \emph{Neuron}, 69\penalty0 (6):\penalty0 1204--1215, 2011.

\bibitem[Finn et~al.(2017)Finn, Abbeel, and Levine]{finn2017model}
Finn, C., Abbeel, P., and Levine, S.
\newblock Model-agnostic meta-learning for fast adaptation of deep networks.
\newblock In \emph{International Conference on Machine Learning}, pp.\
  1126--1135. PMLR, 2017.

\bibitem[French(1991)]{french1991using}
French, R.~M.
\newblock Using semi-distributed representations to overcome catastrophic
  forgetting in connectionist networks.
\newblock In \emph{Proceedings of the 13th annual cognitive science society
  conference}, volume~1, pp.\  173--178, 1991.

\bibitem[French(1999)]{french1999catastrophic}
French, R.~M.
\newblock Catastrophic forgetting in connectionist networks.
\newblock \emph{Trends in cognitive sciences}, 3\penalty0 (4):\penalty0
  128--135, 1999.

\bibitem[Goodfellow et~al.(2013)Goodfellow, Mirza, Xiao, Courville, and
  Bengio]{goodfellow2013empirical}
Goodfellow, I.~J., Mirza, M., Xiao, D., Courville, A., and Bengio, Y.
\newblock An empirical investigation of catastrophic forgetting in
  gradient-based neural networks.
\newblock \emph{arXiv preprint arXiv:1312.6211}, 2013.

\bibitem[Hafner et~al.(2019{\natexlab{a}})Hafner, Lillicrap, Ba, and
  Norouzi]{hafner2019dream}
Hafner, D., Lillicrap, T., Ba, J., and Norouzi, M.
\newblock Dream to control: Learning behaviors by latent imagination.
\newblock \emph{arXiv preprint arXiv:1912.01603}, 2019{\natexlab{a}}.

\bibitem[Hafner et~al.(2019{\natexlab{b}})Hafner, Lillicrap, Fischer, Villegas,
  Ha, Lee, and Davidson]{hafner2019learning}
Hafner, D., Lillicrap, T., Fischer, I., Villegas, R., Ha, D., Lee, H., and
  Davidson, J.
\newblock Learning latent dynamics for planning from pixels.
\newblock In \emph{International Conference on Machine Learning}, pp.\
  2555--2565. PMLR, 2019{\natexlab{b}}.

\bibitem[Hafner et~al.(2020)Hafner, Lillicrap, Norouzi, and
  Ba]{hafner2020mastering}
Hafner, D., Lillicrap, T., Norouzi, M., and Ba, J.
\newblock Mastering atari with discrete world models.
\newblock \emph{arXiv preprint arXiv:2010.02193}, 2020.

\bibitem[Huang et~al.(2020)Huang, Xie, Bharadhwaj, and
  Shkurti]{huang2020continual}
Huang, Y., Xie, K., Bharadhwaj, H., and Shkurti, F.
\newblock Continual model-based reinforcement learning with hypernetworks.
\newblock \emph{arXiv preprint arXiv:2009.11997}, 2020.

\bibitem[Huisman et~al.(2021)Huisman, van Rijn, and Plaat]{huisman2021survey}
Huisman, M., van Rijn, J.~N., and Plaat, A.
\newblock A survey of deep meta-learning.
\newblock \emph{Artificial Intelligence Review}, pp.\  1--59, 2021.

\bibitem[Kemker et~al.(2018)Kemker, McClure, Abitino, Hayes, and
  Kanan]{kemker2018measuring}
Kemker, R., McClure, M., Abitino, A., Hayes, T., and Kanan, C.
\newblock Measuring catastrophic forgetting in neural networks.
\newblock In \emph{Proceedings of the AAAI Conference on Artificial
  Intelligence}, volume~32, 2018.

\bibitem[Kessler et~al.(2021)Kessler, Parker-Holder, Ball, Zohren, and
  Roberts]{kessler2021same}
Kessler, S., Parker-Holder, J., Ball, P., Zohren, S., and Roberts, S.~J.
\newblock Same state, different task: Continual reinforcement learning without
  interference.
\newblock \emph{arXiv preprint arXiv:2106.02940}, 2021.

\bibitem[Khetarpal et~al.(2020)Khetarpal, Riemer, Rish, and
  Precup]{khetarpal2020towards}
Khetarpal, K., Riemer, M., Rish, I., and Precup, D.
\newblock Towards continual reinforcement learning: A review and perspectives.
\newblock \emph{arXiv preprint arXiv:2012.13490}, 2020.

\bibitem[Kirkpatrick et~al.(2017)Kirkpatrick, Pascanu, Rabinowitz, Veness,
  Desjardins, Rusu, Milan, Quan, Ramalho, Grabska-Barwinska,
  et~al.]{kirkpatrick2017overcoming}
Kirkpatrick, J., Pascanu, R., Rabinowitz, N., Veness, J., Desjardins, G., Rusu,
  A.~A., Milan, K., Quan, J., Ramalho, T., Grabska-Barwinska, A., et~al.
\newblock Overcoming catastrophic forgetting in neural networks.
\newblock \emph{Proceedings of the national academy of sciences}, 114\penalty0
  (13):\penalty0 3521--3526, 2017.

\bibitem[Konidaris et~al.(2012)Konidaris, Scheidwasser, and
  Barto]{konidaris2012transfer}
Konidaris, G., Scheidwasser, I., and Barto, A.~G.
\newblock Transfer in reinforcement learning via shared features.
\newblock 2012.

\bibitem[Lazaric(2012)]{lazaric2012transfer}
Lazaric, A.
\newblock Transfer in reinforcement learning: a framework and a survey.
\newblock In \emph{Reinforcement Learning}, pp.\  143--173. Springer, 2012.

\bibitem[Lu et~al.(2020)Lu, Grover, Abbeel, and Mordatch]{lu2020reset}
Lu, K., Grover, A., Abbeel, P., and Mordatch, I.
\newblock Reset-free lifelong learning with skill-space planning.
\newblock \emph{arXiv preprint arXiv:2012.03548}, 2020.

\bibitem[McCloskey \& Cohen(1989)McCloskey and
  Cohen]{mccloskey1989catastrophic}
McCloskey, M. and Cohen, N.~J.
\newblock Catastrophic interference in connectionist networks: The sequential
  learning problem.
\newblock In \emph{Psychology of learning and motivation}, volume~24, pp.\
  109--165. Elsevier, 1989.

\bibitem[Moerland et~al.(2020)Moerland, Broekens, and
  Jonker]{moerland2020model}
Moerland, T.~M., Broekens, J., and Jonker, C.~M.
\newblock Model-based reinforcement learning: A survey.
\newblock \emph{arXiv preprint arXiv:2006.16712}, 2020.

\bibitem[Moore(1990)]{moore1990efficient}
Moore, A.~W.
\newblock Efficient memory-based learning for robot control.
\newblock 1990.

\bibitem[Nguyen et~al.(2012)Nguyen, Silander, and
  Leong]{nguyen2012transferring}
Nguyen, T., Silander, T., and Leong, T.
\newblock Transferring expectations in model-based reinforcement learning.
\newblock \emph{Advances in Neural Information Processing Systems}, 25, 2012.

\bibitem[Paszke et~al.(2019)Paszke, Gross, Massa, Lerer, Bradbury, Chanan,
  Killeen, Lin, Gimelshein, Antiga, et~al.]{paszke2019pytorch}
Paszke, A., Gross, S., Massa, F., Lerer, A., Bradbury, J., Chanan, G., Killeen,
  T., Lin, Z., Gimelshein, N., Antiga, L., et~al.
\newblock Pytorch: An imperative style, high-performance deep learning library.
\newblock \emph{Advances in neural information processing systems},
  32:\penalty0 8026--8037, 2019.

\bibitem[Robins(1995)]{robins1995catastrophic}
Robins, A.
\newblock Catastrophic forgetting, rehearsal and pseudorehearsal.
\newblock \emph{Connection Science}, 7\penalty0 (2):\penalty0 123--146, 1995.

\bibitem[Schrittwieser et~al.(2019)Schrittwieser, Antonoglou, Hubert, Simonyan,
  Sifre, Schmitt, Guez, Lockhart, Hassabis, Graepel,
  et~al.]{schrittwieser2019mastering}
Schrittwieser, J., Antonoglou, I., Hubert, T., Simonyan, K., Sifre, L.,
  Schmitt, S., Guez, A., Lockhart, E., Hassabis, D., Graepel, T., et~al.
\newblock Mastering atari, go, chess and shogi by planning with a learned
  model.
\newblock \emph{arXiv preprint arXiv:1911.08265}, 2019.

\bibitem[Sutton \& Barto(2018)Sutton and Barto]{sutton2018reinforcement}
Sutton, R.~S. and Barto, A.~G.
\newblock \emph{Reinforcement learning: An introduction}.
\newblock MIT press, 2018.

\bibitem[Sutton et~al.(1999)Sutton, Precup, and Singh]{sutton1999between}
Sutton, R.~S., Precup, D., and Singh, S.
\newblock Between mdps and semi-mdps: A framework for temporal abstraction in
  reinforcement learning.
\newblock \emph{Artificial intelligence}, 112\penalty0 (1-2):\penalty0
  181--211, 1999.

\bibitem[Sutton et~al.(2012)Sutton, Szepesv{\'a}ri, Geramifard, and
  Bowling]{sutton2012dyna}
Sutton, R.~S., Szepesv{\'a}ri, C., Geramifard, A., and Bowling, M.~P.
\newblock Dyna-style planning with linear function approximation and
  prioritized sweeping.
\newblock \emph{arXiv preprint arXiv:1206.3285}, 2012.

\bibitem[Tassa et~al.(2018)Tassa, Doron, Muldal, Erez, Li, Casas, Budden,
  Abdolmaleki, Merel, Lefrancq, et~al.]{tassa2018deepmind}
Tassa, Y., Doron, Y., Muldal, A., Erez, T., Li, Y., Casas, D. d.~L., Budden,
  D., Abdolmaleki, A., Merel, J., Lefrancq, A., et~al.
\newblock Deepmind control suite.
\newblock \emph{arXiv preprint arXiv:1801.00690}, 2018.

\bibitem[Taylor \& Stone(2009)Taylor and Stone]{taylor2009transfer}
Taylor, M.~E. and Stone, P.
\newblock Transfer learning for reinforcement learning domains: A survey.
\newblock \emph{Journal of Machine Learning Research}, 10\penalty0 (7), 2009.

\bibitem[van Seijen \& Sutton(2015)van Seijen and Sutton]{vanseijen2015deeper}
van Seijen, H. and Sutton, R.
\newblock A deeper look at planning as learning from replay.
\newblock In \emph{International conference on machine learning}, pp.\
  2314--2322, 2015.

\bibitem[Van~Seijen et~al.(2020)Van~Seijen, Nekoei, Racah, and
  Chandar]{vanseijen-LoCA}
Van~Seijen, H., Nekoei, H., Racah, E., and Chandar, S.
\newblock The loca regret: A consistent metric to evaluate model-based behavior
  in reinforcement learning.
\newblock In Larochelle, H., Ranzato, M., Hadsell, R., Balcan, M.~F., and Lin,
  H. (eds.), \emph{Advances in Neural Information Processing Systems},
  volume~33, pp.\  6562--6572. Curran Associates, Inc., 2020.
\newblock URL
  \url{https://proceedings.neurips.cc/paper/2020/file/48db71587df6c7c442e5b76cc723169a-Paper.pdf}.

\bibitem[Wan et~al.(2019)Wan, Abbas, White, White, and Sutton]{wan2019planning}
Wan, Y., Abbas, Z., White, A., White, M., and Sutton, R.~S.
\newblock Planning with expectation models.
\newblock \emph{arXiv preprint arXiv:1904.01191}, 2019.

\bibitem[Wang et~al.(2019)Wang, Bao, Clavera, Hoang, Wen, Langlois, Zhang,
  Zhang, Abbeel, and Ba]{wang2019benchmarking}
Wang, T., Bao, X., Clavera, I., Hoang, J., Wen, Y., Langlois, E., Zhang, S.,
  Zhang, G., Abbeel, P., and Ba, J.
\newblock Benchmarking model-based reinforcement learning.
\newblock \emph{arXiv preprint arXiv:1907.02057}, 2019.

\bibitem[Zhang et~al.(2019)Zhang, Vikram, Smith, Abbeel, Johnson, and
  Levine]{zhang2019solar}
Zhang, M., Vikram, S., Smith, L., Abbeel, P., Johnson, M., and Levine, S.
\newblock Solar: Deep structured representations for model-based reinforcement
  learning.
\newblock In \emph{International Conference on Machine Learning}, pp.\
  7444--7453. PMLR, 2019.

\bibitem[Zhu et~al.(2020)Zhu, Lin, and Zhou]{zhu2020transfer}
Zhu, Z., Lin, K., and Zhou, J.
\newblock Transfer learning in deep reinforcement learning: A survey.
\newblock \emph{arXiv preprint arXiv:2009.07888}, 2020.

\end{thebibliography}
\bibliographystyle{icml2022}

\onecolumn
\appendix
%%%%%%%%%%%%%%%%%%%%%%

\counterwithin{figure}{section}
\counterwithin{table}{section}
\counterwithin{theorem}{section}
\counterwithin{equation}{section}

\section{
Supplementary Details for Experiments in Section\ \ref{sec: tabular}
}\label{app: tabular experiments}

\subsection{The method \mbox{\emph{mb-2-r}}}

The method \mbox{\emph{mb-2-r}} maintains an estimate of the 2-step transition model $\hat p_2(s''|s,a)$, which estimates a distribution of the state 2 time steps in the future, as well as a 2-step reward model $\hat r_2(s,a)$, which estimates the expected discounted sum of rewards over the next 2 time steps. These models are updated according to
\begin{eqnarray*}
\hat r_2(S_t,A_t) &\leftarrow& \hat r_2(S_t,A_t) + \alpha \big( R_t + \gamma R_{t+1}  - \hat r_2(S_t,A_t)\big)\,, \\
\hat p_2(\cdot|S_t,A_t) &\leftarrow& \hat p_2(\cdot|S_t,A_t) + \alpha \big( \langle  S_{t+2} \rangle  - \hat p_2(\cdot|S_t,A_t)\big)\,,
\end{eqnarray*}
with $\alpha$ the fixed learning rate and $\langle  S_{t+2} \rangle$ a one-hot encoding of state $S_{t+2}$. Both $\langle  S_{t+2} \rangle$ and $p(\cdot|S_t,A_t)$ are vectors of length $N$, the total number of states. To be able to make these updates, \mbox{\emph{mb-2-r}} needs to store at least the last two samples and preforms updates with a small delay, as occurs for any \mbox{n-step} method (for more details on that, see Chapter 7 of \citet{sutton2018reinforcement}).

Planning consists of performing a single state-value update at each time step based on these model estimates:
$$ V(s) \leftarrow \max_{a} \Big(\hat r_2(s,a)  + \gamma^2 \sum_{s''} \hat p_2(s''|s,a) V(s'') \Big)\,.$$

\subsection{Details of the Tabular Experiment}\label{app: Experiment Procedure of the tabular LoCA setup}

\begin{table}[h!]
    \centering
    \begin{tabular}{c|c|c}
        \hline
        \multirow{6}{5em}{Initial distributions} & Phase 1 training  & Uniform distribution over the entire state space \\
        & Phase 1 evaluation  & Uniform distribution over the entire state space \\
        & Phase 2 training  & Uniform distribution over states within T1-zone\\
        & Phase 2 evaluation  & Uniform distribution over the entire state space \\
        & Phase 3 training  & Uniform distribution over the entire state space  \\
        & Phase 3 evaluation  & Uniform distribution over the entire state space \\
        \hline
        \multirow{3}{5em}{Training steps} & Phase 1 steps & $100,000$\\
        & Phase 2 steps & $50,000$\\
        & Phase 3 steps & $50,000$\\
        \hline
        \multirow{3}{5em}{Other details} & Training steps between two evaluations & 500\\
        & Number of runs & 10\\
        & Number of evaluation episodes & 100 \\
    \hline
    \end{tabular}
    \caption{Details of the tabular experiment.}
    \label{tab: Details of the tabular experiment}
\end{table}

For the GridWorldLoCA domain, we used the same discount factor as used in \citep{vanseijen-LoCA}, $\gamma = 0.97$, but used stochastic transitions. In particular, with 25\% probability the action outcome is that of a random action instead of the selected action.

For all methods, we used an $\epsilon$-greedy behavior policy with $\epsilon = 0.1$. In addition, we used optimistic initialization to encourage high exploration during initial training in Phase 1. Specifically, for Sarsa($\lambda$) we set the initial action-values to $8$ (twice the value of the maximum reward in Phase 1). And for all model-based methods, we set the initial state-values and initial reward-function estimates to 8.  For Sarsa($\lambda$), we used $\lambda = 0.95$. We roughly optimized the step-size for each method. This resulted in a step-size of 0.2 for all model-based method and a step-size of 0.03 for Sarsa($\lambda$). The reason the best step-size for Sarsa($\lambda$) is substantially lower than the one for the model-based methods is that a $\lambda$ of 0.95 combined with the relatively high environment stochasticity leads to high variance updates for Sarsa($\lambda$). Hence, to counter this variance and get accurate estimates a small step-size is needed.

\subsection{MuZero and its relation to Failure Mode \#1}\label{app: MuZero}

We start by summarizing some key elements of MuZero's implementation. At its core, MuZero learns a 1-step model $g$ that takes as input a hidden state $h_t$ (derived from past observations) and action $a_t$, and outputs an estimate of the reward and next hidden state $h_{t+1}$. Simultaneously, it learns a value-function $v$ that maps a hidden state $s_t$ to an estimate of the expected return. At every time step, a MCTS routine is performed starting from the current hidden state, by leveraging the learned $g$ model as the agent does not have access to the true environment model. MCTS is performed up to a certain maximum horizon k (k=5 for the Atari domain). At the leaf nodes of the tree, MuZero's MCTS bootstraps from $v$ rather than performing a sample rollout, as normally occurs in MCTS. The output of MuZero's MCTS routine is a (local) policy $\pi_t(h_t)$, according to which action $a_t$ is selected, and a target value $v_t^{target}$. This target value is an improved estimate of $v(h_t)$, as it is computed from the bootstrapped MCTS routine. The target value $v_t^{target}$ is stored in a replay buffer, in addition to the current observation $o_t$, action $a_t$, reward $r_{t+1}$ and policy $\pi_t(h_t)$. The value function $v$ is updated by selecting a past trajectory, computing the hidden state $h_t$ for some time step $t$ within the trajectory and updating its value $v(h_t)$ using the $n$-step return $r_{t+1} + \gamma r_{t+2} +   ... \gamma^n \,v_{t+n}^{target}$. Crucially, the $n$-step return bootstraps not from $v(h_{t+n})$, but from the target value of the corresponding time step, $v_{t+n}^{target}$. These target values are not updated once they are in the replay buffer. Hence, despite that MuZero employs experience replay to update the value function $v$, during Phase 2 the update targets outside the T1-zone remain the same and do not adapt to the newly observed reward. This can be viewed as a variation of Failure Mode \#1.

\section{
Supplementary Details and Results for Experiments in Section\ \ref{sec: planet and dreamer}
}

\subsection{Setup of the DreamerV2 Experiment}\label{app: Experiment Setup of the DreamerV2 Experiment}

\begin{table}[h!]
    \centering
    \begin{tabular}{c|c|c}
        \hline
        \multirow{6}{5em}{Initial distributions} & Phase 1 training  & Uniform distribution over the entire state space \\
        & Phase 1 evaluation  & Uniform distribution over the entire states outside T1-zone\\
        & Phase 2 training  & Uniform distribution over states within T1-zone\\
        & Phase 2 evaluation  & Uniform distribution over the entire states outside T1-zone\\
        & Phase 3 training  & Uniform distribution over the entire state space \\
        & Phase 3 evaluation  & Uniform distribution over the entire states outside T1-zone\\
        \hline
        \multirow{3}{5em}{Training steps} & Phase 1 steps & Planet: $3 \times 10^5$, DreamerV2: $10^6$\\
        & Phase 2 steps & Planet: $2.5 \times 10^5$, DreamerV2: $5 \times 10^5$\\
        & Phase 3 steps & Planet: $2.5 \times 10^5$, DreamerV2: $10^6$\\
        \hline
        \multirow{3}{5em}{Other details} & Training steps between two evaluations & PlaNet: $10^4$, DreamerV2 $5 \times 10^4$\\
        & Number of runs & 5\\
        & Number of evaluation episodes & 5 \\
    \hline
    \end{tabular}
    \caption{Details of DreamerV2 and PlaNet experiments.}
    \label{tab: Phase 1 Value and Model}
\end{table}

\begin{figure*}[h!]
\centering
    \begin{subfigure}{.4\textwidth}
        \centering
        \includegraphics[width=0.95\textwidth]{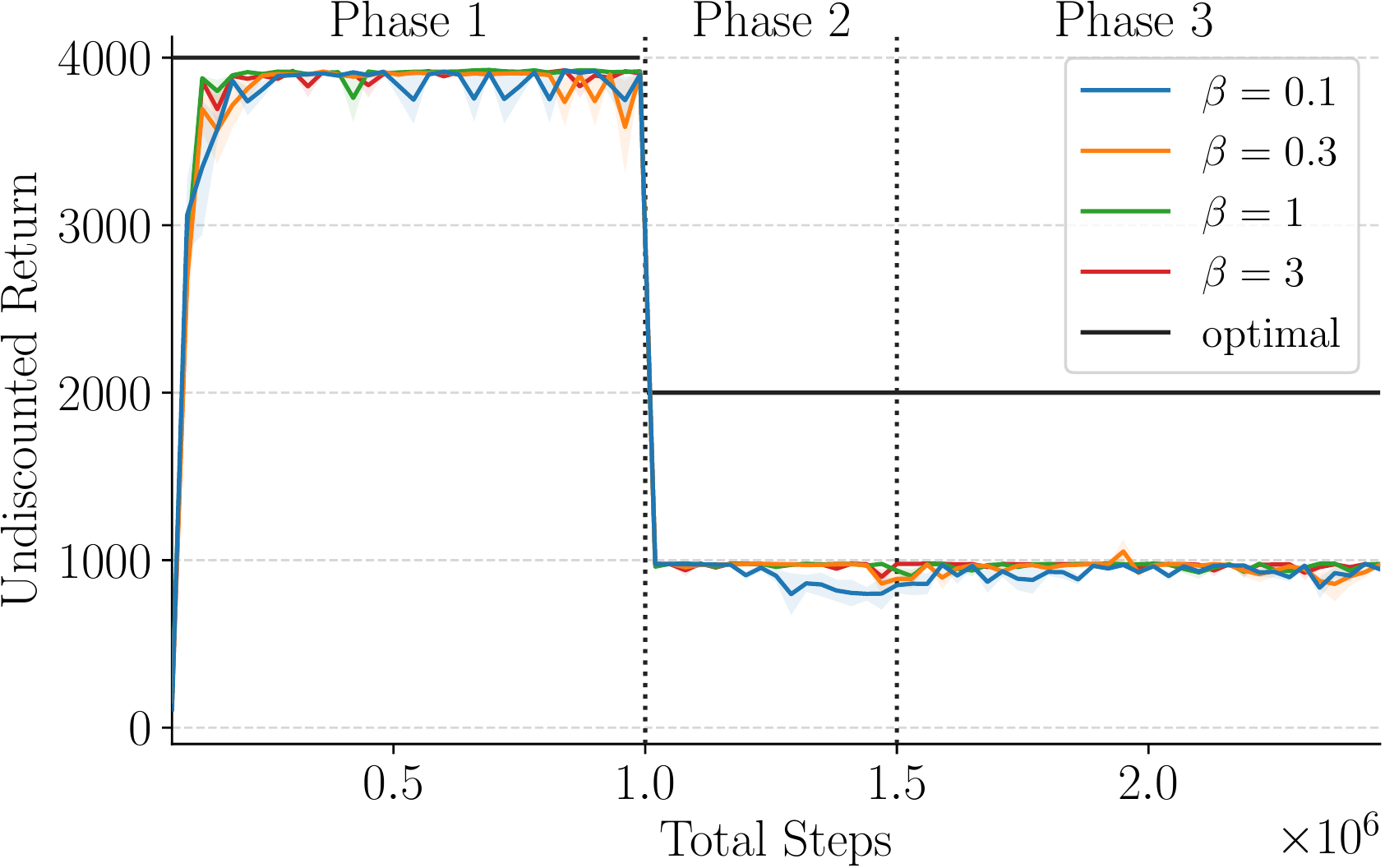}
    \end{subfigure}%
    \begin{subfigure}{.4\textwidth}
        \centering
        \includegraphics[width=0.95\textwidth]{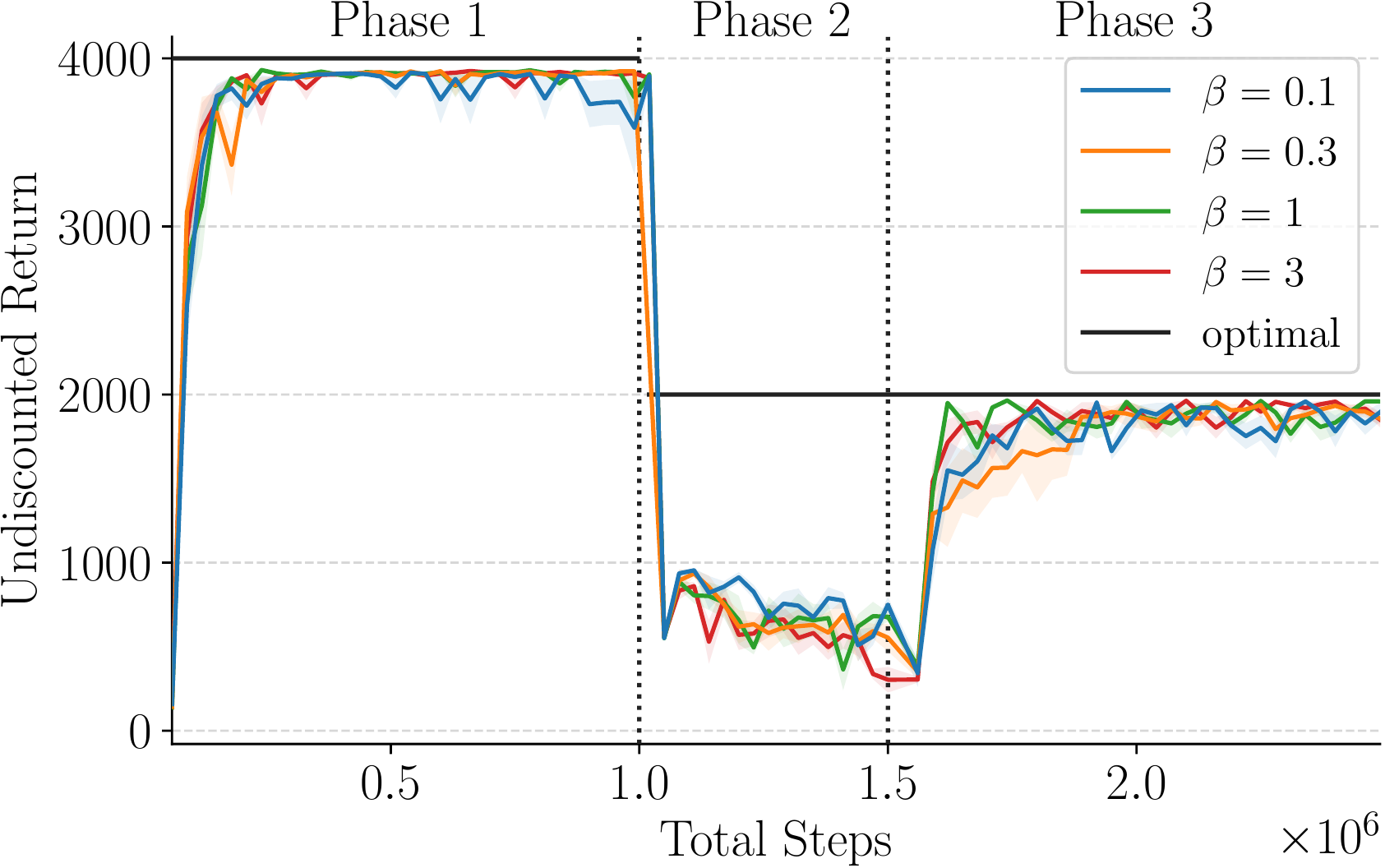}
    \end{subfigure}%
    
    \begin{subfigure}{.4\textwidth}
        \centering
        \includegraphics[width=0.95\textwidth]{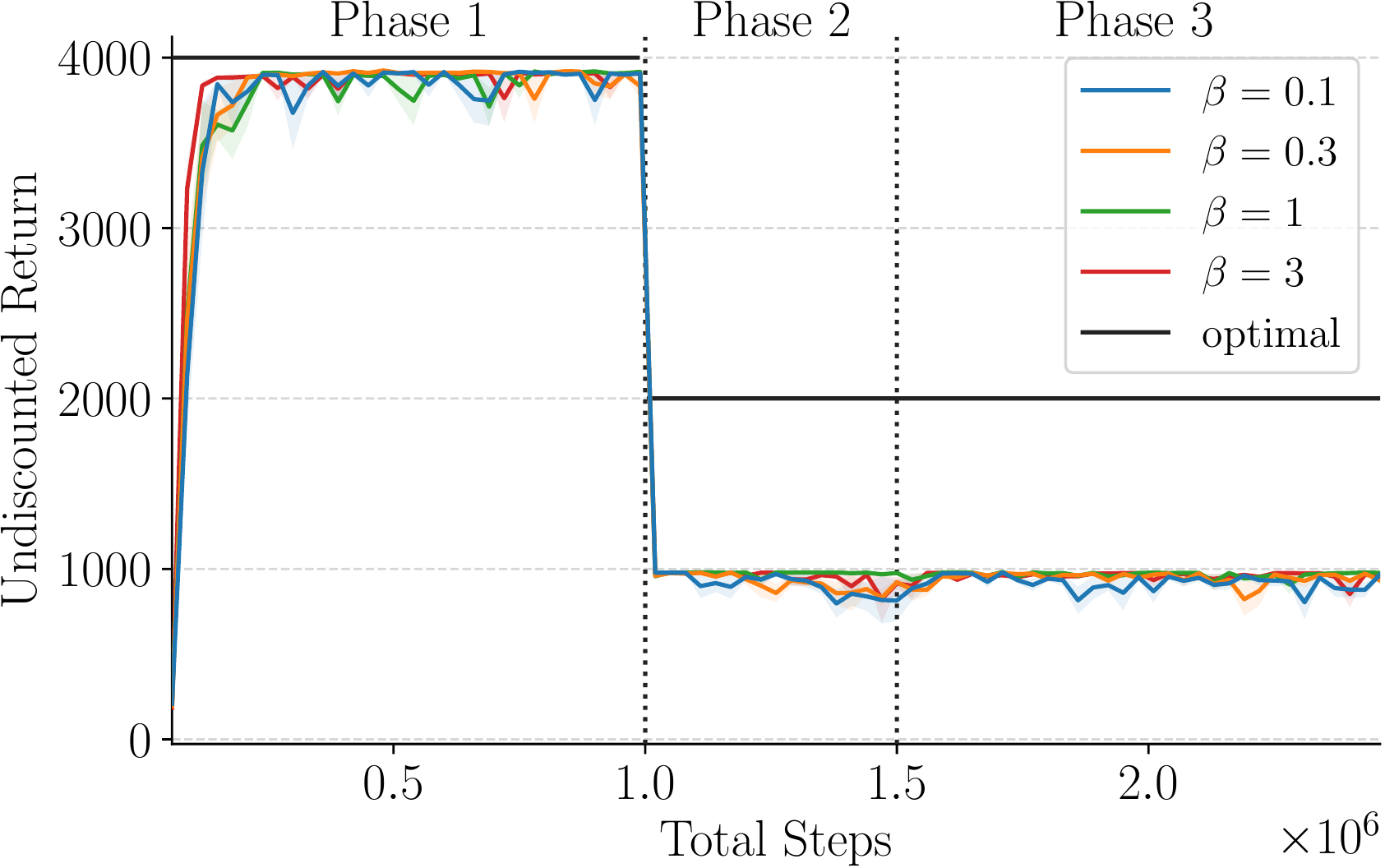}
    \end{subfigure}%
    \begin{subfigure}{.4\textwidth}
        \centering
        \includegraphics[width=0.95\textwidth]{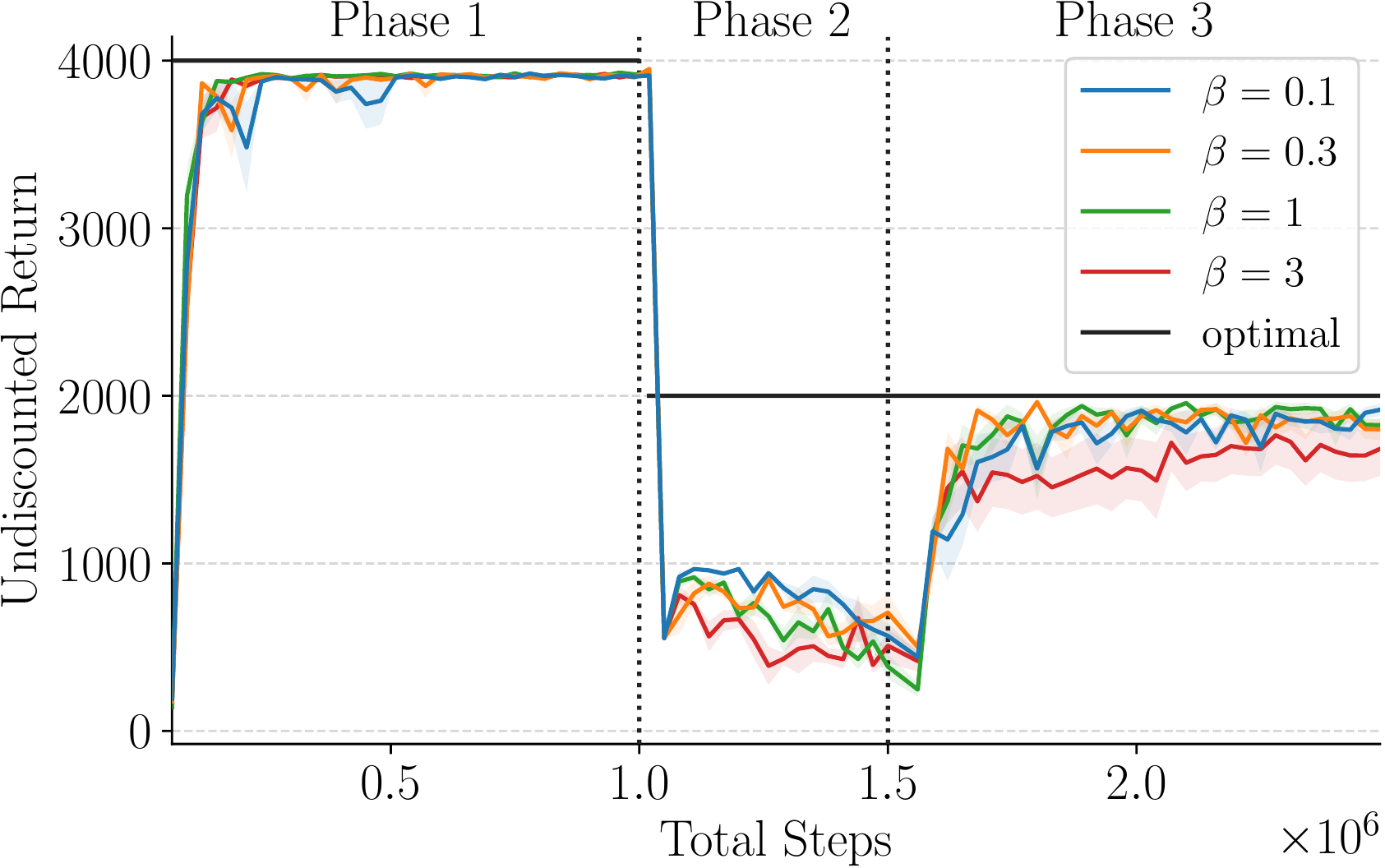}
    \end{subfigure}%
    
    \begin{subfigure}{.4\textwidth}
        \centering
        \includegraphics[width=0.95\textwidth]{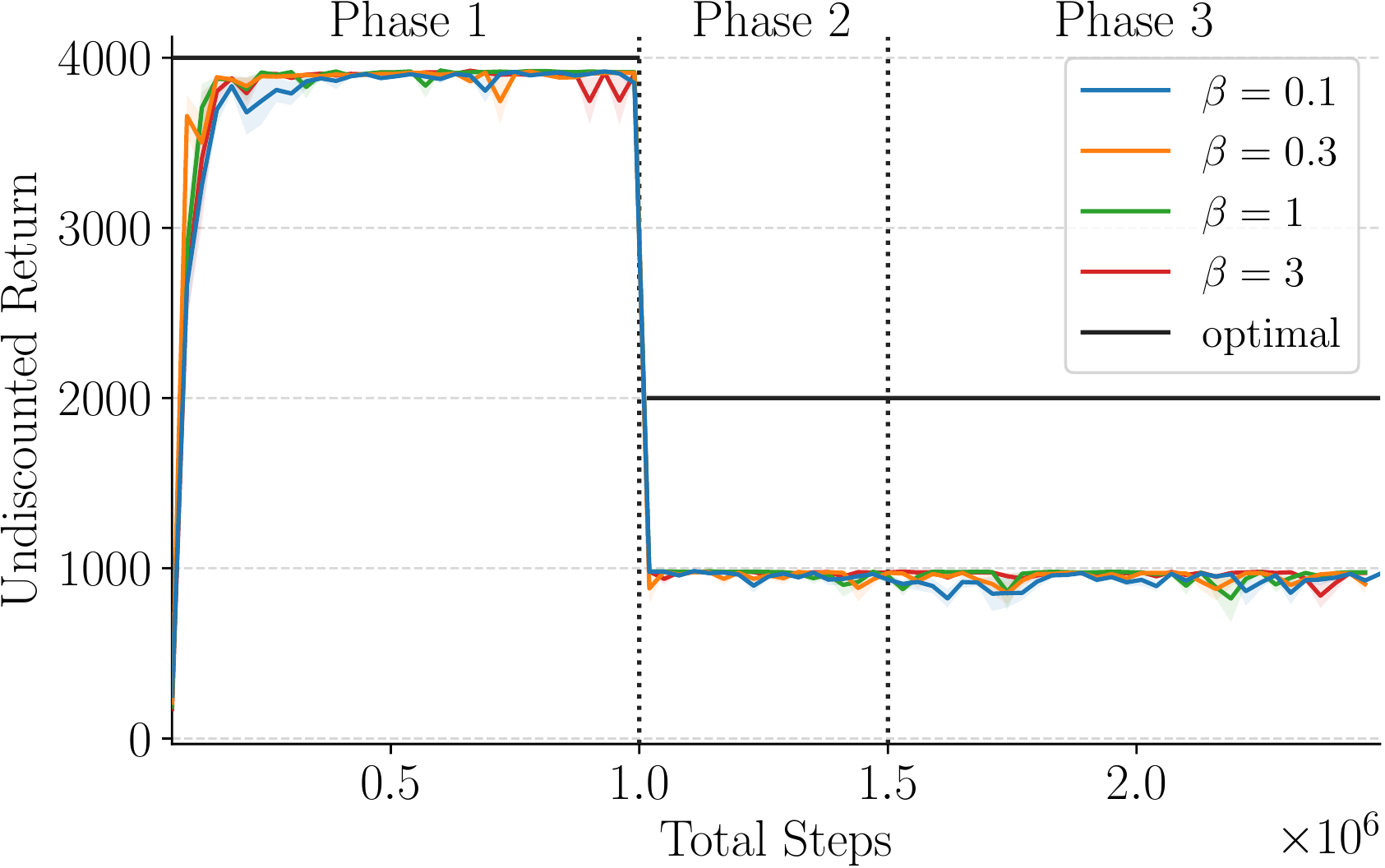}
    \end{subfigure}%
    \begin{subfigure}{.4\textwidth}
        \centering
        \includegraphics[width=0.95\textwidth]{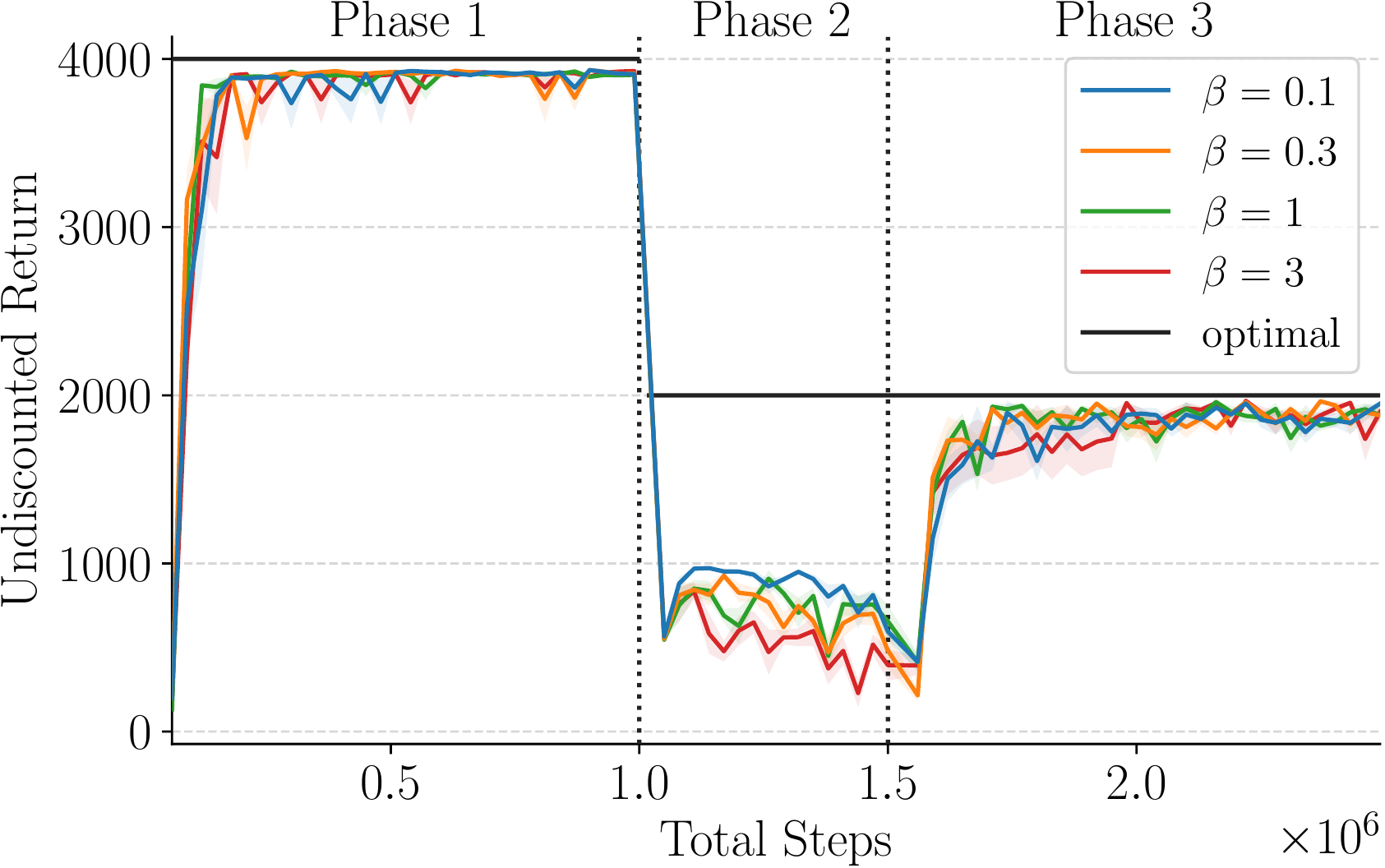}
    \end{subfigure}%
    
    \begin{subfigure}{.4\textwidth}
        \centering
        \includegraphics[width=0.95\textwidth]{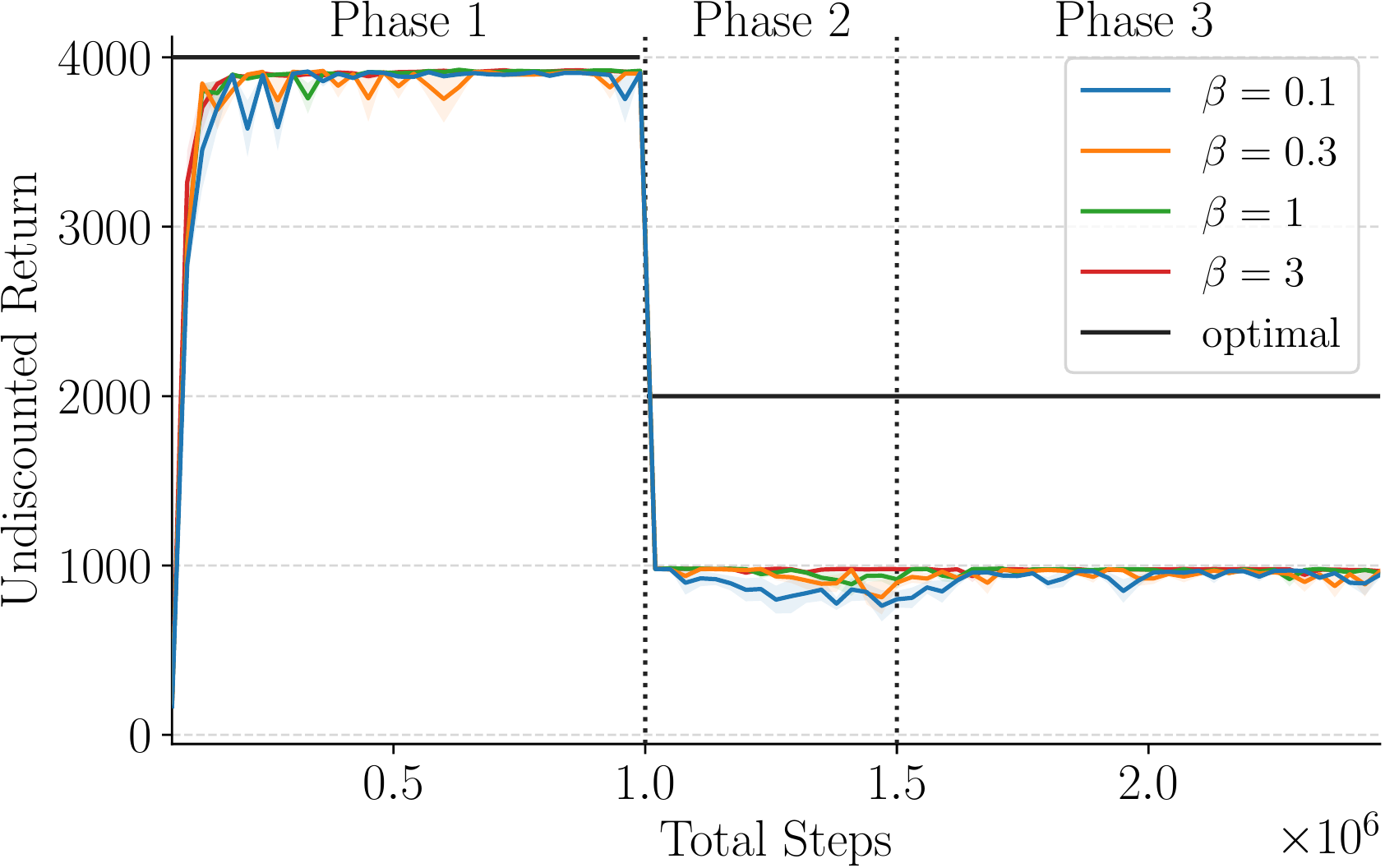}
    \end{subfigure}%
    \begin{subfigure}{.4\textwidth}
        \centering
        \includegraphics[width=0.95\textwidth]{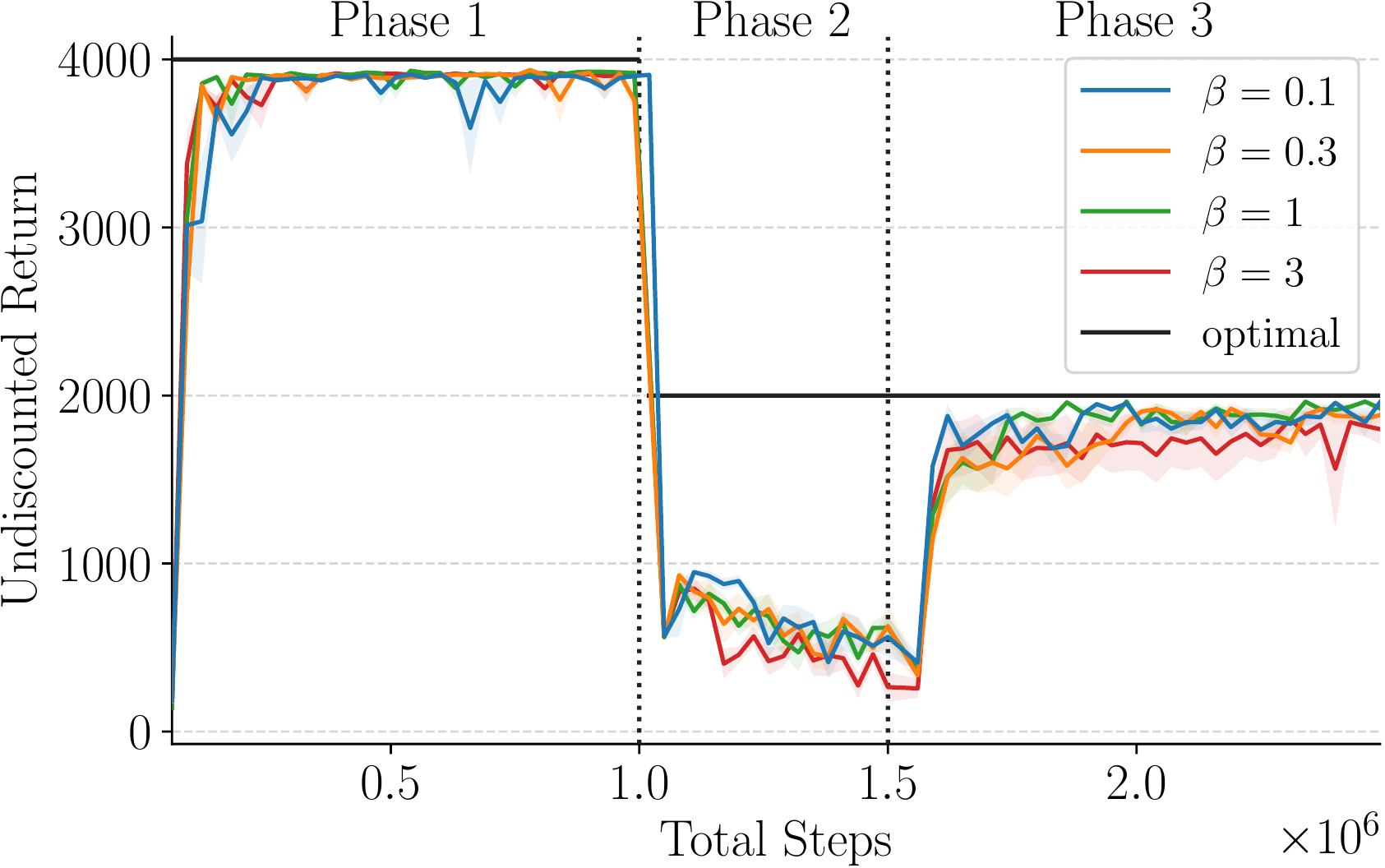}
    \end{subfigure}%
\caption{Plots showing the performance of DreamerV2 with different hyperparameter settings. The value of each point in a curve is the average return over 5 evaluation episodes. We searched over the critical hyperparameters: 1) the KL loss scale $\beta$, 2) the actor entropy loss scale $\eta$, and 3) the discount factor $\gamma$. In this figure, $\gamma = 0.99$. $\eta = 10^{-3}, 3 \times 10^{-4}, 10^{-4}, 10^{-5}$ in four different rows respectively. The right column corresponds to the buffer reinitializing case, and the left column corresponds to the other case.}
\label{fig: dreamer comp results 1}
\end{figure*}

\begin{figure*}[h!]
\centering
    \begin{subfigure}{.4\textwidth}
        \centering
        \includegraphics[width=0.95\textwidth]{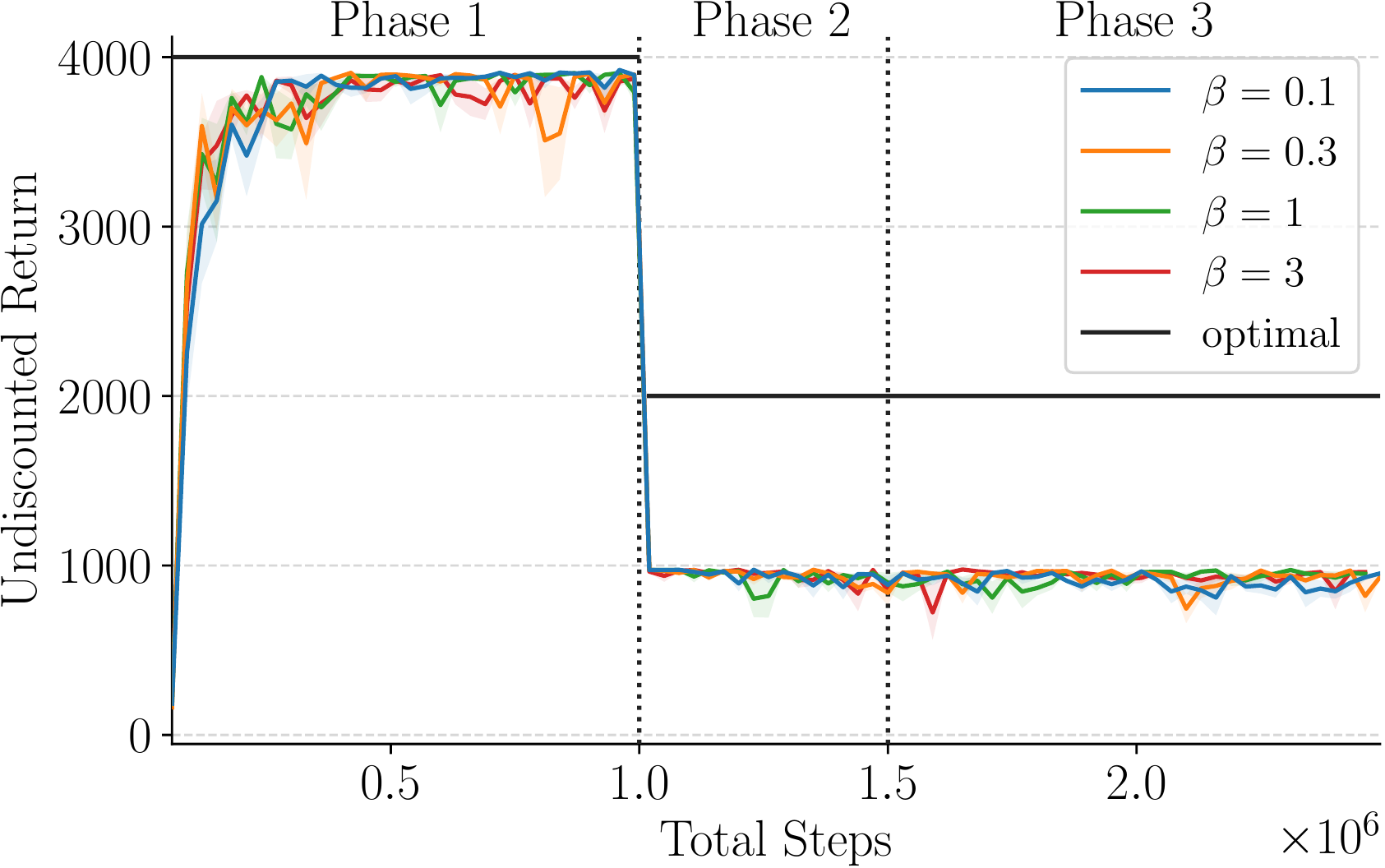}
    \end{subfigure}%
    \begin{subfigure}{.4\textwidth}
        \centering
        \includegraphics[width=0.95\textwidth]{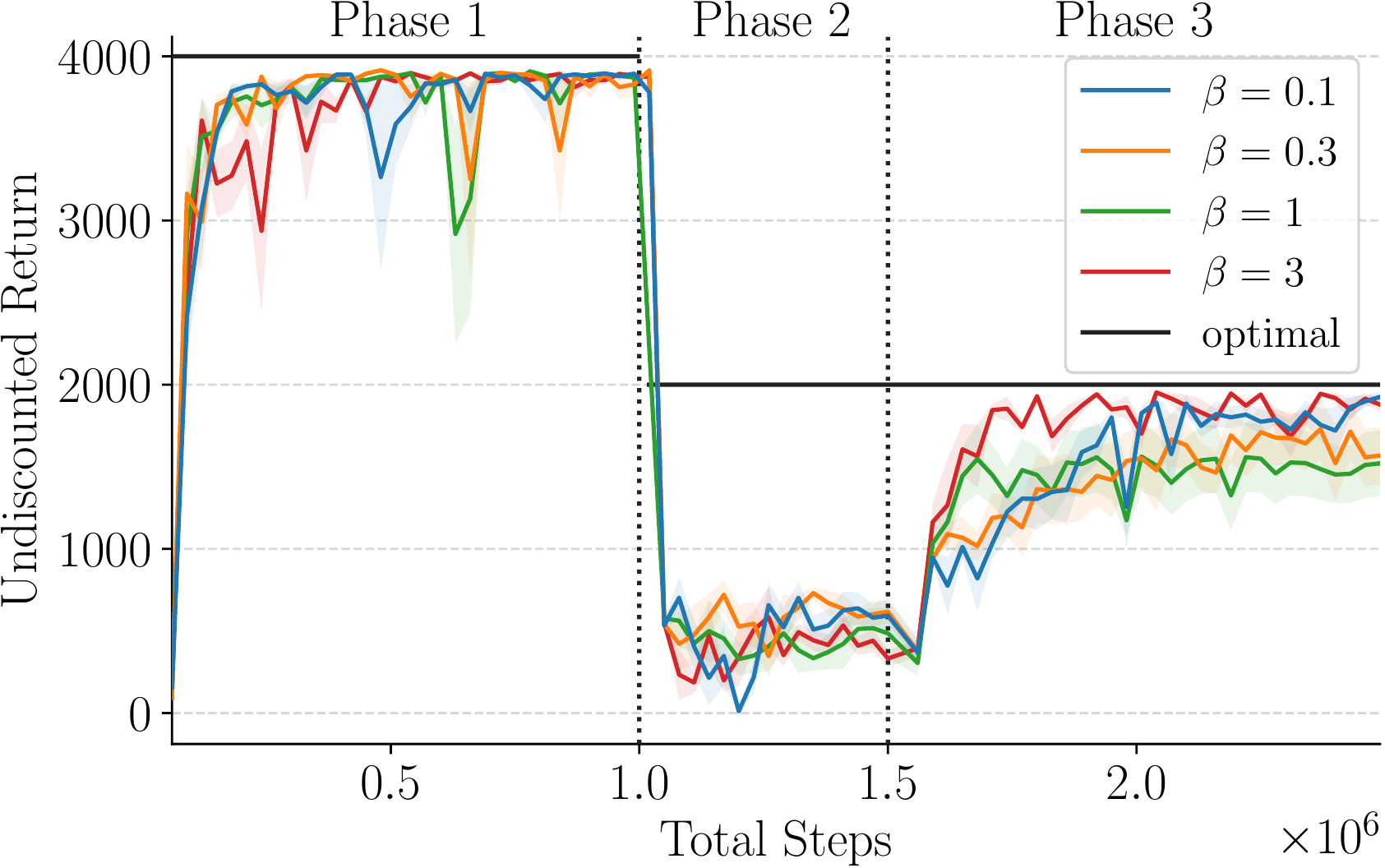}
    \end{subfigure}%
    
    \begin{subfigure}{.4\textwidth}
        \centering
        \includegraphics[width=0.95\textwidth]{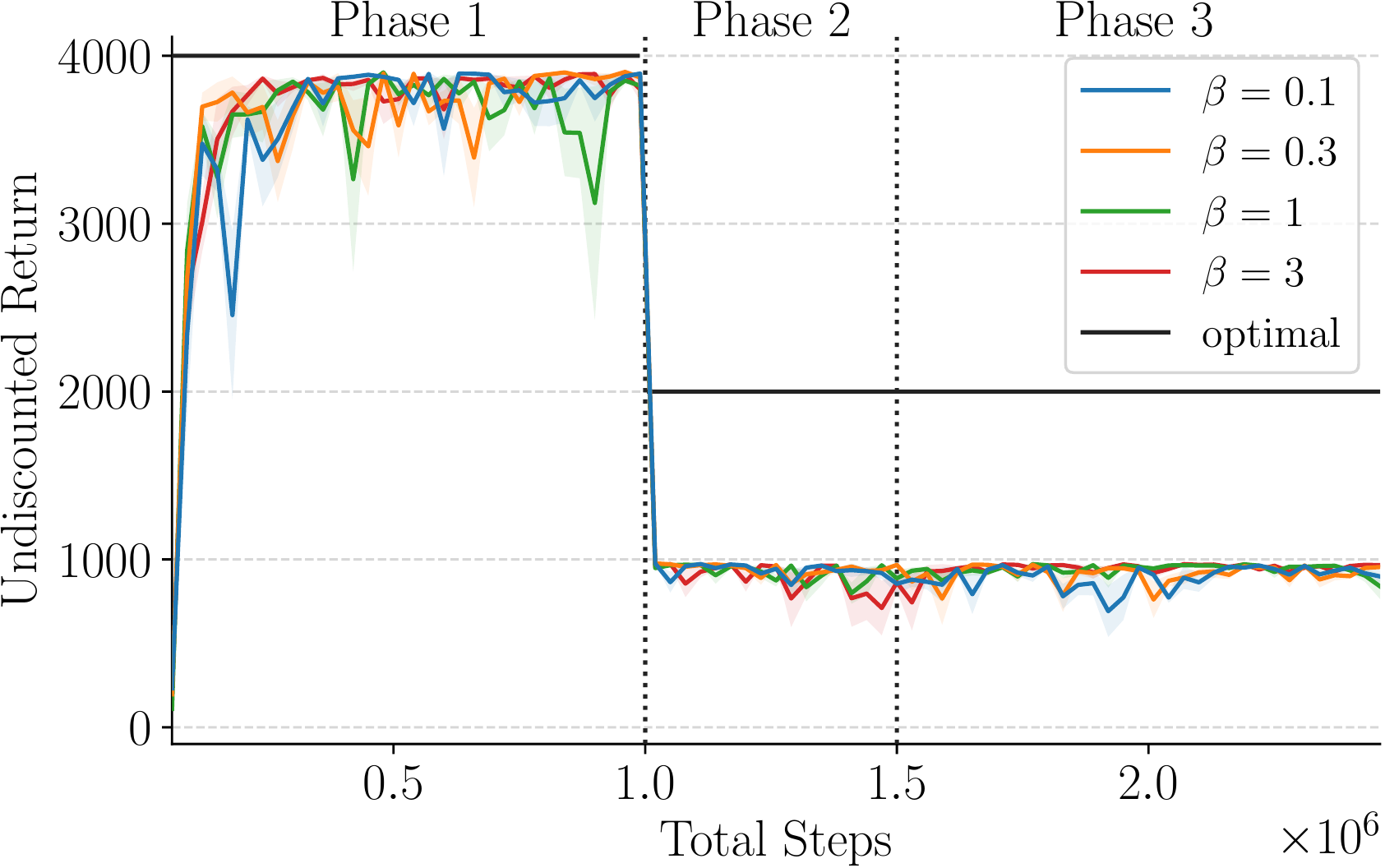}
    \end{subfigure}%
    \begin{subfigure}{.4\textwidth}
        \centering
        \includegraphics[width=0.95\textwidth]{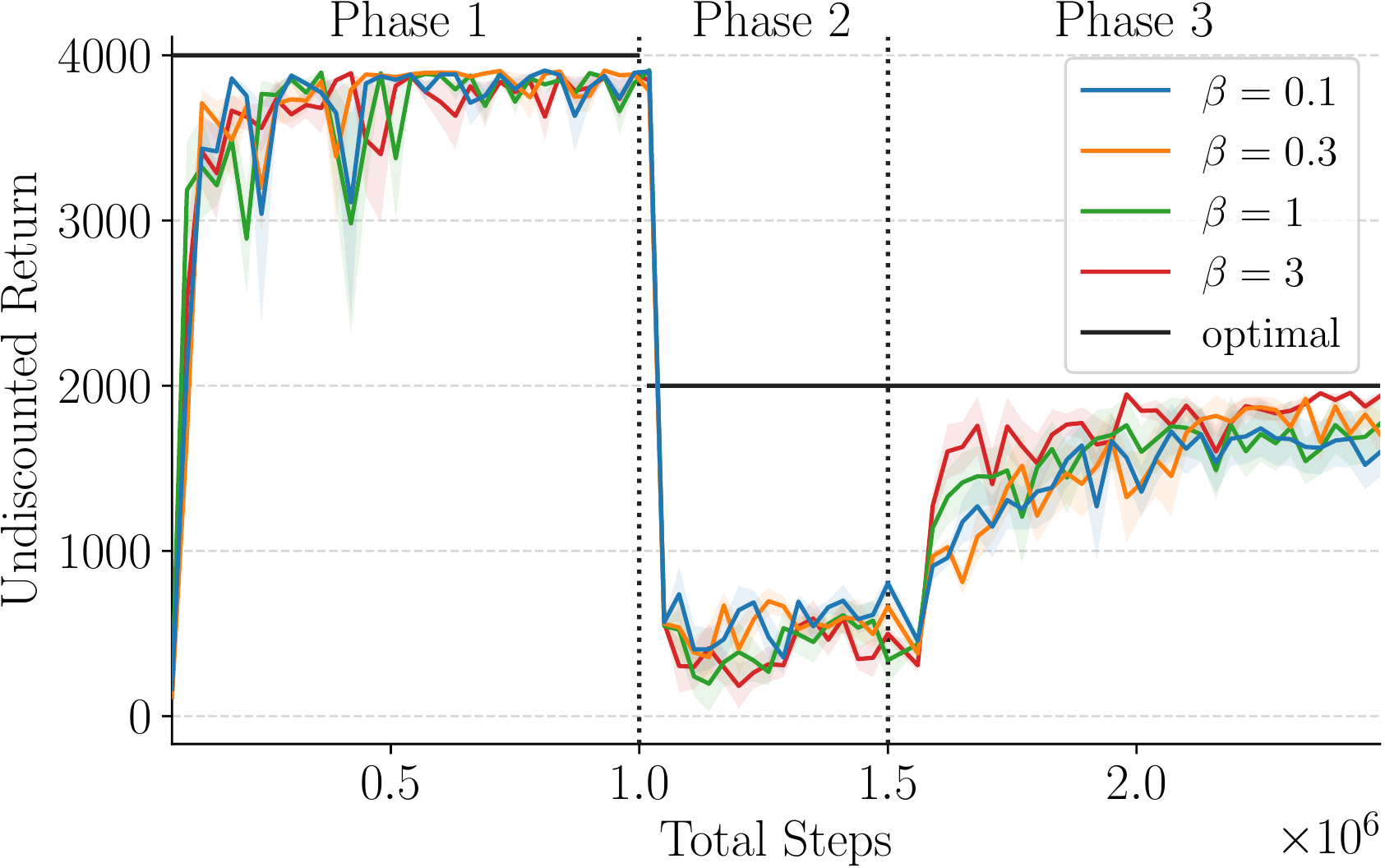}
    \end{subfigure}%
    
    \begin{subfigure}{.4\textwidth}
        \centering
        \includegraphics[width=0.95\textwidth]{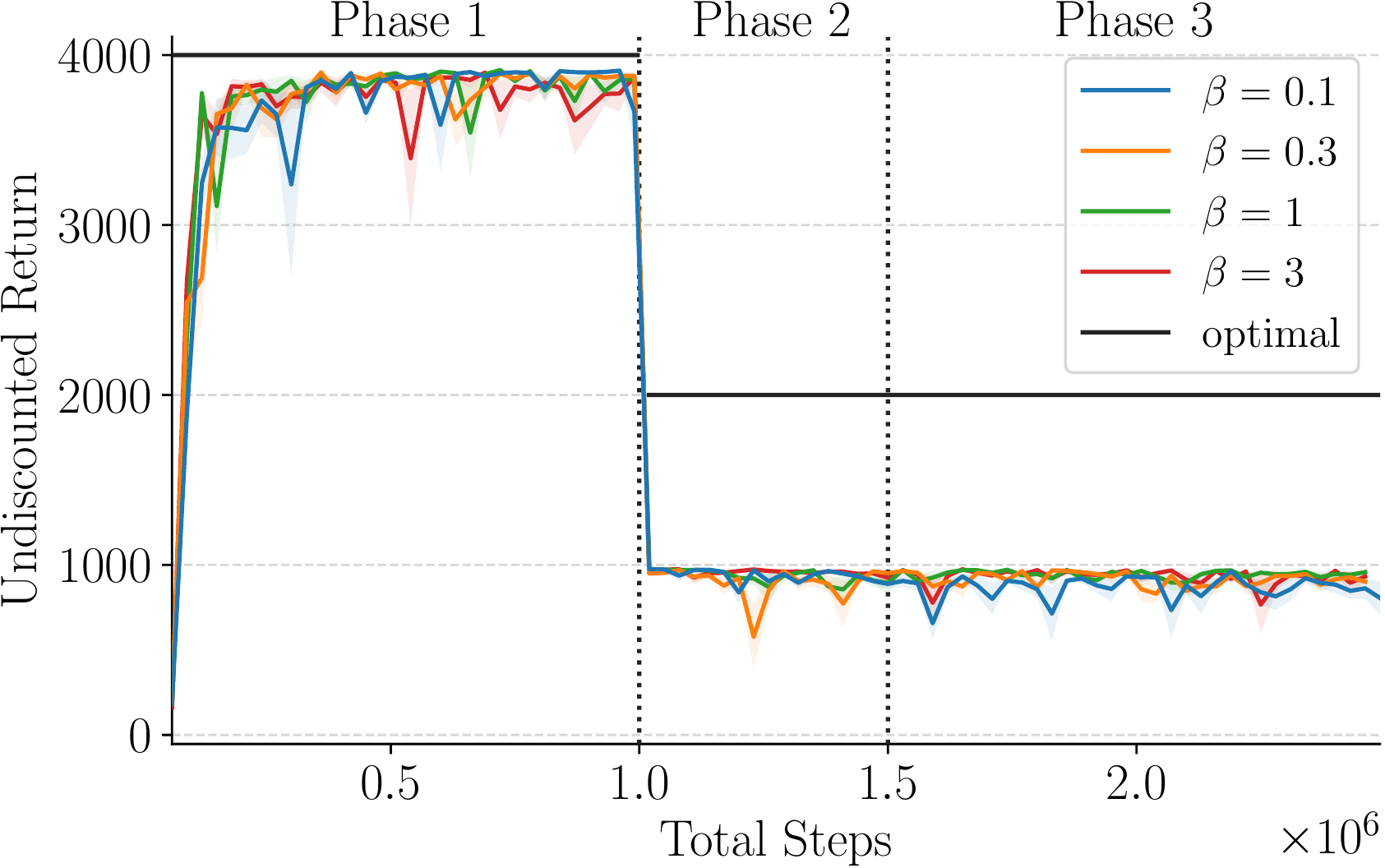}
    \end{subfigure}%
    \begin{subfigure}{.4\textwidth}
        \centering
        \includegraphics[width=0.95\textwidth]{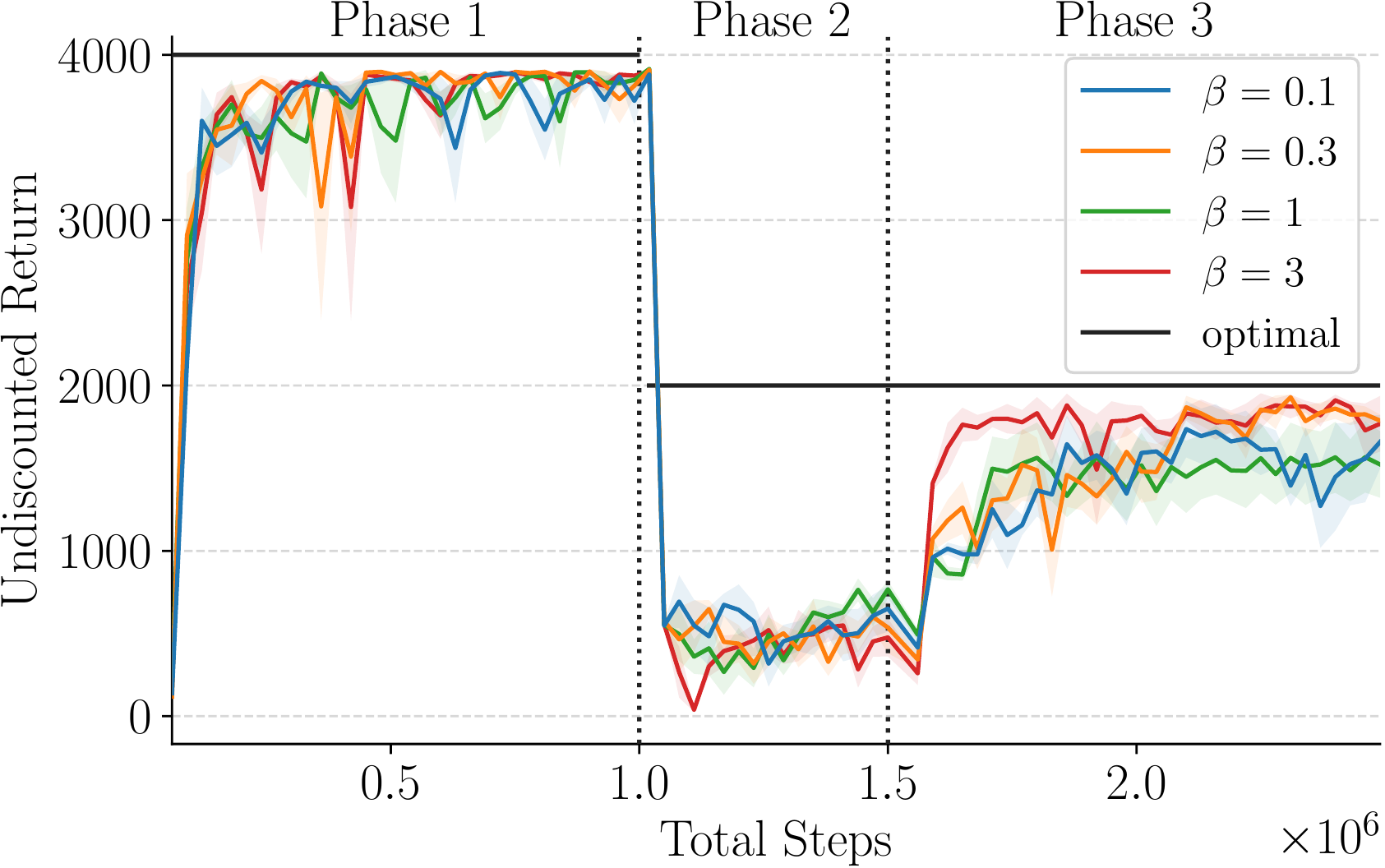}
    \end{subfigure}%
    
    \begin{subfigure}{.4\textwidth}
        \centering
        \includegraphics[width=0.95\textwidth]{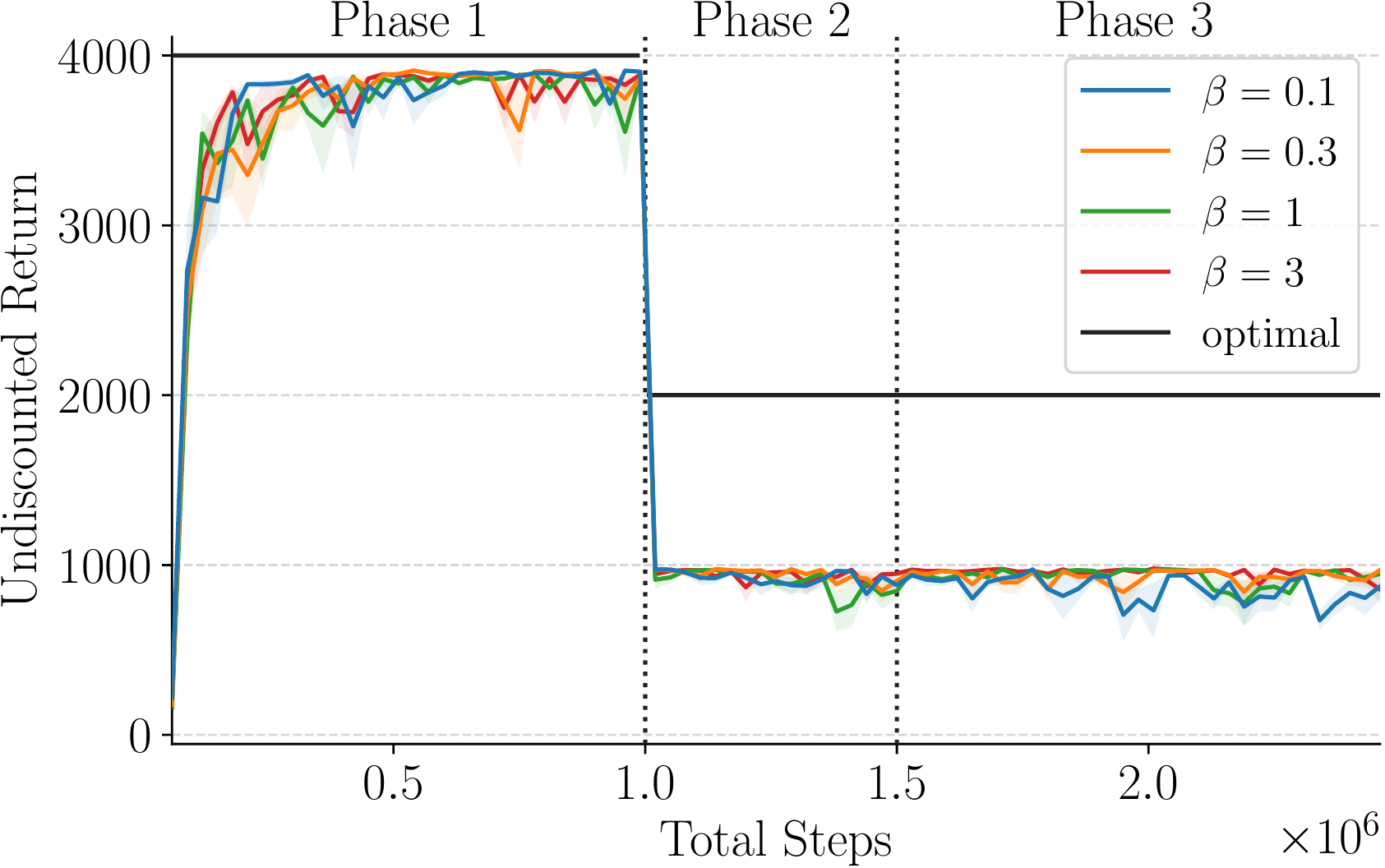}
    \end{subfigure}%
    \begin{subfigure}{.4\textwidth}
        \centering
        \includegraphics[width=0.95\textwidth]{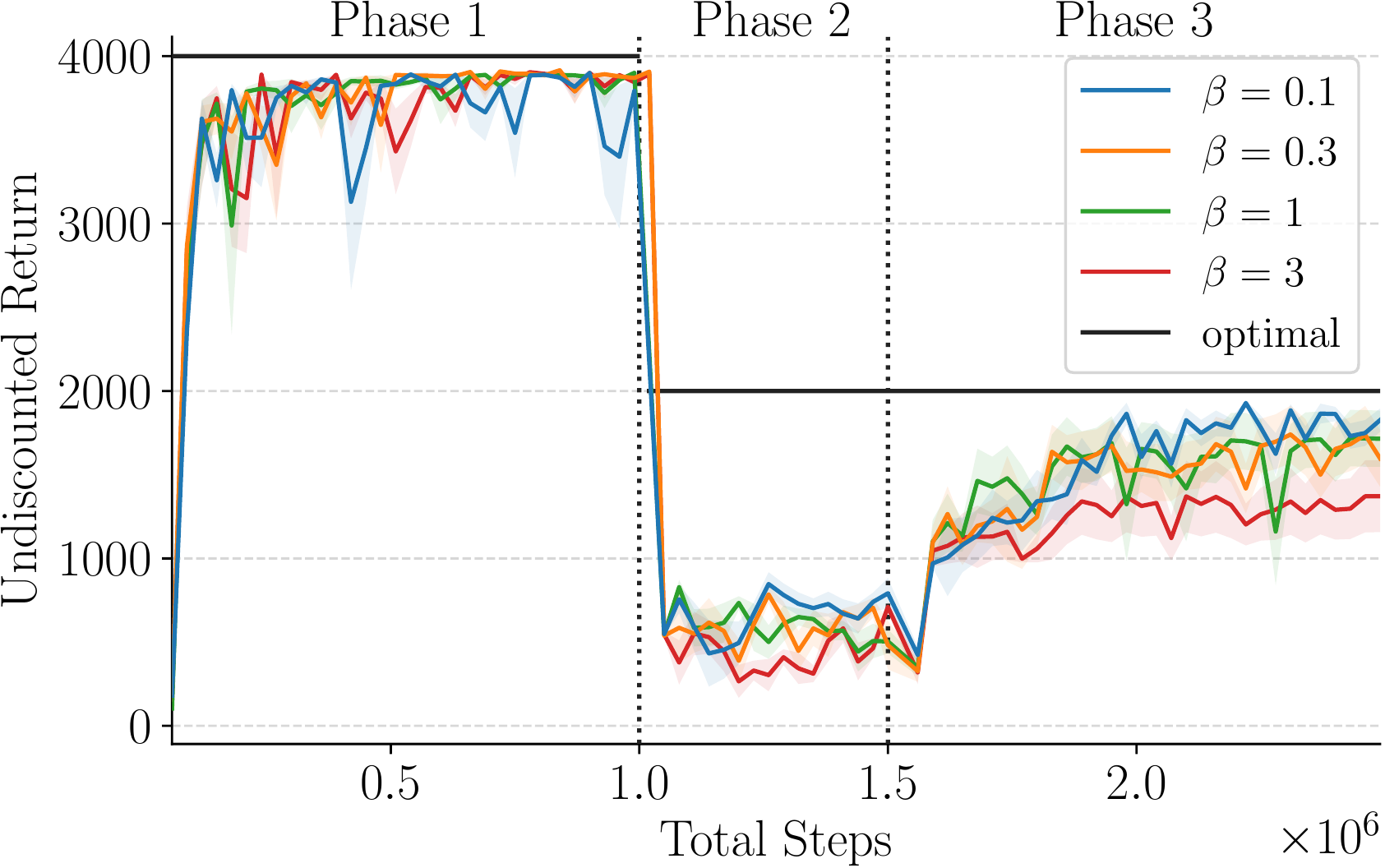}
    \end{subfigure}%
\caption{Plots showing the performance of DreamerV2 with different hyperparameter settings. We searched over the critical hyperparameters: 1) the KL loss scale $\beta$, 2) the actor entropy loss scale $\eta$, and 3) the discount factor $\gamma$. In this figure $\gamma = 0.999$. $\eta = 10^{-3}, 3 \times 10^{-4}, 10^{-4}, 10^{-5}$ in four different rows respectively. The right column corresponds to the buffer reinitializing case, and the left column corresponds to the other case.}
\label{fig: dreamer comp results 2}
\end{figure*}
\begin{figure*}[h!]
\centering
    \begin{subfigure}[h]{.95\textwidth}
        \centering
        \includegraphics[width=0.8\textwidth]{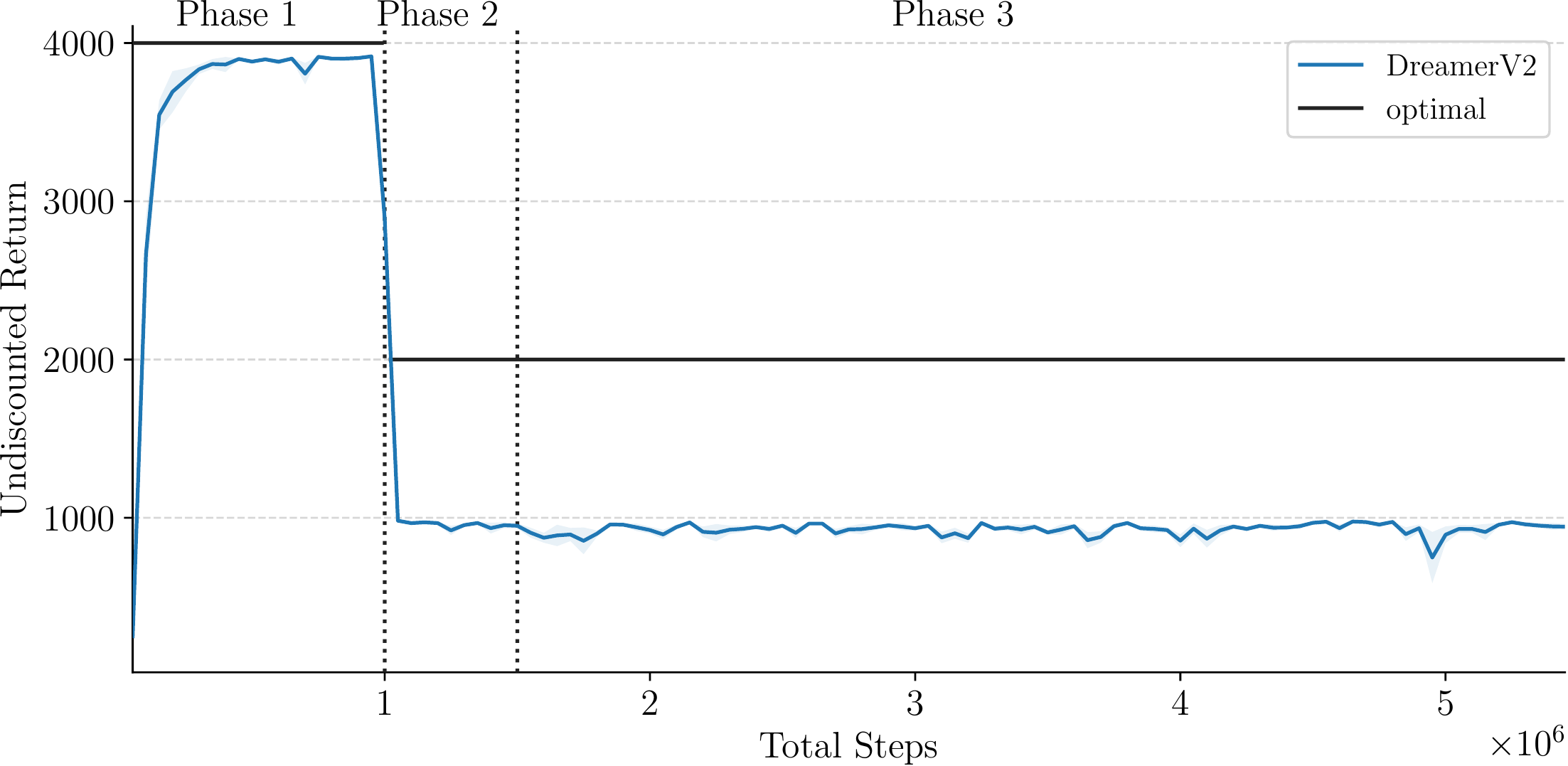}
    \end{subfigure}%

    \caption{Plot showing the learning curve of DreamerV2 without buffer reinitialization. We show the maximum achievable undiscounted reward at each phase as a baseline. To check if the agent's performance could improve through more time, we increased the total environments steps in Phase 3. The curve shows that the agent fails to escape from the sub-optimal target even with much longer training in Phase 3.}
    \label{fig: dreamer no clear buffer results}
\end{figure*}

In the ReacherLoCA domain, if the agent is within the T1-zone and chooses an action that takes it out of the T1-zone, the agent's state remains unchanged. 
In general, T1-zone can be implemented without touching native domain code as long as the domain code offers an API function to set or reload the state. A wrapper around the domain can be implemented that resets the state to the previous one whenever the agent takes an action that moves it out of the T1-zone. 

We tried two variants of the DreamerV2 algorithm. One reinitializes the replay buffer at the beginning of Phase 2 and the other one does not. When initializing (at the beginning of Phase 1) or reinitializing the replay buffer, the agent performs 50 episodes following a uniformly random policy and stores these episodes in the buffer (the same initialization strategy is also adopted in DreamerV2, although fewer number of episodes are produced there). We found that increasing the number of random episodes helps the agent to find the optimal target, not the sub-optimal one. 

We searched over the KL loss scale $\beta \in \{0.1, 0.3, 1, 3\}$, the actor entropy loss scale $\eta \in \{10^{-5}, 10^{-4}, 3 \times 10^{-4}, 10^{-3}\}$, and the discount factor $\gamma \in \{0.99, 0.999\}$. Note that the discount factor tested here is not the problem's discount factor (the problem's discount factor is 1). Instead, it is part of the solution method. 

\subsection{Additional Information about the DreamerV2 Result Shown in the Middle Panel of Figure \ref{fig: deep rl results}}\label{app: Additional Information about the DreamerV2 Result Shown in the Middle Panel of Figure DeepRL Results}

The hyperparameters used to generate the plotted learning curves are chosen in the following way. We first collect the set of hyperparameter settings that, in both Phases 1 and 3, result in average returns (over the last five evaluation runs) being more than 80\% of the optimal average returns. If the set of hyperparameter settings is empty, we chose the setting that achieved the highest average return in Phase 3 (over the last five evaluation runs). Otherwise, from the set of hyperparameter settings, we picked the setting that achieved the highest average return over all evaluation runs in Phase 2. Each point in a learning curve is the average of undiscounted cumulative reward over 5 evaluation episodes. The shading area shows the standard error. 

The best hyperparamerter setting for DreamerV2 with replay buffer reinitialization is $\beta = 0.1$, $\eta= 10^{-4}$, and $\gamma = 0.99$. Without replay buffer reinitialization, none of the runs achieved more than 80\% of optimal average return in Phase 3 over the last five evaluation runs. The setting that achieved the highest performance in Phase 3 is $\beta = 1$, $\eta= 3 \times 10^{-4}$, and $\gamma = 0.99$. 

Figure \ref{fig: dreamer reward estimates} is produced using the best hyperparameter setting with replay buffer reinitialization. The seed used is randomly picked. Furthermore, we have tested other random seeds and hyperparameter settings (not shown in the paper) and obtained similar results as those shown in Figure \ref{fig: dreamer reward estimates}.

\subsection{Additional DreamerV2 Results}\label{app: Additional DreamerV2 Results}
Figure \ref{fig: dreamer comp results 1} and \ref{fig: dreamer comp results 2} show learning curves with different hyperparameter settings for DreamerV2.

Figure \ref{fig: dreamer no clear buffer results} shows DreamerV2's (no reinitialization) performance with longer training in Phase 3. We observe that even with a much longer training in Phase 3, the agent did not escape from the sub-optimal target.

\subsection{Setup of the PlaNet Experiment}\label{app: Experiment Setup of the PlaNet Experiment}

PlaNet uses the cross-entropy method (CEM) for planning. We empirically found that some of the runs would get stuck in the local optima at the end of Phase 1 when using the suggested hyperparameters. Through more experiments, we found that increasing the CEM hyperparameters avoids poor performance in phase 1 for all the random seeds. We increased the horizon length $H$ to 50, optimization iterations $I$ to 25, and candidate samples $J$ to 4000. 

Just like the two tested DreamerV2 variants, the two tested PlaNet variants are one that reinitializes the replay buffer and one that does not. The reinitializing strategy is the same as what we apply to DreamerV2 shown in Section \ref{app: Experiment Setup of the DreamerV2 Experiment}.

We searched over the KL-divergence scale $\beta \in \{0, 1, 3, 10\}$, and the step-size $\in \{10^{-4}, 3 \times 10^{-4}, 10^{-3}\}$.

\subsection{Additional Information about the PlaNet Result Shown in the Right Panel of Figure \ref{fig: deep rl results}}\label{app: Additional Information about the PlaNet Result Shown in the Right Panel of Figure DeepRL Results}

The best hyperparameter setting for PlaNet with buffer reinitialization is $\beta = 1$, and the step-size $=10^{-3}$. For PlaNet without buffer reinitialization, similar to DreamerV2, none of the runs achieved more than 80\% of optimal average return in Phase 3. The setting that achieved the highest average return in Phase 3 is $\beta = 0$, and the step-size $=10^{-3}$.

\subsection{Additional PlaNet Results}\label{app: Additional PlaNet Results}

\begin{figure*}[h!]
\centering
    \begin{subfigure}{.4\textwidth}
        \centering
        \includegraphics[width=0.95\textwidth]{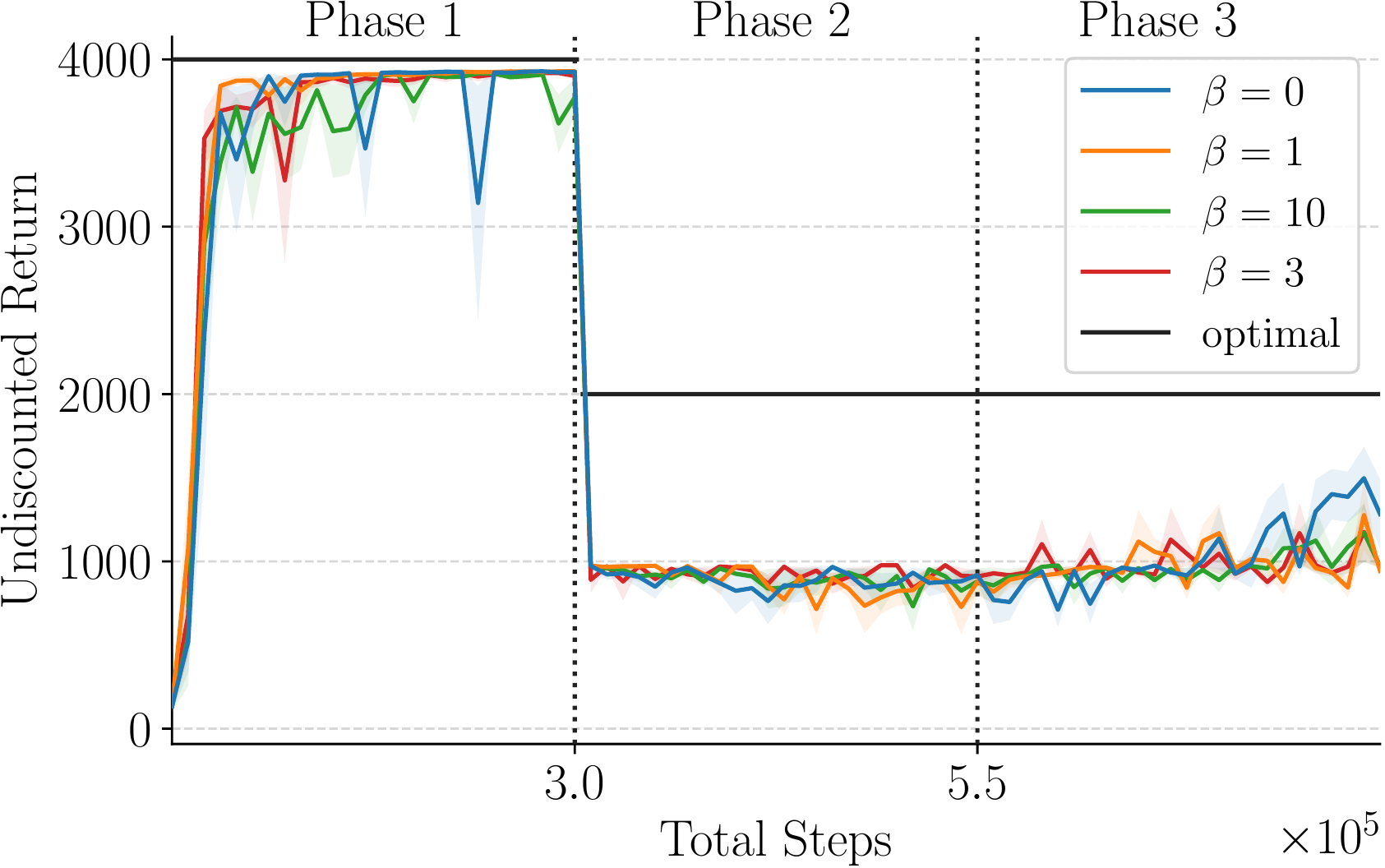}
    \end{subfigure}
    \begin{subfigure}{.4\textwidth}
        \centering
        \includegraphics[width=0.95\textwidth]{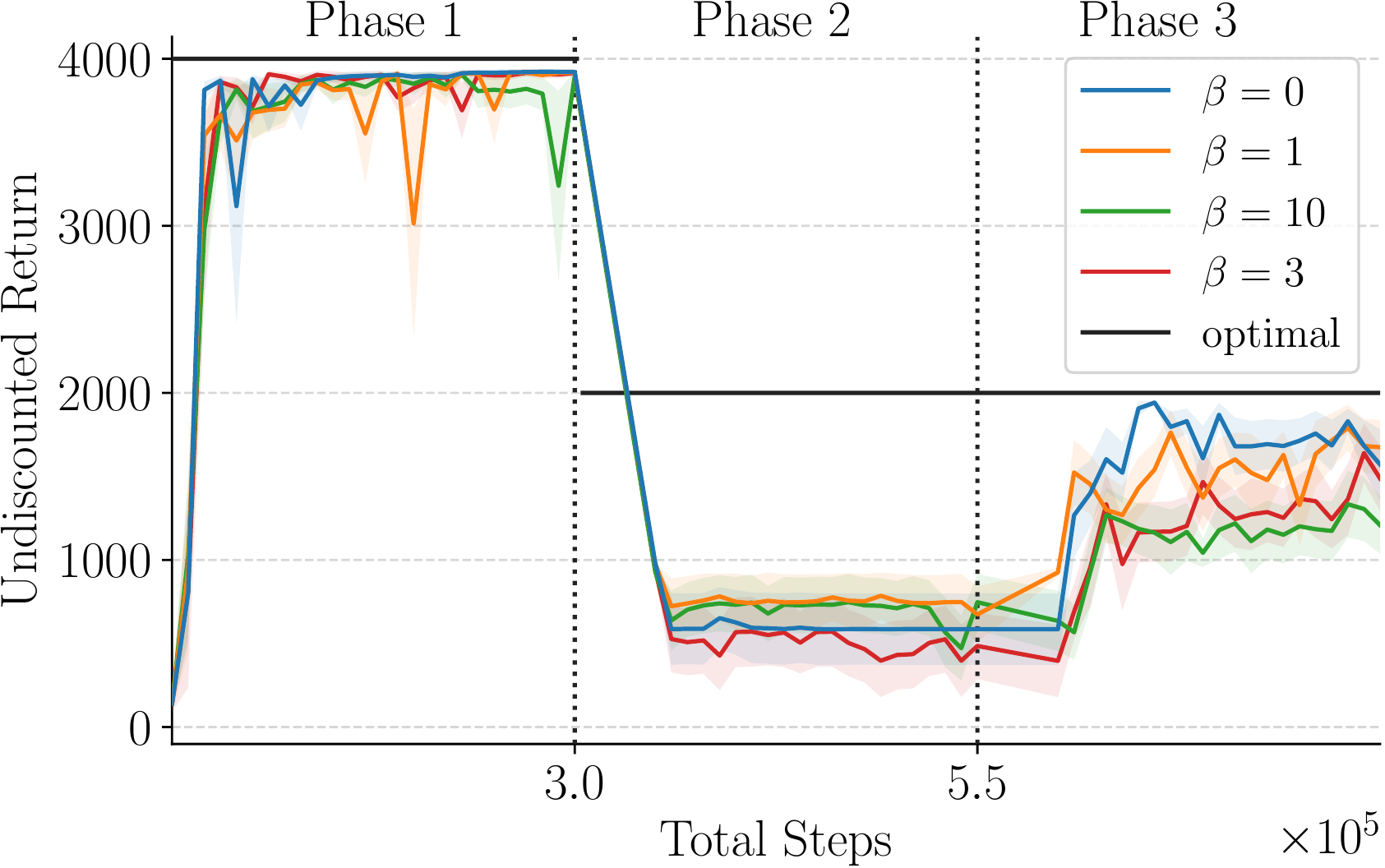}
    \end{subfigure}
    
    \begin{subfigure}{.4\textwidth}
        \centering
        \includegraphics[width=0.95\textwidth]{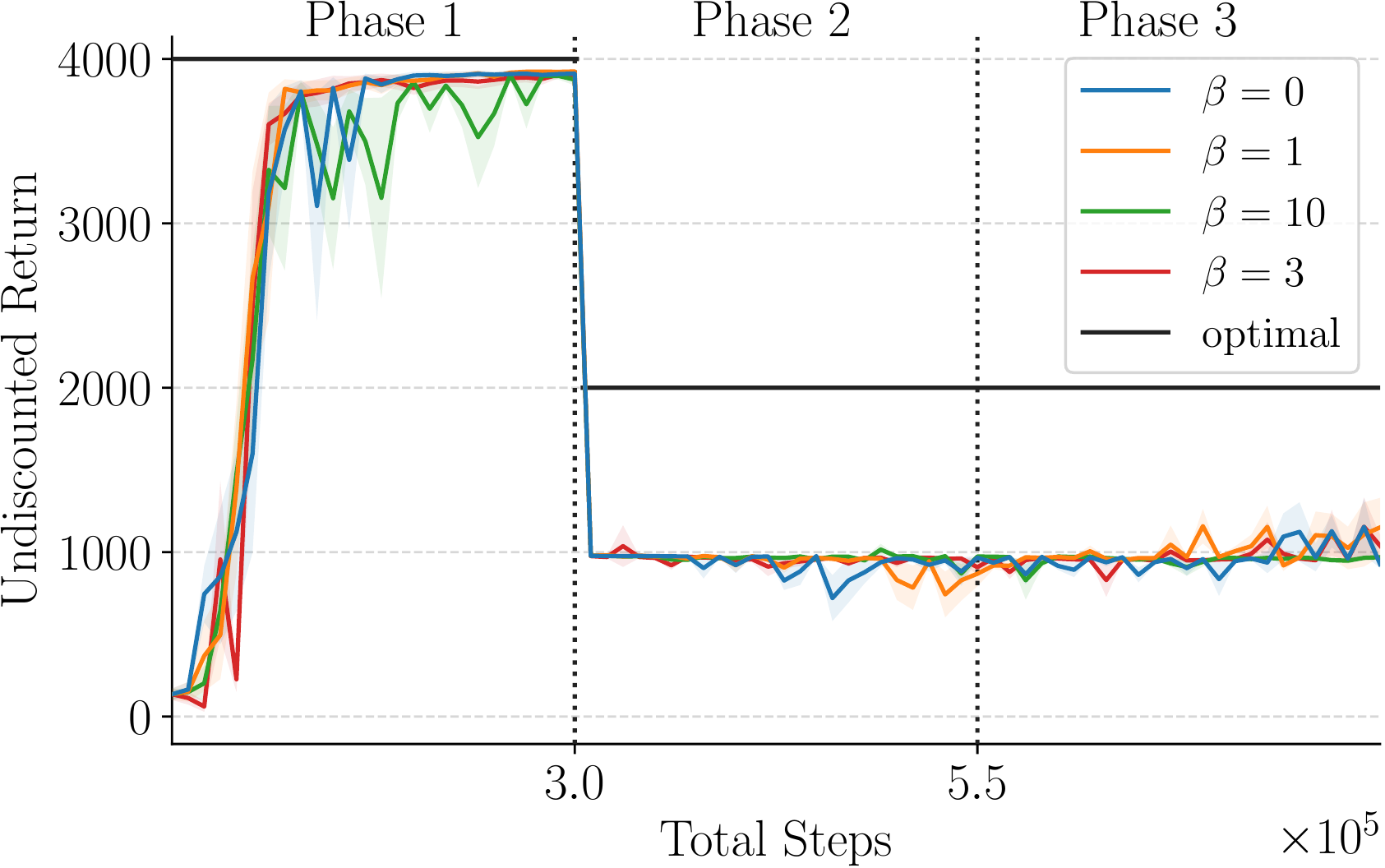}
    \end{subfigure}
    \begin{subfigure}{.4\textwidth}
        \centering
        \includegraphics[width=0.95\textwidth]{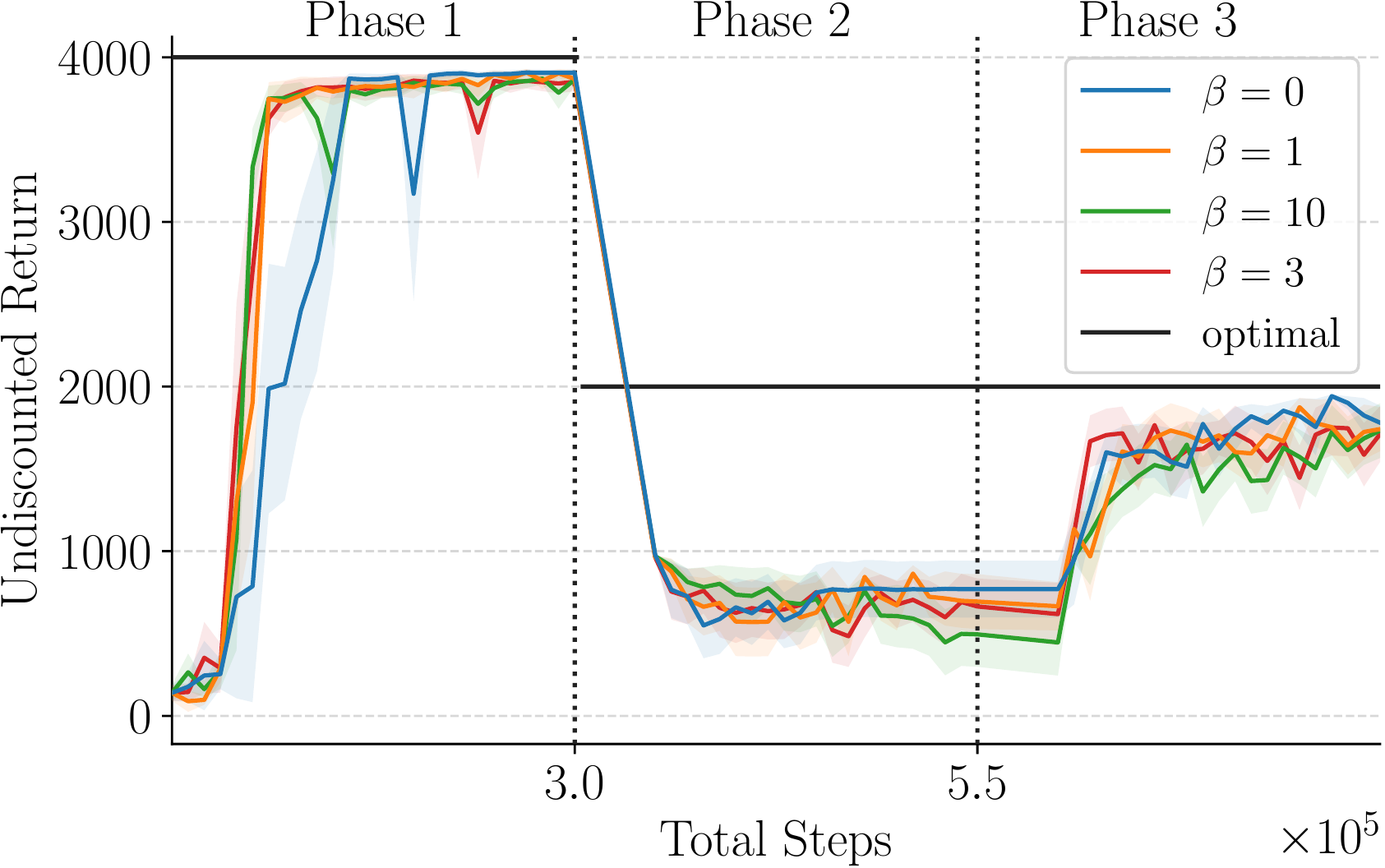}
    \end{subfigure}
    
    \begin{subfigure}{.4\textwidth}
        \centering
        \includegraphics[width=0.95\textwidth]{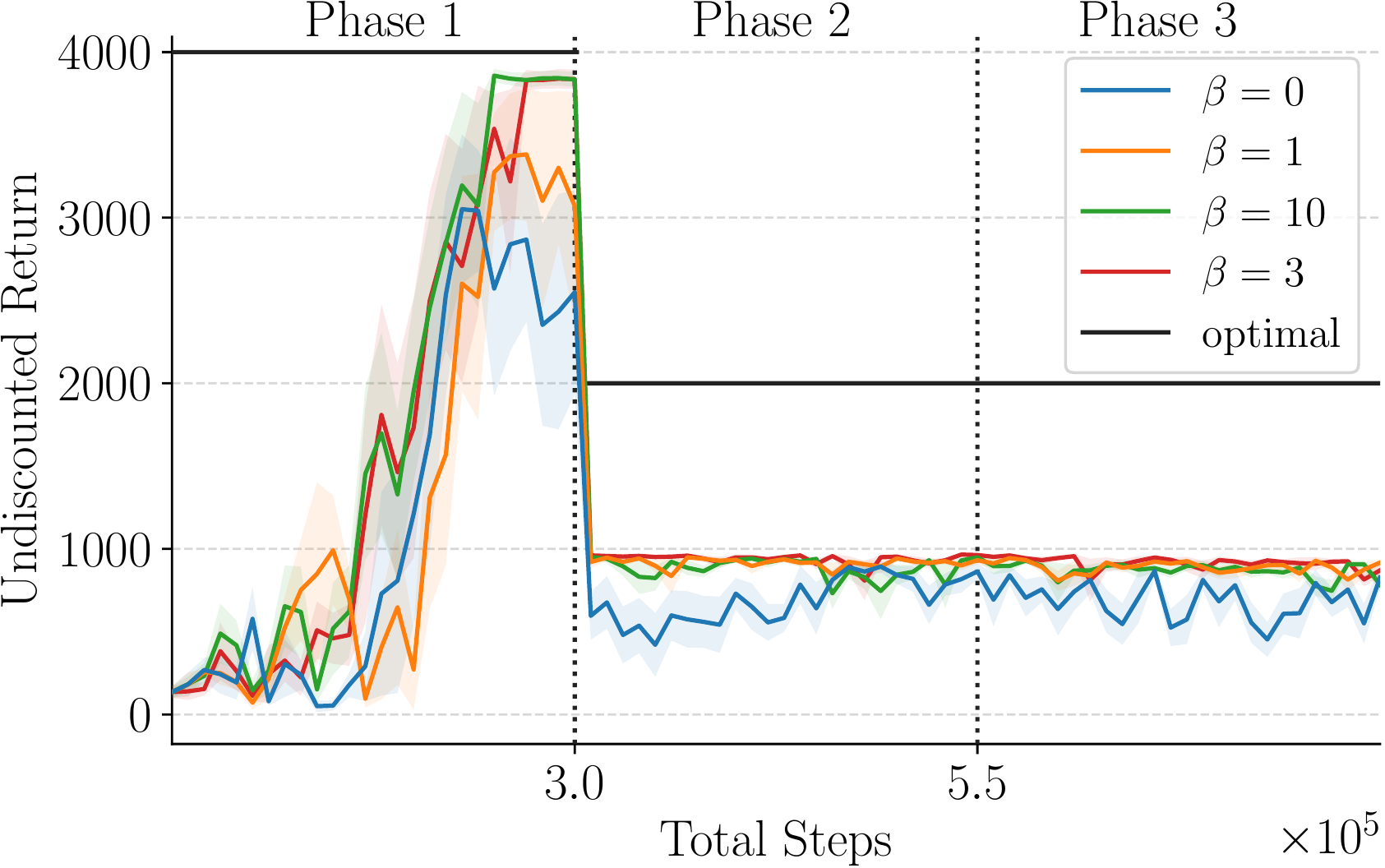}
    \end{subfigure}
    \begin{subfigure}{.4\textwidth}
        \centering
        \includegraphics[width=0.95\textwidth]{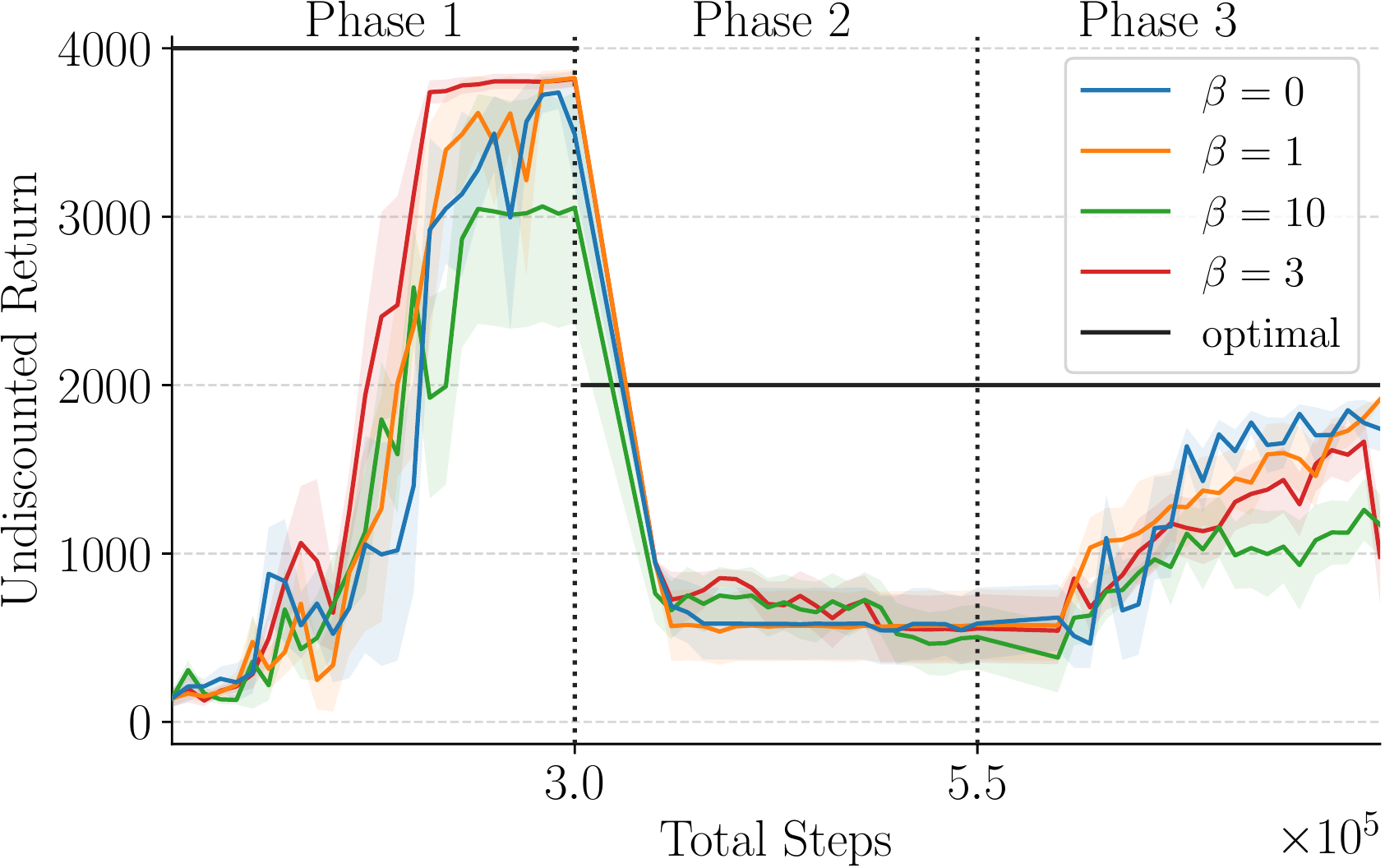}
    \end{subfigure}
\caption{Plots showing the performance of DreamerV2 with different hyperparameter settings.  We searched over the critical hyperparameters: 1) the KL-divergence scales $\beta$, and 2) the step size. The step size is $10^{-3}, 3 \times 10^{-4}, 10^{-4}$ in three different rows respectively. The right column corresponds to the buffer reinitializing case, and the left column corresponds to the other case.}
\label{fig: planet comp results}
\end{figure*}

Figure \ref{fig: planet comp results} shows learning curves with different hyperparameter settings for PlaNet.

\begin{figure*}[h!]
\centering
\begin{subfigure}[b]{.2\textwidth}

        \centering
        \includegraphics[width=0.99\textwidth]{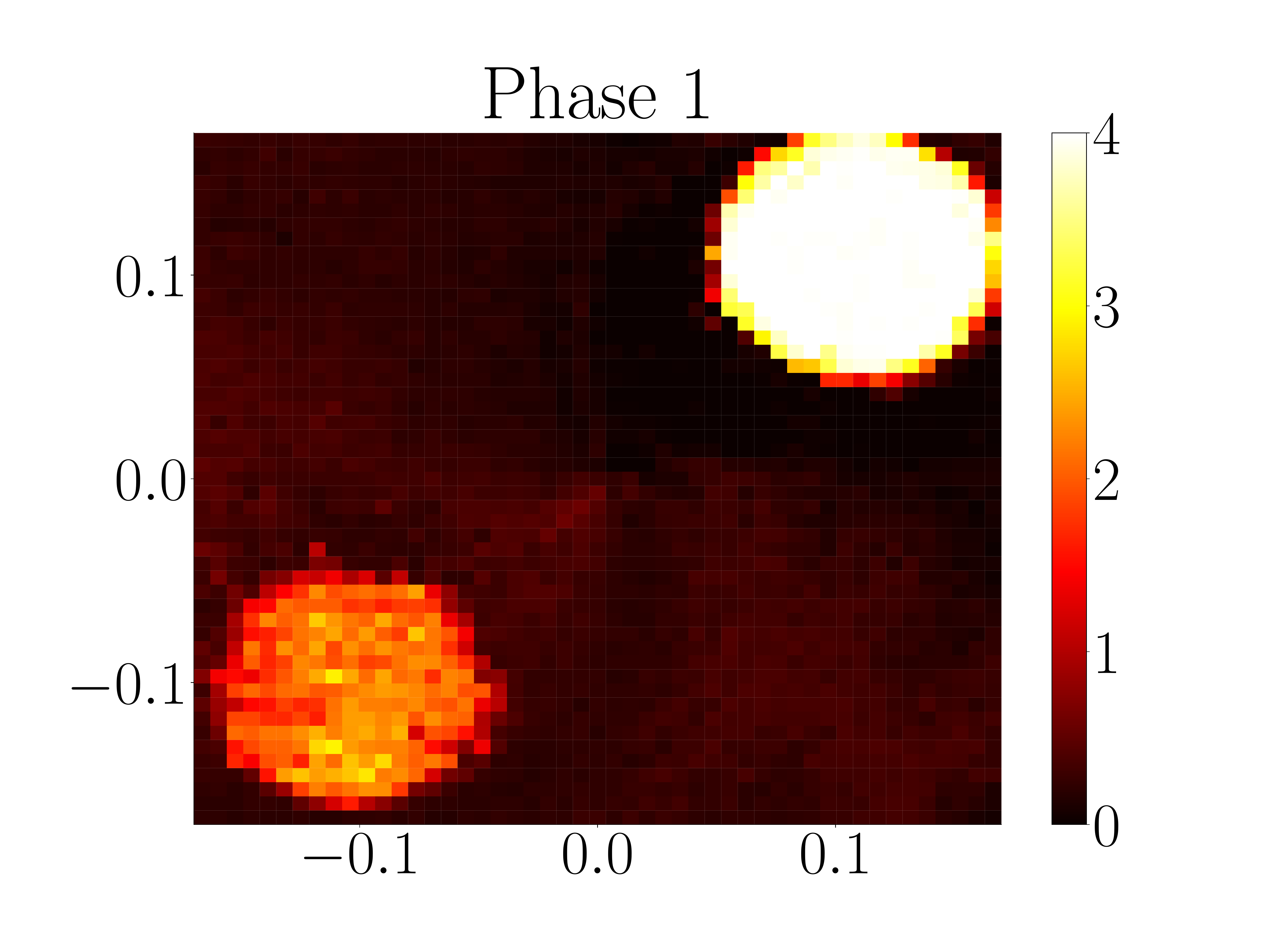}
    \end{subfigure}%
\begin{subfigure}[b]{0.2\textwidth}
        \centering
        \includegraphics[width=0.99\textwidth]{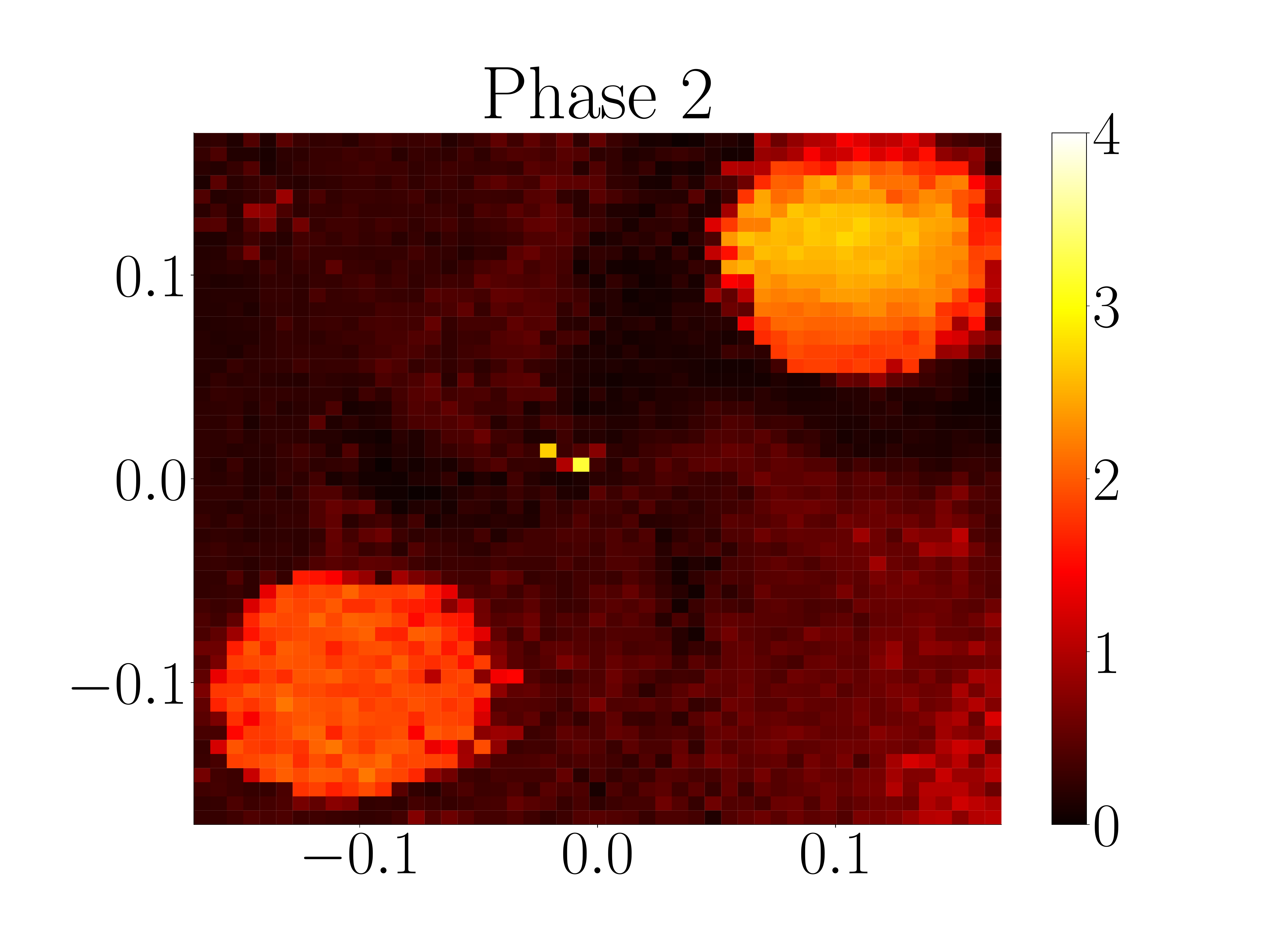}
    \end{subfigure}%
    \begin{subfigure}[b]{0.2\textwidth}
        \centering
        \includegraphics[width=0.99\textwidth]{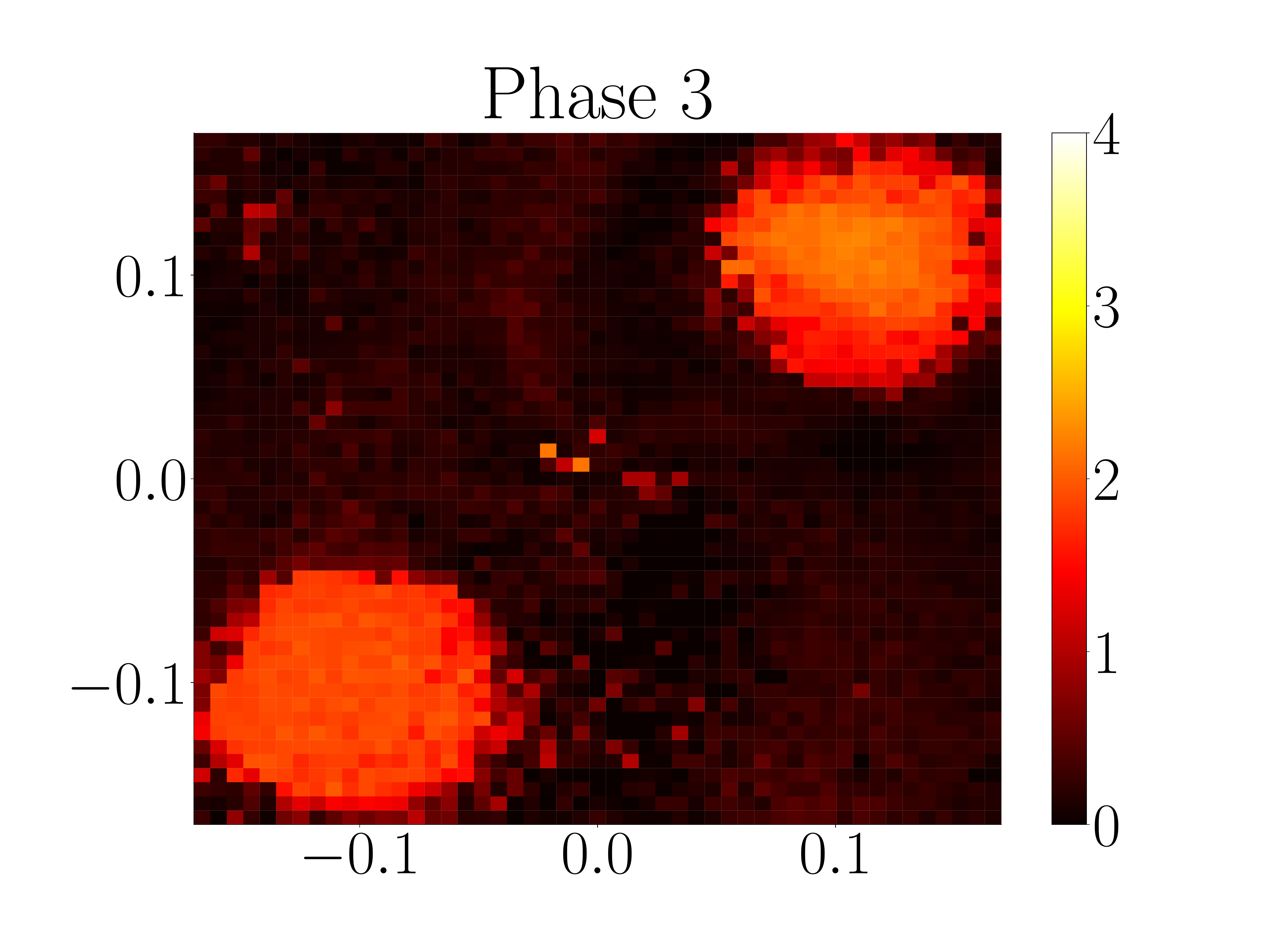}
    \end{subfigure}%
    \begin{subfigure}[b]{.2\textwidth}
        \centering
        \includegraphics[width=0.99\textwidth]{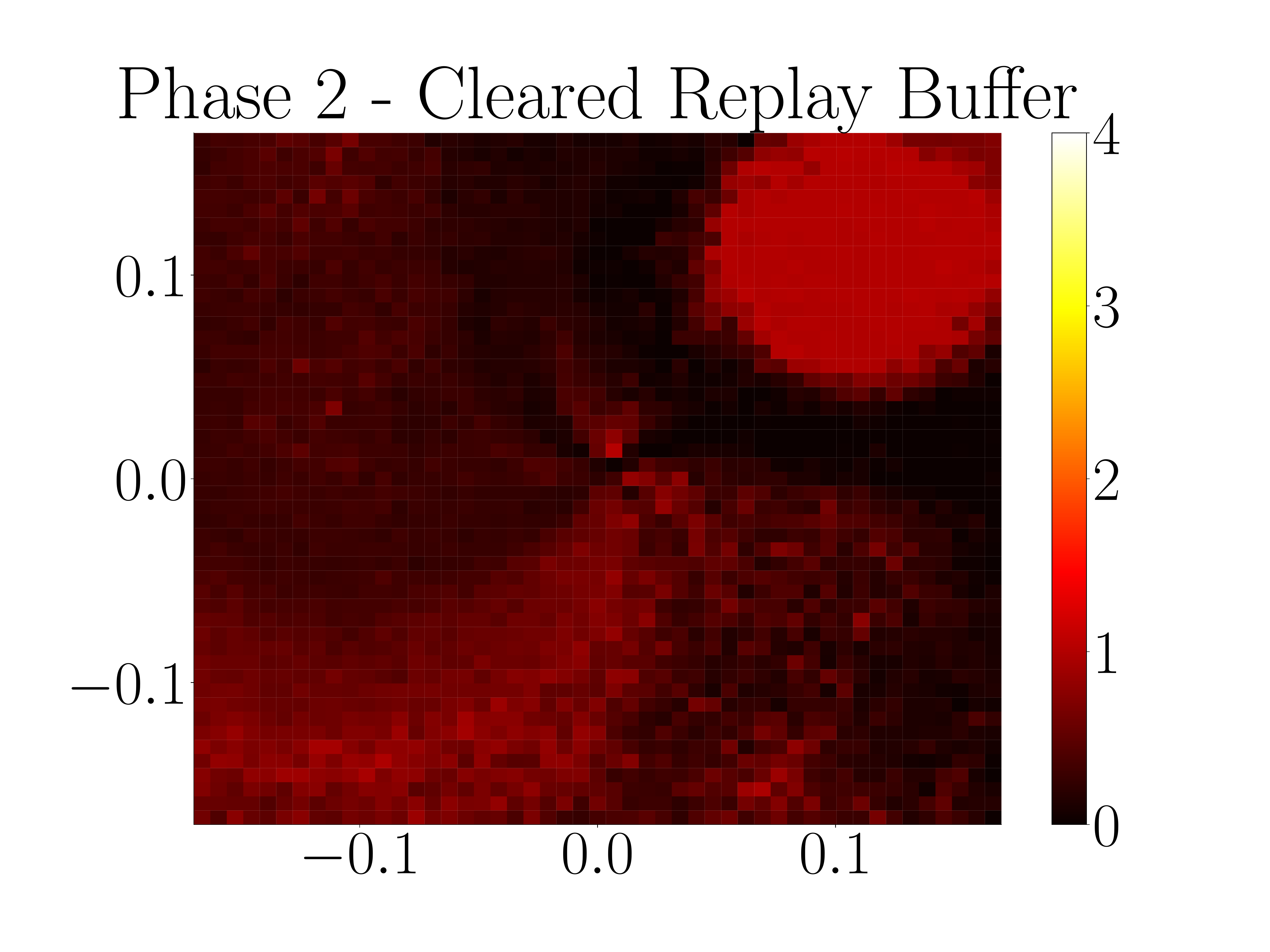}
    \end{subfigure}%
    \begin{subfigure}[b]{0.2\textwidth}
        \centering
        \includegraphics[width=0.99\textwidth]{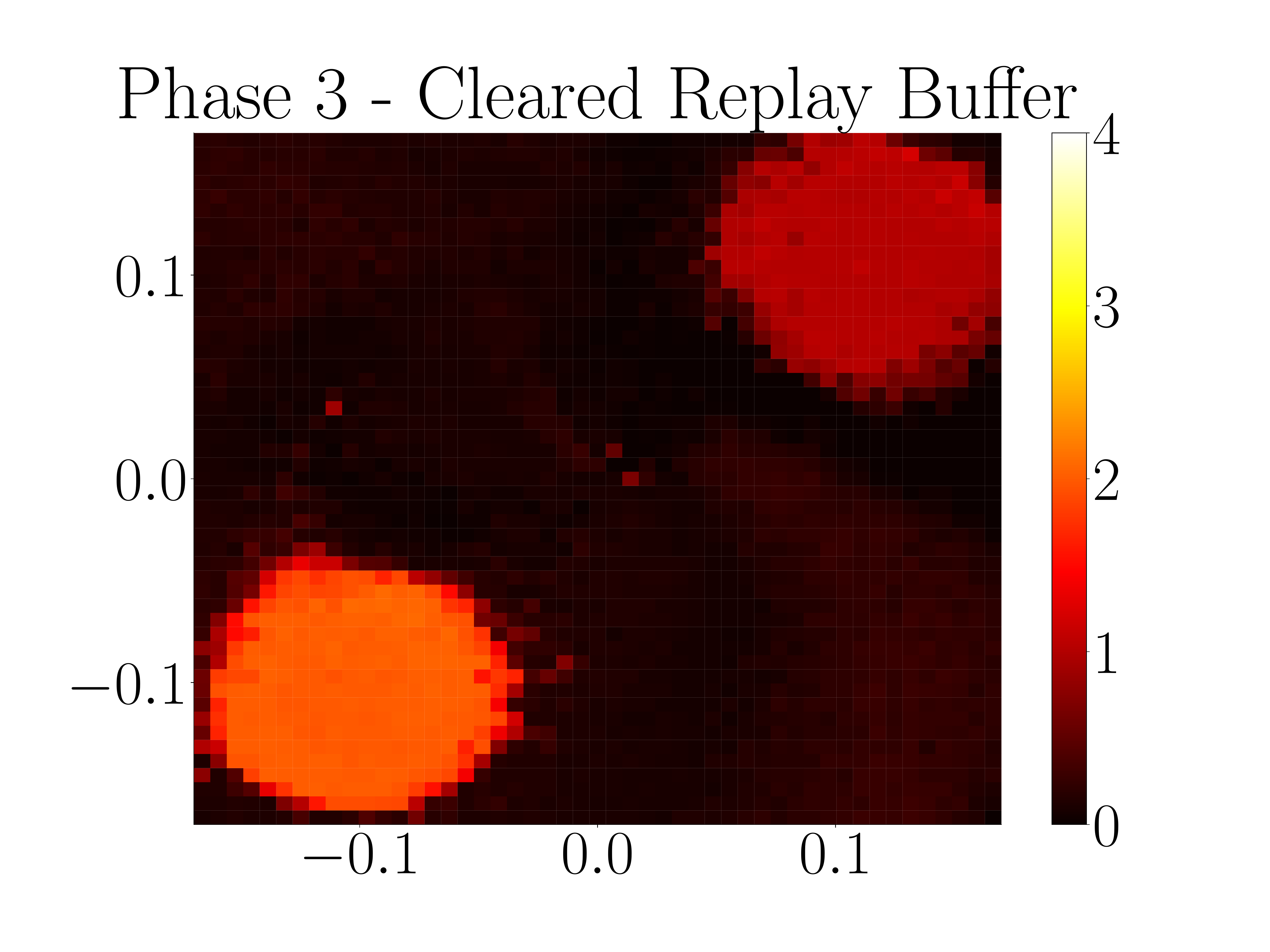}
    \end{subfigure}%
\caption{Visualization of the PlaNet agent's estimated reward model generated at the end of each phase. The $x$ and $y$ axes of each heatmap represent the agent's position in the ReacherLoCA domain. These heatmaps show that in Phase 2 PlaNet incorrectly estimate the reward model regardless of the buffer being reinitialized or not.}
\label{fig: planet reward estimates}
\end{figure*}

We also plotted the reward estimations of PlaNet's world model at the end of each phase in Figure \ref{fig: planet reward estimates}. The results are similar to those of DreamerV2's (see Figure \ref{fig: dreamer reward estimates}).

\newpage
\newpage
\section{Supplementary Details and Results for Experiments in Section\ \ref{sec: Dyna}} \label{app: Additional Experiment Setup Details}

\subsection{Setup of the Adaptive Linear Dyna Experiment}
\label{app: Setups for the Adaptive Linear Dyna Experiment}

% \subsection{The MountainCarLoCA Domain}\label{sec: The MountainCarLoCA Domain}

We tested our algorithm on a variant of the MountainCar domain. Our variant of the MountainCar domain extends the variant of the same domain used by \citet{vanseijen-LoCA}, by introducing stochasticity. Specifically, with probability $1 - p$, the agent's action is followed and with probability $p$, the state evolves just as a random action was taken. We tested $p \in [0, 0.3, 0.5]$. The discount factor is 0.99. 
% Other details of the experiment setup are summarized in Section \ref{app: Setups for the Adaptive Linear Dyna Experiment}.
The MountainCarLoCA domain has two terminal states, one of which locates at the top of the mountain (position $> 0.5$, velocity $> 0$) and the other one locates at the valley $((\text{position} + 0.52)^2 + 100 * \text{velocity}^2) <= 0.07^2$). The dynamics of the environment is described in Section \ref{app: Setups for the Adaptive Linear Dyna Experiment}. The small region where the agent is initialized during evaluation is roughly at the middle of the two terminal states ($-0.2 \leq$ position $\leq -0.1$, $-0.01 \leq$ velocity $\leq 0.01$).

The linear Dyna algorithm we tested is basically Algorithm 4 by \citet{sutton2012dyna}. However, we found that Algorithm 4 takes too much memory and computation, making it difficult for testing. To reduce the memory and computation usage, the tested linear Dyna algorithm slightly modifies Algorithm 4 in the following two ways. First, while Algorithm 4 does not specify the maximum size of the priority queue, our algorithm sets that maximum to be 400000 (maximum training steps) and removes 20\% oldest elements in the queue whenever the queue is full. Second, in each planning step, Algorithm 4 updates for all predecessor features of the first element in the priority queue and add all predecessor features to the priority queue, our algorithm only updates 5 randomly chosen predecessor features and adds them to the queue. These two modifications we made significantly reduce memory usage and computation. 

We summarize in Table \ref{tab: Linear Dyna hyperparameters.} tested hyperparameters for linear Dyna and adaptive linear Dyna.

\begin{table}[h!]
    \centering
    \begin{tabular}{c|c}
    \hline
        Number of Tilings and Tiling Size & [3, 18 by 18; 5, 14 by 14; 10, 10 by 10] \\
        Exploration parameter $\epsilon$ & [0.1, 0.5, 1.0]\\
        Stepsize $\alpha$ & [0.001, 0.005, 0.01, 0.05, 0.1] / number of tilings\\
        Model Stepsize $\beta$ & [0.001, 0.005, 0.01, 0.05, 0.1] / number of tilings\\
        Number of planning steps & 5\\
        Size of the planning buffer $n_p$ (adaptive linear Dyna) & [10000, 200000, 4000000]\\
    \hline
    \end{tabular}
    \caption{Tested hyperparameters in the adaptive linear Dyna experiment.}
    \label{tab: Linear Dyna hyperparameters.}
\end{table}

For Sarsa($\lambda$), the tile coding setups and the choices of stepsize $\alpha$ are the same as those used for adaptive linear Dyna. The exploration parameter was fixed to be 0.1. We also tested three $\lambda$ values: 0, 0.5, and 0.9. 

Table \ref{tab: Experiment Procedure Linear Dyna and Sarsa lambda} shows the experiment setup used to test linear Dyna, adaptive linear Dyna (Algorithm \ref{algo:adaptive linear dyna}), and Sarsa($\lambda$).

\begin{table}[h!]
    \centering
    \begin{tabular}{c|c|c}
        \hline
        \multirow{4}{5em}{Initial distributions} & Phase 1 training  & Uniform distribution over the entire state space\\
        & Phase 1 evaluation  & Uniform distribution over a small region\\
        & Phase 2 training  & Uniform distribution over states within T1-zone\\
        & Phase 2 evaluation  & Uniform distribution over a small region\\
        \hline
        \multirow{2}{5em}{Training steps} & Phase 1 steps & $2 \times 10^6$\\
        & Phase 2 steps & $2 \times 10^6$\\
        \hline
        \multirow{3}{5em}{Other details} & Training steps between two evaluations & $10^4$\\
        & Number of runs & 10\\
        & Number of evaluation episodes & 10\\
    \hline
    \end{tabular}
    \caption{Experiment setup used to test linear Dyna, adaptive linear Dyna, and Sarsa($\lambda$).}
    \label{tab: Experiment Procedure Linear Dyna and Sarsa lambda}
\end{table}

\subsection{Additional Information about the Result Shown in Figure \ref{fig: linear dyna}} \label{app: Additional Information of Result Shown in Figure Linear Dyna}

The hyperparameter setting chosen to draw curves in Figure \ref{fig: linear dyna} was chosen in the following way. We first collected hyperparameter settings that achieved 90\% of the optimal average return over the last 5 evaluations in Phase 1. From these hyperparameter settings, we picked one that achieved the highest average return over the last 5 evaluations in Phase 2 and drew its learning curve. For stochasticity = 0.5, no hyperparameter setting for Sarsa($\lambda$) achieved 90\% optimal performance in the first phase and we picked the one that achieved more than 80\% optimality in the first phase.

We summarize in Table\ \ref{tab: best parameter settings in linear experiments} the best hyperparameter settings under different level of stochasticities for each algorithm tested in Section\ \ref{sec: Linear Dyna Experiments}.
\begin{table}[h!]
    \centering
    \begin{tabular}{c|c|c|c|c}
        \hline
        Algorithm & Hyperparameter & Deterministic & Stochasticity = 0.3 & Stochasticity = 0.5\\
        \hline
        \multirow{5}{10em}{Adaptive linear Dyna} & Exploration parameter ($\epsilon$) & 0.5 & 1.0 & 1.0\\
        & Step-size ($\alpha$) & 0.05 & 0.05 & 0.1\\
        & Step-size ($\beta$) & 0.01 & 0.01 & 0.01\\
        & Number of tilings & 5 & 5 & 5\\
        & Size of the planning buffer & 4000000 & 4000000 & 4000000\\
        \hline
        \multirow{3}{10em}{Sarsa($\lambda$)} 
        & Step-size ($\alpha$) & 0.1 & 0.05 & 0.05\\
        & $\lambda$ & 0.5 & 0.5 & 0.9\\
        & Number of tilings & 10 & 10 & 10\\
        \hline
        \multirow{4}{10em}{Linear Dyna} & Exploration parameter ($\epsilon$) & 0.5 & &\\
        & Step-size ($\alpha$) & 0.5 & &\\
        & Number of tilings & 5 & &\\
        \hline
    \end{tabular}
    \caption{Best hyperparameter settings in the adaptive linear Dyna experiment.}
    \label{tab: best parameter settings in linear experiments}
\end{table}

\subsection{Additional Results of the Adaptive Linear Dyna Experiment} \label{app: Additional Results of the Adaptive Linear Dyna Experiment}

\begin{figure*}[h!]
\centering
\begin{subfigure}{.33\textwidth}
    \centering
    \includegraphics[width=0.95\textwidth]{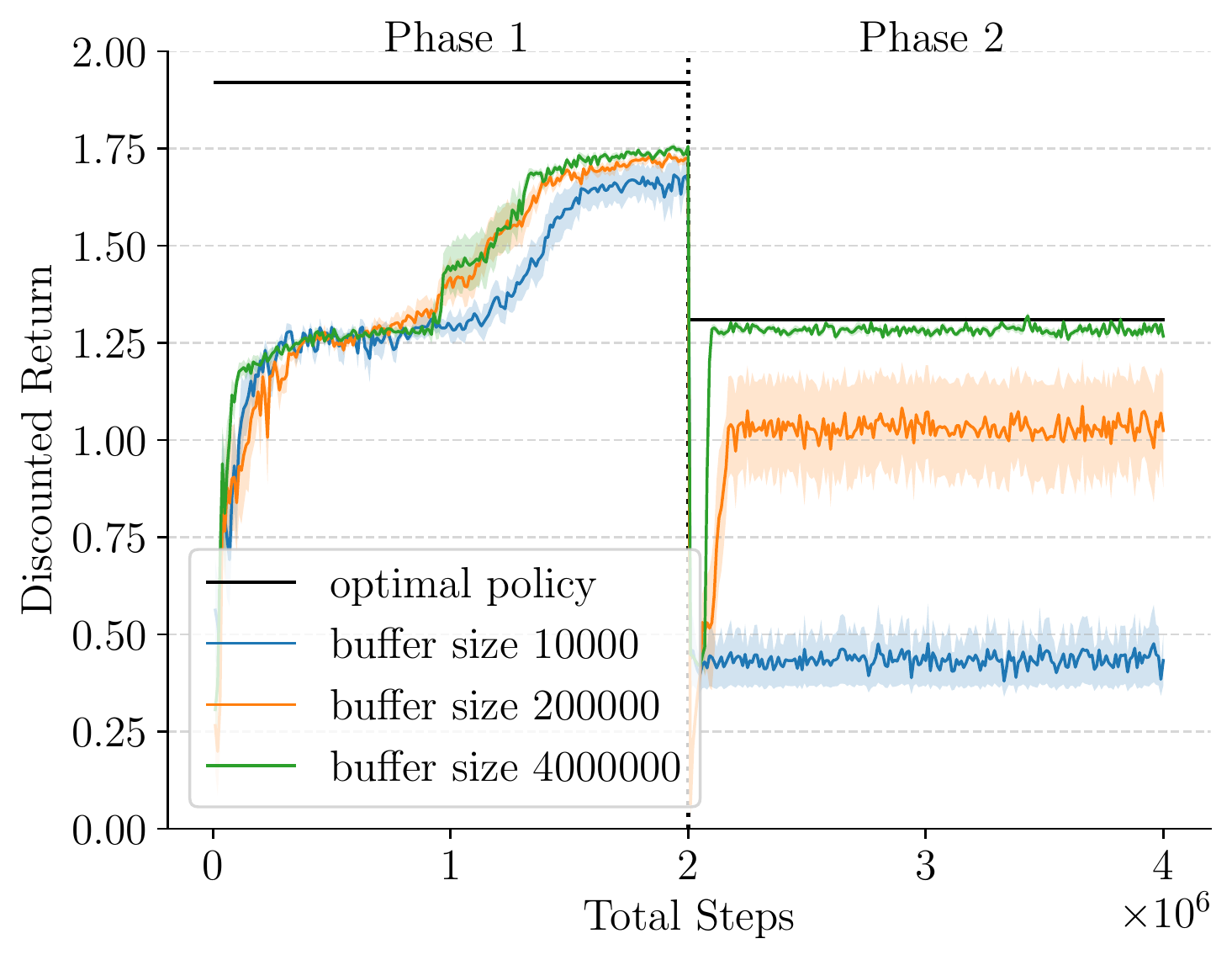}
    \caption{Buffer size}
\end{subfigure}%
\begin{subfigure}{.33\textwidth}
    \centering
    \includegraphics[width=0.95\textwidth]{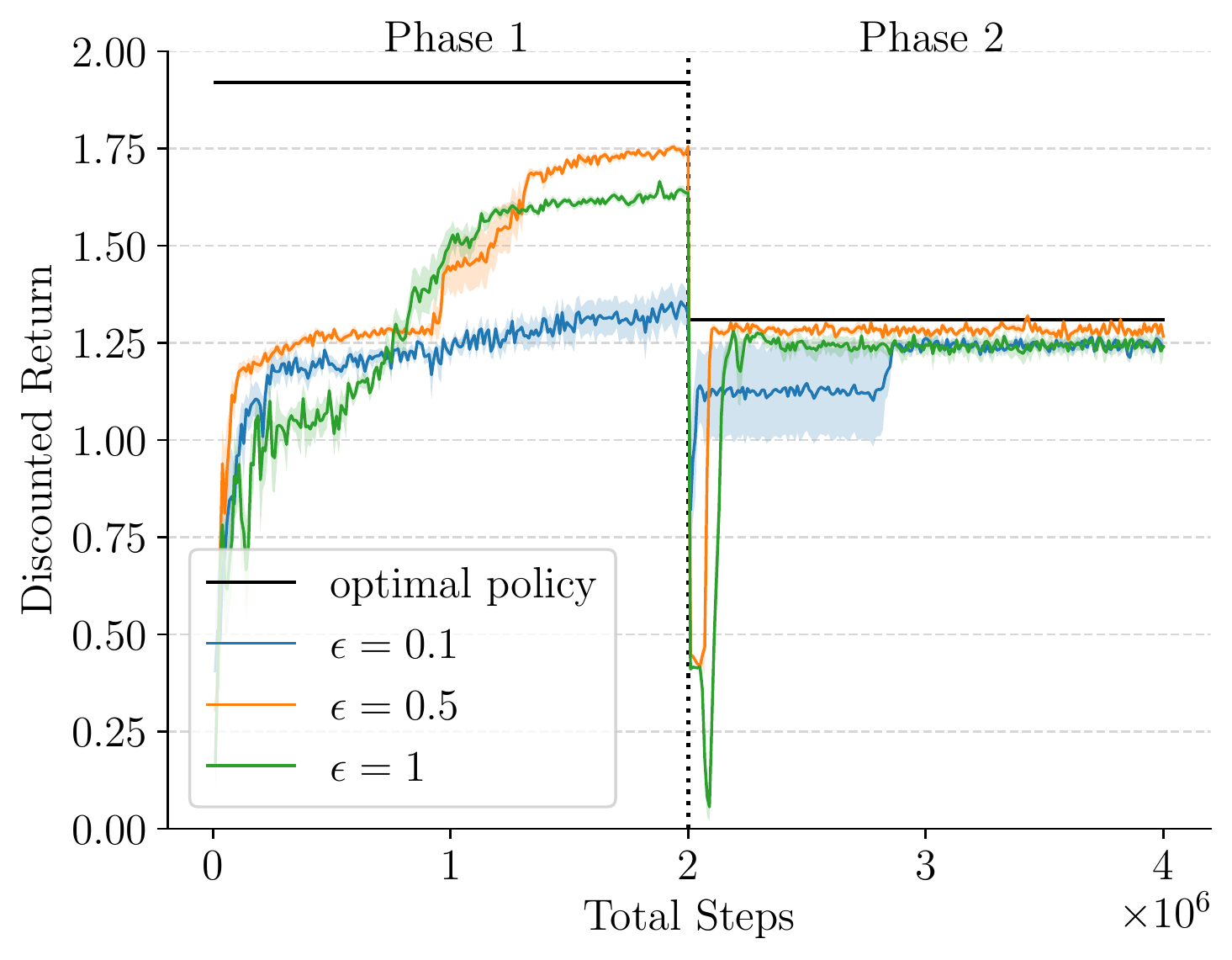}
    \caption{$\epsilon$}
\end{subfigure}%
\begin{subfigure}{.33\textwidth}
    \centering
    \includegraphics[width=0.95\textwidth]{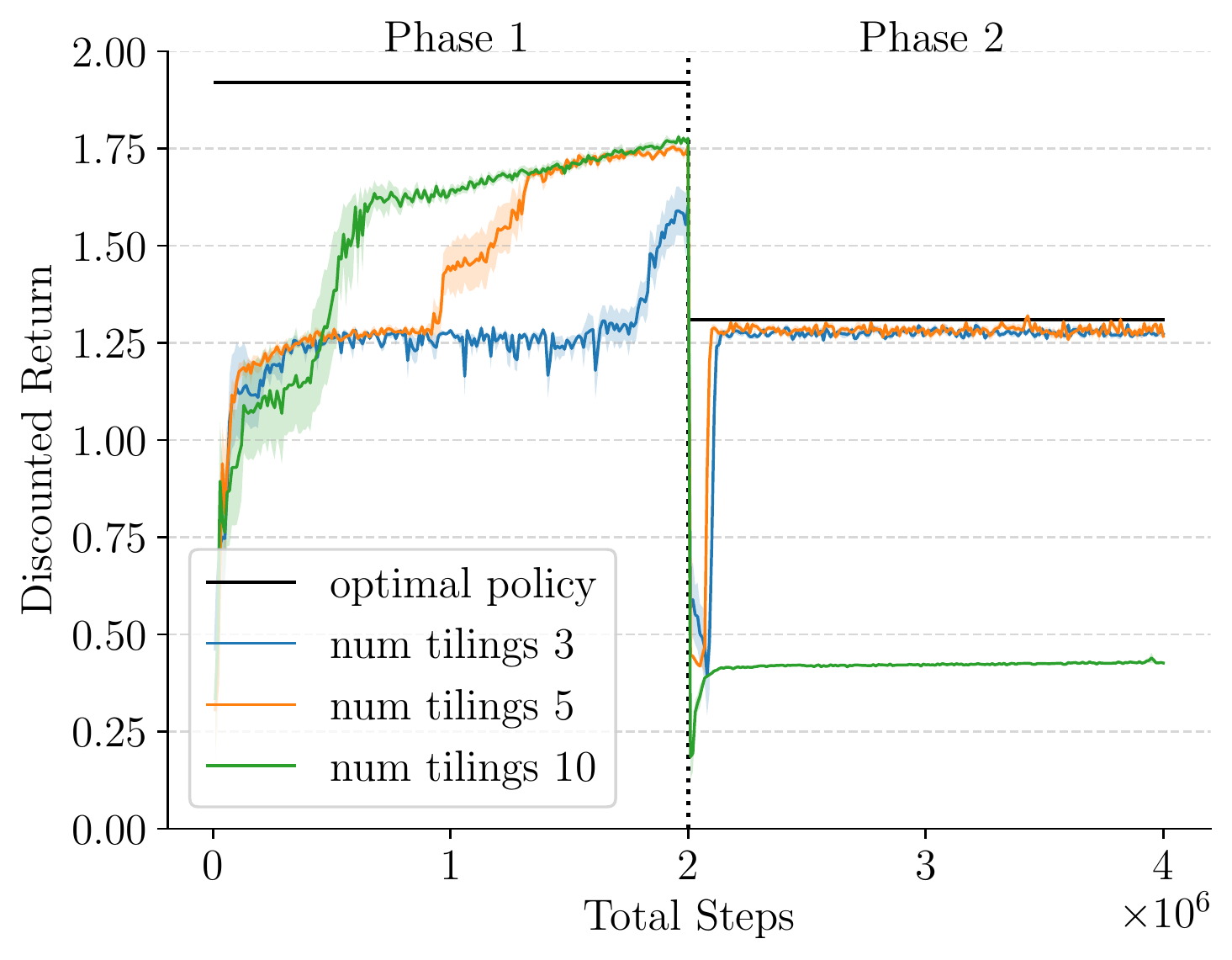}
    \caption{Number of Tilings}
\end{subfigure}\\
\begin{subfigure}{.33\textwidth}
    \centering
    \includegraphics[width=0.95\textwidth]{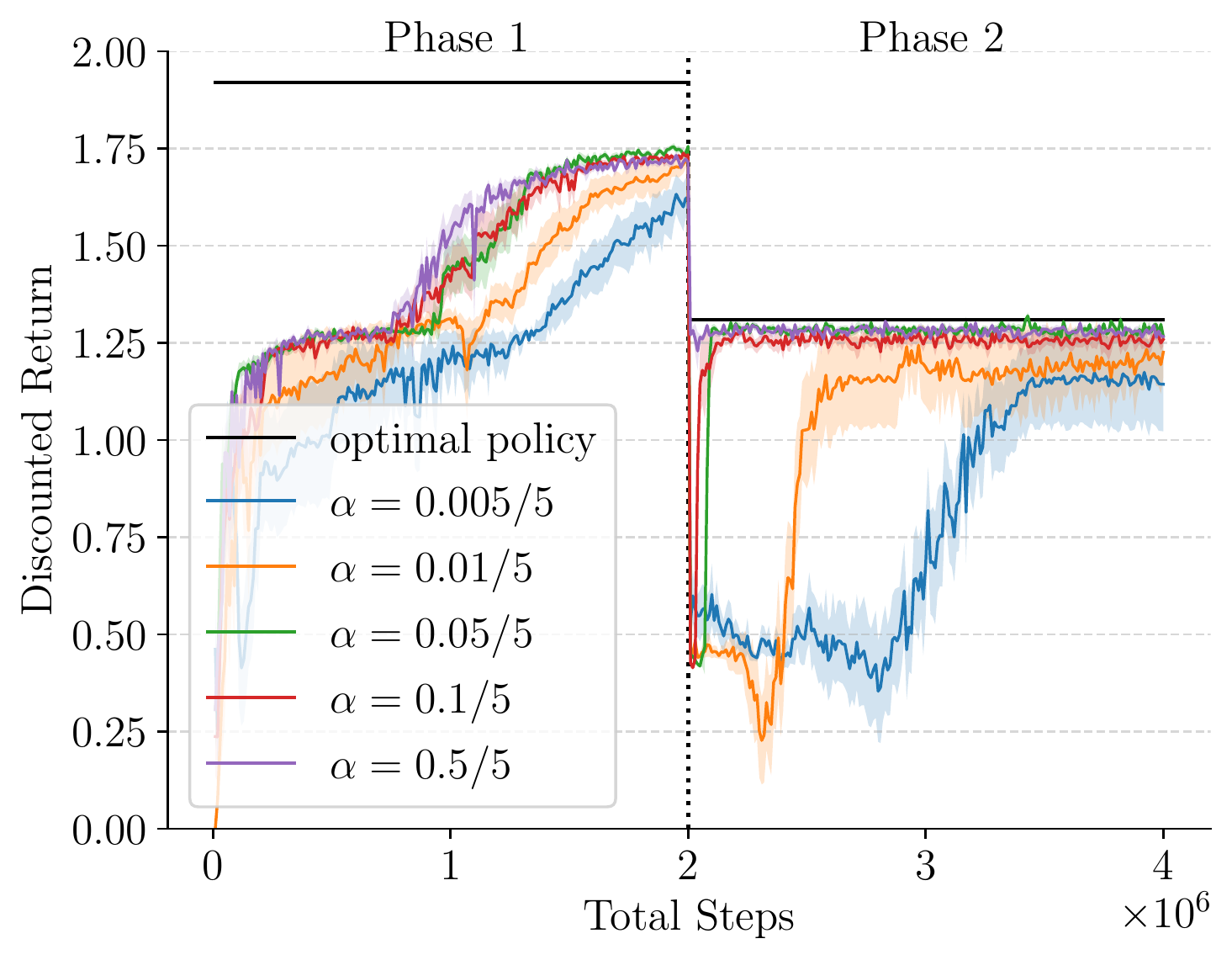}
    \caption{Stepsize}
\end{subfigure}%
\begin{subfigure}{.33\textwidth}
    \centering
    \includegraphics[width=0.95\textwidth]{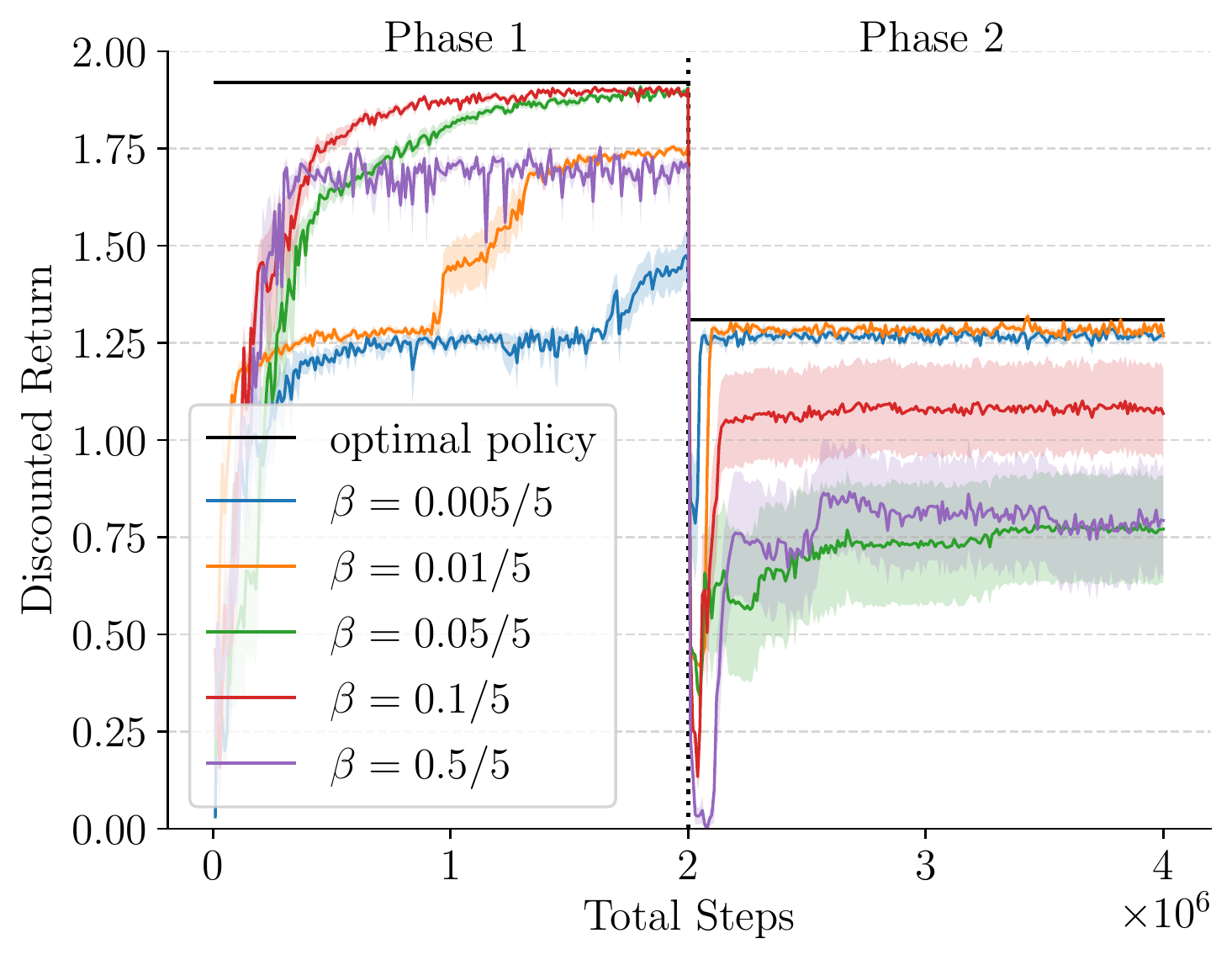}
    \caption{Model Stepsize}
\end{subfigure}%
\caption{Plots showing the sensitivity of the adaptive linear Dyna algorithm to its hyperparameters in the deterministic MountainCarLoCA  domain. 
}
\label{fig: linear dyna stoch 0}
\end{figure*}

\begin{figure*}[h!]
\centering
\begin{subfigure}{.33\textwidth}
    \centering
    \includegraphics[width=0.95\textwidth]{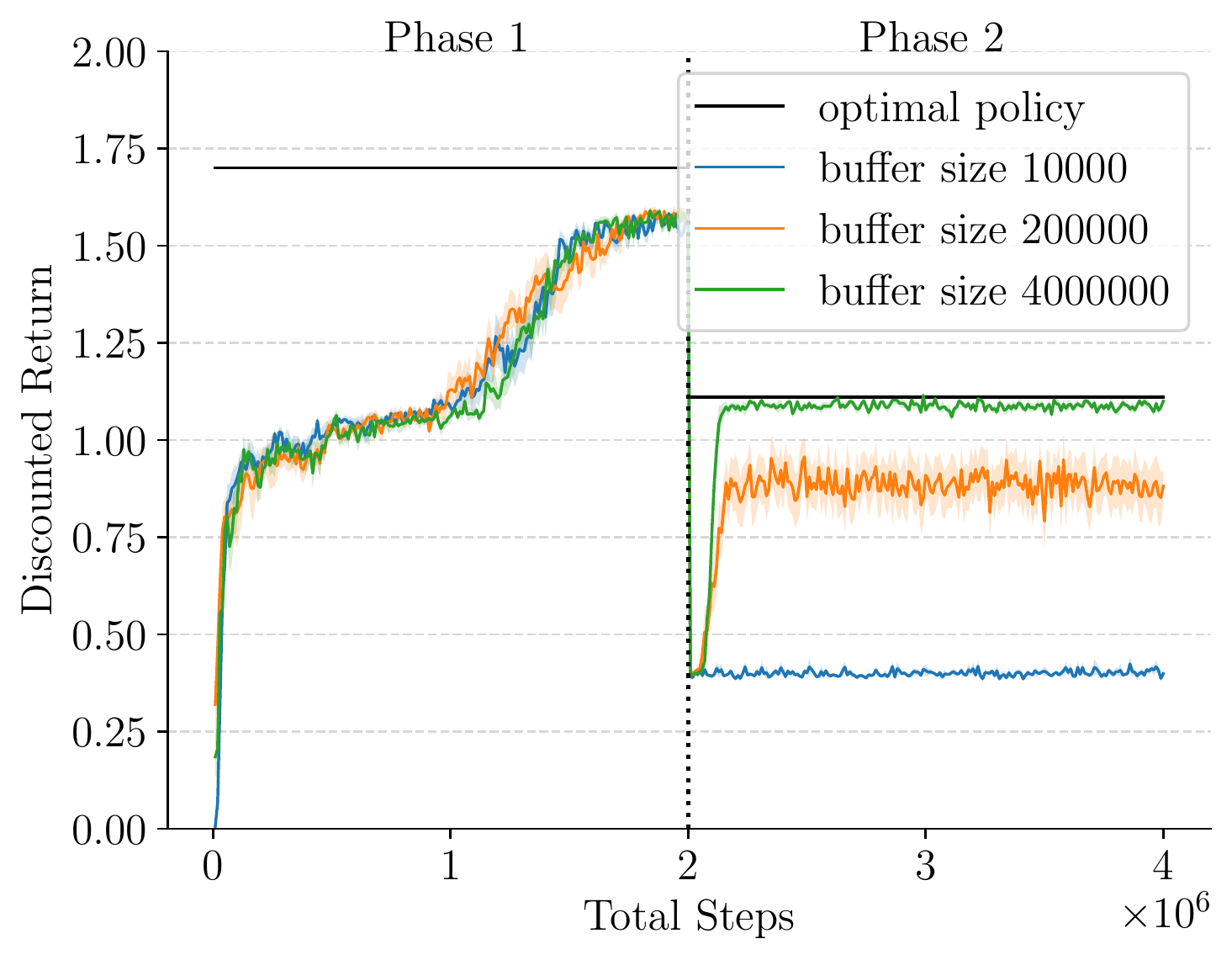}
    \caption{Buffer size}
\end{subfigure}%
\begin{subfigure}{.33\textwidth}
    \centering
    \includegraphics[width=0.95\textwidth]{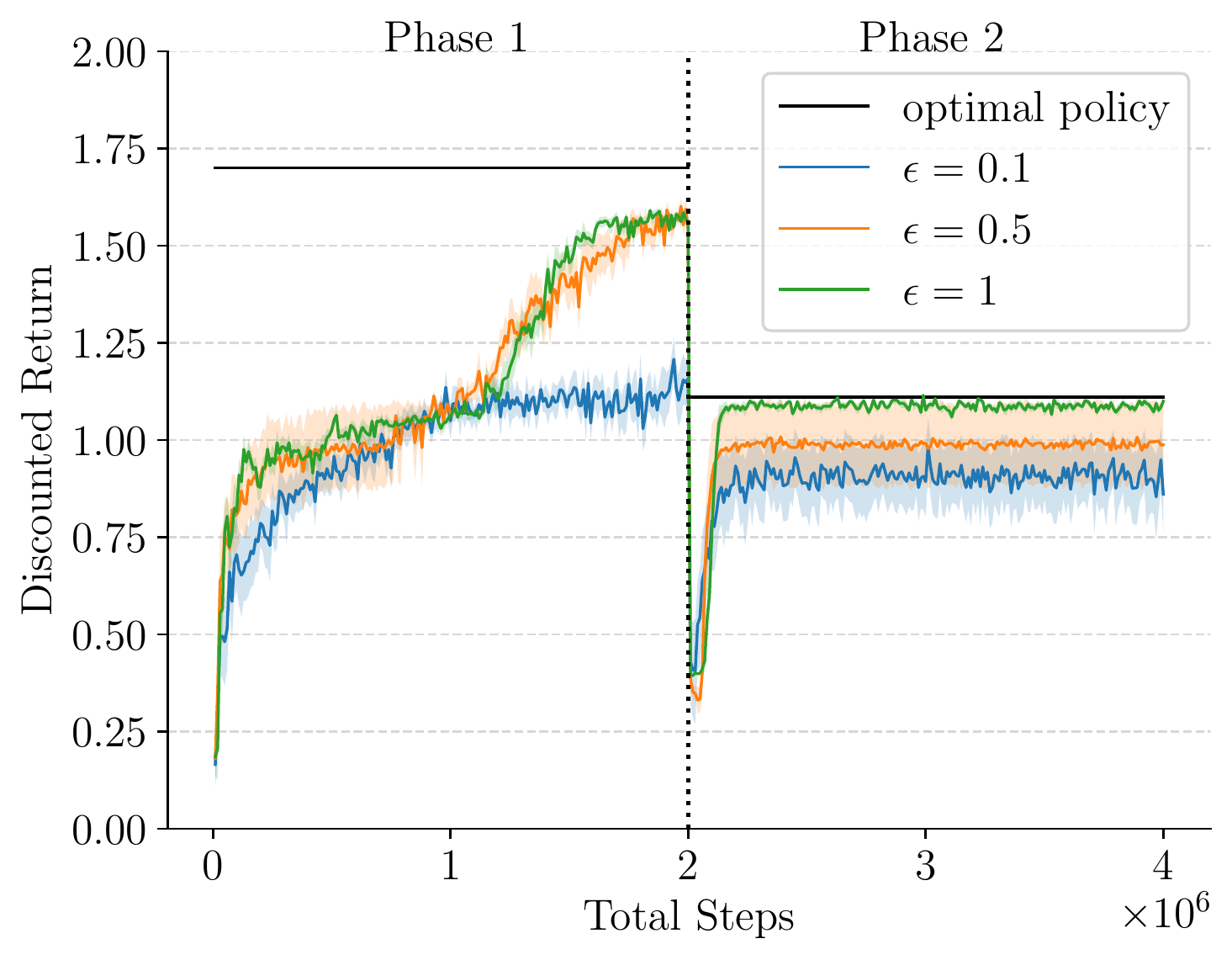}
    \caption{$\epsilon$}
\end{subfigure}%
\begin{subfigure}{.33\textwidth}
    \centering
    \includegraphics[width=0.95\textwidth]{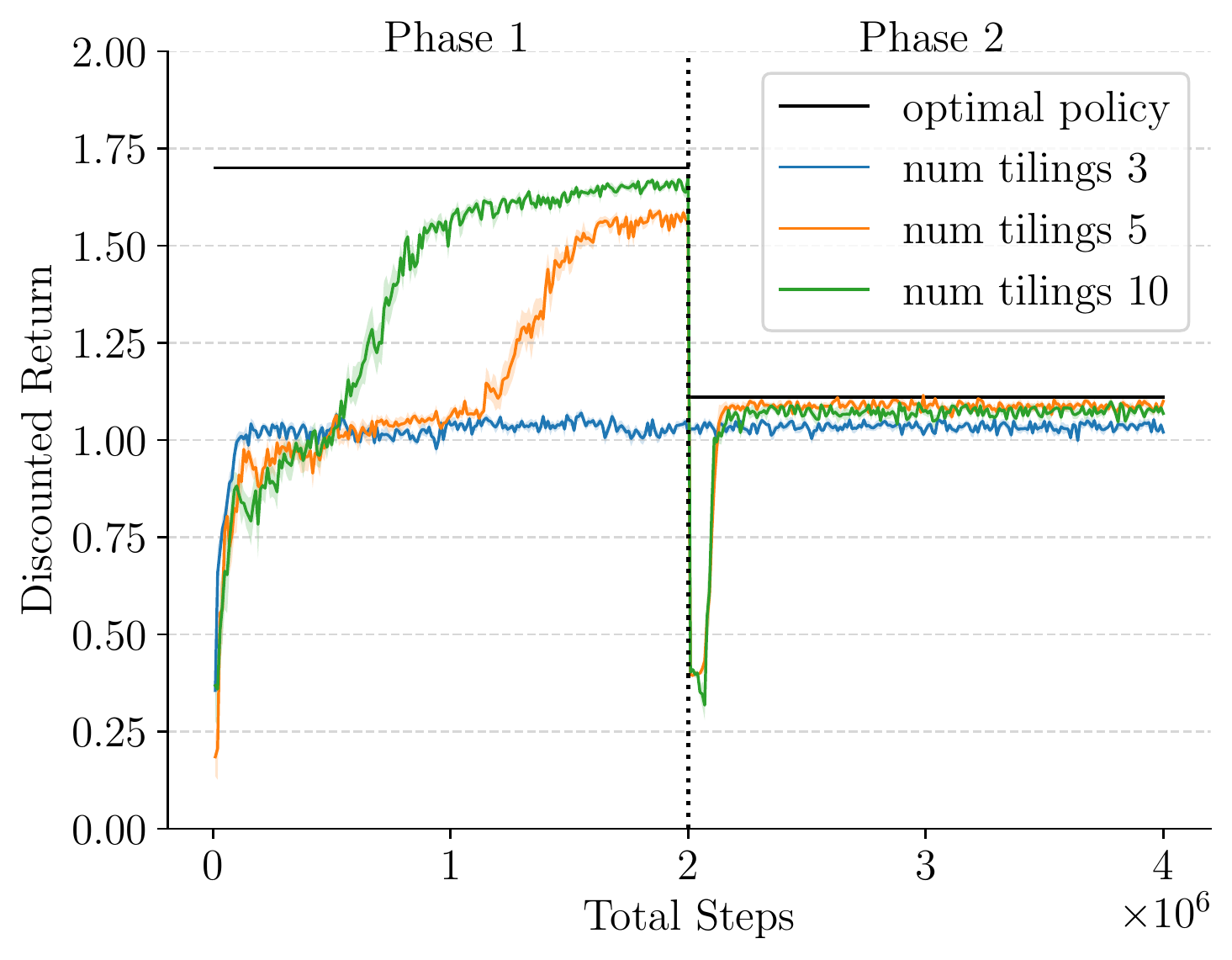}
    \caption{Number of Tilings}
\end{subfigure}\\
\begin{subfigure}{.33\textwidth}
    \centering
    \includegraphics[width=0.95\textwidth]{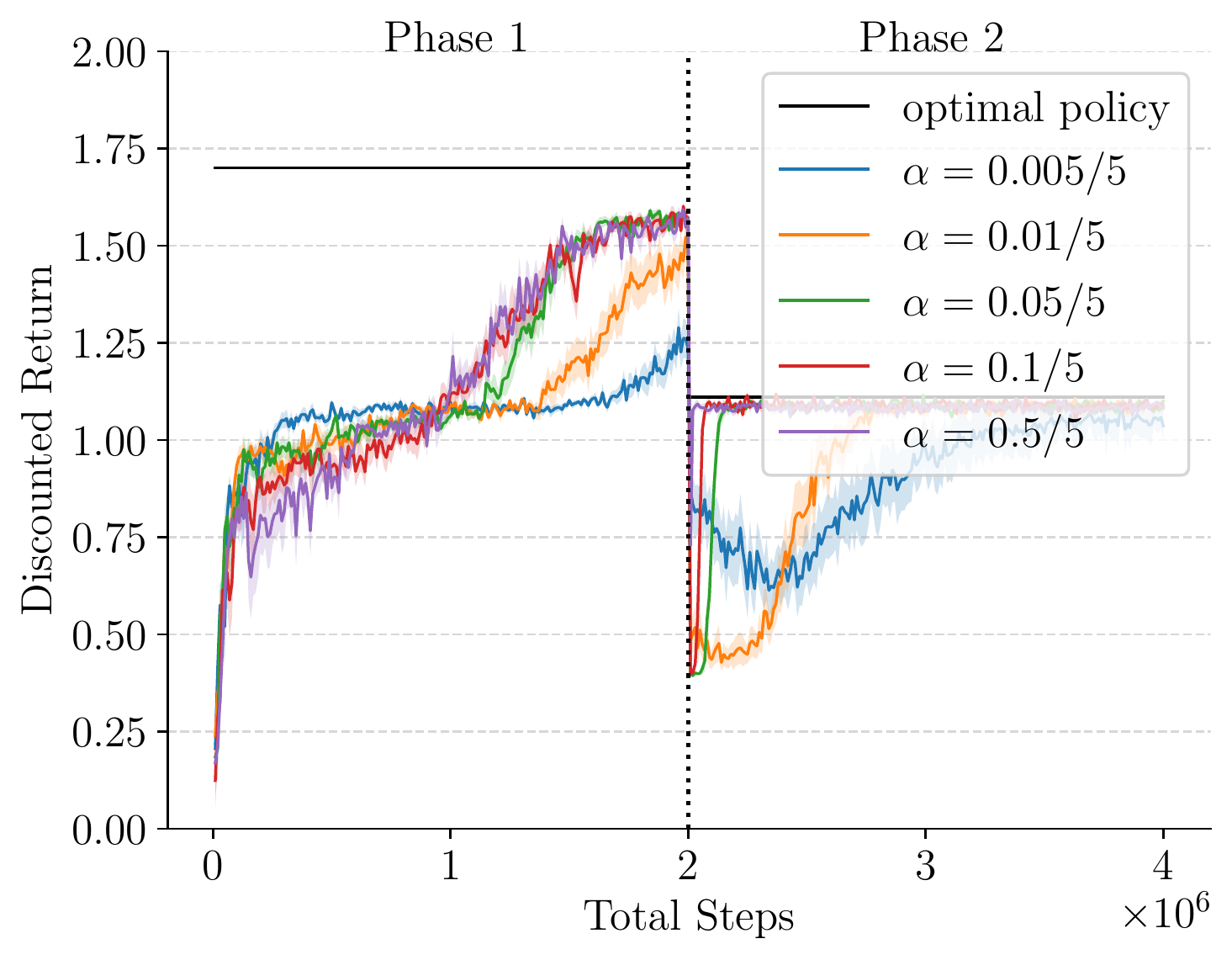}
    \caption{Stepsize}
\end{subfigure}%
\begin{subfigure}{.33\textwidth}
    \centering
    \includegraphics[width=0.95\textwidth]{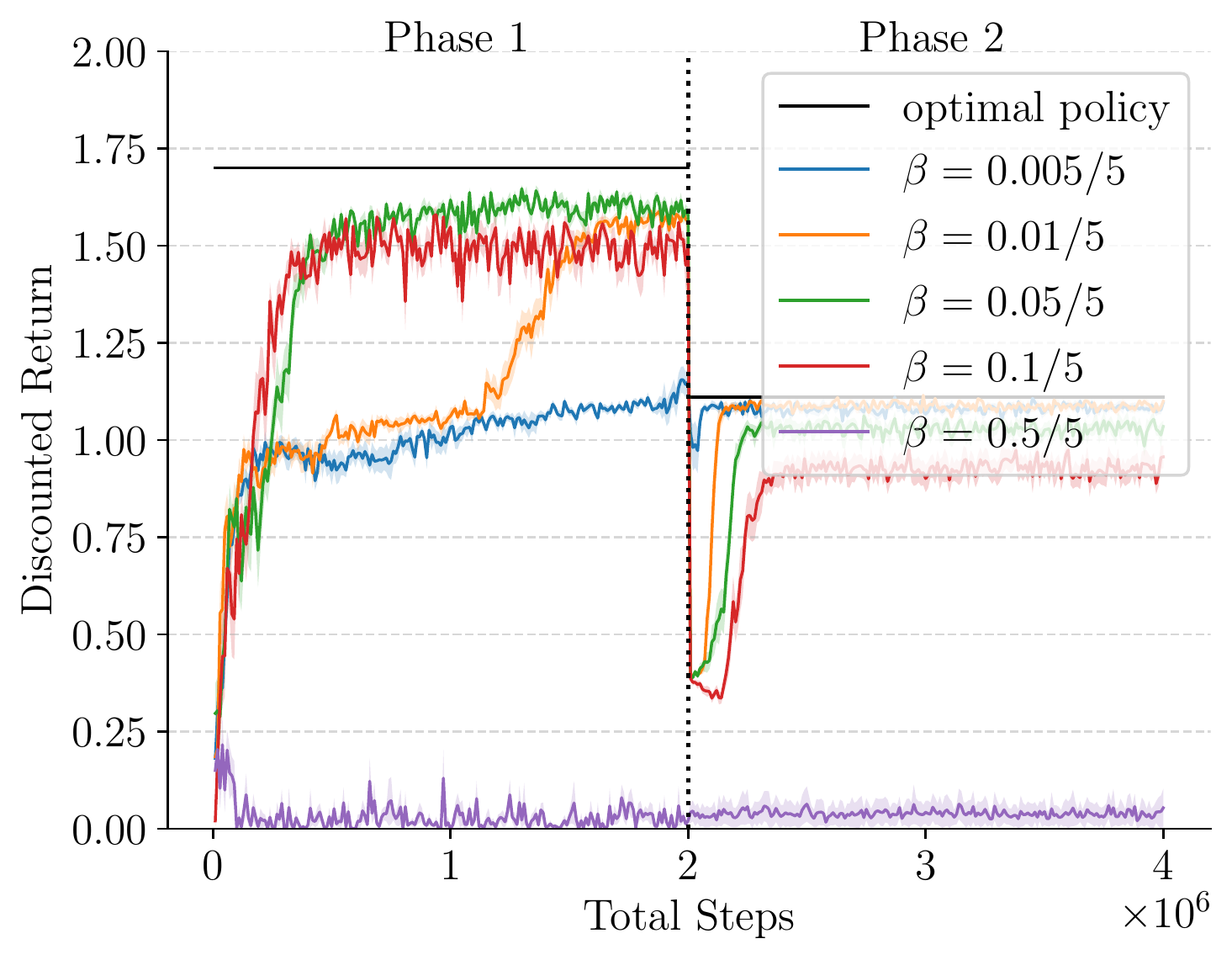}
    \caption{Model Stepsize}
\end{subfigure}%
\caption{Plots showing the sensitivity of the adaptive linear Dyna algorithm to its hyperparameters in the stochastic MountainCarLoCA domain (stochasticity = 0.3). 
}
\label{fig: linear dyna stoch 0.3}
\end{figure*}

\begin{figure*}[h!]
\centering
\begin{subfigure}{.33\textwidth}
    \centering
    \includegraphics[width=0.95\textwidth]{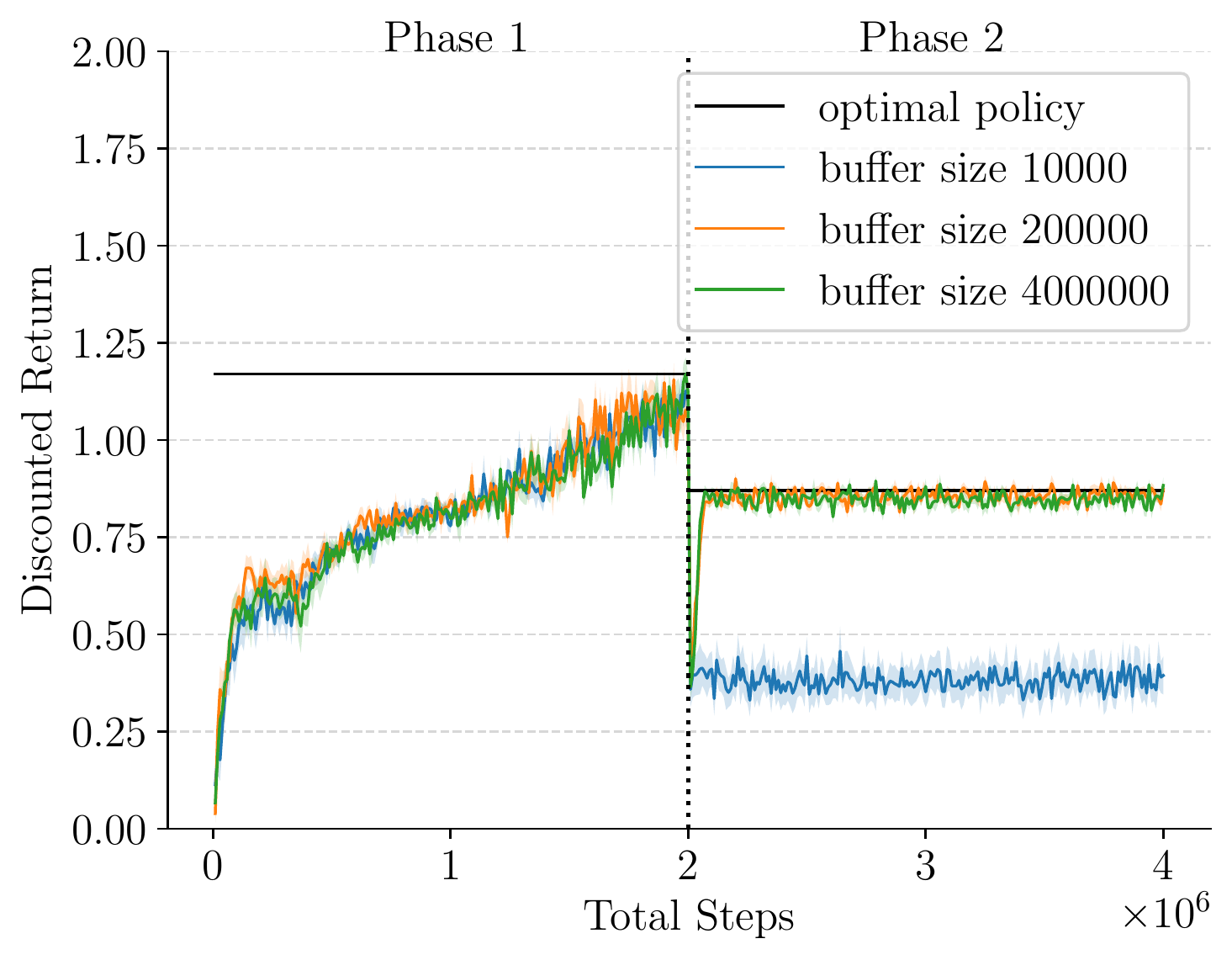}
    \caption{Buffer size}
\end{subfigure}%
\begin{subfigure}{.33\textwidth}
    \centering
    \includegraphics[width=0.95\textwidth]{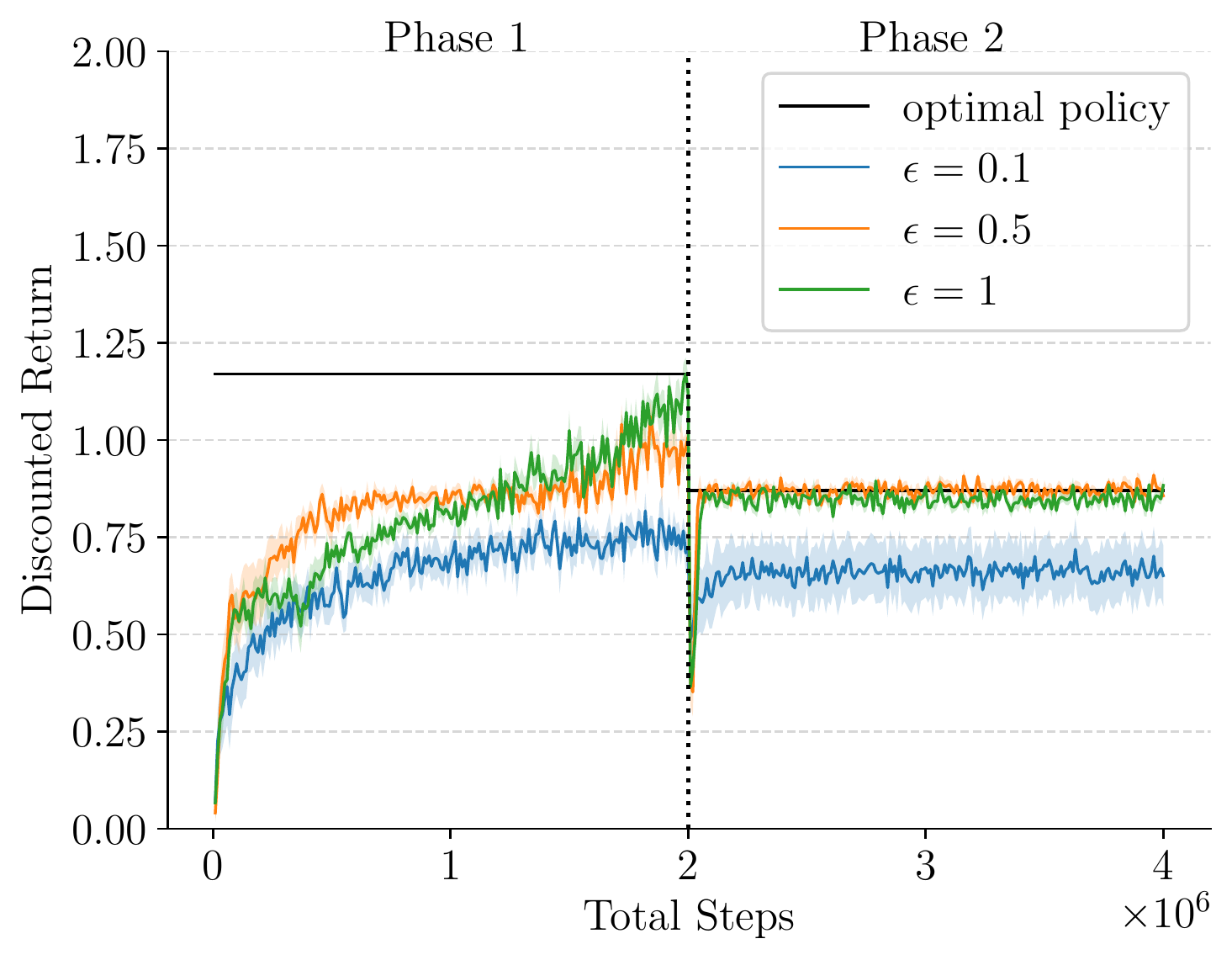}
    \caption{$\epsilon$}
\end{subfigure}%
\begin{subfigure}{.33\textwidth}
    \centering
    \includegraphics[width=0.95\textwidth]{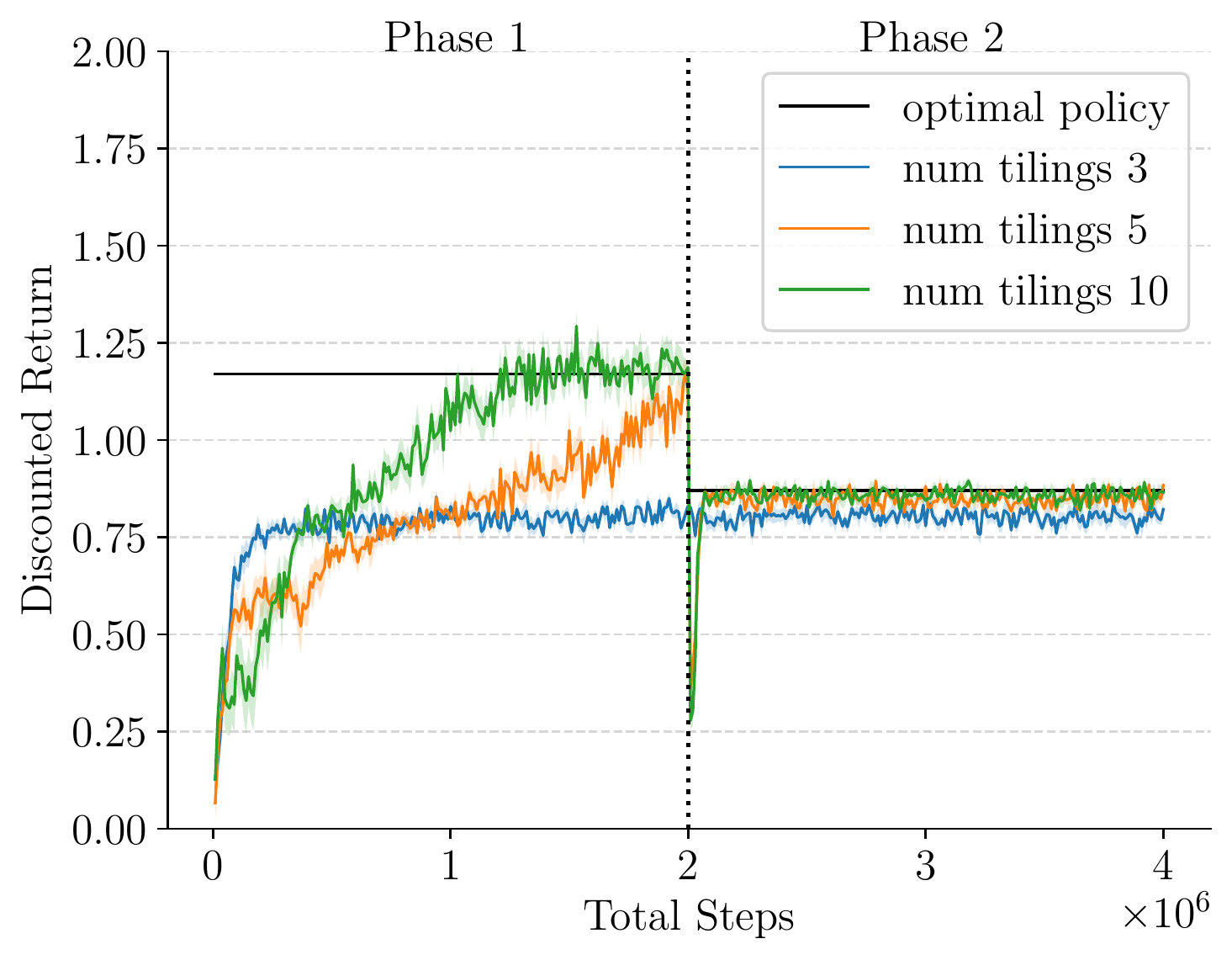}
    \caption{Number of Tilings}
\end{subfigure}\\
\begin{subfigure}{.33\textwidth}
    \centering
    \includegraphics[width=0.95\textwidth]{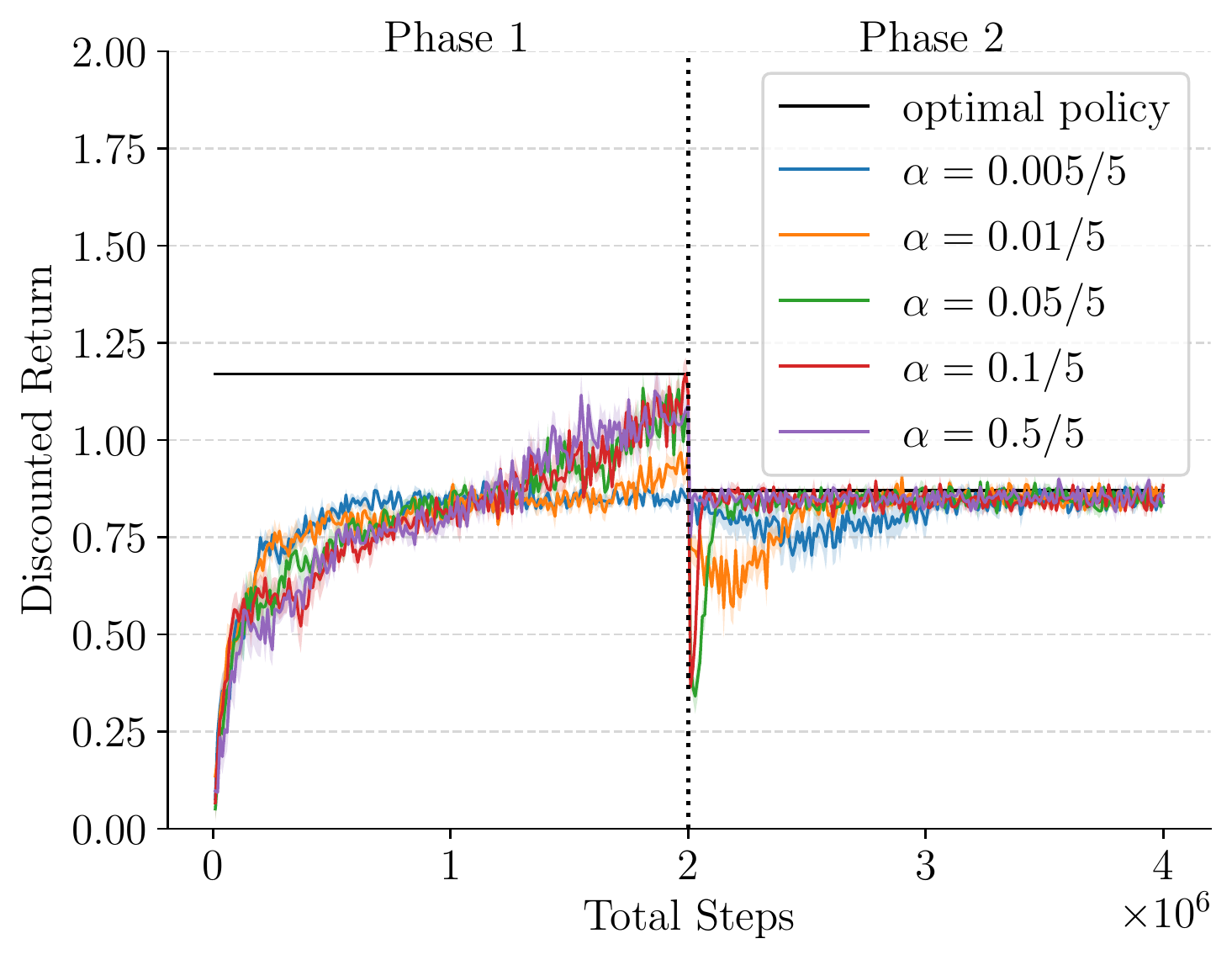}
    \caption{Stepsize}
\end{subfigure}%
\begin{subfigure}{.33\textwidth}
    \centering
    \includegraphics[width=0.95\textwidth]{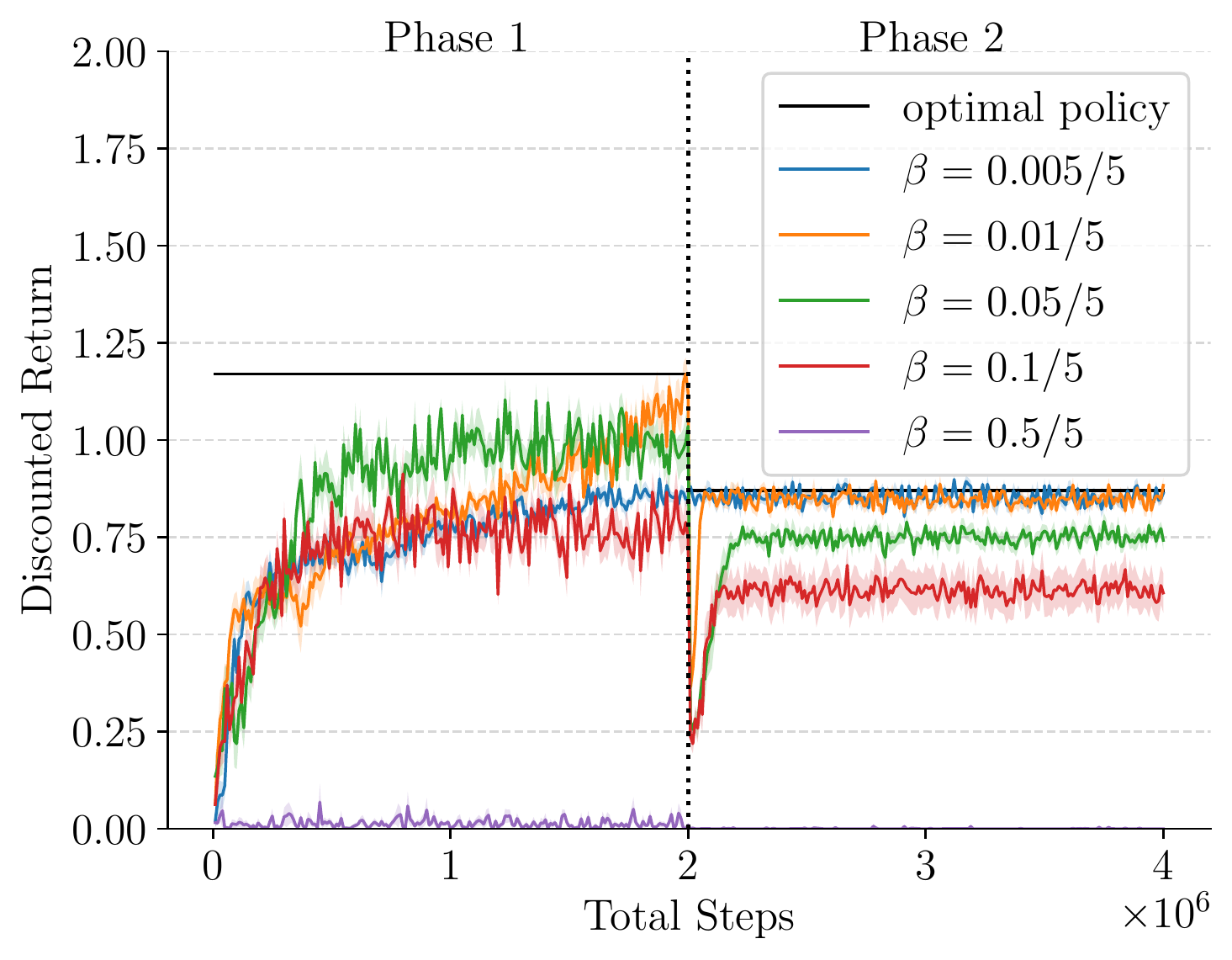}
    \caption{Model Stepsize}
\end{subfigure}%
\caption{Plots showing the sensitivity of the adaptive linear Dyna algorithm to its hyperparameters in the deterministic MountainCarLoCA domain. 
}
\label{fig: linear dyna stoch 0.5}
\end{figure*}

To demonstrate adaptive linear Dyna's sensitivity to a hyperparameter, we fixed all of the hyperparameters, except the one being tested, to be those in the best hyperparameter setting (Table \ref{tab: best parameter settings in linear experiments}). We then varied the tested hyperparameter and plotted the corresponding learning curves. Learning curves for stochasticity = 0, 0.3, and 0.5 are shown in Figure~\ref{fig: linear dyna stoch 0}, Figure~\ref{fig: linear dyna stoch 0.3}, and Figure~\ref{fig: linear dyna stoch 0.5} respectively.

We now use Figure~\ref{fig: linear dyna stoch 0} as an example to understand the algorithm's sensitivity to its hyperparameters. Sub-figure (a) shows that the algorithm needs a sufficiently large replay buffer to achieve fast adaptation. This is not surprising because the collected data in Phase 2 would only cover a small region of the state space. The Phase 2 data quickly occupy the entire planning buffer, after which planning in Phase 2 would only update states in that small region. Sub-figure (b) shows that the algorithm worked well when a relatively large exploration parameter is used. $\epsilon = 0.1$ performed well in the second phase but it was not close to optimality in the first phase. A deeper look at the experimental data (not shown here) suggested that, when $\epsilon = 0.1$, for most of runs, the agent only moved to the sub-optimal terminal state in the first phase, presumably as a result of weak exploration. The sub-optimal terminal state in Phase 1 is the optimal terminal state in Phase 2 and thus the agent didn't have to change its policy to adapt to the task change. A first look at Sub-figure (c) seems to suggest that the algorithm only adapted well when the number of tilings $k$ is small (3 or 5). By checking experimental results of more hyperparameter settings, we found that the failure when $k = 10$ is actually a result of other hyperparameters being chosen to be the best for $k = 5$. There are also quite a few hyperparameter settings that achieved fast adaptation in Phase 2 for $k = 10$ (not shown here). In fact, Figure\ \ref{fig: linear dyna stoch 0.3}, and \ref{fig: linear dyna stoch 0.5} show that $k = 10$ performed pretty well for stochasticity = 0.3 and 0.5. Sub-figure (g) shows the sensitivity of the algorithm to the stepsize. As one would expect, lower stepsize results in slower learning in both Phases 1 and 2. Sub-figure (h) is particularly interesting. It shows that when the model stepsize is large (equal or larger than 0.05/5) the performance in Phase 2 has a high variance. 
The model stepsize can not be too small either: with $\beta = 0.005/5$ the agent didn't learn to move to the right terminal in the first phase.

\begin{figure*}[h!]
\centering
\begin{subfigure}{.25\textwidth}
    \centering
    \includegraphics[width=0.95\textwidth]{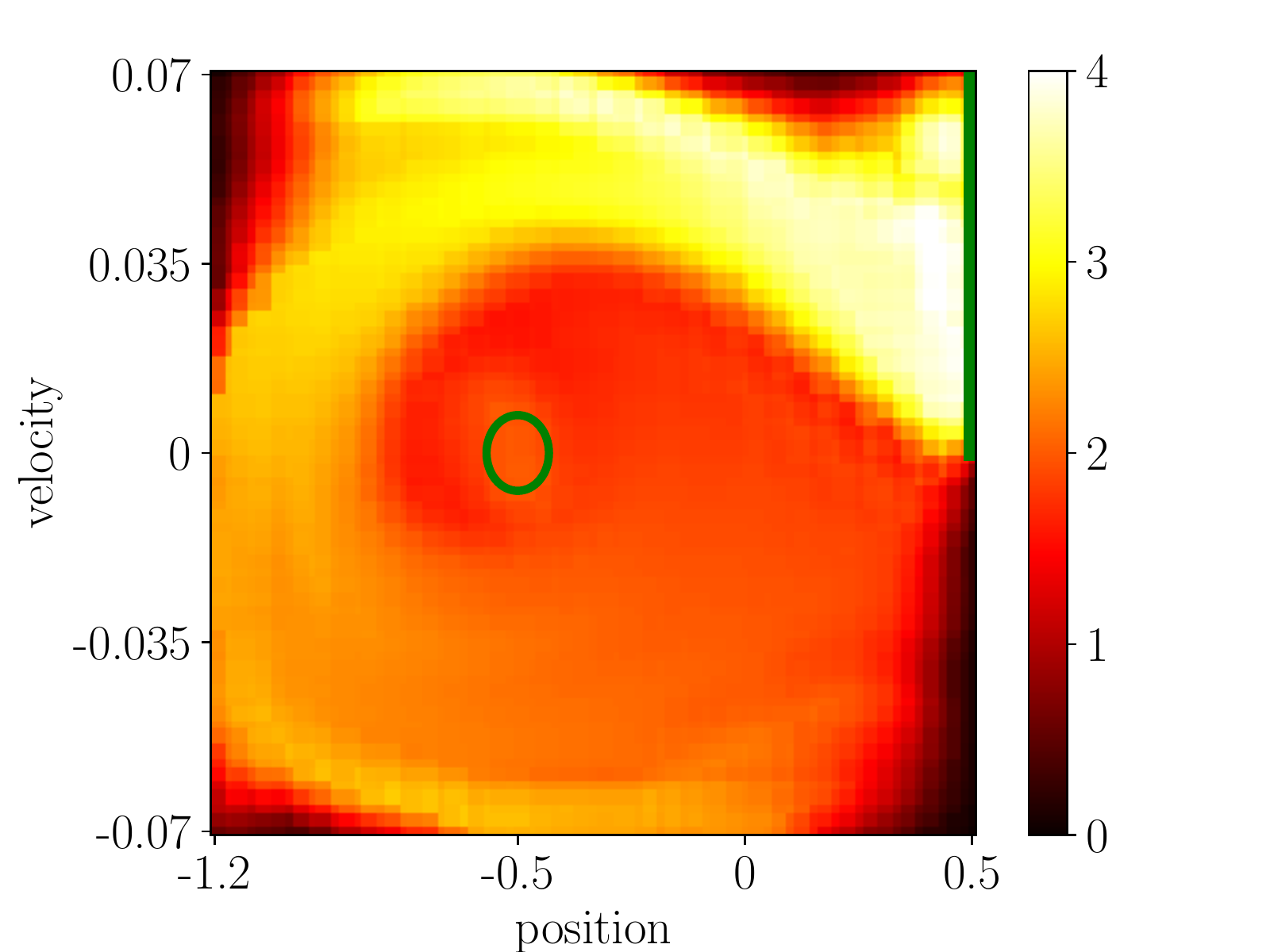}
    \caption{Values}
\end{subfigure}%
\begin{subfigure}{.25\textwidth}
    \centering
    \includegraphics[width=0.95\textwidth]{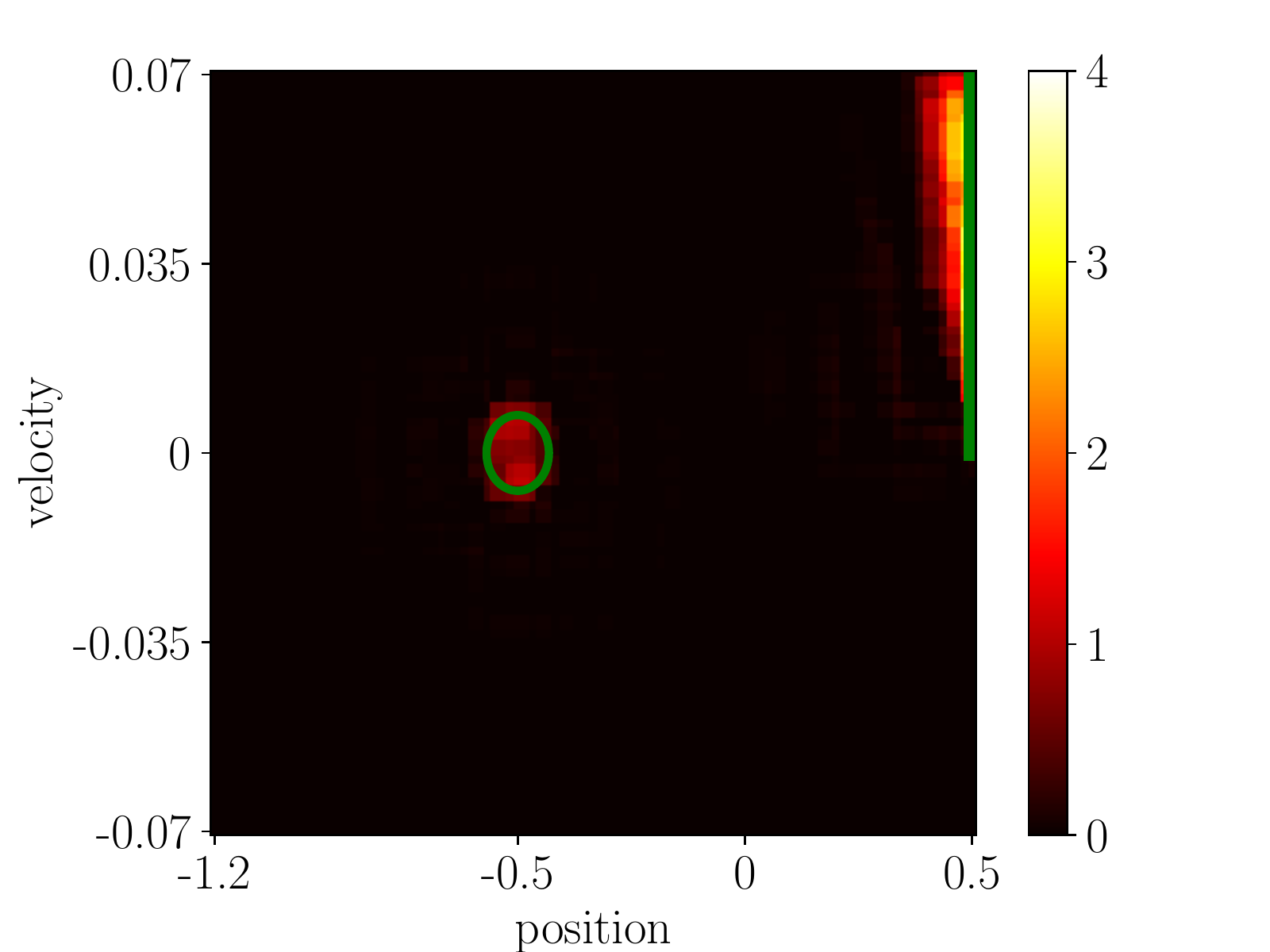}
    \caption{Rewards}
\end{subfigure}%
\begin{subfigure}{.25\textwidth}
    \centering
    \includegraphics[width=0.95\textwidth]{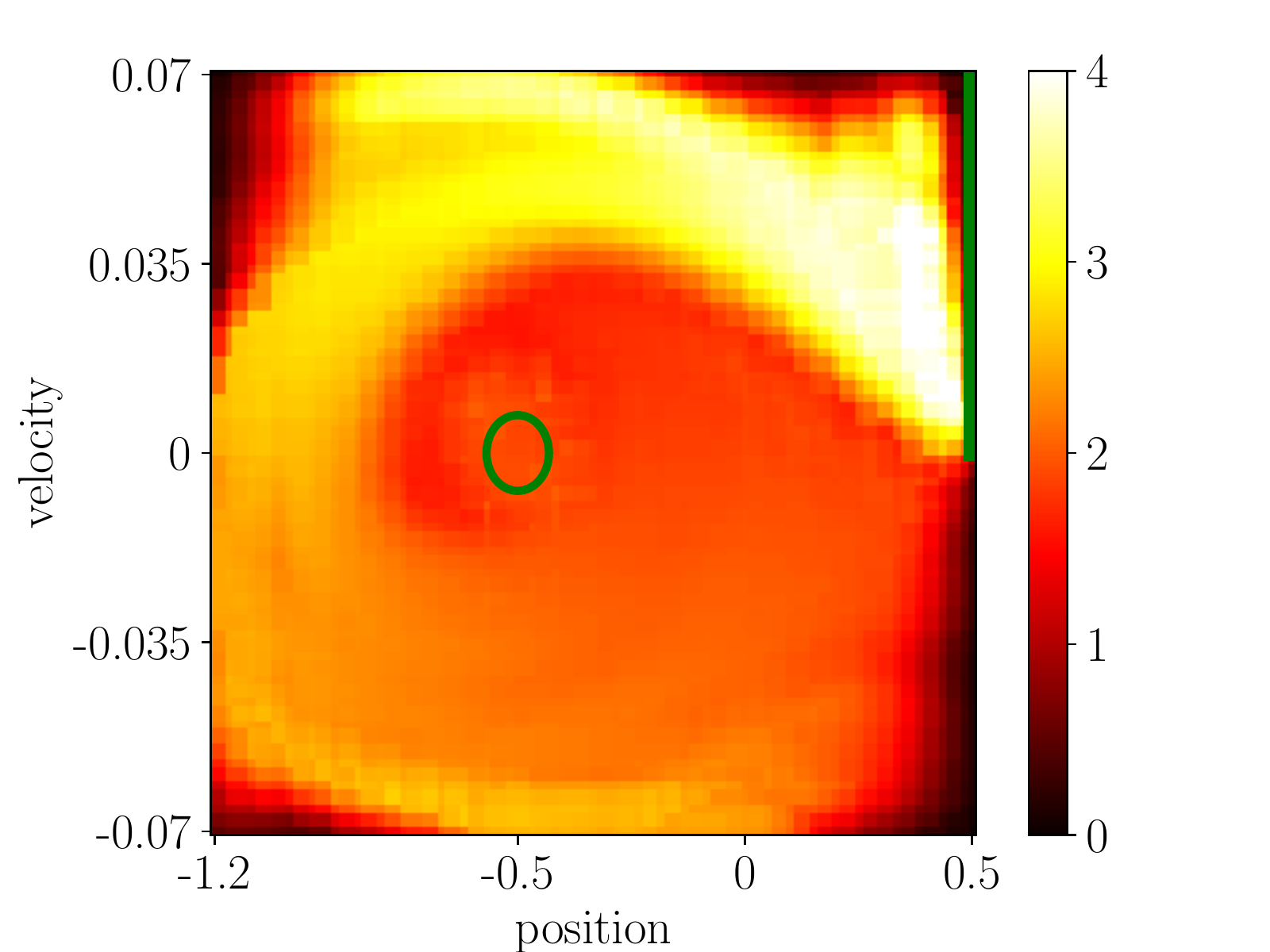}
    \caption{Expected Next Values}
\end{subfigure}%
\begin{subfigure}{.2\textwidth}
    \centering
    \includegraphics[width=0.95\textwidth]{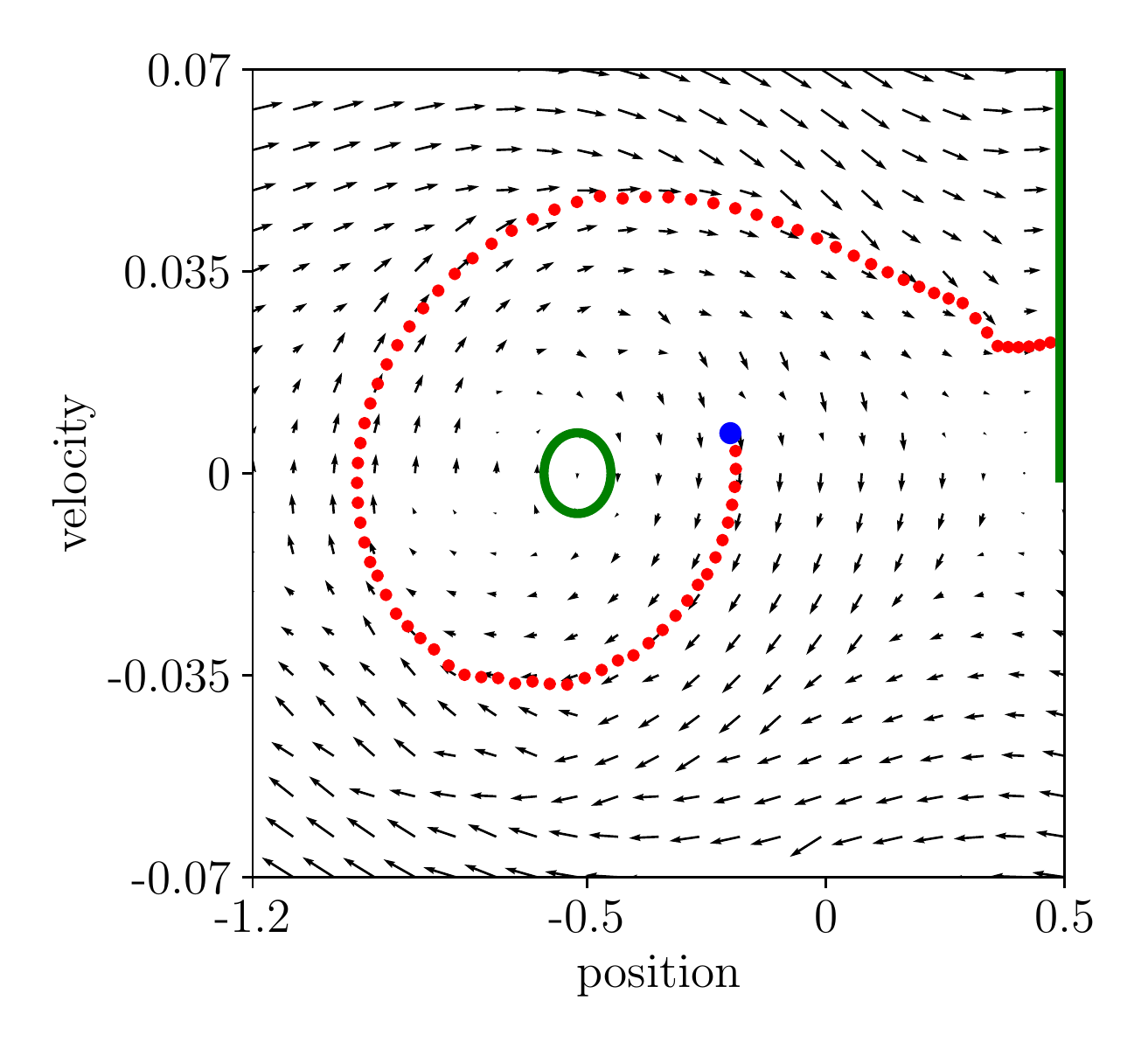}
    \caption{Policy}
\end{subfigure}\\
\begin{subfigure}{.25\textwidth}
    \centering
    \includegraphics[width=0.95\textwidth]{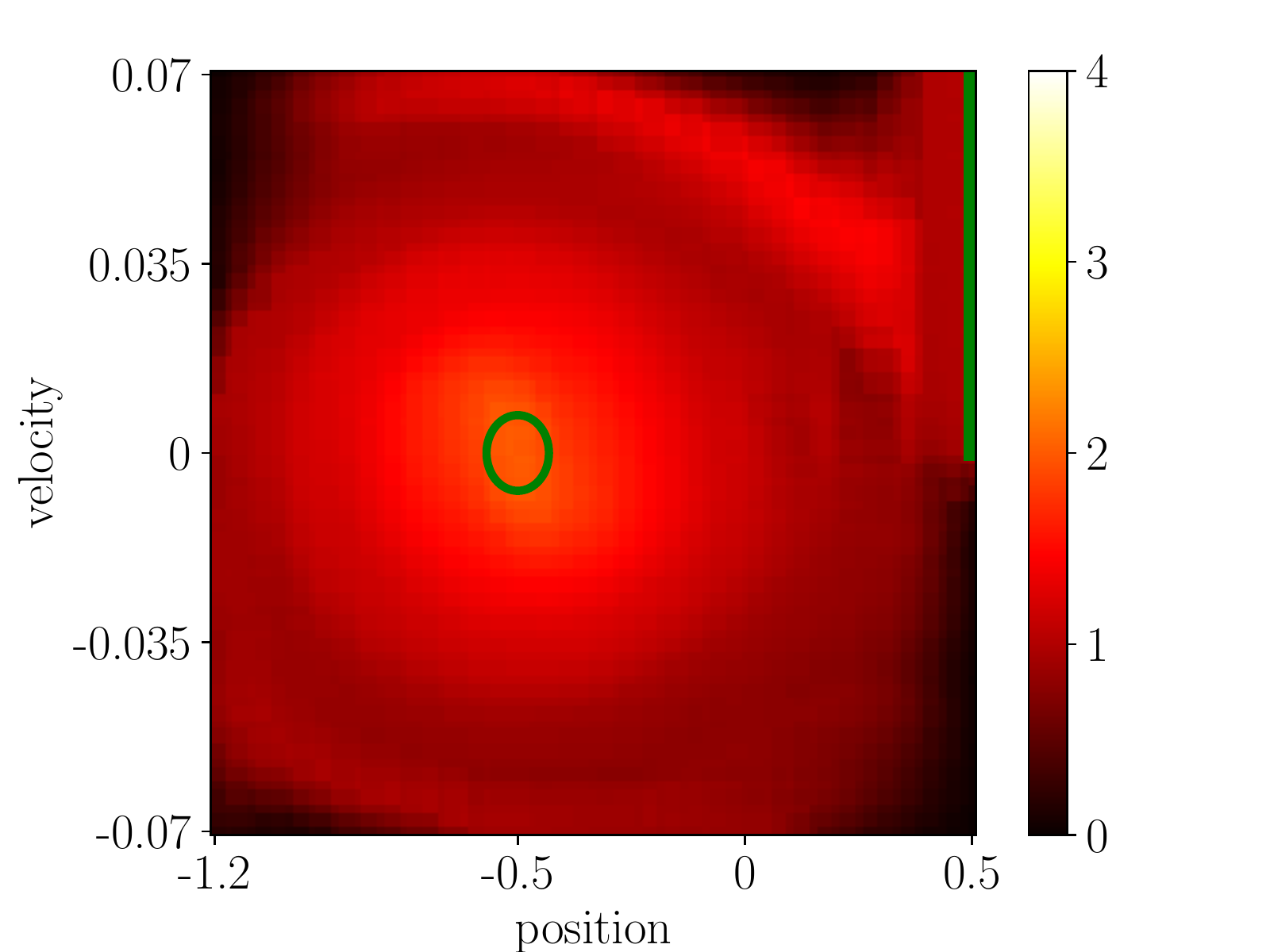}
    \caption{Values}
\end{subfigure}%
\begin{subfigure}{.25\textwidth}
    \centering
    \includegraphics[width=0.95\textwidth]{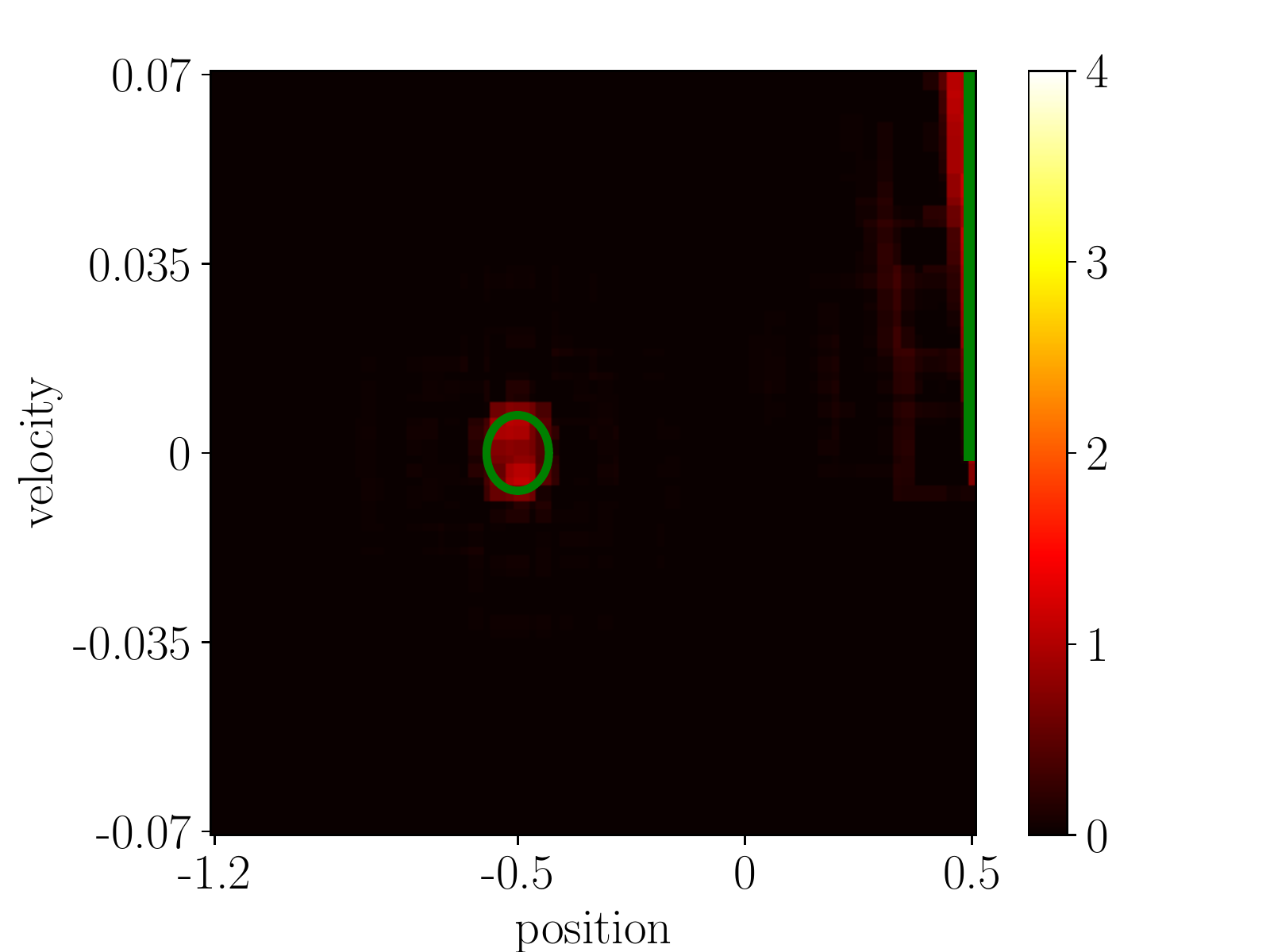}
    \caption{Rewards}
\end{subfigure}%
\begin{subfigure}{.25\textwidth}
    \centering
    \includegraphics[width=0.95\textwidth]{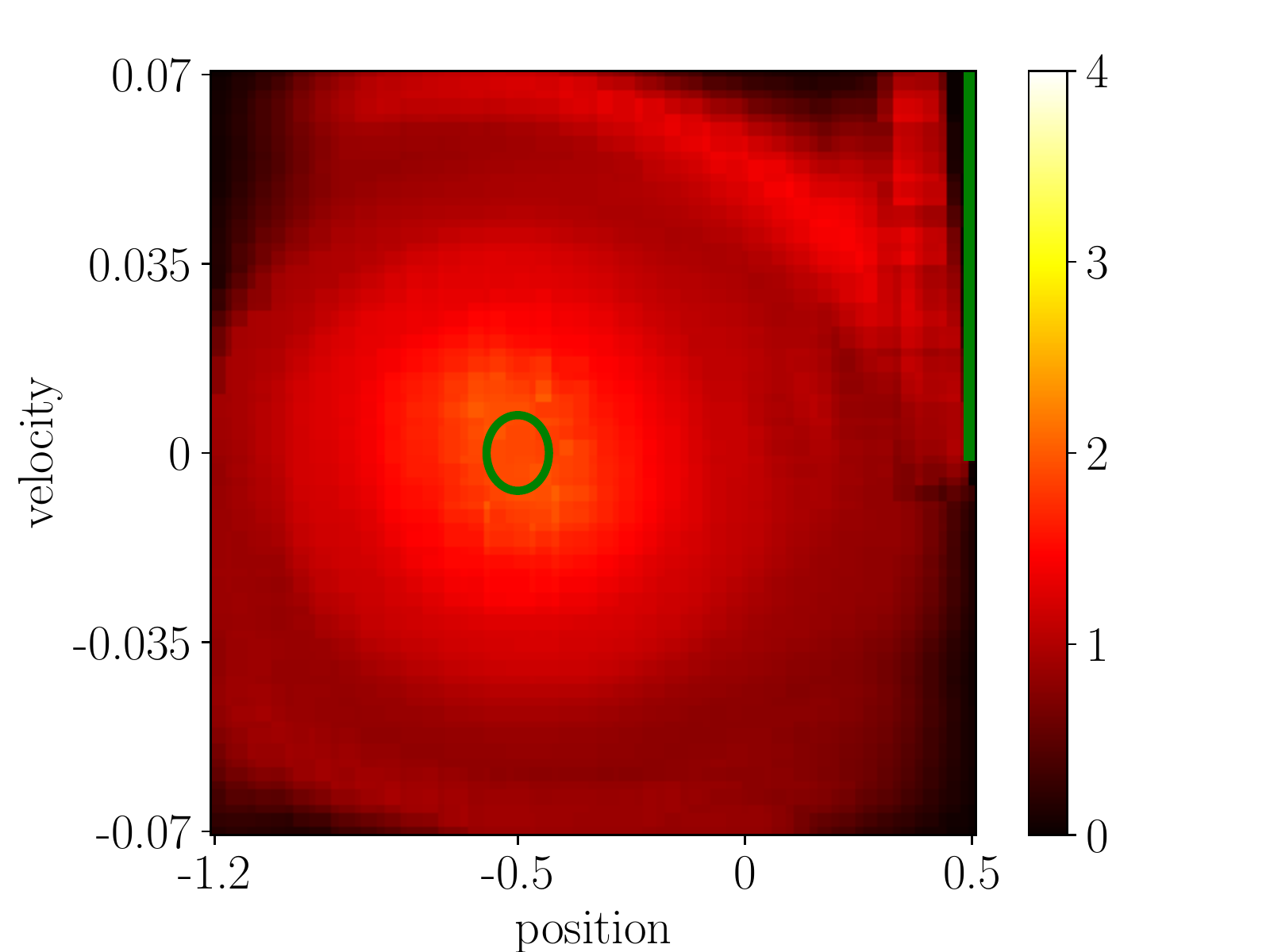}
    \caption{Expected Next Values}
\end{subfigure}%
\begin{subfigure}{.2\textwidth}
    \centering
    \includegraphics[width=0.95\textwidth]{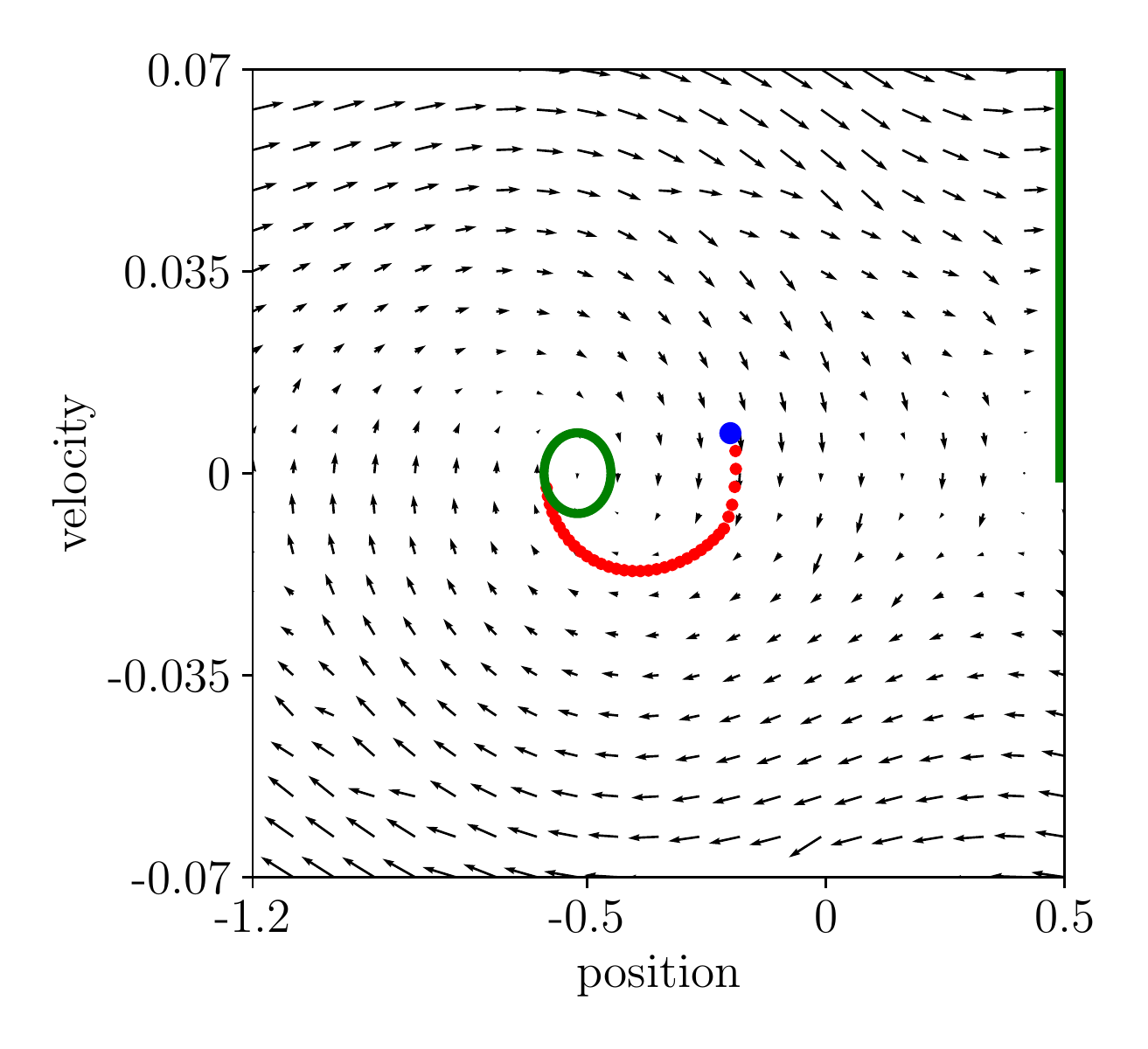}
    \caption{Policy}
\end{subfigure}%
\caption{Plots showing estimated values, reward model, the value of the predicted next state produced by the model, and the learned policy of adaptive linear Dyna in the deterministic MountainCarLoCA domain, with the best hyperparameter setting (Table\ \ref{tab: best parameter settings in linear experiments}). The first and second row correspond to the results at the end of Phases 1 and 2 respectively. In each sub-figure, the $x$-axis is the position of the car, the $y$-axis is the velocity of the car. For each state (position and velocity), we plotted the estimated value in the first column. The second corresponds to the maximum of estimated immediate reward. The third column corresponds to the maximum of the value of the model's predicted next state. In both cases, the maximum was taken over different actions. In the fourth column, we drew a vector field, where each vector represents the evolution of a state under the learned policy. We further drew an sequence of red dots, illustrating evolution of states in one particular evaluation episode. A large blue dot marks the start state of the episode. Comparing Sub-figure (a), (e) and Sub-figure a), c) in Figure \ref{fig:optimal heatmaps} show that resulting values are close to optimal values for most of states. }
\label{fig: linear dyna plots}
\end{figure*}

\begin{figure*}[h!]
\centering
\begin{subfigure}{.25\textwidth}
    \centering
    \includegraphics[width=0.95\textwidth]{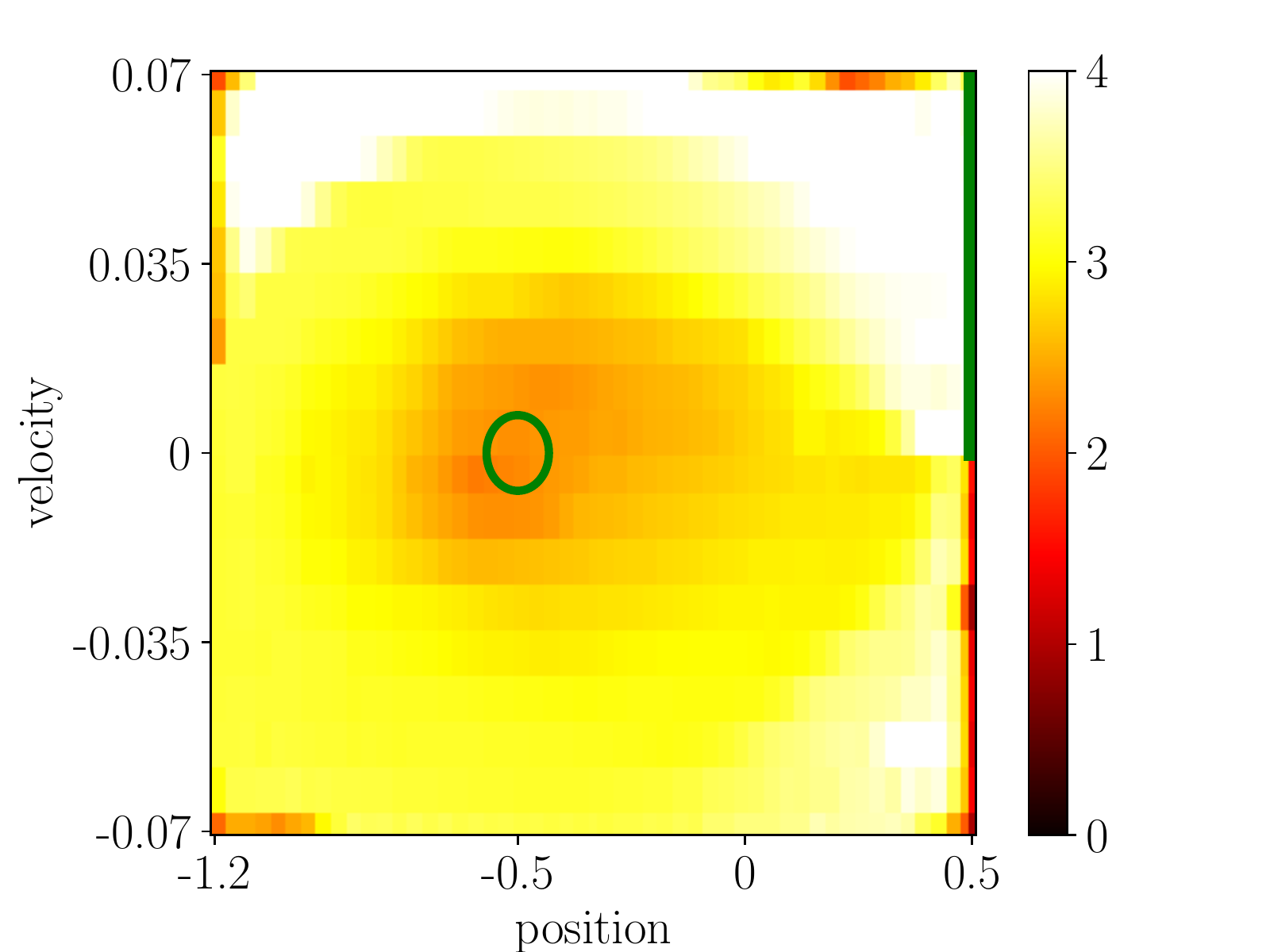}
    \caption{Values}
\end{subfigure}%
\begin{subfigure}{.25\textwidth}
    \centering
    \includegraphics[width=0.95\textwidth]{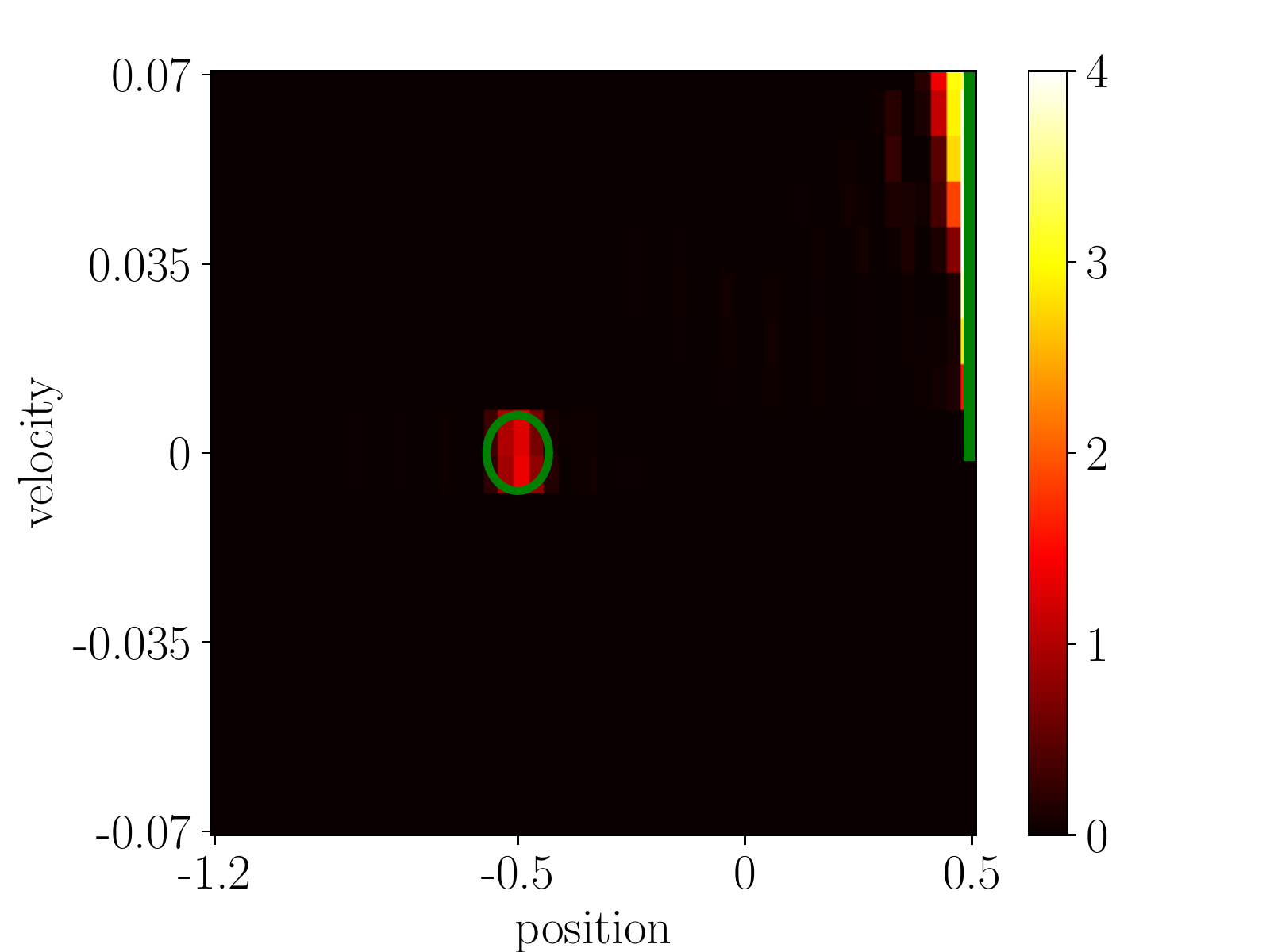}
    \caption{Rewards}
\end{subfigure}%
\begin{subfigure}{.25\textwidth}
    \centering
    \includegraphics[width=0.95\textwidth]{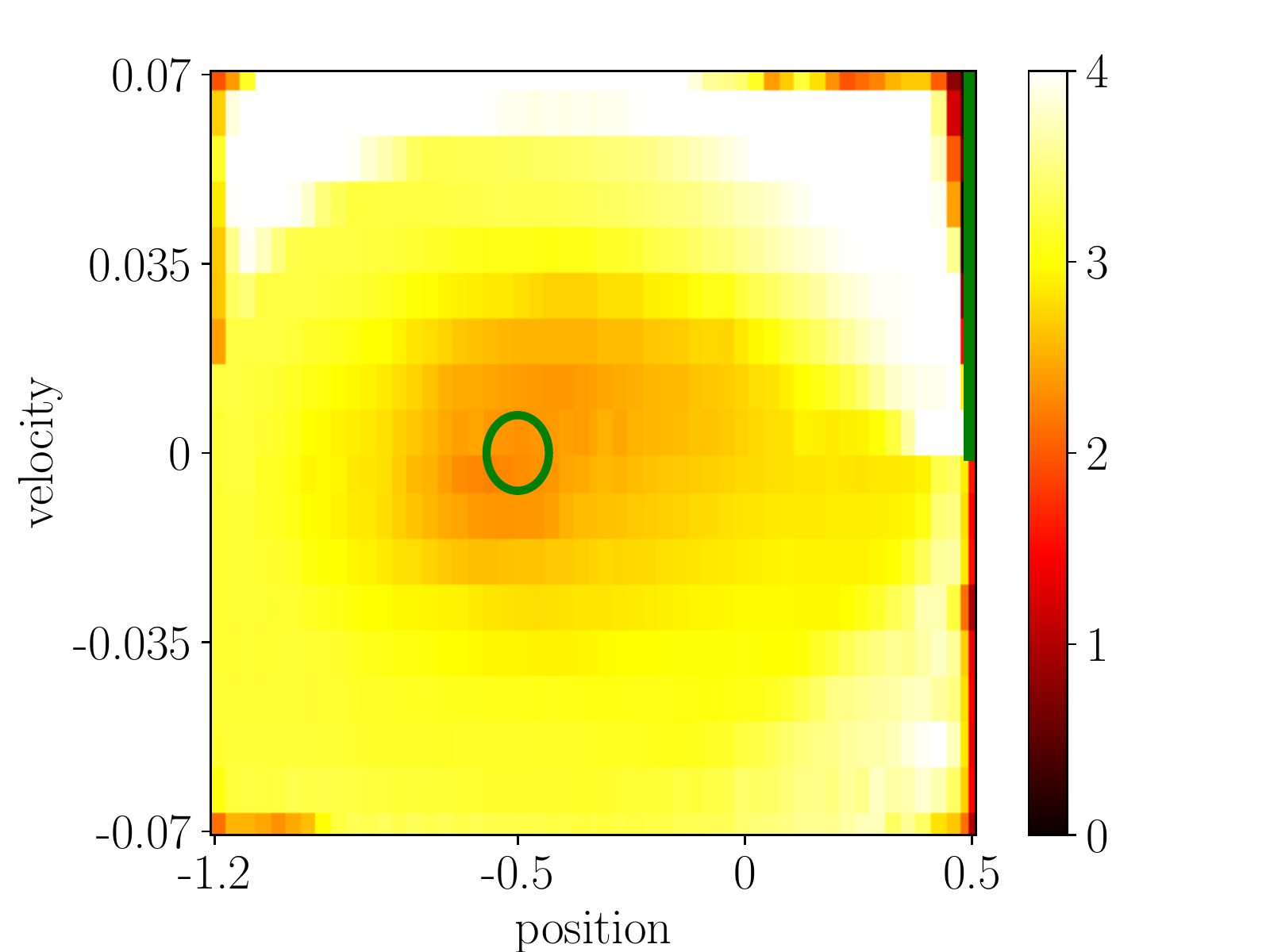}
    \caption{Expected Next Values}
\end{subfigure}%
\begin{subfigure}{.205\textwidth}
    \centering
    \includegraphics[width=0.95\textwidth]{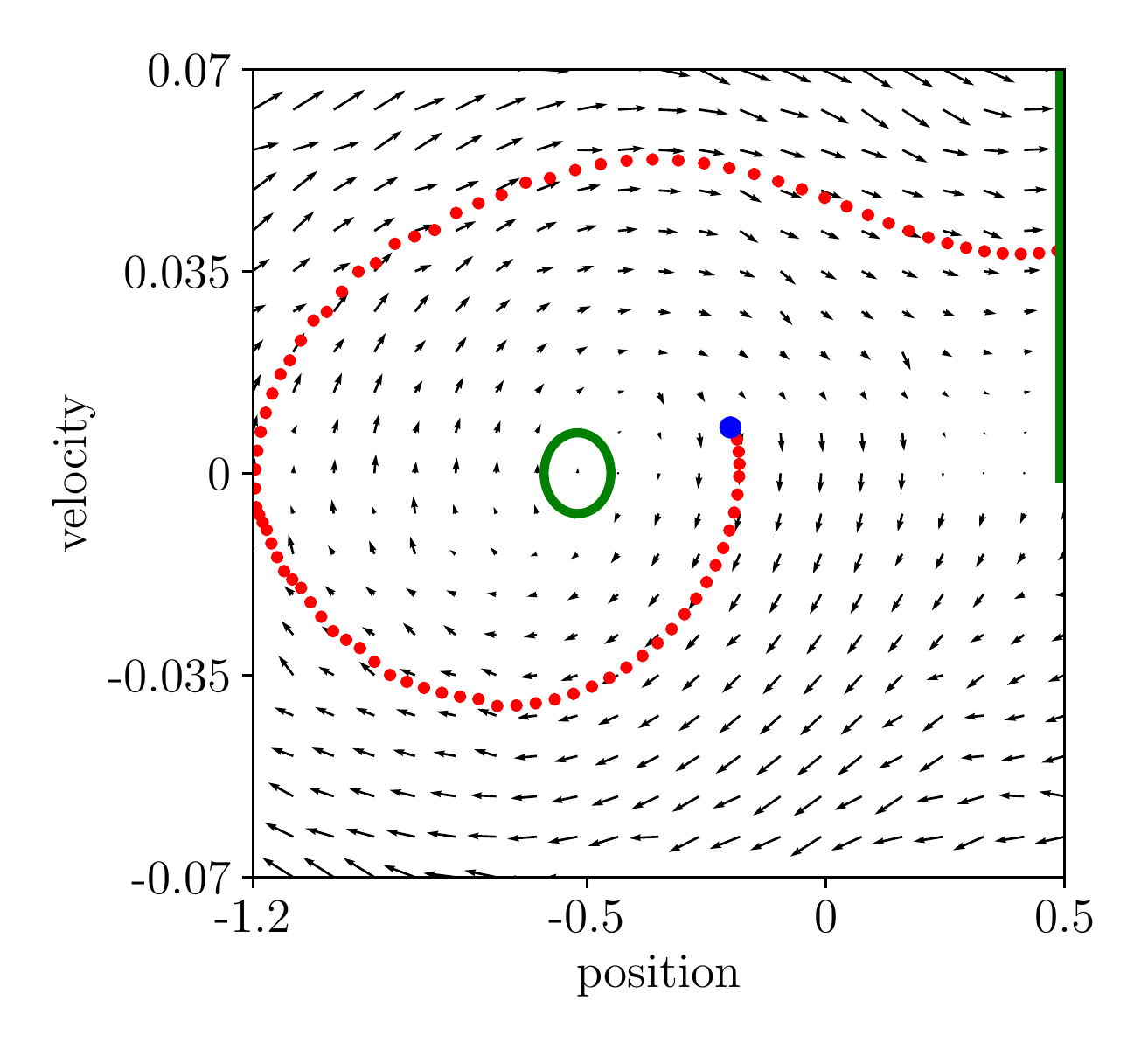}
    \caption{Policy}
\end{subfigure}\\
\begin{subfigure}{.25\textwidth}
    \centering
    \includegraphics[width=0.95\textwidth]{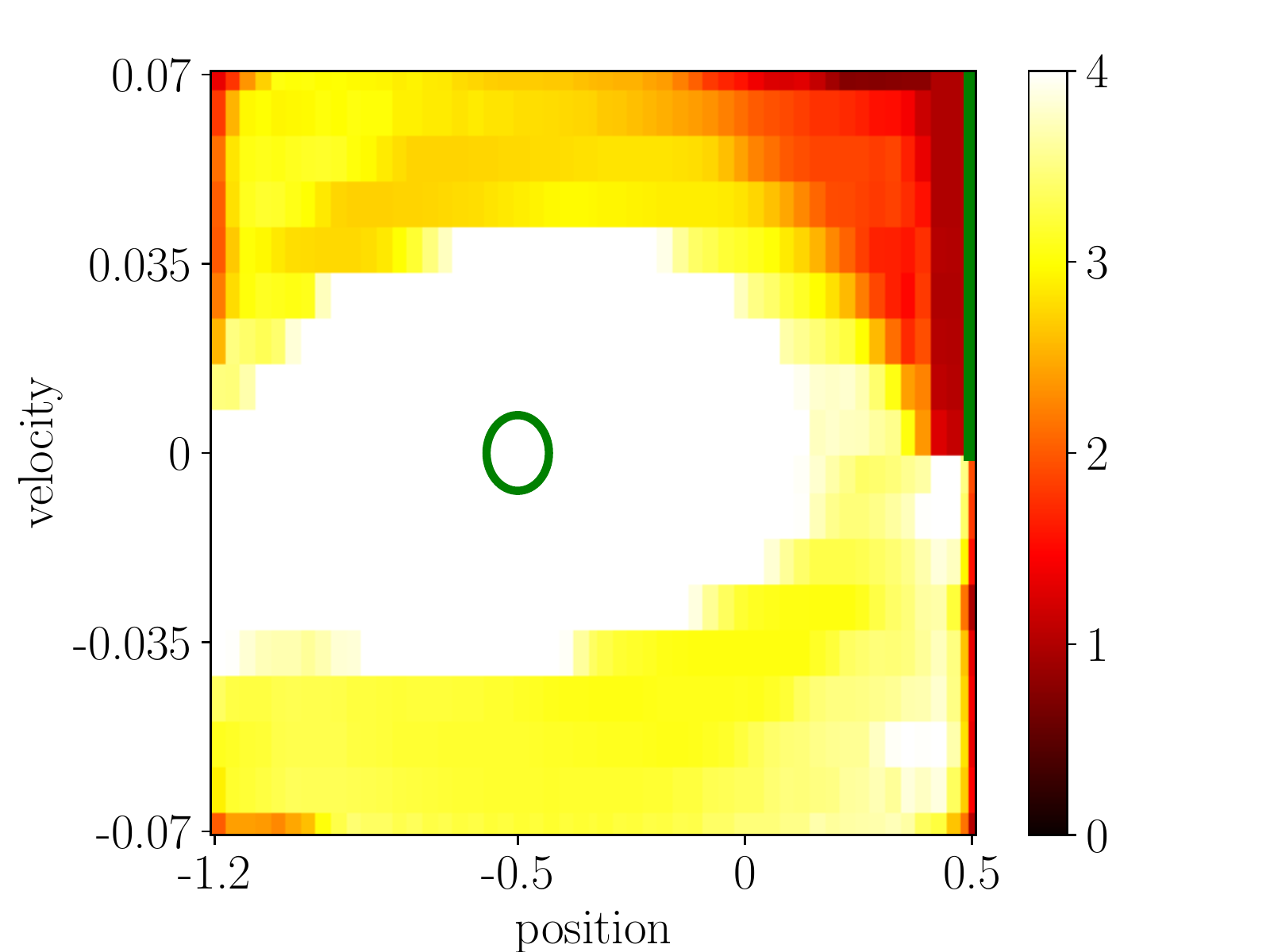}
    \caption{Values}
\end{subfigure}%
\begin{subfigure}{.25\textwidth}
    \centering
    \includegraphics[width=0.95\textwidth]{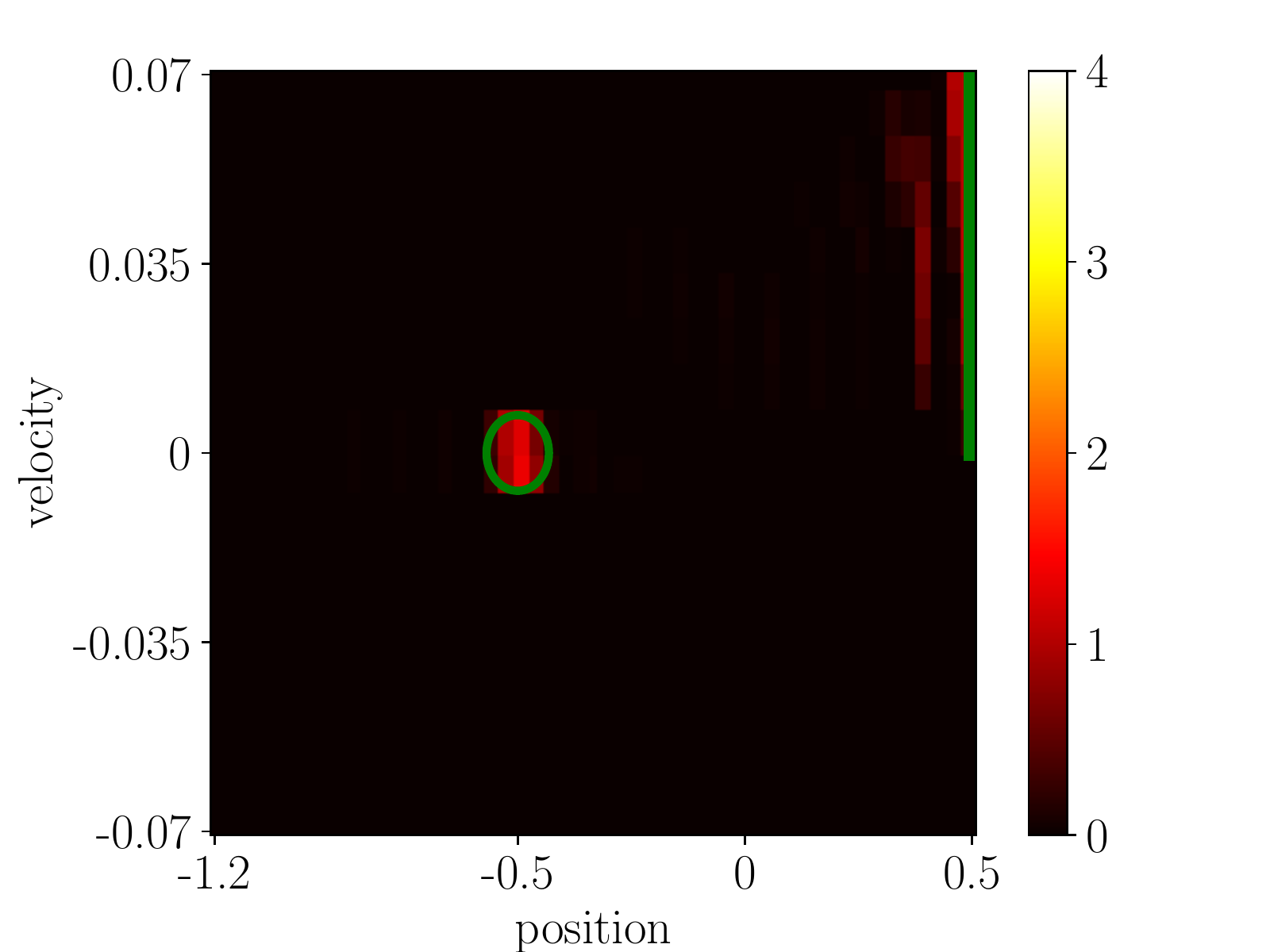}
    \caption{Rewards}
\end{subfigure}%
\begin{subfigure}{.25\textwidth}
    \centering
    \includegraphics[width=0.95\textwidth]{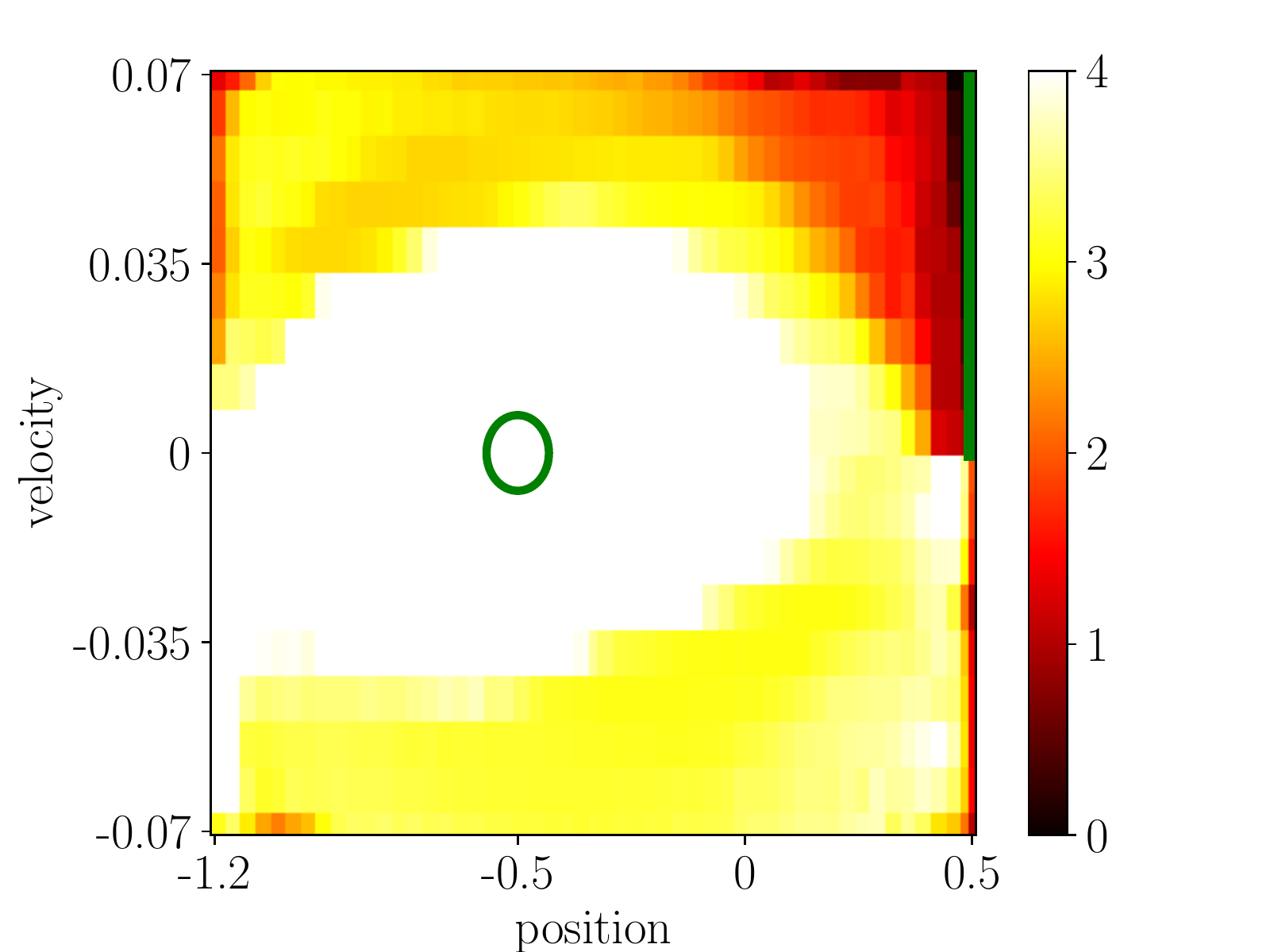}
    \caption{Expected Next Values}
\end{subfigure}%
\begin{subfigure}{.205\textwidth}
    \centering
    \includegraphics[width=0.95\textwidth]{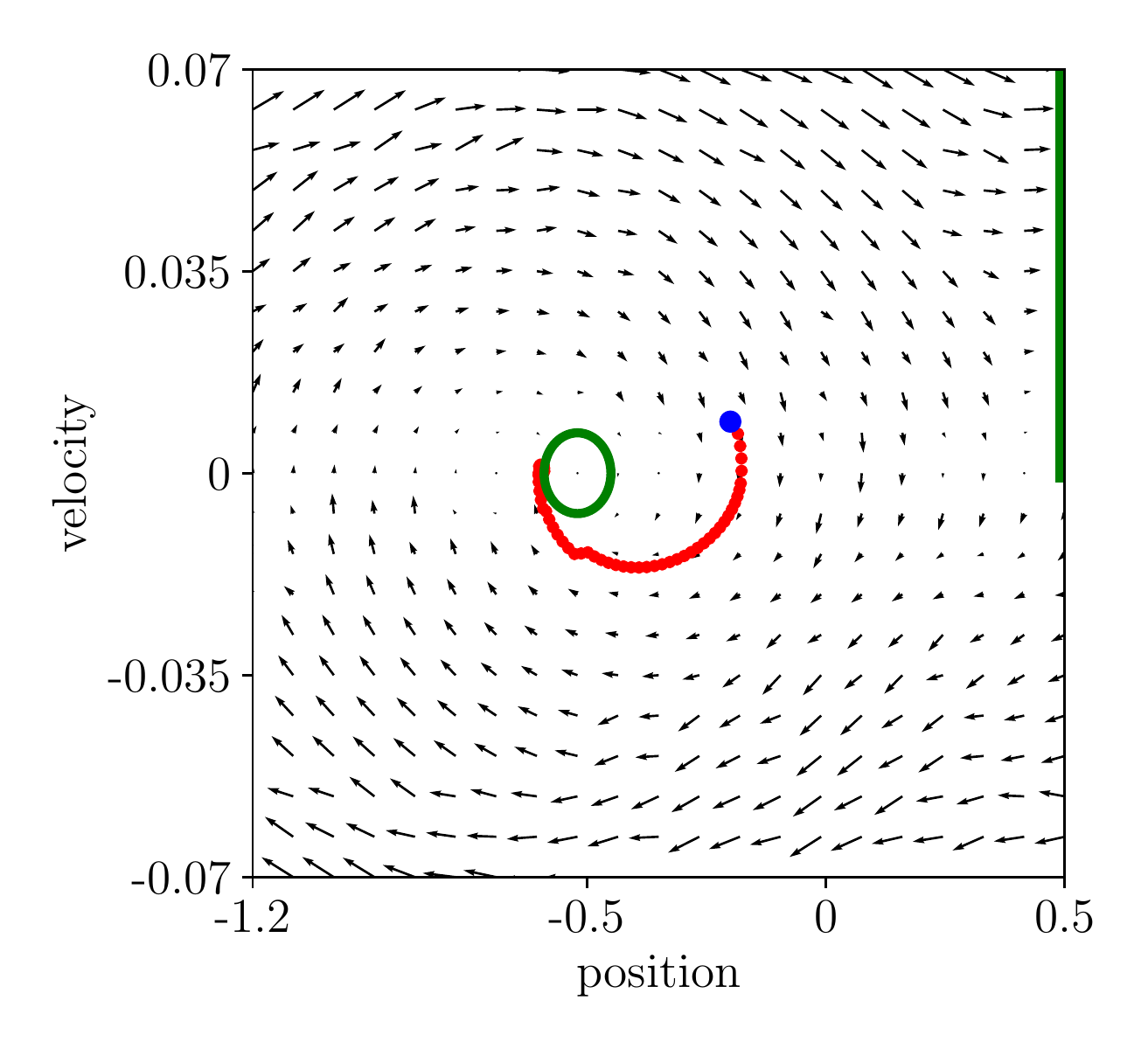}
    \caption{Policy}
\end{subfigure}%
\caption{Plots showing estimated values, reward model, the value of the predicted next state produced by the model, and the learned policy of linear Dyna in the deterministic MountainCarLoCA domain. Comparing Sub-figure (a), (e) and Sub-figure a), c) in Figure \ref{fig:optimal heatmaps} show that resulting values are far from optimal values.}
\label{fig: linear dyna plots tabular}
\end{figure*}

\begin{figure*}[h!]
\centering
\begin{subfigure}{.25\textwidth}
    \centering
    \includegraphics[width=0.95\textwidth]{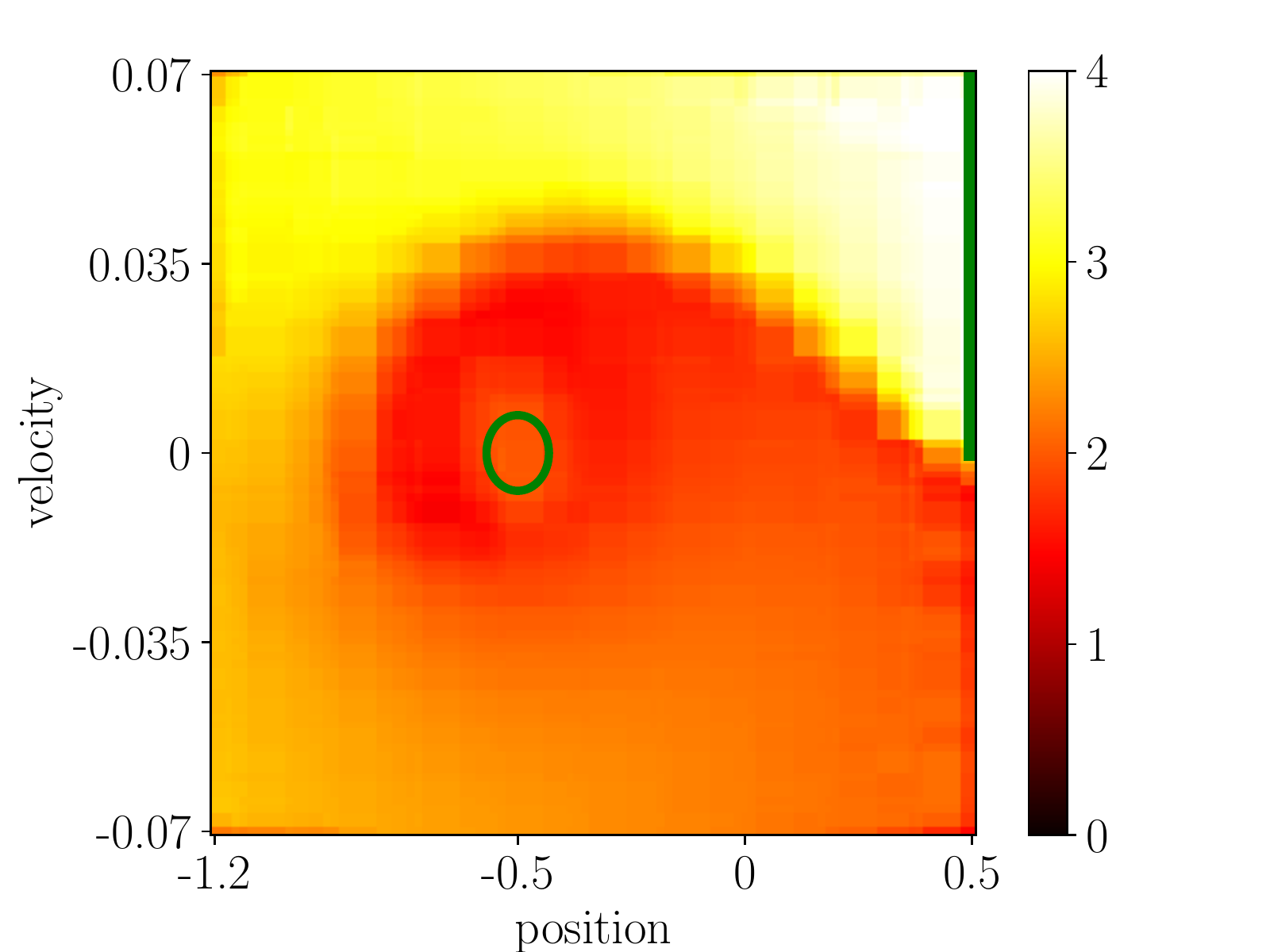}
    \caption{Estimated Values (Phase 1)}
\end{subfigure}%
\centering
\begin{subfigure}{.205\textwidth}
    \centering
    \includegraphics[width=0.95\textwidth]{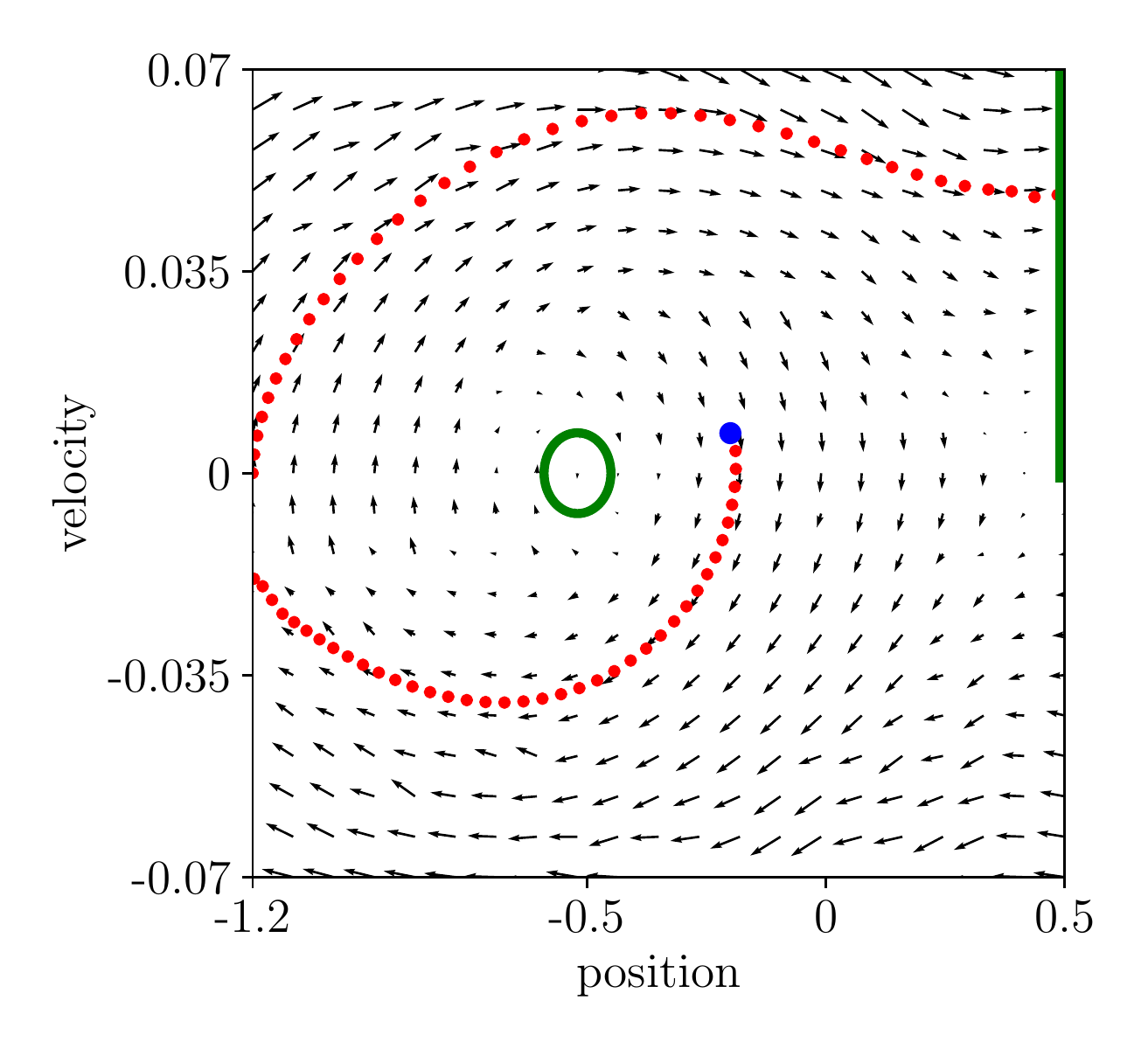}
    \caption{Policy (Phase 1)}
\end{subfigure}%
\begin{subfigure}{.25\textwidth}
    \centering
    \includegraphics[width=0.95\textwidth]{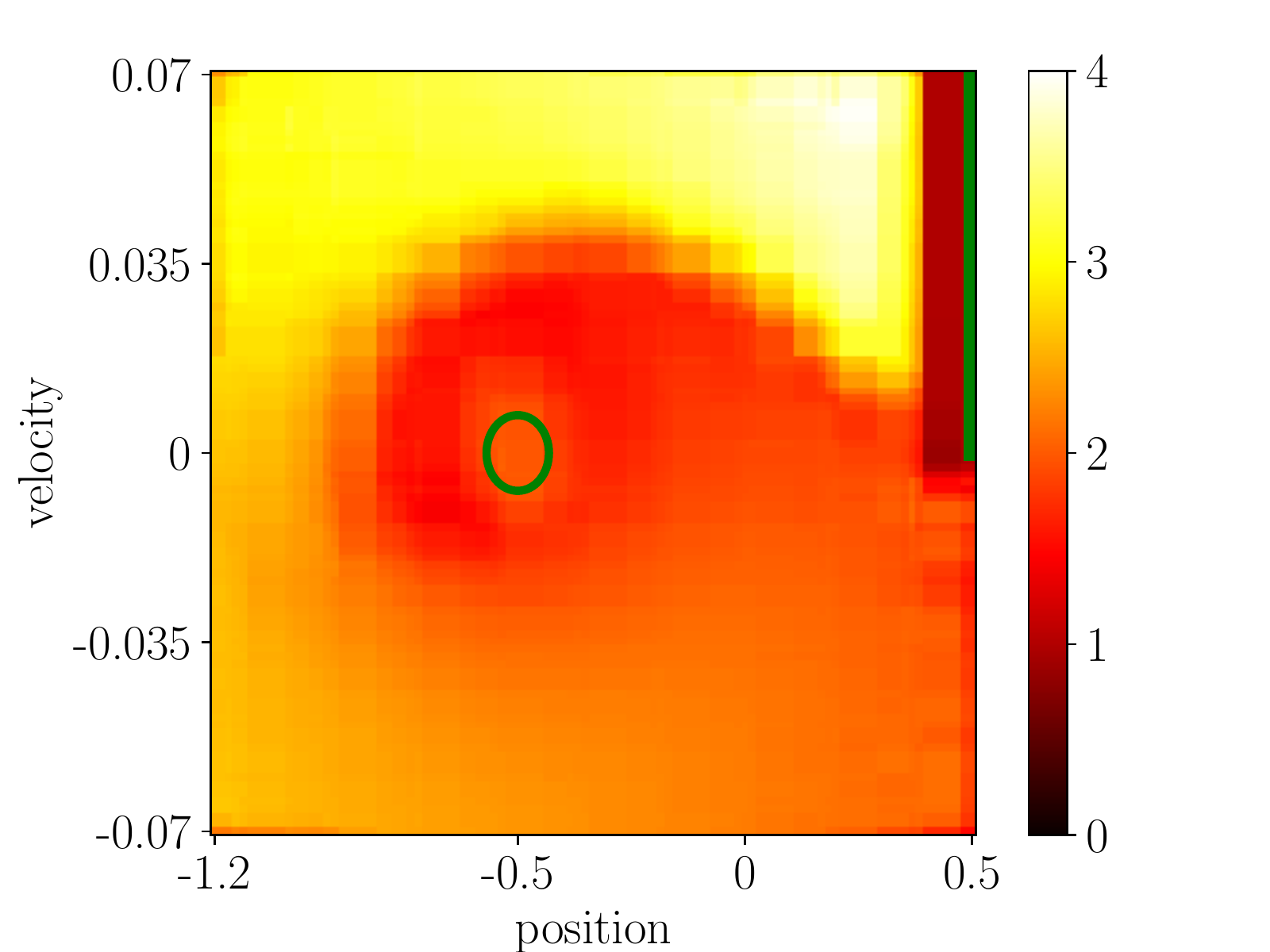}
    \caption{Estimated Values (Phase 2)}
\end{subfigure}
\begin{subfigure}{.205\textwidth}
    \centering
    \includegraphics[width=0.95\textwidth]{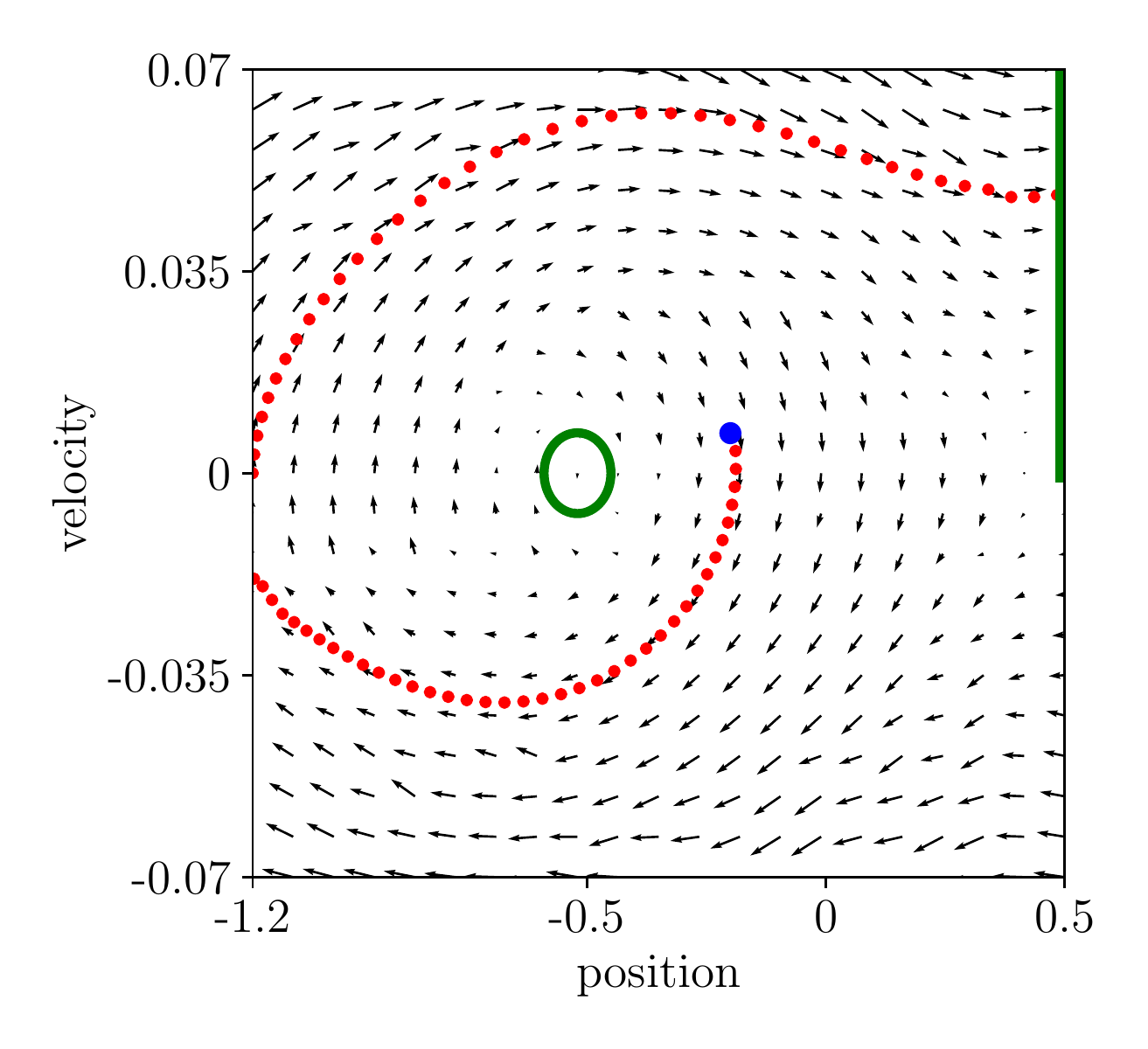}
    \caption{Policy (Phase 2)}
\end{subfigure}
\caption{Plots showing learning curves of linear Sarsa($\lambda$) in the deterministic MountainCarLoCA domain. It can be seen that in the second phase, only the action values of a local region were updated because only data in that region were collected. And thus the policy is not updated for most of states.}
\label{fig: linear Sarsa plots}
\end{figure*}

\begin{figure*}[h!]
\centering
\begin{subfigure}{.25\textwidth}
    \centering
    \includegraphics[width=0.95\textwidth]{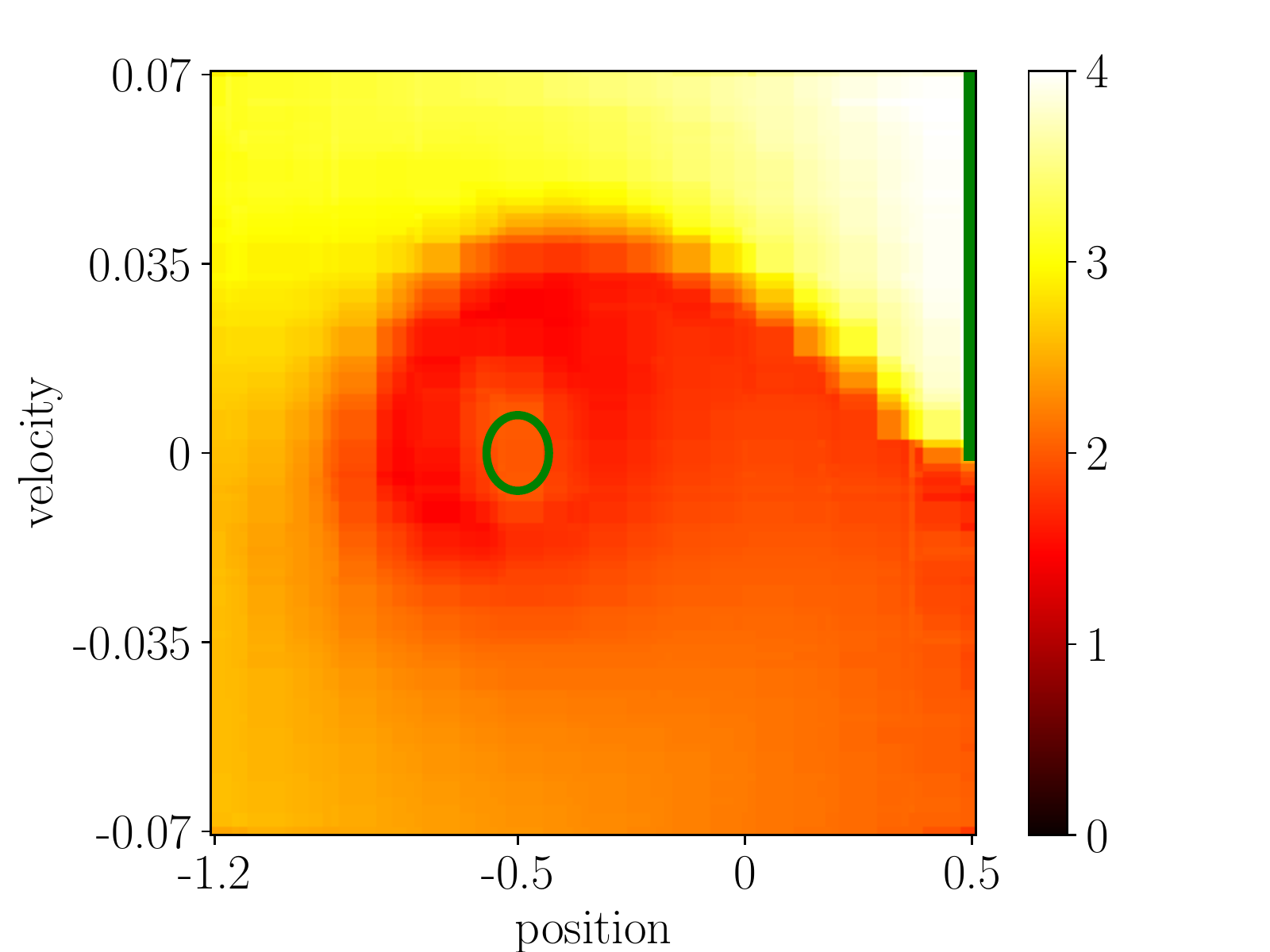}
    \caption{Optimal Values (Task 1)}
\end{subfigure}%
\centering
\begin{subfigure}{.205\textwidth}
    \centering
    \includegraphics[width=0.95\textwidth]{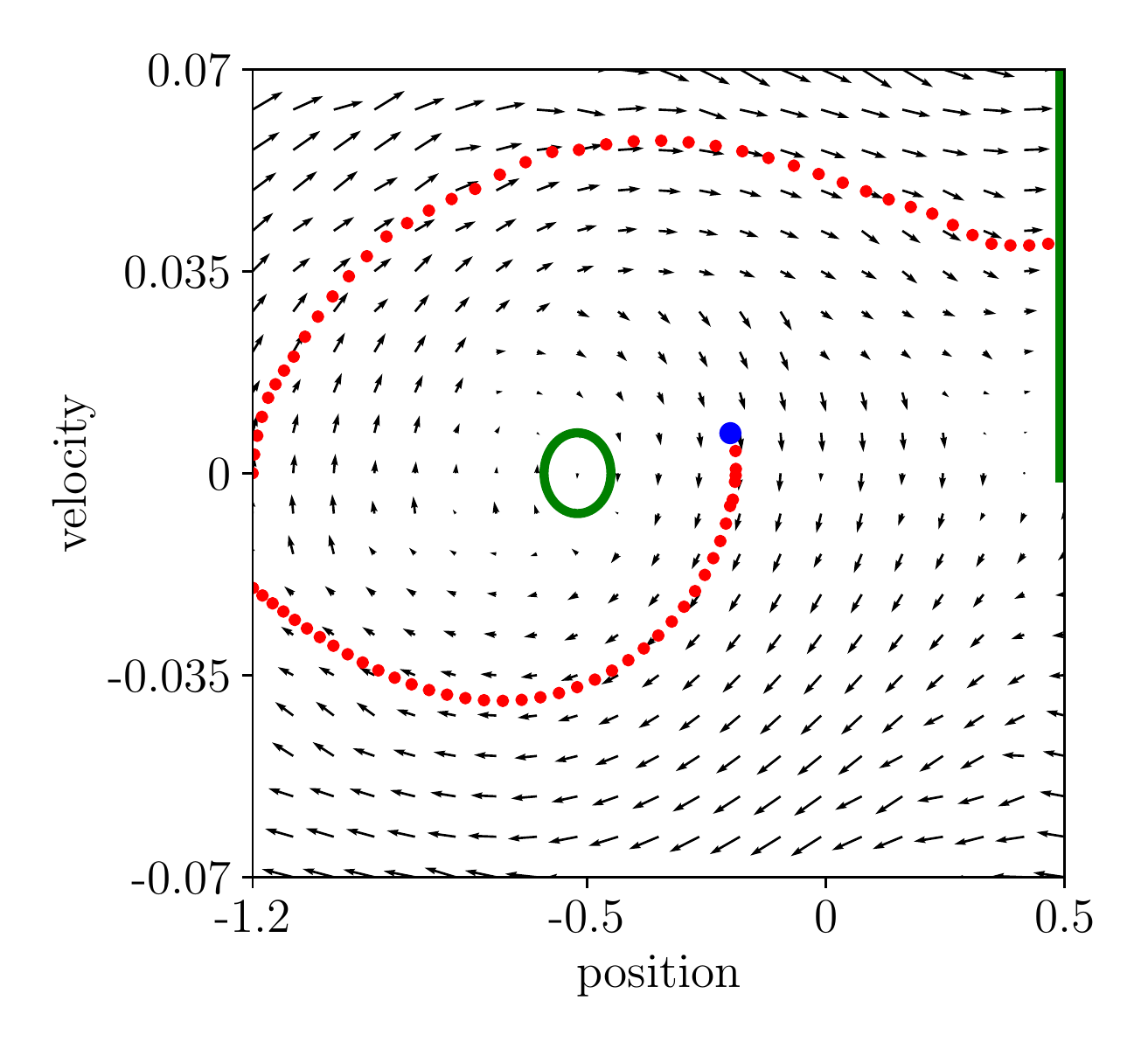}
    \caption{Optimal Policy (Task 1)}
\end{subfigure}%
\begin{subfigure}{.25\textwidth}
    \centering
    \includegraphics[width=0.95\textwidth]{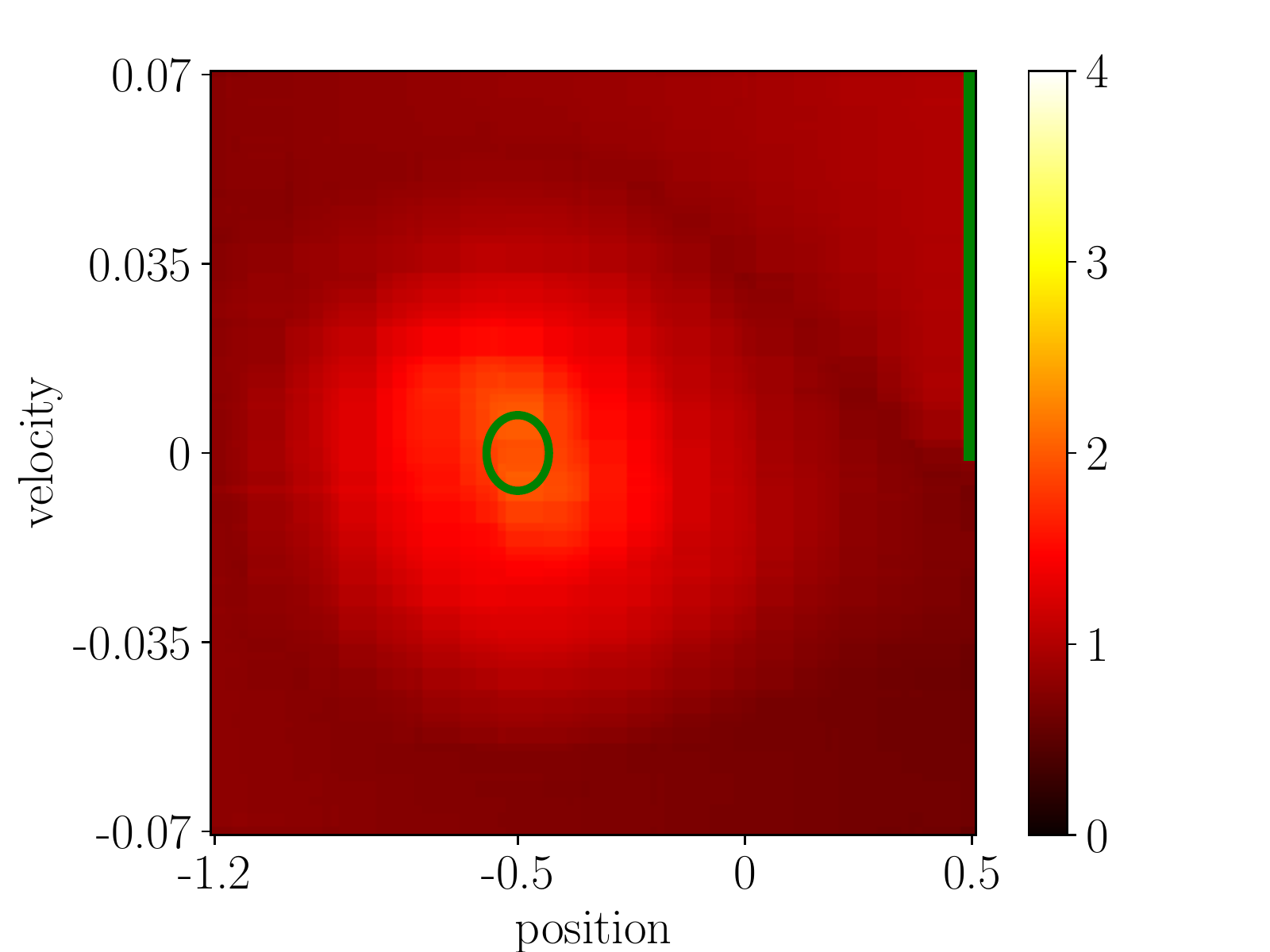}
    \caption{Optimal Values (Task 2)}
\end{subfigure}
\begin{subfigure}{.205\textwidth}
    \centering
    \includegraphics[width=0.95\textwidth]{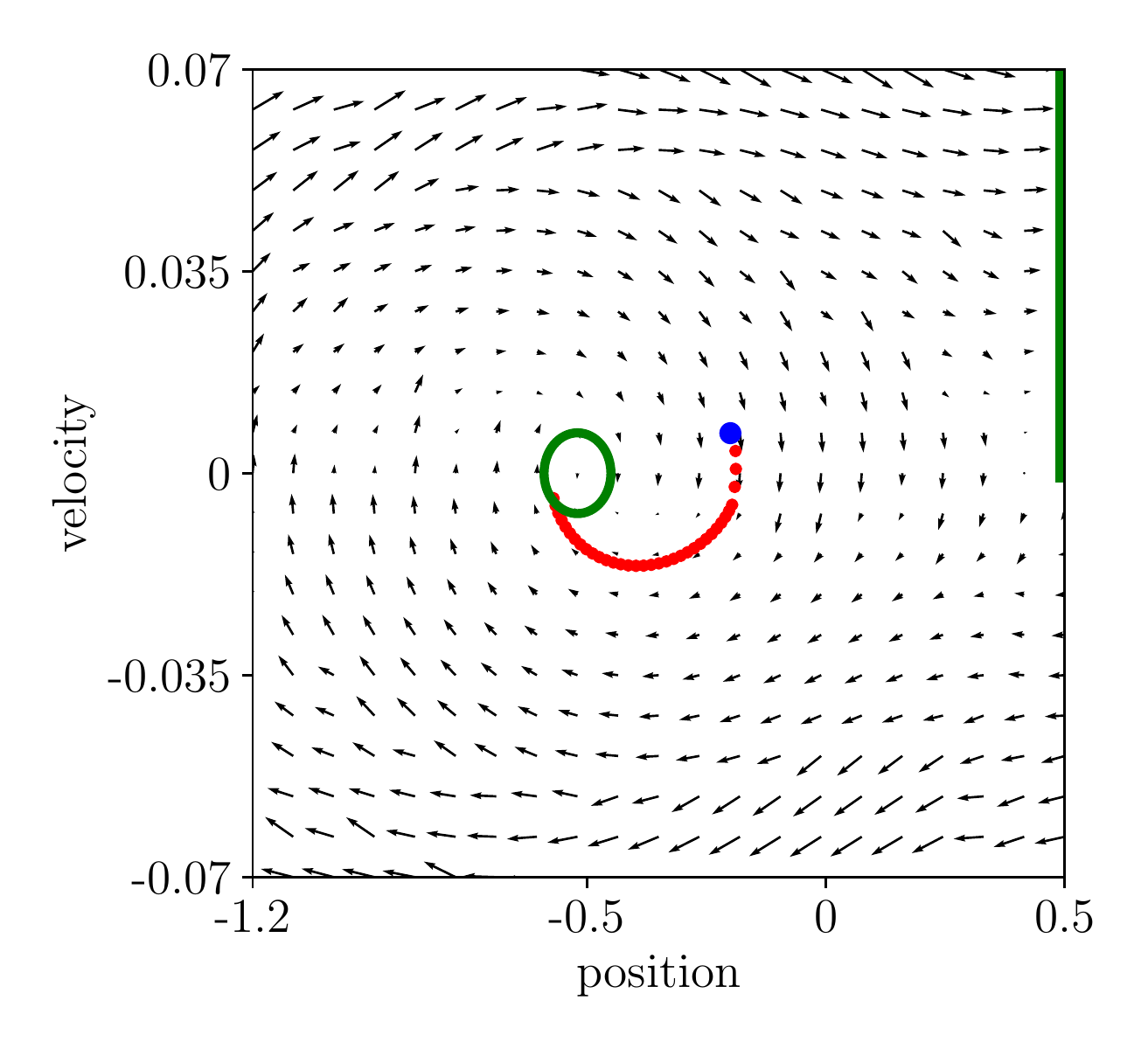}
    \caption{Optimal Policy (Task 2)}
\end{subfigure}
\caption{Plots showing "optimal" values and policies in the two tasks obtained by running linear Sarsa($\lambda$) for sufficiently long in the deterministic MountainCarLoCA domain.}
\label{fig:optimal heatmaps}
\end{figure*}

For better understanding of the tested linear algorithms, we plotted their learned values, policies, reward model, and predicted next value for adaptive linear Dyna (Figure\ \ref{fig: linear dyna plots}) and linear Dyna (Figure\ \ref{fig: linear dyna plots tabular}). 

Plots for Sarsa($\lambda$) are summarized in Figure\ \ref{fig: linear Sarsa plots}. We also plotted learned values and policies by the Sarsa($\lambda$) algorithm in the two tasks when the algorithm is applied for sufficiently long.

First, we note that the model learning part of the adaptive linear Dyna algorithm with sparse tile-coded features does not suffer from catastrophic forgetting. This can be seen by comparing sub-figures (b) and (f) in Figure\ \ref{fig: linear dyna plots} or those in Figure\ \ref{fig: linear dyna plots tabular}. 
We also note that, in Phase 1, linear Dyna produced inferior value estimates (the upper part of sub-figure (a) should be less than 4 because of discounting while it is not) even though the greedy policy w.r.t. these values is not apparently inferior (sub-figure (d)). In Phase 2, the inferior values are even more significant, and the policy did not produce a trajectory that leads to T2. We believe that incorrect value estimate in Phase 1 is not as severe as it in Phase 2 because the model-free learning part of the linear Dyna algorithm helps value estimation for most of states in Phase 1, but only for states within T1-zone in Phase 2.

\newpage
\subsection{Setup of the Nonlinear Dyna Q Experiment}\label{app: Setups for the Nonlinear Dyna Q Experiment}

We summarize in Table \ref{tab: Tested hyperparameters in the nonlinear Dyna Q experiment} tested hyperparameters for the nonlinear Dyna Q algorithm.
\begin{table}[h!]
    \centering
    \begin{tabular}{c|c|c}
        \hline
        \multirow{4}{4em}{Neural networks} & Dynamics model  & $64 \times 64 \times 64 \times 64$, tanh\\
        & Reward model  & $64 \times 64 \times 64 \times 64$, tanh\\
        & Termination model  & $64 \times 64 \times 64 \times 64$, tanh\\
        & Action-value estimator  & $64 \times 64 \times 64 \times 64$, tanh\\
        \hline
        \multirow{3}{4em}{Optimizer} & Value optimizer & \begin{tabular}[h]{@{}c@{}}Adam, step-size\\$\alpha \in [5e-7, 1e-6, 5e-6, 1e-5, 5e-5]$\end{tabular}\\
        & Model optimizer & \begin{tabular}[h]{@{}c@{}}Adam, step-size\\$\beta \in [1e-6, 5e-6, 1e-5, 5e-5, 1e-4]$\end{tabular}\\
        & Other value and model optimizer parameters & Default choices in Pytorch \citep{paszke2019pytorch}\\
        \hline
        \multirow{8}{4em}{Other details} & Exploration parameter ($\epsilon$) & $[0.1, 0.5, 1.0]$\\
        & Target network update frequency ($k$) & 500\\
        & Number of model learning steps ($m$) & 5\\
        & Number of planning steps ($m$) & 5\\
        & Size of learning buffer ($n_l$) & $[300000, 3000000]$\\
        & Size of planning buffer ($n_p$) & $3000000$\\
        & Mini-batch size of model learning ($m_l$) & 32\\
        & Mini-batch size of planning ($m_p$) & 32\\
    \hline
    \end{tabular}
    \caption{Tested hyperparameters in the nonlinear Dyna Q experiment.}
    \label{tab: Tested hyperparameters in the nonlinear Dyna Q experiment}
\end{table}

Table \ref{tab: Experiment Procedure of the LoCA Setup to Test Non-linear Dyna Q} shows the experiment setup used to test the nonlinear Dyna Q algorithm (Algorithm \ref{algo:Non-Linear Dyna Q}).

\begin{table}[h!]
    \centering
    \begin{tabular}{c|c|c}
        \hline
        \multirow{4}{5em}{Initial distributions} & Phase 1 training  & Uniform distribution over the entire state space\\
        & Phase 1 evaluation  & Uniform distribution over a small region\\
        & Phase 2 training  & Uniform distribution over states within T1-zone\\
        & Phase 2 evaluation  & Uniform distribution over a small region\\
        \hline
        \multirow{2}{5em}{Training steps} & Phase 1 steps & $1.5 \times 10^6$\\
        & Phase 2 steps & $1.5 \times 10^6$\\
        \hline
        \multirow{3}{5em}{Other details} & Training steps between two evaluations & $10^4$\\
        & Number of runs & $5$\\
        & Number of evaluation episodes & $10$\\
    \hline
    \end{tabular}
    \caption{Experiment details used to test nonlinear Dyna Q.}
    \label{tab: Experiment Procedure of the LoCA Setup to Test Non-linear Dyna Q}
\end{table}

\subsection{Additional Information about the Result Shown in Figure \ref{fig: nonlineardyna}}\label{app: Additional Information about the Result Shown in Figure Nonlineardyna}

The set of hyperparameters used to generate learning curves in Figure \ref{fig: linear dyna} is chosen in the way the set of hyperparameters chosen in the adaptive linear Dyna experiment. The best hyperparameter setting of nonlinear Dyna Q in this experiment is summarized in Table\ \ref{tab: Best Parameter Setting for Non-linear Dyna Q.}.

\begin{table}[h!]
    \centering
    \begin{tabular}{c|c|c}
        \hline
        \multirow{2}{8em}{Optimizer} & Value optimizer & Adam, step-size $\alpha = 5e-6$\\
        & Model optimizer & Adam, step-size $\beta = 5e-5$\\
        \hline
        \multirow{2}{8em}{Other hyperparameters} & Exploration parameter ($\epsilon$) & $0.5$\\
        & Size of learning buffer ($n_l$) & $ 3000000$\\
    \hline
    \end{tabular}
    \caption{Best hyperparameter setting for nonlinear Dyna Q.}
    \label{tab: Best Parameter Setting for Non-linear Dyna Q.}
\end{table}

\subsection{Additional Results of the Nonlinear Dyna Q Experiment}\label{app: Additional Results of the Nonlinear Dyna Q Experiment}

\begin{figure*}[h!]
\centering
\begin{subfigure}{.25\textwidth}
    \centering
    \includegraphics[width=0.95\textwidth]{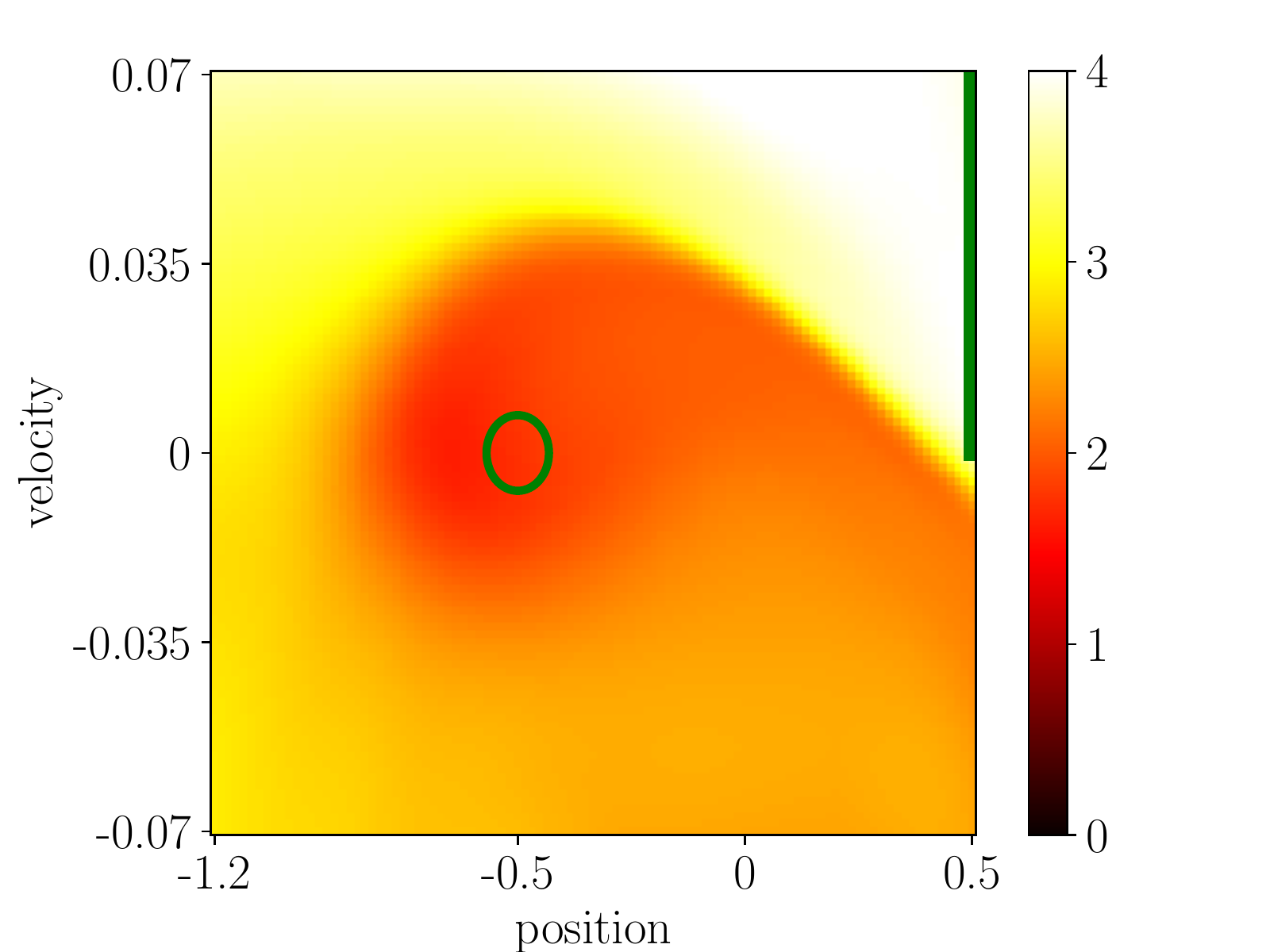}
    \caption{Estimated Values}
\end{subfigure}%
\begin{subfigure}{.25\textwidth}
    \centering
    \includegraphics[width=0.95\textwidth]{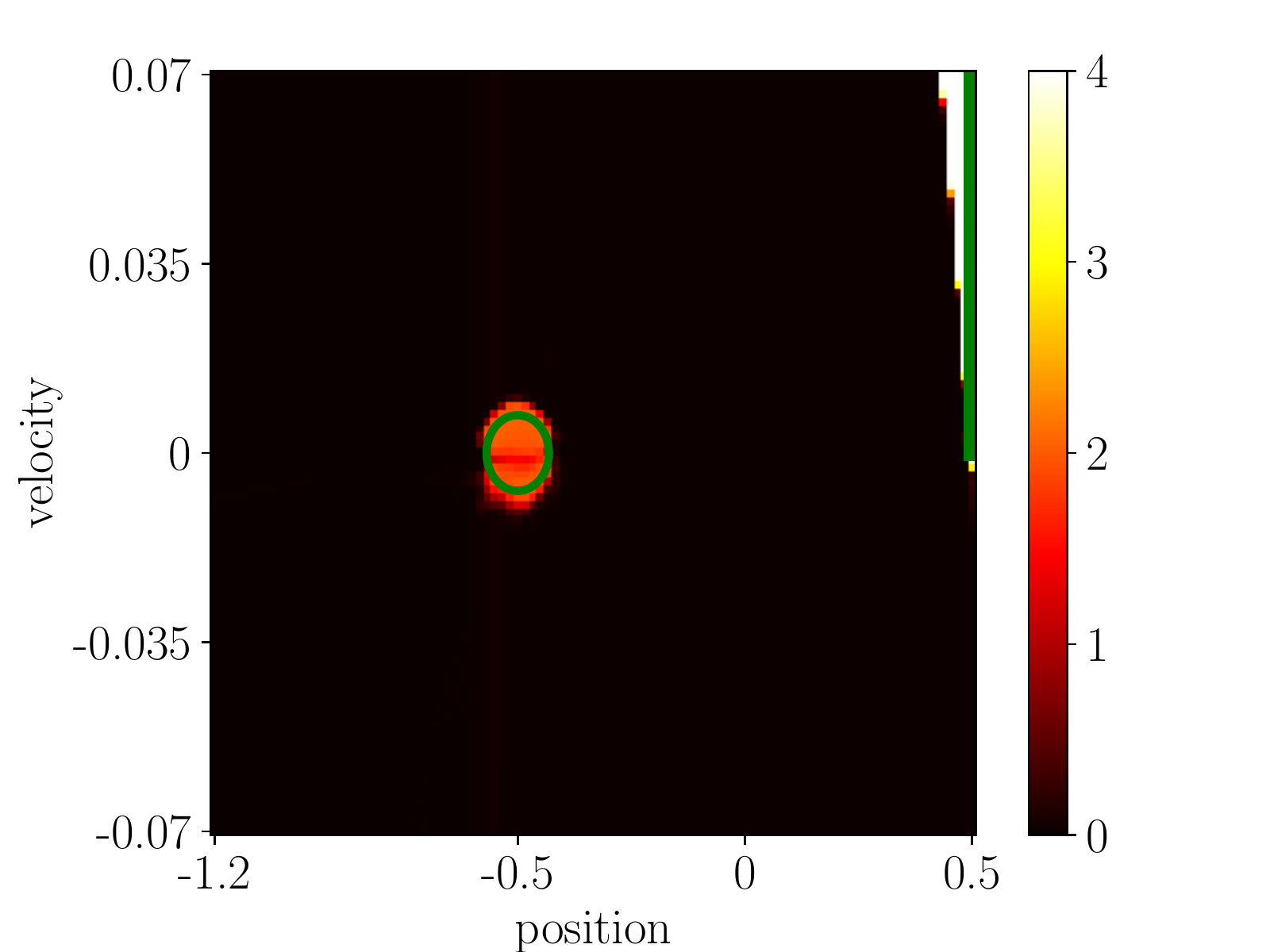}
    \caption{Estimated Rewards}
\end{subfigure}%
\begin{subfigure}{.25\textwidth}
    \centering
    \includegraphics[width=0.95\textwidth]{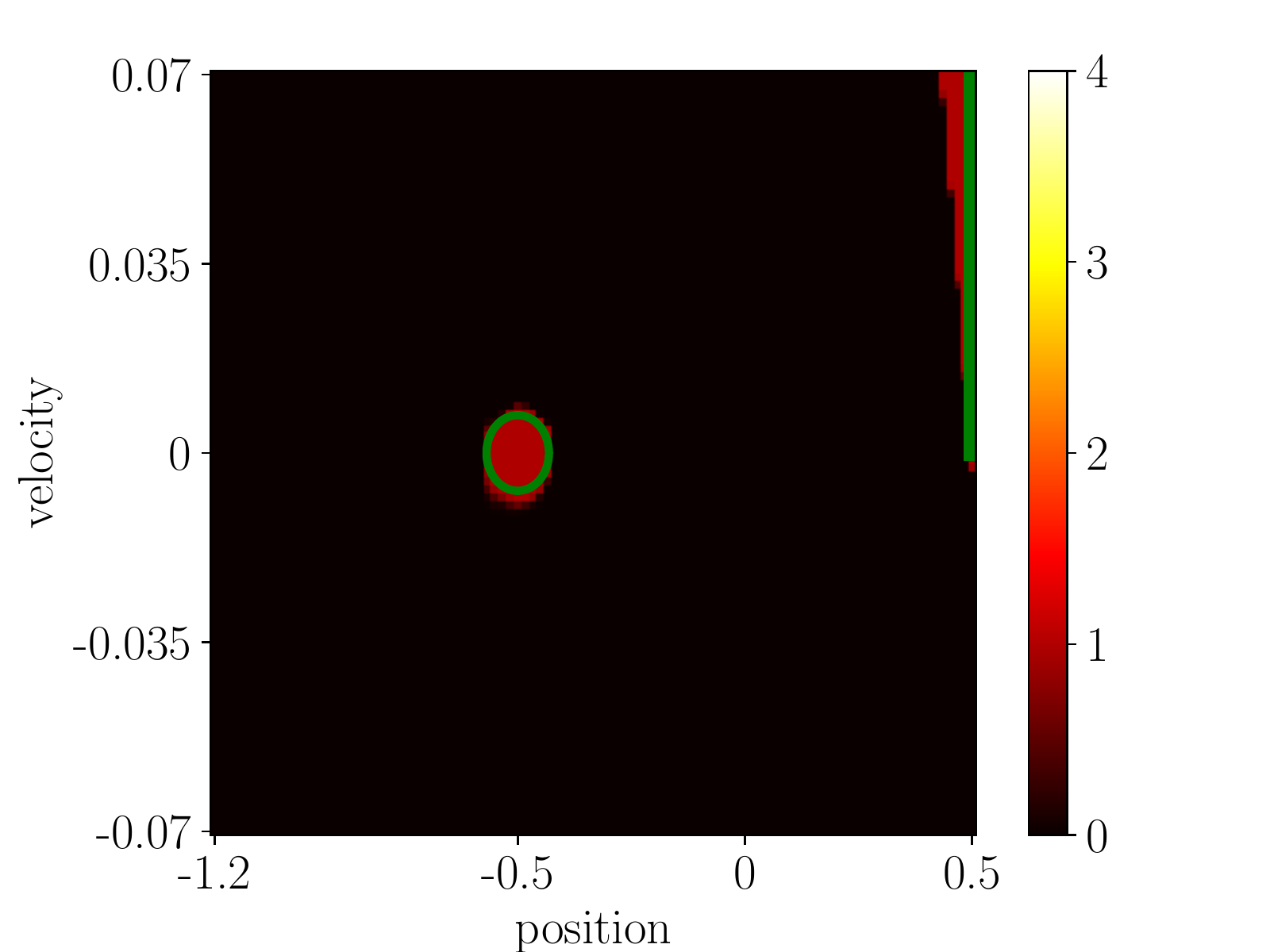}
    \vspace{-0.6mm}
    \caption{Estimated Terminations}
\end{subfigure}%
\begin{subfigure}{.2\textwidth}
    \centering
    \vspace{1.5mm}
    \includegraphics[width=0.95\textwidth]{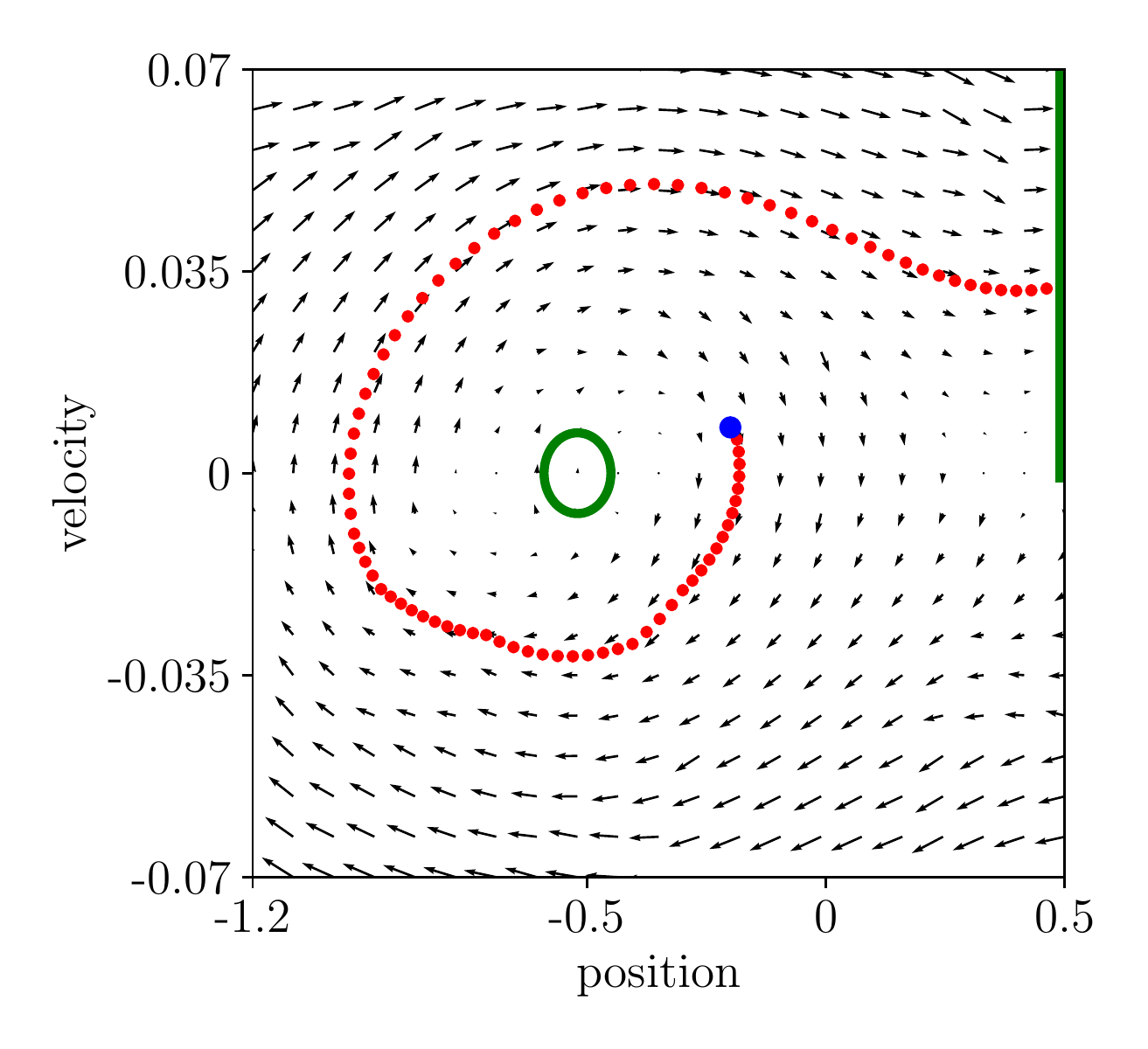}
    \vspace{-1.5mm}
    \caption{Policy}
\end{subfigure}\\
\begin{subfigure}{.25\textwidth}
    \centering
    \includegraphics[width=0.95\textwidth]{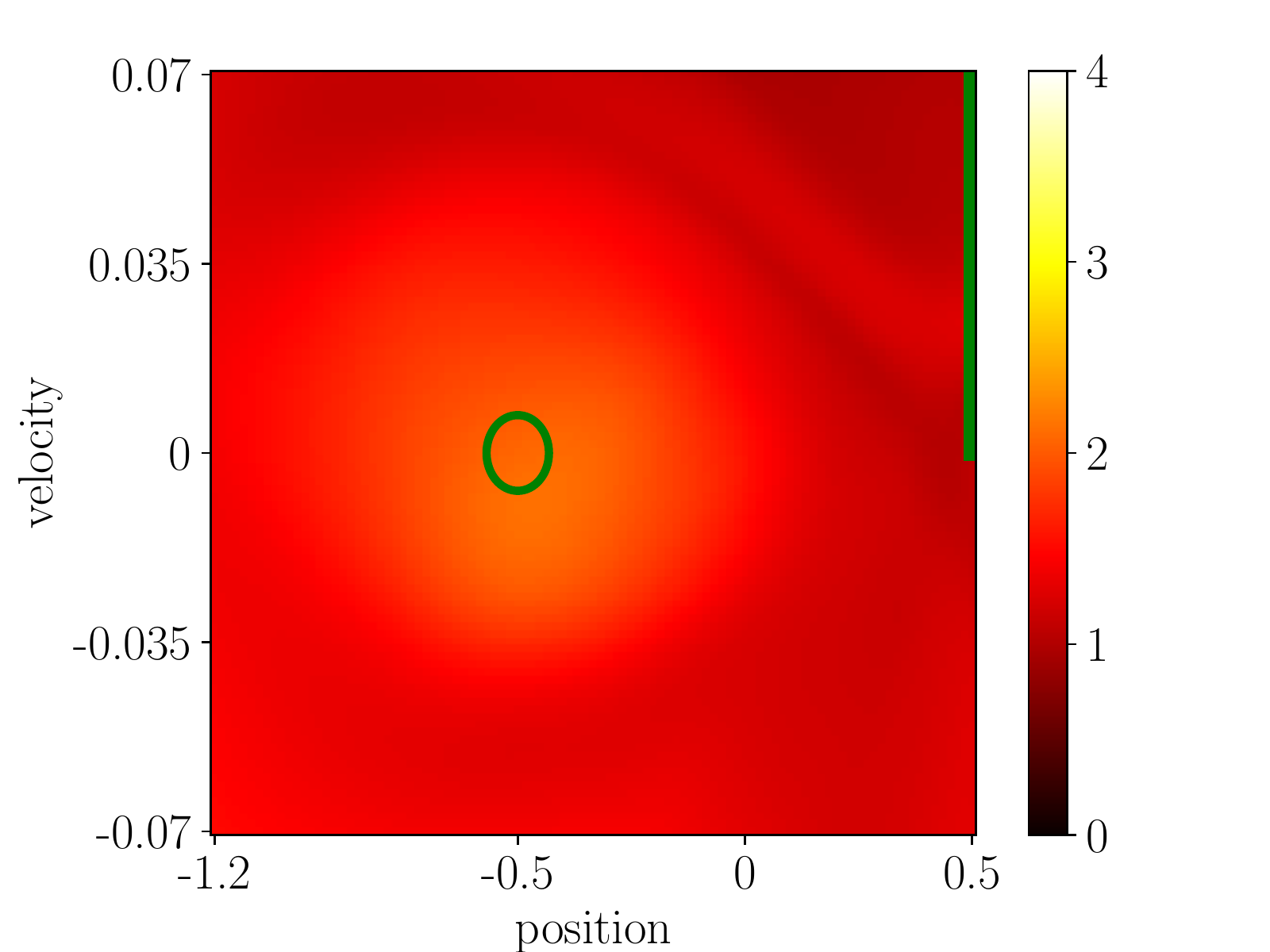}
    \caption{Estimated Values}
\end{subfigure}%
\begin{subfigure}{.25\textwidth}
    \centering
    \includegraphics[width=0.95\textwidth]{ICML2022_figures/472_3000000_reward.pdf}
    \caption{Estimated Rewards}
\end{subfigure}%
\begin{subfigure}{.25\textwidth}
    \centering
    \includegraphics[width=0.95\textwidth]{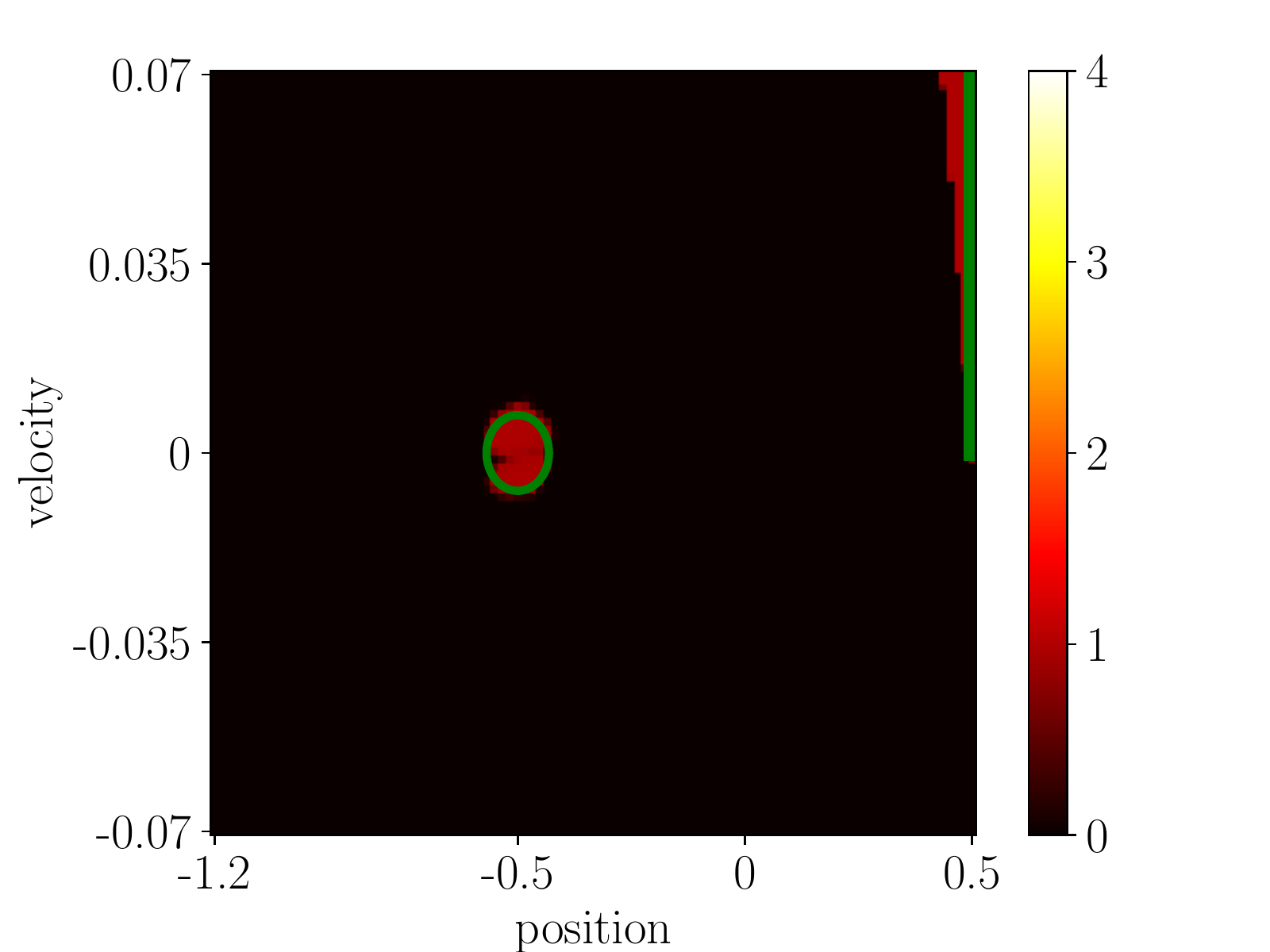}
    \vspace{-0.6mm}
    \caption{Estimated Terminations}
\end{subfigure}%
\begin{subfigure}{.2\textwidth}
    \centering
    \vspace{1.5mm}
    \includegraphics[width=0.95\textwidth]{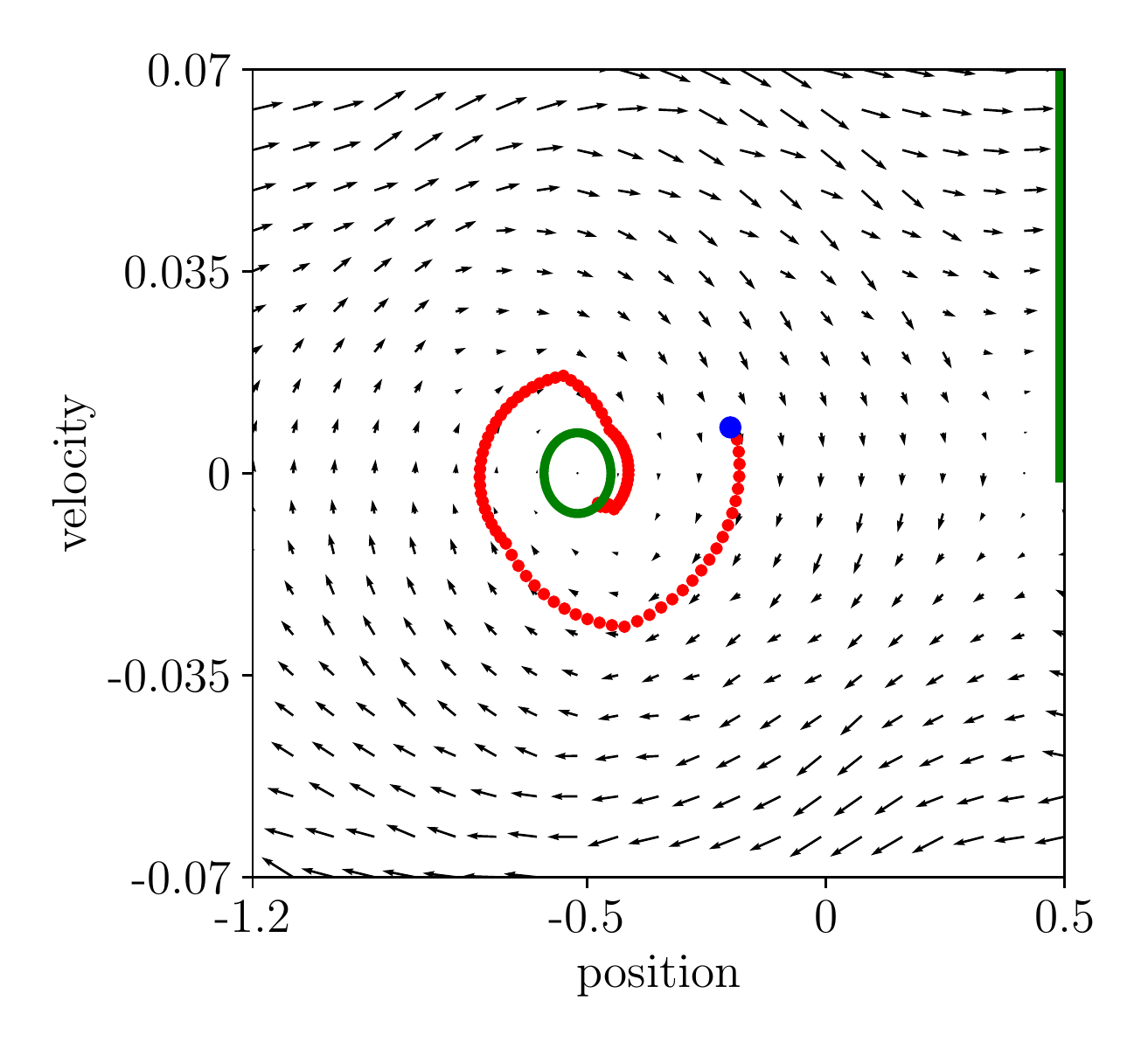}
    \vspace{-1.5mm}
    \caption{Policy}
\end{subfigure}
\caption{Plots showing learning curves, estimated values, rewards, and learned policy of nonlinear Dyna Q in the deterministic MountainCarLoCA domain. 
}
\label{fig: nonlineardyna 2}
\end{figure*}

Just as what we did for linear algorithm, we plot the estimated values, reward and termination models, as well as the learned policy in Figure\ \ref{fig: nonlineardyna 2}.

Comparing sub-figures (b) and (f) shows that the middle orange region changes to two smaller red regions, indicating that the algorithm forgets the reward model as more and more training goes on for the T1 part. Although sub-figure (e) still seems to be correct, but a deeper look at the learned policy shows that the trajectory generated by the policy (sub-figure (h)) is not optimal (c.f. sub-figure (h) in Figure\ \ref{fig: linear dyna plots}).

\newpage

\section{Pseudo Code} \label{app: Pseudo Code}
We provide the pseudo code of the adaptive linear Dyna algorithm, and the nonlinear Dyna Q algorithm in Algorithm \ref{algo:adaptive linear dyna} and \ref{algo:Non-Linear Dyna Q} respectively.

In addition to maintaining a learning buffer and a planning buffer, there are two other ways in which the nonlinear algorithm is different from the linear algorithm. First, the nonlinear algorithm needs to predict the termination of each episode, while the linear algorithm can simply assign the terminal state with a zero feature vector. In addition, for the nonlinear algorithm, we chose to estimate the action-value function and maintained separate networks for different actions (i.e., no parameters are shared across different actions). In this way, interference among different actions is removed.

\begin{algorithm}[H]
% \DontPrintSemicolon
\SetAlgoLined
\KwIn{exploration parameter $\epsilon$, stepsize $\alpha$, a feature mapping $\phi: \calS \to \bbR^m$ that produces from a state $s$ the corresponding $m$-sized feature vector $\phi(s)$, number of planning steps $n$, and the size of the planning buffer $n_p$.}
\SetKwInput{AP}{Algorithm parameters}
\SetKwRepeat{Do}{do}{while}
Initialize weights of the approximate value function $\theta \in \bbR^m$, weights of the linear dynamics model for each action $F_a \in \bbR^m \times \bbR^m$, and weights of the linear reward model for each action $b_a \in \bbR^m, \forall a \in \calA$ arbitrarily (e.g., to zero)\\
Obtain initial state $S$, Compute feature vector $\phi$ from $S$.\\
\While{still time to train}
{
    $A \gets \epsilon \text{-greedy}_a (b^\top_a \phi + \gamma \theta^\top F^\top_a \phi)$\\
    Take action $A$, receive reward $R$, next state $S'$, and episode termination signal $Z$ ($Z = 1$ means the current episode terminates. $Z = 0$ means the episode continues). \\
    Add $\phi$ to the planning buffer (if the buffer is full, remove the oldest element and then add $\phi$).\\
    Compute the next feature vector $\phi'$ from $S'$.\\
    \If{$A$ is greedy}
    {
        $\delta \gets R + \gamma (1 - Z) \theta^\top \phi' - \theta^\top \phi$\\
        $\theta \gets \theta + \alpha \delta \phi$
    }
    $F_A \gets F_A + \beta ((1 - Z)\phi' - F_A \phi) \phi^\top$\\
    $b_A \gets b_A + \beta (r - b_A^\top \phi) \phi$\\
    \For{all of the $n$ planning steps}
    {
        Generate a start feature vector $x$ by randomly sampling from the planning buffer.\\
        $\delta \gets \max_a (b_a^\top x + \gamma  \theta^\top F_a^\top x) - \theta^\top x$\\
        $\theta \gets \theta + \alpha \delta x$
    }
    $\phi \gets \phi'$
}
\caption{Adaptive Linear Dyna}
\label{algo:adaptive linear dyna}
\end{algorithm}

\begin{algorithm}[H]
% \DontPrintSemicolon
\SetAlgoLined
\KwIn{exploration parameter $\epsilon$, value step-size $\alpha$, model step-size $\beta$, target network update frequency $k$, parameterized value estimator $\hat q_\bfw$, parameterized dynamics model $f_\bfu$, parameterized reward model $b_\bfv$, parameterized termination model $t_\bfz$, number of model learning steps $m$, number of planning steps $n$, size of the learning buffer $n_l$, size of the planning buffer $n_p$, mini-batch size of model-learning $m_l$, mini-batch size of planning $m_p$.}
\SetKwInput{AP}{Algorithm parameters}
\SetKwRepeat{Do}{do}{while}
Randomly initialize $\bfw, \bfu, \bfv, \bfz$. Initialize target network parameters $\tilde \bfw = \bfw$. Initialize value optimizer with step-size $\alpha$ and parameters $\bfw$, model optimizer with step-size $\beta$ and parameters $\bfu, \bfv, \bfz$.\\
Obtain the vector representation of the initial state $\bfs$.\\
\While{still time to train}
{
    $A \gets \epsilon \text{-greedy}_a (\hat q_\bfw(\bfs, \cdot))$\\
    Take action $A$, receive reward $R$, vector representation of the next state $\bfs'$, and termination $Z$. \\
    Add $(\bfs, A, R, \bfs', Z)$ to the learning buffer. If the learning buffer is full, replace the oldest element by $(\bfs, A, R, \bfs', Z)$.\\
    Add $\bfs$ to the planning buffer. If the planning buffer is full, replace the oldest element by $\bfs$.\\
    \For{all of the $m$ model learning steps}{
    Sample a $m_l$-sized mini-batch $(\bfs_i, a_i, r_i, \bfs'_i, z_i)$ from the learning buffer.\\
    Apply model optimizer to the loss $\sum_i \frac{1 - z_i}{2} (\bfs'_i - f_\bfu(\bfs_i, a_i))^2 + \frac{1}{2} (r_i - b_\bfv(\bfs_i, a_i))^2 + \frac{1}{2} (z_i - t_\bfz(\bfs_i, a_i))^2$\\
    }
    \For{all of the $n$ planning steps}
    {
        Sample a $m_p$-sized mini-batch of feature vectors $\{\bfx_i\}$ from the planning buffer.\\
        Sample a $m_p$-sized mini-batch of actions $\{\bfa_i\}$ uniformly.\\
        Apply value optimizer to the loss $\sum_i \frac{1}{2}(b_\bfv(\bfx_i, \bfa_i) + \gamma (1 - t_\bfz(\bfx_i, \bfa_i)) \max_a \hat q_{\tilde \bfw} (f_\bfu(\bfx_i, \bfa_i), a) - \hat q_\bfw(\bfx_i, \bfa_i))^2$
    }
    $\bfs \gets \bfs'$\\
    $t \gets t + 1$\\
    \If{$t \% k == 0$}
    {
    $\tilde \bfw = \bfw$
    }
}
\caption{Nonlinear Dyna Q}
\label{algo:Non-Linear Dyna Q}
\end{algorithm}

\end{document}